\documentclass{article}

\PassOptionsToPackage{numbers,compress}{natbib}
\PassOptionsToPackage{table}{xcolor}

\usepackage[preprint]{neurips_2026}

\usepackage[utf8]{inputenc}
\usepackage[T1]{fontenc}

\usepackage{amsmath}
\usepackage{amsfonts}
\usepackage{nicefrac}

\usepackage{graphicx}
\usepackage{xcolor}
\usepackage{caption}
\usepackage{adjustbox}
\usepackage{booktabs}
\usepackage{multirow}
\usepackage{makecell}

\usepackage{algorithm}
\usepackage{algorithmic}

\usepackage{microtype}
\usepackage{tcolorbox}
\usepackage{xspace}
\usepackage{url}

\usepackage{hyperref}
\usepackage{cleveref}

\usepackage{amsmath,amsfonts,bm}

\def\eqref#1{equation~\ref{#1}}

\def\1{\bm{1}}

\DeclareMathAlphabet{\mathsfit}{\encodingdefault}{\sfdefault}{m}{sl}
\SetMathAlphabet{\mathsfit}{bold}{\encodingdefault}{\sfdefault}{bx}{n}

\newcommand{\eg}{\textit{e.g.}\xspace}

\newif\ifreview 
\newif\ifarxiv 
\newif\ifcamera 
\newif\ifrebuttal

\newcommand{\ourBench}{LU-500\xspace}
\newcommand{\ourBenchEasy}{LUex-500\xspace}
\newcommand{\ourBenchHard}{LUim-500\xspace}
\newcommand{\ourBaseline}{ProLU\xspace}
\newcommand{\ourSD}{stable-diffusion-3-medium\xspace}
\newcommand{\numBench}{9584\xspace}
\newcommand{\numBenchEasy}{4748\xspace}
\newcommand{\numBenchHard}{4836\xspace}

\newcommand{\metricone}{{CLIPScore}\xspace}
\newcommand{\metrictwo}{{LogoScore}\xspace}
\newcommand{\metricthree}{{LogoSSIM}\xspace}
\newcommand{\metricfour}{{ImageScore}\xspace}
\newcommand{\metricfive}{{ImageSSIM}\xspace}
\newcommand{\sldhalf}{{SLD\_v2}\xspace}
\newcommand{\sldone}{{SLD\_v3}\xspace}
\newcommand{\sldzero}{{SLD\_v1}\xspace}
\newcommand{\segahalf}{{SEGA\_v2}\xspace}
\newcommand{\segaone}{{SEGA\_v3}\xspace}
\newcommand{\segazero}{{SEGA\_v1}\xspace}

\makeatletter
\renewcommand{\@notice}{}
\makeatother

\title{\ourBench: A Logo Benchmark for Concept Unlearning}

\author{%
  Keyu Li \\
  Shanghai Jiao Tong University \\
  Shanghai, China \\
  \texttt{chlorophyll@sjtu.edu.cn}
  \And
  Jin Gao \\
  Shanghai Jiao Tong University \\
  Shanghai, China \\
  \texttt{gaojin@sjtu.edu.cn}
  \AND
  Jialing Zhang \\
  Shanghai Jiao Tong University \\
  Shanghai, China \\
  \texttt{jialingzhang@sjtu.edu.cn}
  \And
  Dequan Wang\thanks{%
    Corresponding author.\newline
    LU-500 is available at
    \href{https://github.com/weizhihao1/LU-500}{GitHub}
    and
    \href{https://huggingface.co/datasets/weizhihao1/LU-500}{Hugging Face}.
  } \\
  Shanghai Jiao Tong University \\
  Shanghai, China \\
  \texttt{dequanwang@sjtu.edu.cn}
}

\begin{document}

\maketitle

\begin{abstract}

Concept unlearning is increasingly used to limit the reproduction of protected or unsafe visual concepts in text-to-image models.
Existing evaluations, however, mostly study targets that dominate the whole image, such as styles, broad object categories, or portrait-like identities, leaving company logos comparatively underexamined.
Logos create a different failure mode: a small localized mark can carry the entire protected concept, must be visually precise to remain recognizable, and can be triggered implicitly by products, storefronts, packaging, or advertisements even when the word ``logo'' is absent.
We introduce \ourBench, a logo-unlearning benchmark built from Fortune Global 500 companies to study this localized and semantically entangled setting.
\ourBench contains nearly 10,000 curated text-query and logo-image pairs, with an explicit track (\ourBenchEasy) and an implicit contextual track (\ourBenchHard).
To avoid reducing the task to a binary detector score, we define a multi-grained protocol that evaluates both local logo removal and global image preservation in pixel and latent spaces.
Experiments on representative inference-time methods, including NP, SLD, and SEGA, and compatible fine-tuning-based methods such as ESD and Forget-Me-Not, show that the evaluated methods struggle to remove logo evidence without changing non-target content.
We further analyze \ourBaseline, a prompt-space multi-agent baseline: it improves local erasure by removing logo-inducing semantics, but also illustrates why prompt filtering is not a substitute for weight-level disentanglement.
Correlation analyses over logo area, location, and structural complexity suggest that future logo unlearning may need spatially aware controls, such as SSIM-guided constraints, rather than purely global concept suppression.

\end{abstract}

\begin{figure*}[ht]
\centering
\includegraphics[width=1.0\linewidth]{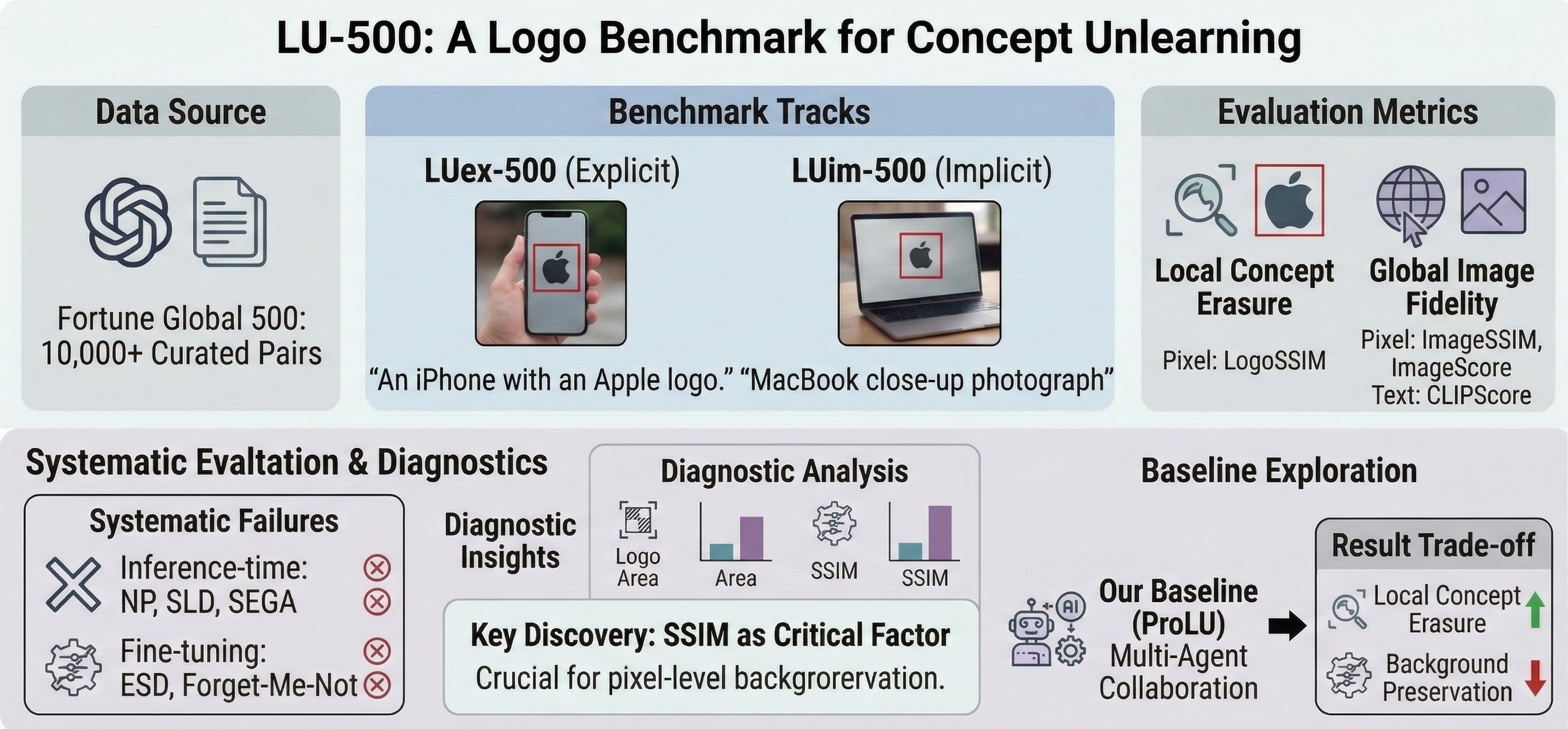}
\captionof{figure}{
\textbf{Overview of \ourBench, a benchmark evaluating logo unlearning on the Fortune Global 500.} We propose multi-grained metrics assessing local concept erasure and global image fidelity. Evaluations reveal existing inference-time approaches (\eg, NP~\citep{np}, SLD~\citep{schramowski2023safe}, SEGA~\citep{brack2023sega}) systematically fail to unlearn these entangled assets. Our prompt-based baseline, \ourBaseline, achieves superior local erasure but exposes a critical trade-off with background preservation. Finally, we diagnose these failures via correlation analysis of intrinsic logo characteristics.
}
\label{fig:teaser}
\end{figure*}

\section{Introduction}
\label{sec:intro}

Concept unlearning~\citep{wu2024unlearning,gao2024meta,nguyen2022survey} has emerged as a practical mechanism for reducing the reproduction of protected, unsafe, or otherwise restricted concepts in text-to-image models~\citep{dou2024avoiding,liu2024shield,meeus2024copyright}.
Most copyright-oriented evaluations, however, are built around targets that affect the whole image, such as artistic styles, broad semantic categories, or portrait-like identities~\citep{ren2024six}.
This leaves an important gap for company logos.
Logos are commercially sensitive symbols that commonly appear in generated product mockups, advertisements, storefronts, screenshots, and synthetic news-like imagery, yet they have not been systematically studied as an unlearning target.

The logo setting changes the nature of the problem.
Unlike a style or a dominant subject, a logo can occupy only a few percent of the image while still determining whether the protected concept appears.
At the same time, it must remain visually precise to be recognizable, so small residual strokes, typography, or color patterns can still carry the brand identity.
Logos are also strongly tied to their context: a prompt for a `MacBook' can induce an Apple mark even without explicitly mentioning an Apple logo.
Thus logo unlearning is not simply keyword blocking, nor is it ordinary image-level suppression; it requires localized removal of a semantically entangled visual mark while preserving the surrounding non-target scene.
This motivates our central question: \emph{how reliably do current concept unlearning methods handle logo leakage in modern text-to-image models?}

To answer this question, we introduce \ourBench, a benchmark for logo unlearning on Fortune Global 500 companies.
\ourBench contains nearly 10,000 curated text-query and logo-image pairs.
Candidate prompts are generated and then filtered through human verification so that retained prompts produce valid images with the assigned logos under \ourSD~\citep{sd3}, yielding a logo-generation success rate above 95\%\footnote{We initially constructed 10,000 prompts and filtered out those failing to generate verifiable logos.}.
We focus on globally recognized companies because they are legally salient, visually diverse, and likely to be represented in web-scale pretraining corpora.
\Cref{tab:different_models} further shows why this evaluation is timely: newer open T2I models, including Stable Diffusion variants~\citep{esser2024scaling,Rombach_2022_CVPR} and Flux~\citep{flux}, increasingly render brand marks that earlier models often failed to reproduce.

\begin{figure}[t]
\centering
\includegraphics[width=0.8\linewidth]{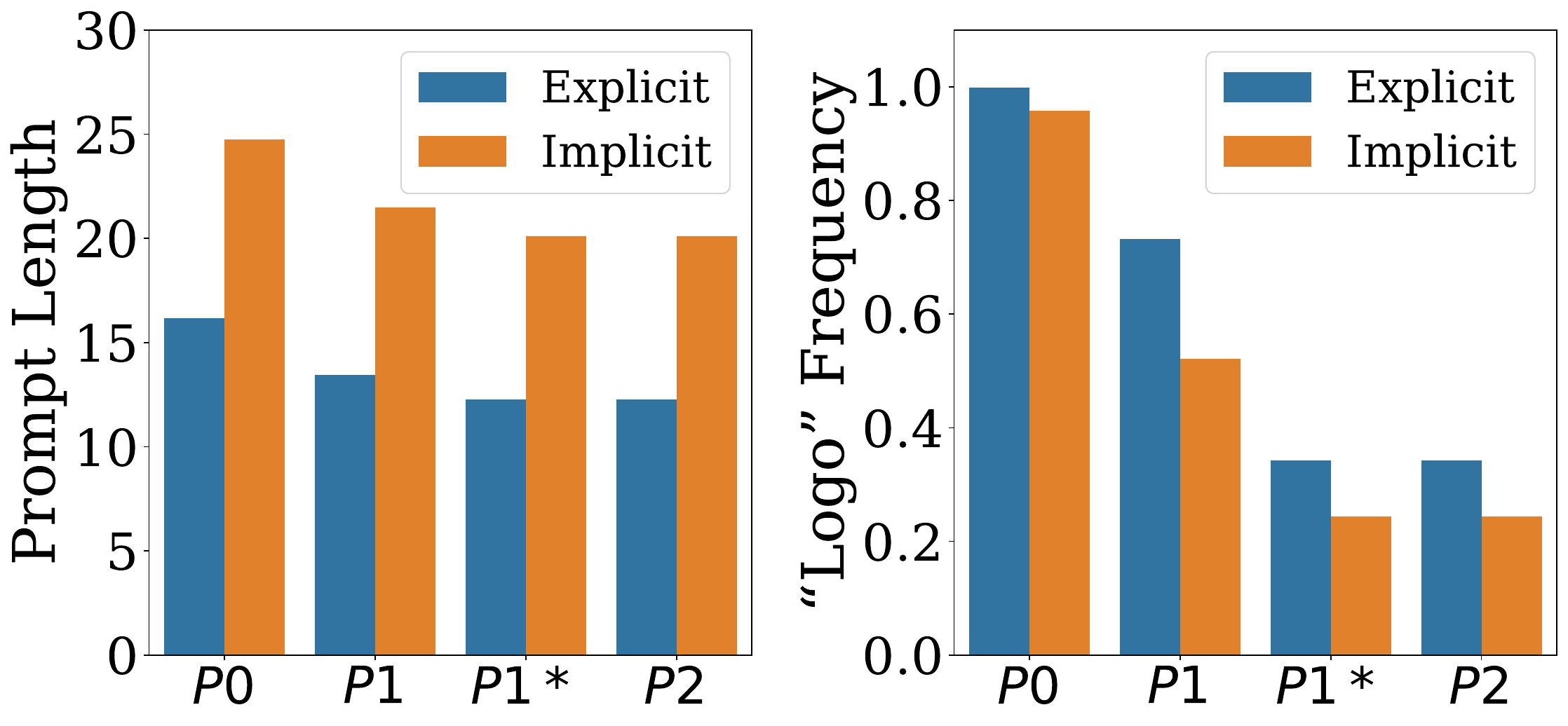}
\caption{\textbf{Prompt distributions across \ourBench tracks and their evolutionary trajectories during semantic unlearning.}
$P0$ is the initial logo-inducing prompt.
$P1$ and $P1^*$ denote intermediate states during iterative concept ablation by the \ourBaseline Reflector agent.
$P2$ is the final sanitized prompt for T2I generation. These statistical profiles highlight the explicit logo references of the Easy track versus the implicit contextual reasoning of the Hard track.
}
\label{fig:dataset}
\end{figure}

\ourBench contains two complementary tracks.
\ourBenchEasy directly names the target logo, testing the basic case where the protected concept is explicit.
\ourBenchHard uses contextual prompts involving products, stores, websites, advertisements, or workplace scenes where the logo may naturally appear without centering the word `logo'.
Prompt statistics are summarized in \Cref{fig:dataset}.
The hard track is not a replacement for a full adaptive attack suite; instead, it tests a realistic intermediate regime where brand cues are conveyed through context rather than direct logo requests.

Our evaluation is designed around the main technical tension in logo unlearning: local erasure versus global preservation.
We primarily test representative inference-time methods on \ourBench and include compatible fine-tuning-based methods on the LUim-500 subset, explicitly scoping conclusions to these evaluated settings because method compatibility varies across backbones.
Rather than using only a binary success rate, we define multi-grained metrics over local and global regions, as well as pixel and latent similarities.
These metrics ask whether a method removes the target mark and whether it preserves the rest of the generated image.

The resulting experiments show that the evaluated inference-time methods, including NP~\citep{np}, SLD~\citep{schramowski2023safe}, and SEGA~\citep{brack2023sega}, do not yet provide reliable logo removal without visible side effects.
Compatible fine-tuning-based methods such as ESD~\citep{gandikota2023erasing} and Forget-Me-Not~\citep{zhang2024forget} show related limitations under their native legacy backbones.
We further introduce \ourBaseline, an exploratory multi-agent prompt-rewriting baseline, to separate prompt-level mitigation from model-level unlearning.
\ourBaseline often improves local erasure by removing logo-inducing semantics, but it also changes the input description and can alter the generated scene, clarifying why prompt filtering alone should not be treated as weight-level disentanglement.
Finally, we analyze correlations between unlearning behavior and image attributes such as logo area, location, edge density, shape count, texture complexity, and fractal dimension.
The results point toward spatially controlled mechanisms, including SSIM~\citep{wang2004image}-guided constraints, as a promising direction for future logo unlearning.
An overview is shown in \Cref{fig:teaser}.

\noindent\textbf{Our contributions are summarized as follows:}
\begin{itemize}
    \item We introduce \ourBench, a benchmark of nearly 10,000 curated text-image pairs for evaluating logo unlearning on Fortune Global 500 companies across explicit and implicit prompt tracks.
    \item We define a multi-grained evaluation protocol that jointly measures local logo erasure and global image preservation, making the core trade-off measurable rather than hidden inside a single success rate.
    \item We benchmark representative inference-time and compatible fine-tuning-based unlearning methods, showing that the evaluated methods struggle to remove localized brand marks without changing non-target content.
    \item We provide \ourBaseline as a diagnostic prompt-space baseline and analyze why semantic rewriting helps in some cases but does not solve localized model-level unlearning.
\end{itemize}

\section{Related Work}
\label{sec:related}

\subsection{Logo Benchmark}

Logo datasets have traditionally supported recognition, retrieval, and detection.
Classification resources such as Logo-2k+~\citep{wang2020logo} and Weblogo-2m~\citep{su2017weblogo} measure brand categorization at scale, while retrieval datasets~\citep{joly2009logo,romberg2011scalable,bhunia2019deep} evaluate the ability to search for logo instances in large image collections.
Detection datasets~\citep{jin2020open,zhang2021multi,hou2021foodlogodet,li2022seetek,hou2023deep}, including LogoDet-3K~\citep{wang2022logodet} and QMUL-OpenLogo~\citep{su2018open}, provide bounding boxes for localizing logos in varied scenes.
In-the-wild datasets~\citep{hoi2015logo,tuzko2017open,yang2023comparative} further stress robustness to scale, viewpoint, lighting, and occlusion.
These benchmarks are valuable, but their goal is to find or classify logos after they appear.
\ourBench asks a different question for generative models: can a method prevent or remove a localized brand mark while preserving the rest of the generated scene?
This turns logos from recognition targets into unlearning targets.

\subsection{Concept Unlearn} 
Concept unlearning for generative models~\citep{li2026aligned,jiang2026davinci,gao2024kan} has been studied for harmful content~\citep{kim2023towards, schramowski2023safe, brack2023sega}, nudity~\citep{gandikota2023erasing,lyu2024one,kim2023towards,gandikota2024unified,lu2024mace,xiong2024editing,schramowski2023safe,brack2023sega}, celebrity likeness~\citep{zhang2024forget,lu2024mace,ma2024dataset}, copyright-related concepts~\citep{lyu2024one, ma2024dataset}, and artistic styles~\citep{kim2023towards,gandikota2023erasing,lyu2024one,zhang2024forget,gandikota2024unified,lu2024mace,xiong2024editing,brack2023sega,zhang2024unlearncanvas,ma2024dataset}.
Logos share the broader copyright motivation of this literature, but they stress a different axis of unlearning.
The target is usually a small region rather than a global visual distribution, and the surrounding scene is often legitimate content that should remain intact.
This creates a narrow operating window: weak interventions leave a recognizable brand trace, while strong interventions can remove useful image content.

Existing methods are often grouped into fine-tuning-based and inference-time approaches~\citep{ren2024six}.
Fine-tuning methods modify model weights and can target internal representations, but they require compatible implementations and additional compute~\citep{gandikota2023erasing, kumari2023ablating,kim2023towards,lyu2024one,gandikota2024unified,lu2024mace,xiong2024editing}.
Inference-time methods avoid retraining and are appealing for large modern generators~\citep{ren2024six,schramowski2023safe,brack2023sega,np}, yet their guidance signals are not naturally localized to small brand marks.
Rather than positioning \ourBench as a proof that every unlearning method fails, we use it as a diagnostic benchmark for representative methods and for future localized approaches.
Motivated by recent work on agentic evaluation and multi-agent systems~\citep{xiao2025limi,li2025datasetresearch,li2026agencybench,wu2025innovatorbench}, we also include \ourBaseline, a prompt-space multi-agent baseline, to make clear what can be gained through semantic rewriting and what remains a weight-level disentanglement problem.

\begin{figure}[t]
\centering
\renewcommand{\arraystretch}{1.2} 
\setlength{\tabcolsep}{0pt} 
\begin{tabular}{cccc}

\multicolumn{1}{c}{\textbf{SD1.5}} &
\multicolumn{1}{c}{\textbf{SD3}} &
\multicolumn{1}{c}{\textbf{SD3.5}} &
\multicolumn{1}{c}{\textbf{FLUX}} \\

\includegraphics[width=0.18\linewidth]{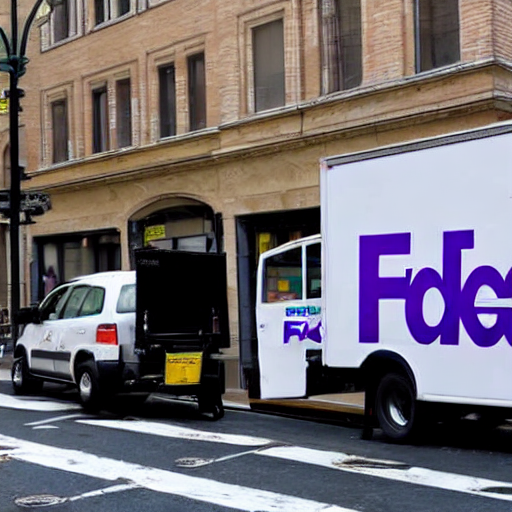} &
\includegraphics[width=0.18\linewidth]{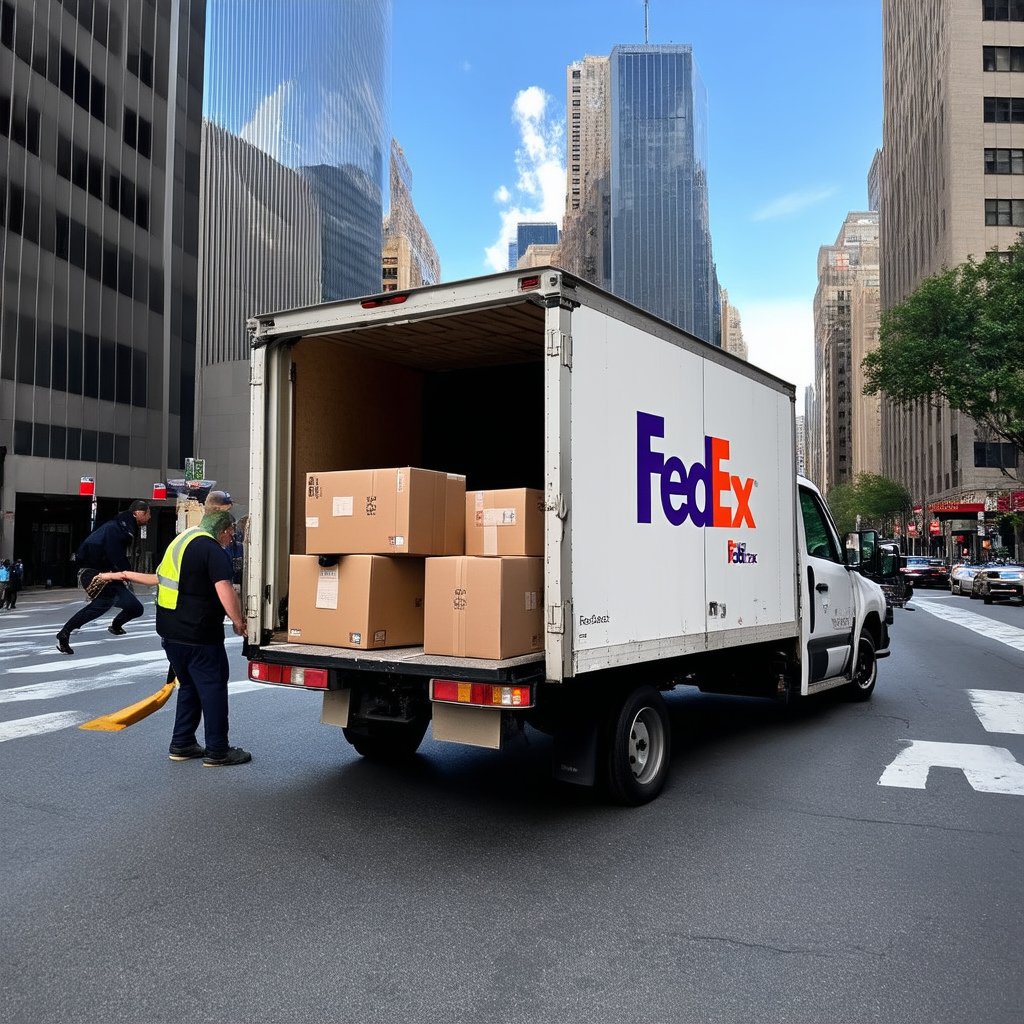} &
\includegraphics[width=0.18\linewidth]{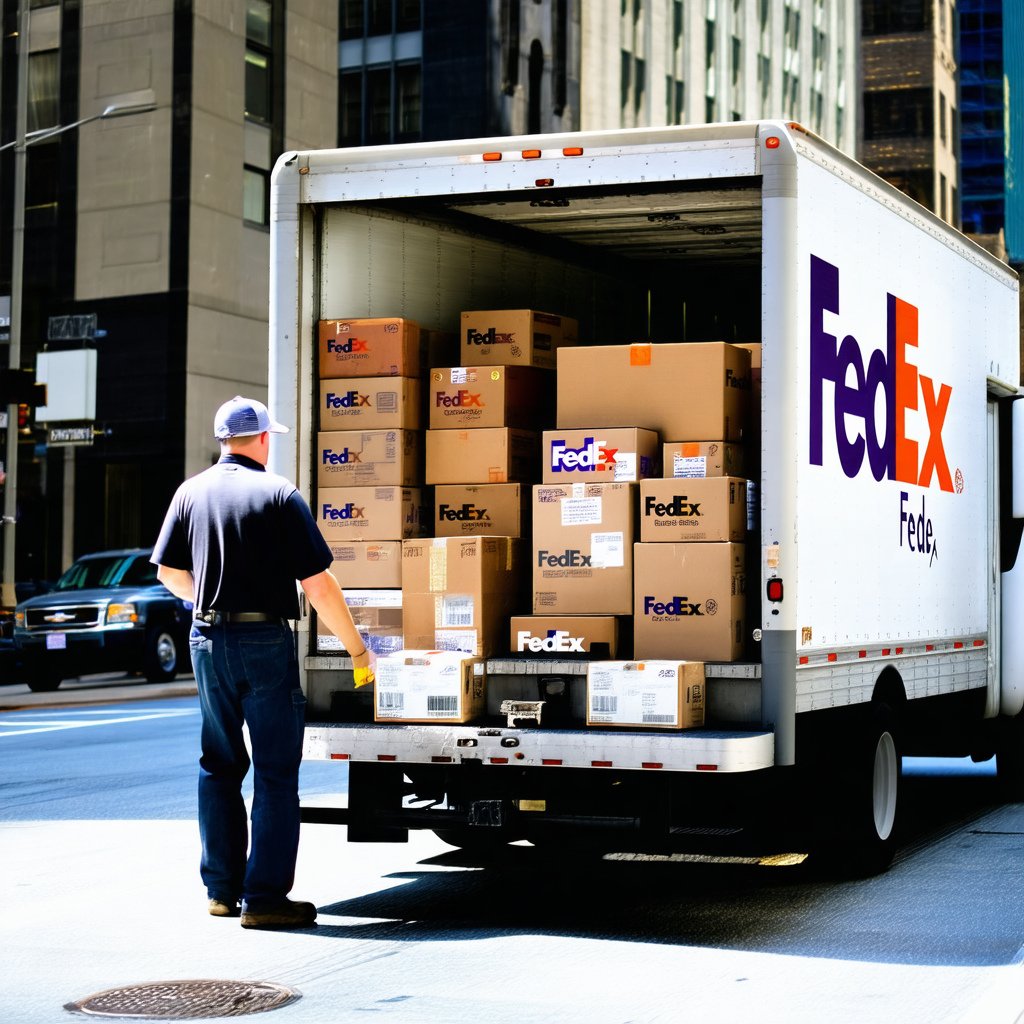} &
\includegraphics[width=0.18\linewidth]{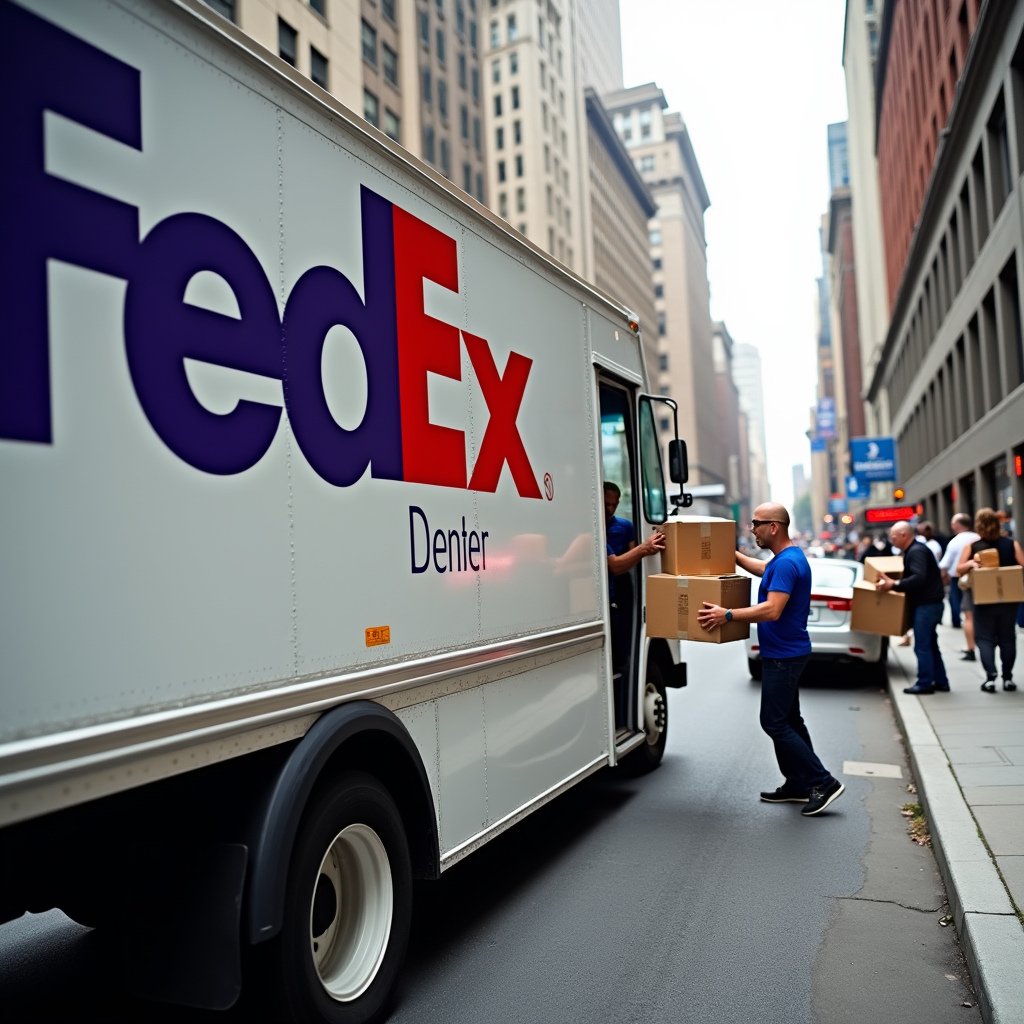}

\end{tabular}
\caption{\textbf{Escalating high-fidelity logo leakage in modern T2I architectures.} 
While legacy models (SD1.5) struggle with complex brand iconography, contemporary models (SD3, SD3.5, FLUX) precisely hallucinate the copyrighted ``FedEx'' logo. This capability jump necessitates \ourBench and justifies selecting SD3 to evaluate concept unlearning.}
\label{tab:different_models}
\end{figure}

\section{\ourBench}
\label{sec:method}
We present \ourBench as a benchmark centered on the two requirements that make logo unlearning difficult: suppressing a localized brand mark and preserving the surrounding scene.
\Cref{subsec:benchmark} describes the data construction and validation procedure.
\Cref{subsec:baseline} introduces \ourBaseline, an exploratory prompt-space baseline used to probe semantic mitigation.
\Cref{subsec:method_evaluation} defines the evaluation protocol.

\subsection{Benchmark}
\label{subsec:benchmark}

The benchmark is built around recognizable corporate identities from the 2024 Fortune Global 500 list.
This choice gives broad coverage across industries and countries while keeping the target concepts sufficiently familiar for modern T2I models to reproduce.
For each company, \ourBench provides two prompt tracks.
\ourBenchEasy contains explicit prompts that directly ask for the target logo, testing the most direct logo-suppression setting.
\ourBenchHard contains contextual prompts built around products, stores, websites, employees, advertisements, and packaging, where the target logo may naturally appear without a simple logo keyword.
Together, the two tracks separate direct concept invocation from contextual brand leakage.

The dataset is constructed through a human-AI pipeline.
We first generate candidate prompts for each company and track with a large language model.
Human reviewers then filter prompts for naturalness, target-company alignment, and validity of the implicit cue in \ourBenchHard.
We also remove prompts whose generated images under the reference model contain ambiguous, unverifiable, or unrelated logos.
After filtering, \ourBench contains \numBench validated prompts, including \numBenchEasy prompts in \ourBenchEasy and \numBenchHard prompts in \ourBenchHard.
The purpose of this validation is not to make prompt generation itself difficult; it is to ensure that each retained sample tests the intended failure mode: a logo appears before unlearning and should be removed without unnecessary scene changes.

\begin{figure}[t]
\centering
\includegraphics[width=1.0\linewidth]{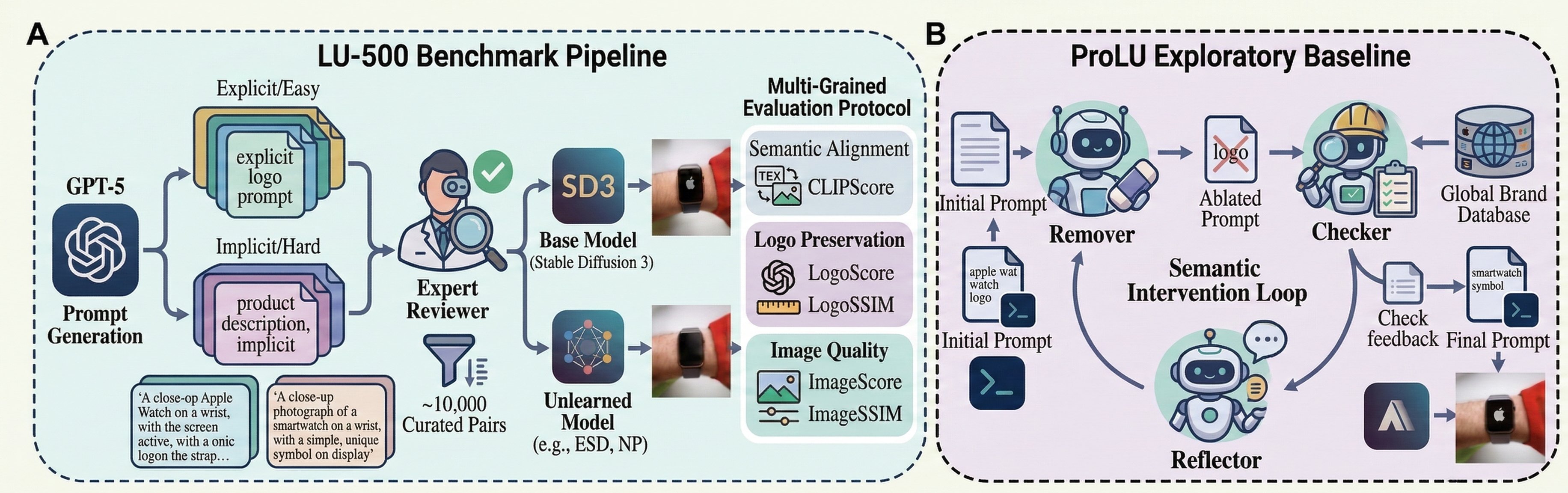}
\caption{\textbf{Overview of the \ourBench pipeline and \ourBaseline exploratory framework.} 
\textbf{Left:} The human-AI collaborative generation of \ourBench. Using Apple as an example, it details explicit and implicit prompt synthesis via GPT-5, multi-stage human validation, and the multi-grained evaluation protocol. 
\textbf{Right:} The semantic intervention logic of \ourBaseline. It orchestrates the \textit{Remover}, \textit{Reflector}, and \textit{Checker} agents to iteratively ablate latent brand triggers in the prompt space while preserving the structural context.}
\label{fig:dataset_generation+ours}
\end{figure}

We use \ourSD~\citep{esser2024scaling,sd3} as the primary T2I backbone for generation and evaluation.
The choice is practical and scoped.
Earlier Stable Diffusion models often do not render complex logos reliably before intervention, which makes unlearning measurement unstable.
Closed-source systems such as Midjourney~\citep{midjourney} and DALLE3~\citep{dalle3} do not provide the access needed for many unlearning interventions.
\ourSD offers an open model with sufficient logo-generation ability and feasible experimental cost, as summarized in \Cref{tab:different_models}.
\Cref{fig:dataset_generation+ours} illustrates the construction and evaluation pipeline.

\subsection{Exploratory Baseline: \ourBaseline}
\label{subsec:baseline}

\ourBaseline is designed as a diagnostic prompt-space baseline rather than a new model-unlearning algorithm.
It tests how much of the benchmark can be addressed by rewriting the user prompt before generation, a mechanism that is easy to deploy but does not modify model weights.
This distinction is important: if prompt rewriting performs well, it indicates that part of the failure is semantic and input-facing; if it changes the scene, it reveals the limits of prompt-space control.

\ourBaseline uses three LLM-based agents: a Remover, a Reflector, and a Checker.
Given an initial prompt $P0$, the Remover removes explicit or implicit logo-inducing cues while trying to preserve the main visual intent, producing $P1$.
The Reflector compares $P1$ with the original prompt and revises the output when the edit either leaves residual brand cues or removes too much non-target context.
The Checker performs a final textual pass for obvious logo references and returns the prompt for revision if needed.
The final prompt $P2$ is then used for generation.
As illustrated in \Cref{fig:dataset_generation+ours}, \ourBaseline serves as a semantic reference point for interpreting the harder weight-level unlearning results.

\begin{figure*}[!t]
\centering
\renewcommand{\arraystretch}{1.0} 
\setlength{\tabcolsep}{0pt} 
\begin{tabular}{ccccccccc}

    \multicolumn{1}{c}{\tiny \textbf{Before}} &
        \multicolumn{1}{c}{\tiny \textbf{NP}} &
        \multicolumn{1}{c}{\tiny \textbf{\sldzero}} &
        \multicolumn{1}{c}{\tiny \textbf{\sldhalf}} &
        \multicolumn{1}{c}{\tiny \textbf{\sldone}} &
        \multicolumn{1}{c}{\tiny \textbf{\segazero}} &
        \multicolumn{1}{c}{\tiny \textbf{\segahalf}} &
        \multicolumn{1}{c}{\tiny \textbf{\segaone}} &
        \multicolumn{1}{c}{\tiny \textbf{\ourBaseline}} \\

\includegraphics[width=0.111\textwidth]{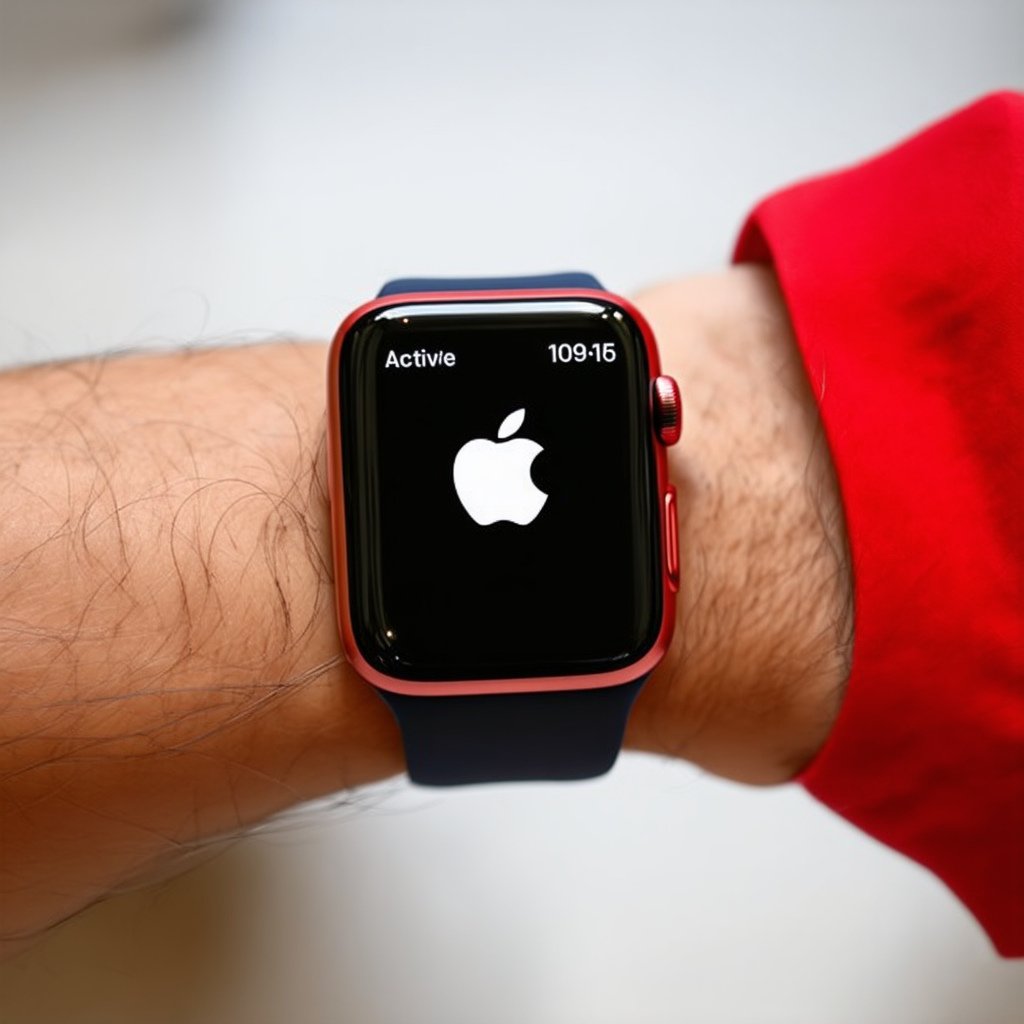} &
\includegraphics[width=0.111\textwidth]{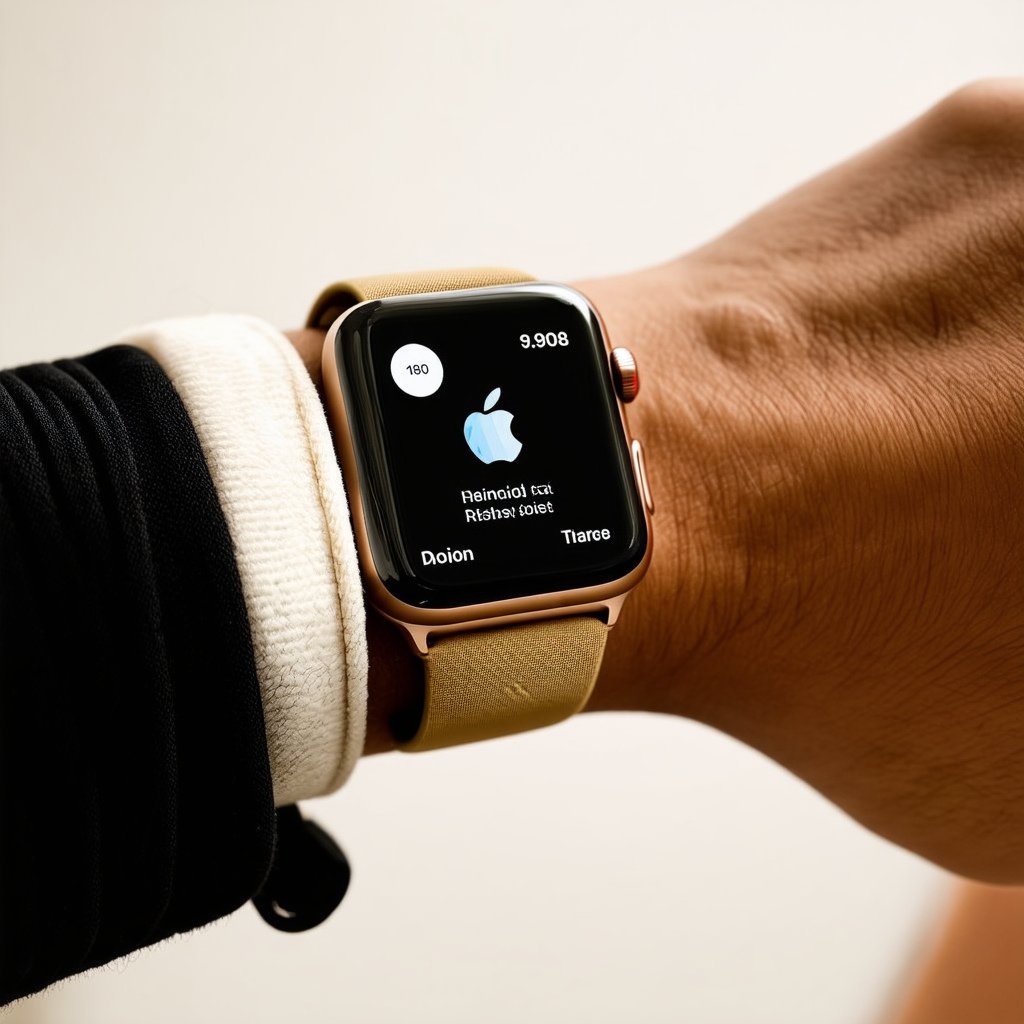} &
\includegraphics[width=0.111\textwidth]{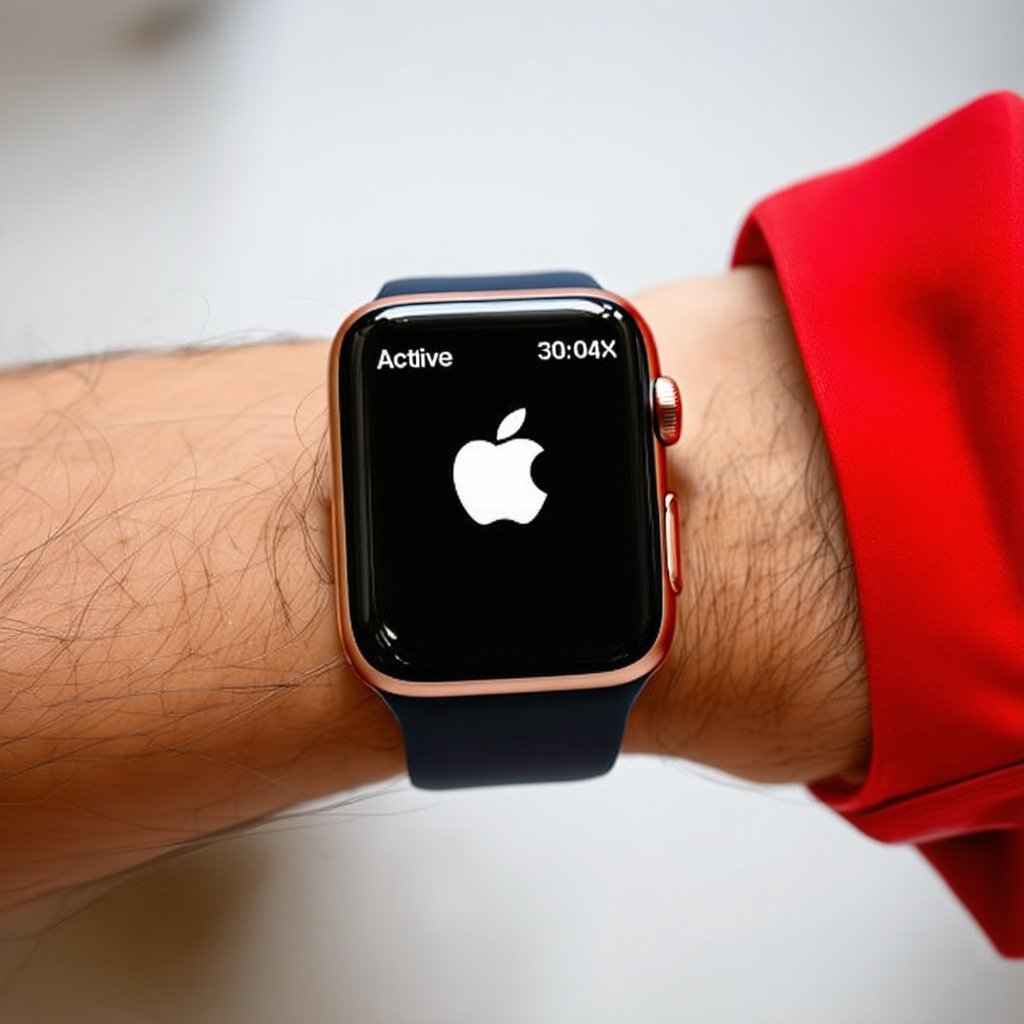} &
\includegraphics[width=0.111\textwidth]{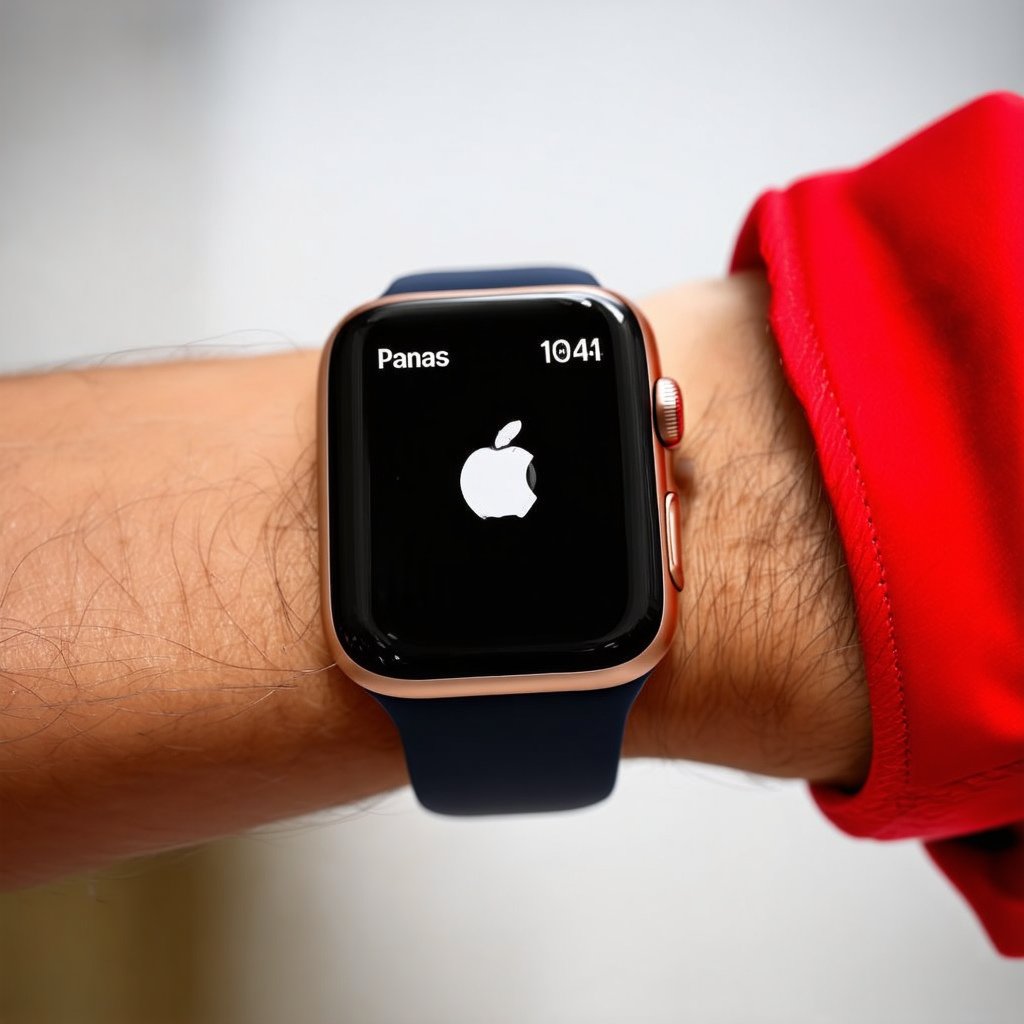} &
\includegraphics[width=0.111\textwidth]{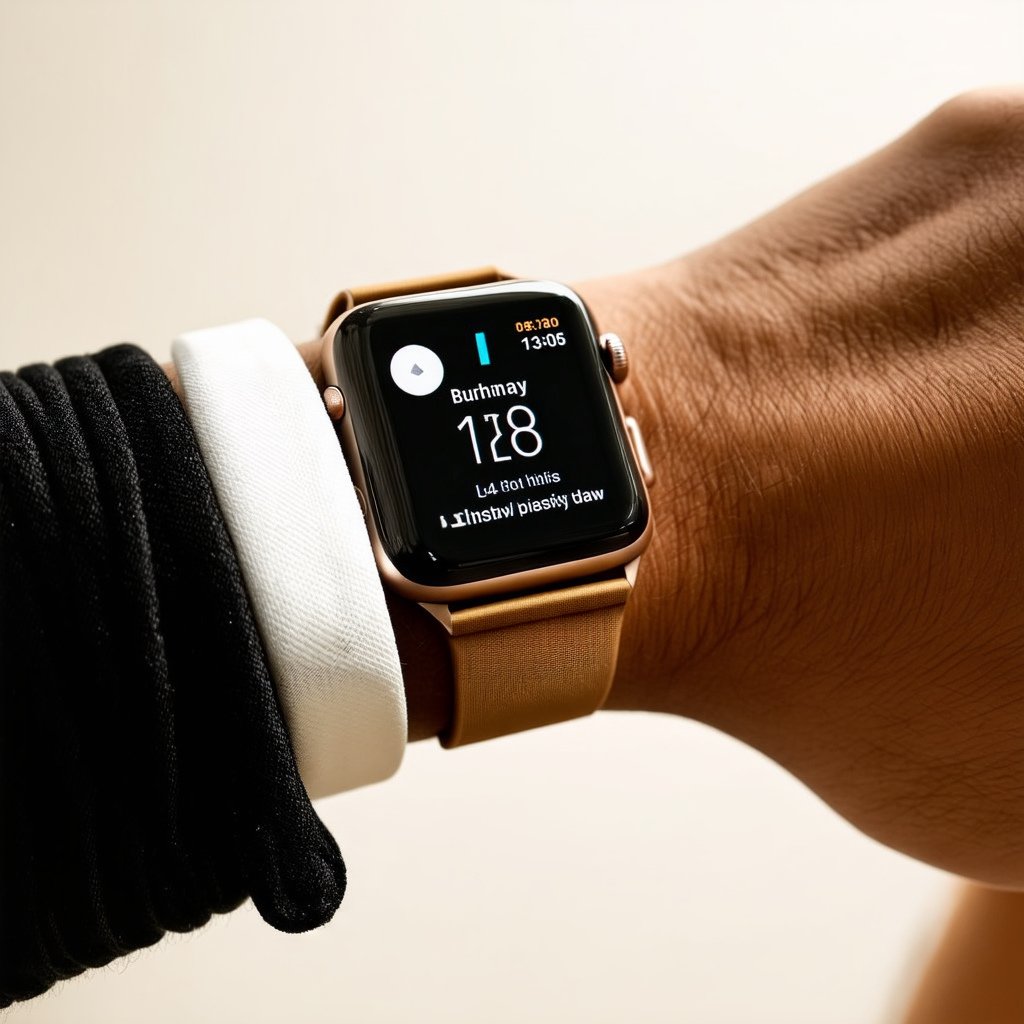} &
\includegraphics[width=0.111\textwidth]{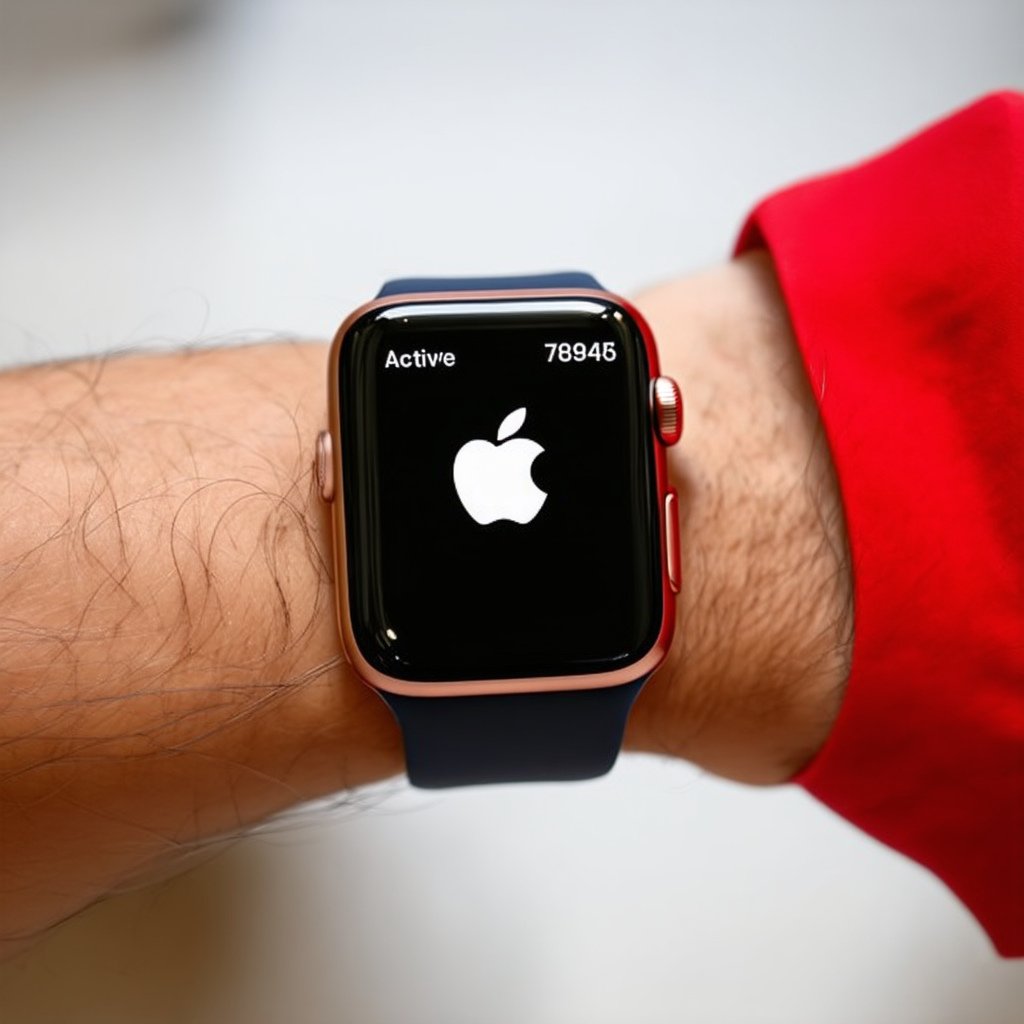} &
\includegraphics[width=0.111\textwidth]{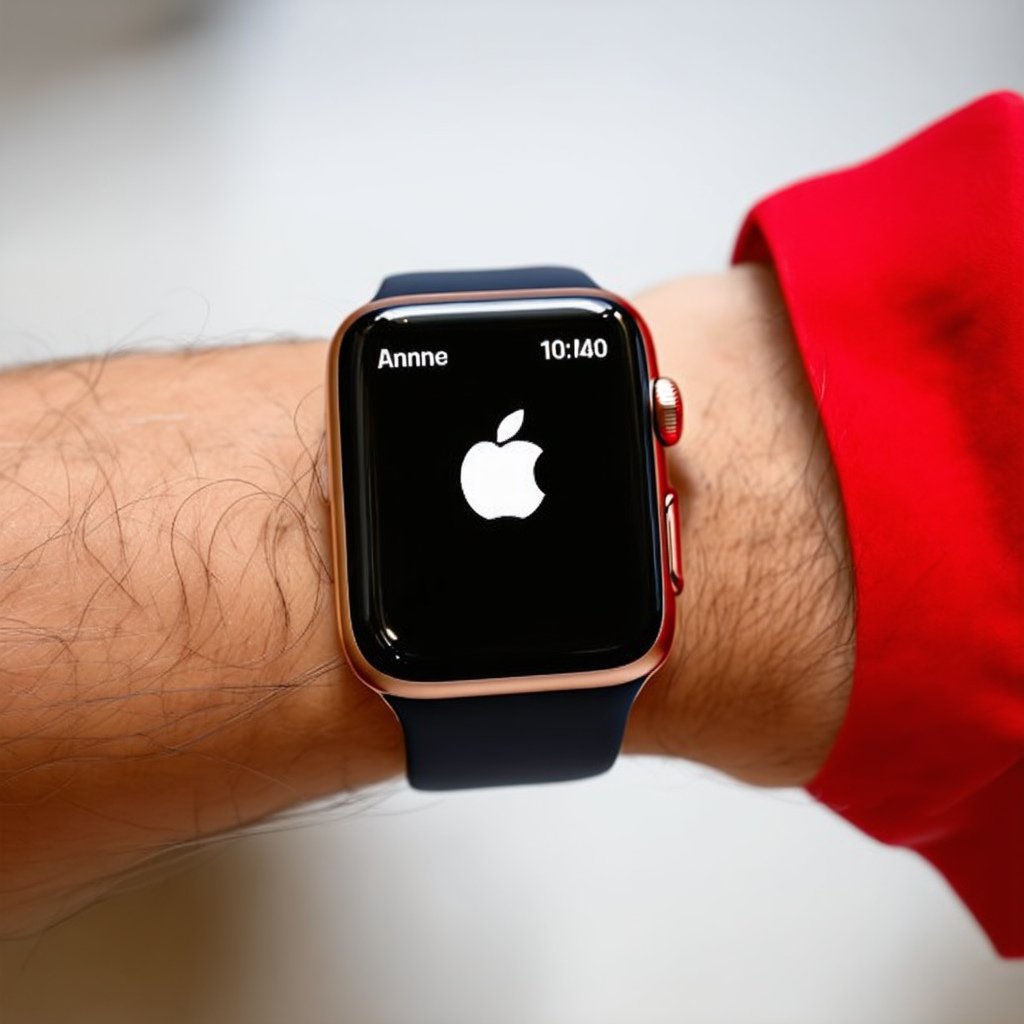} &
\includegraphics[width=0.111\textwidth]{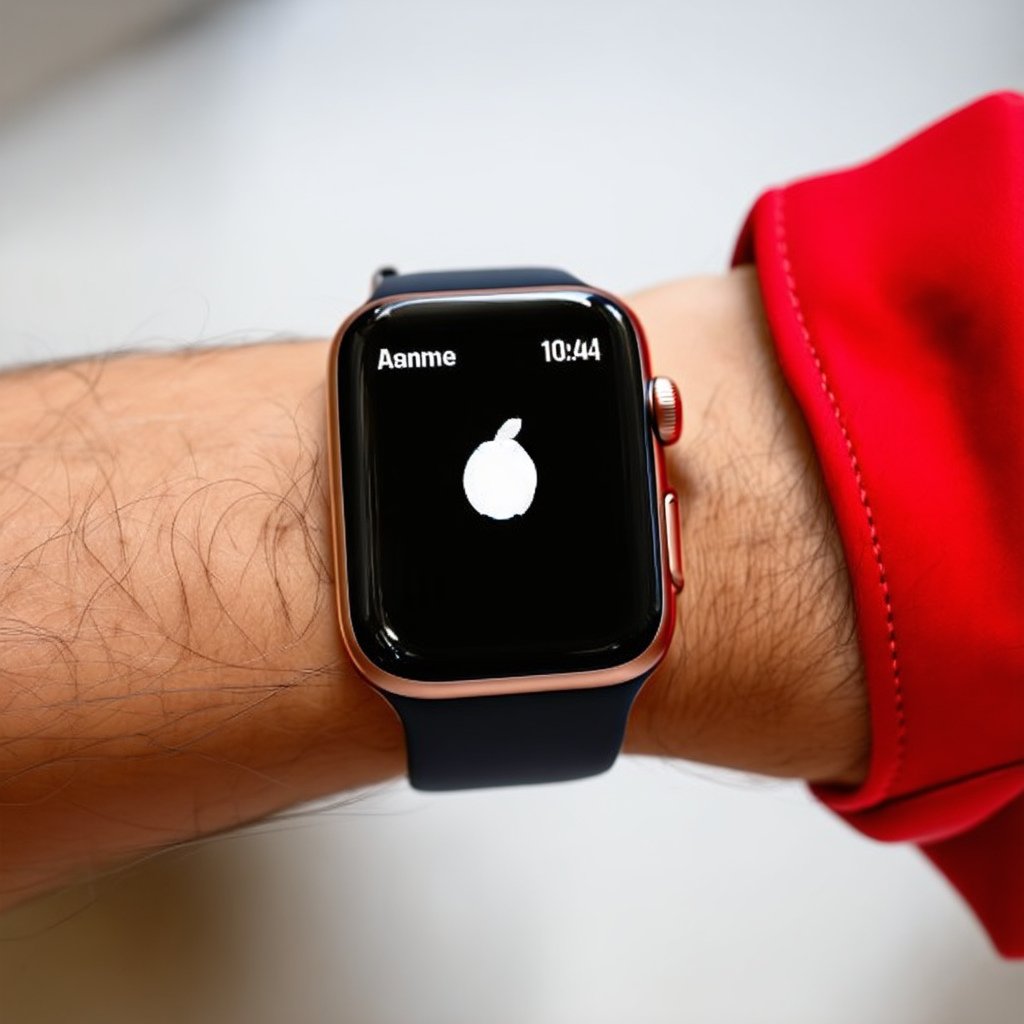} &
\includegraphics[width=0.111\textwidth]{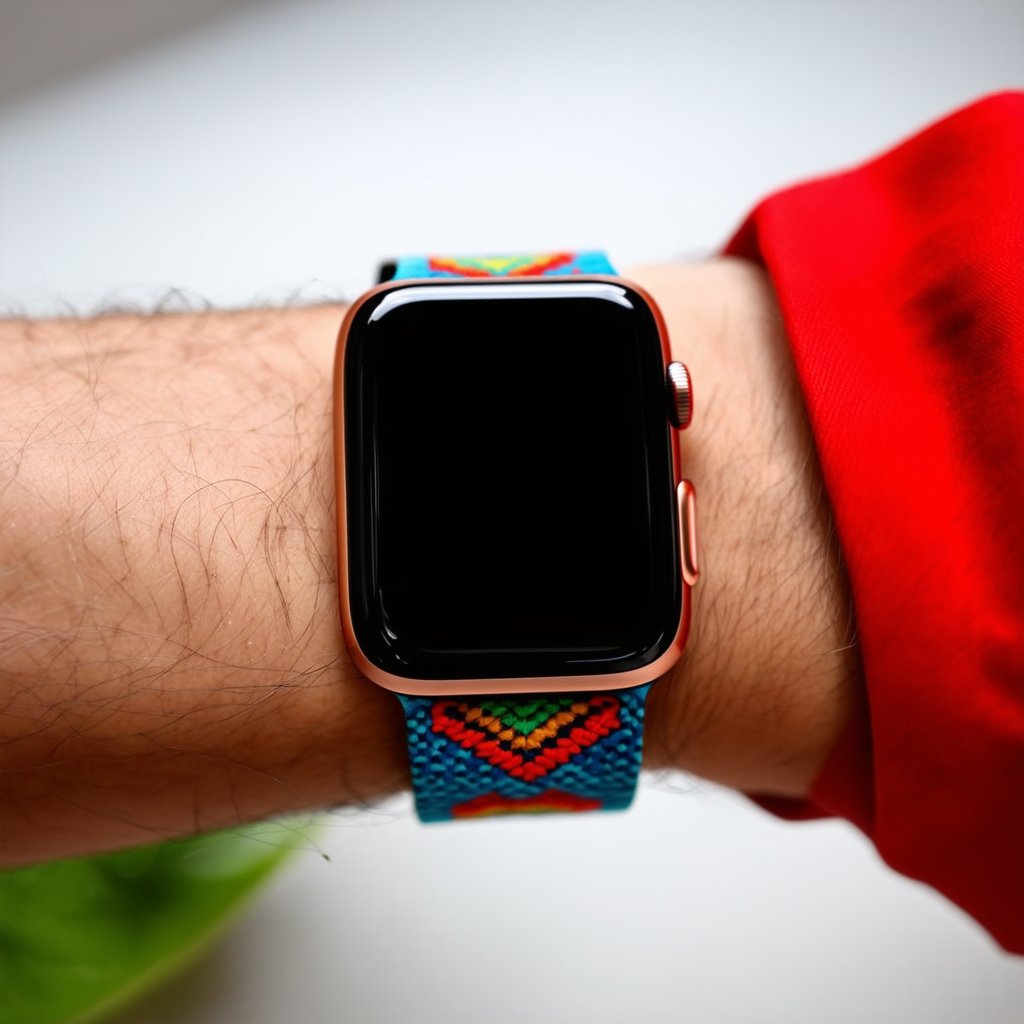} \\

\includegraphics[width=0.111\textwidth]{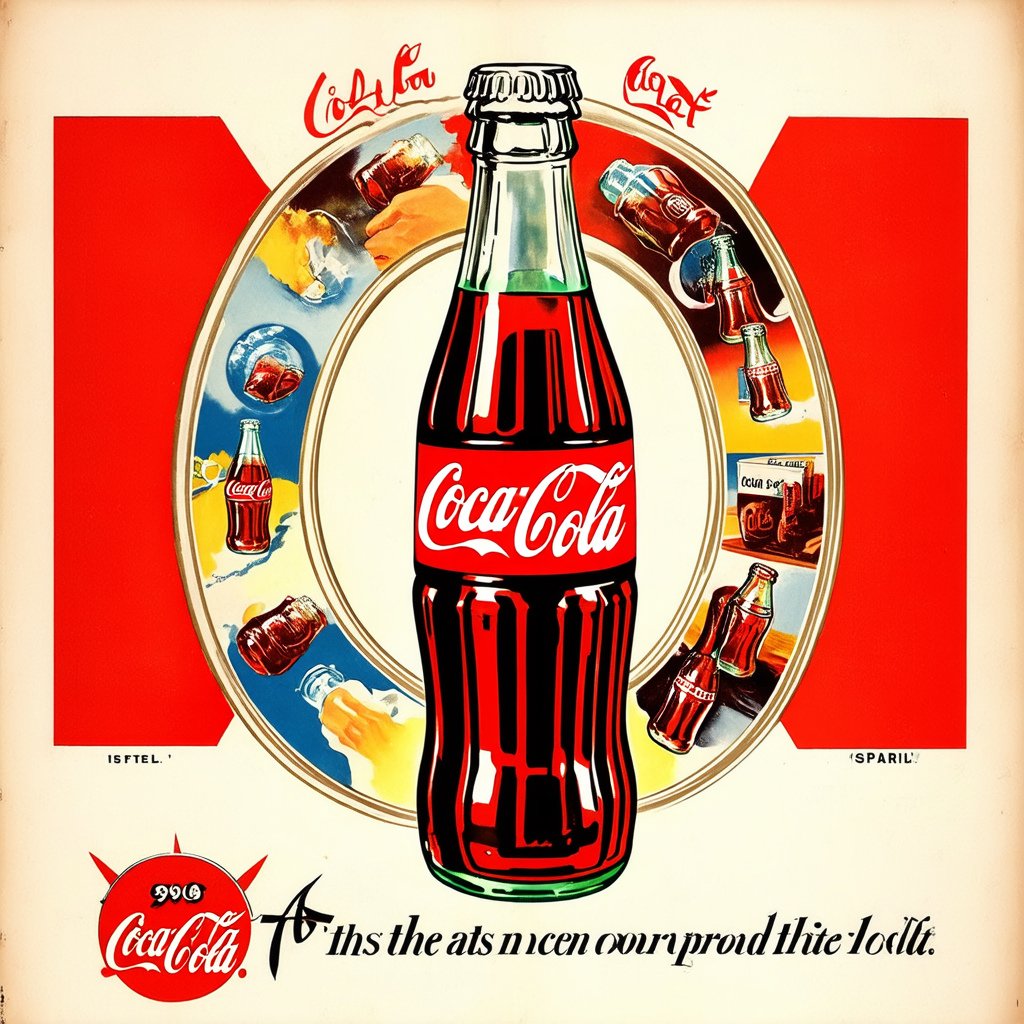} &
\includegraphics[width=0.111\textwidth]{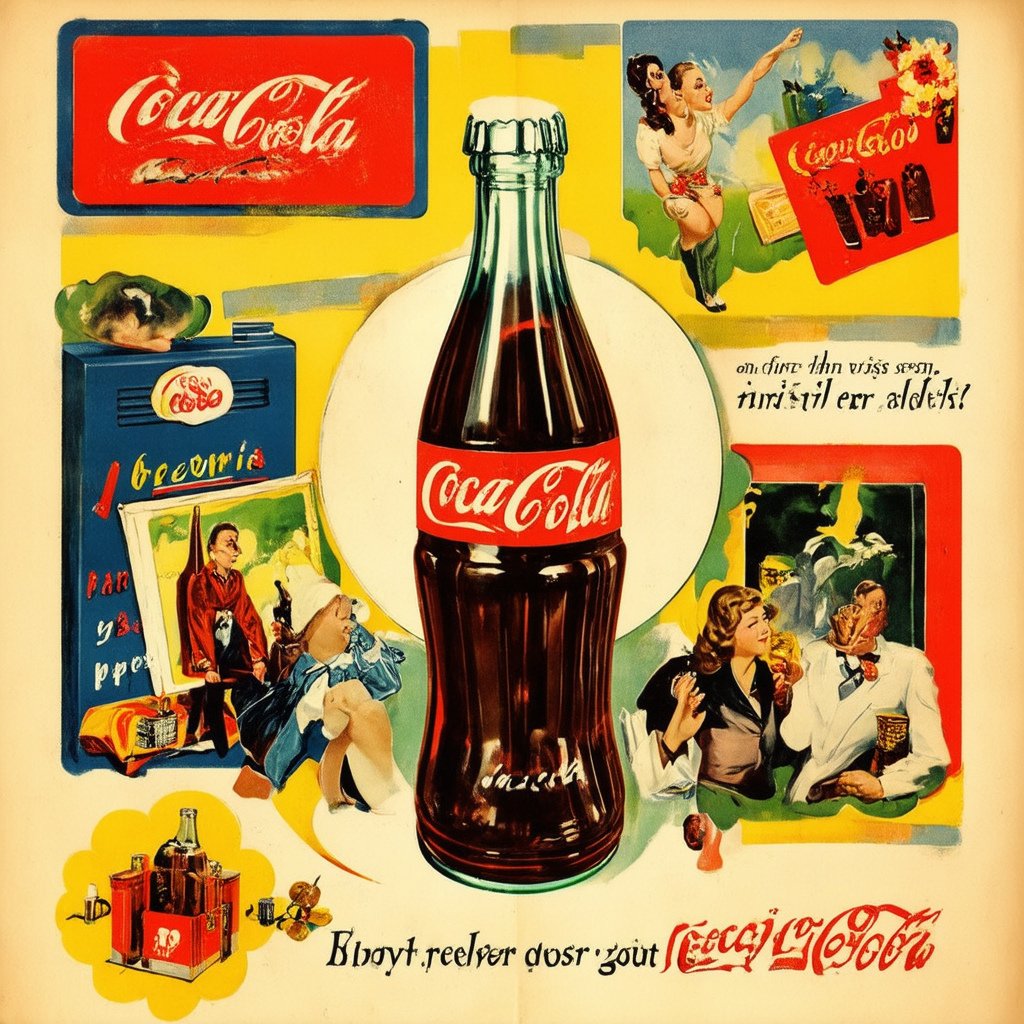} &
\includegraphics[width=0.111\textwidth]{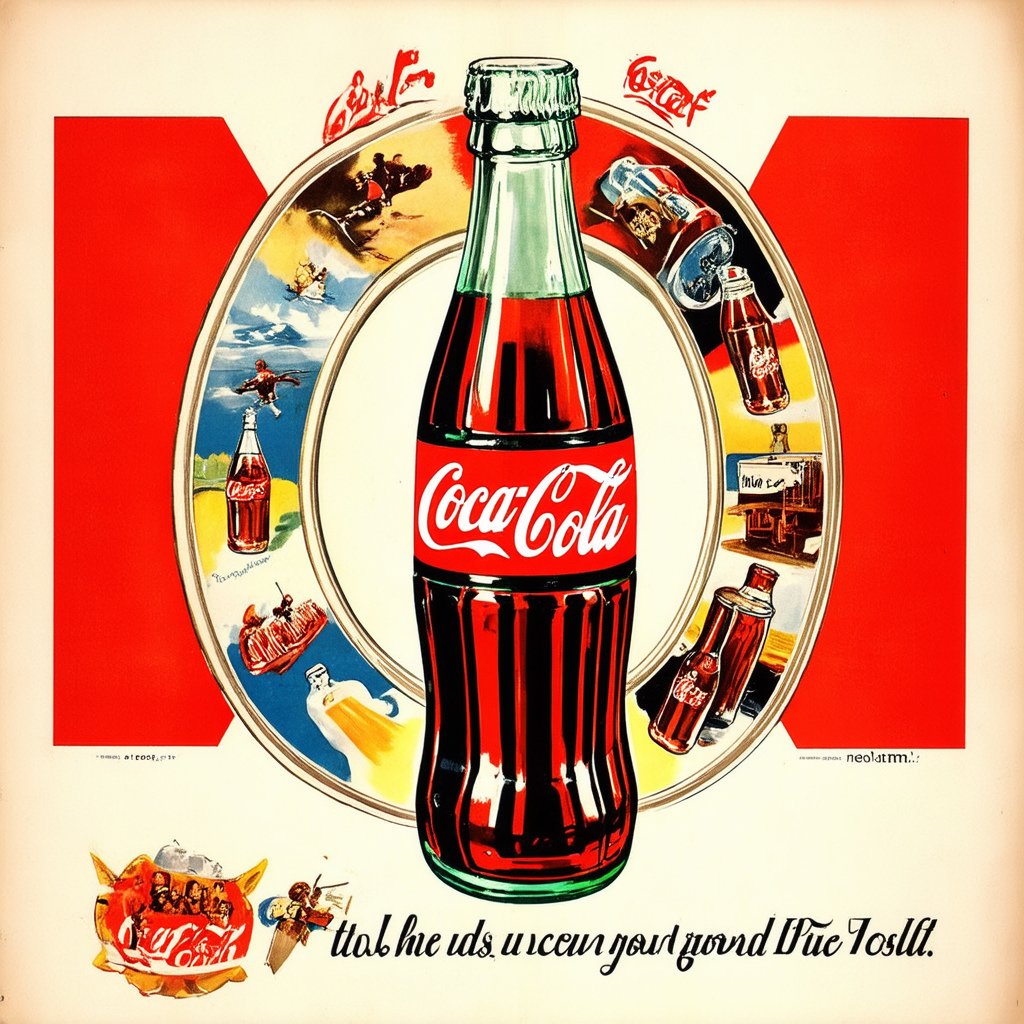} &
\includegraphics[width=0.111\textwidth]{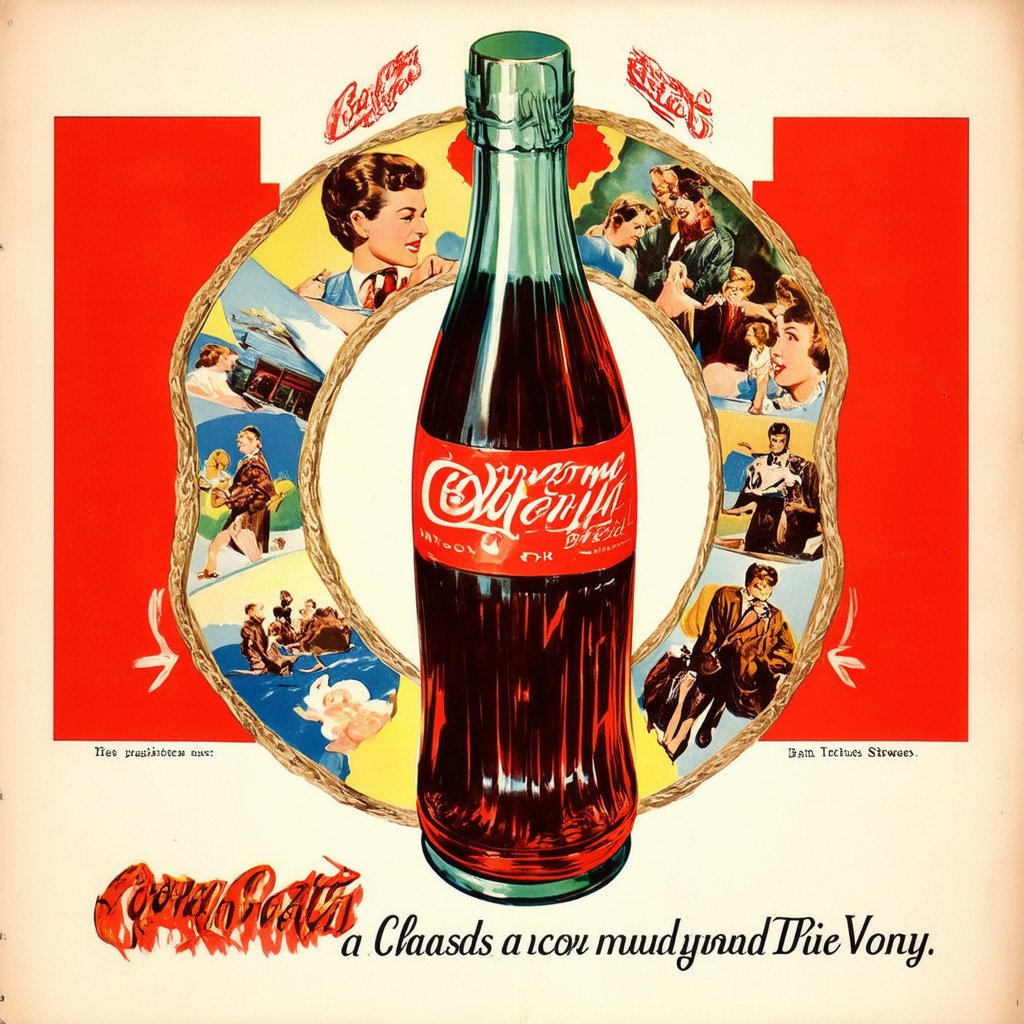} &
\includegraphics[width=0.111\textwidth]{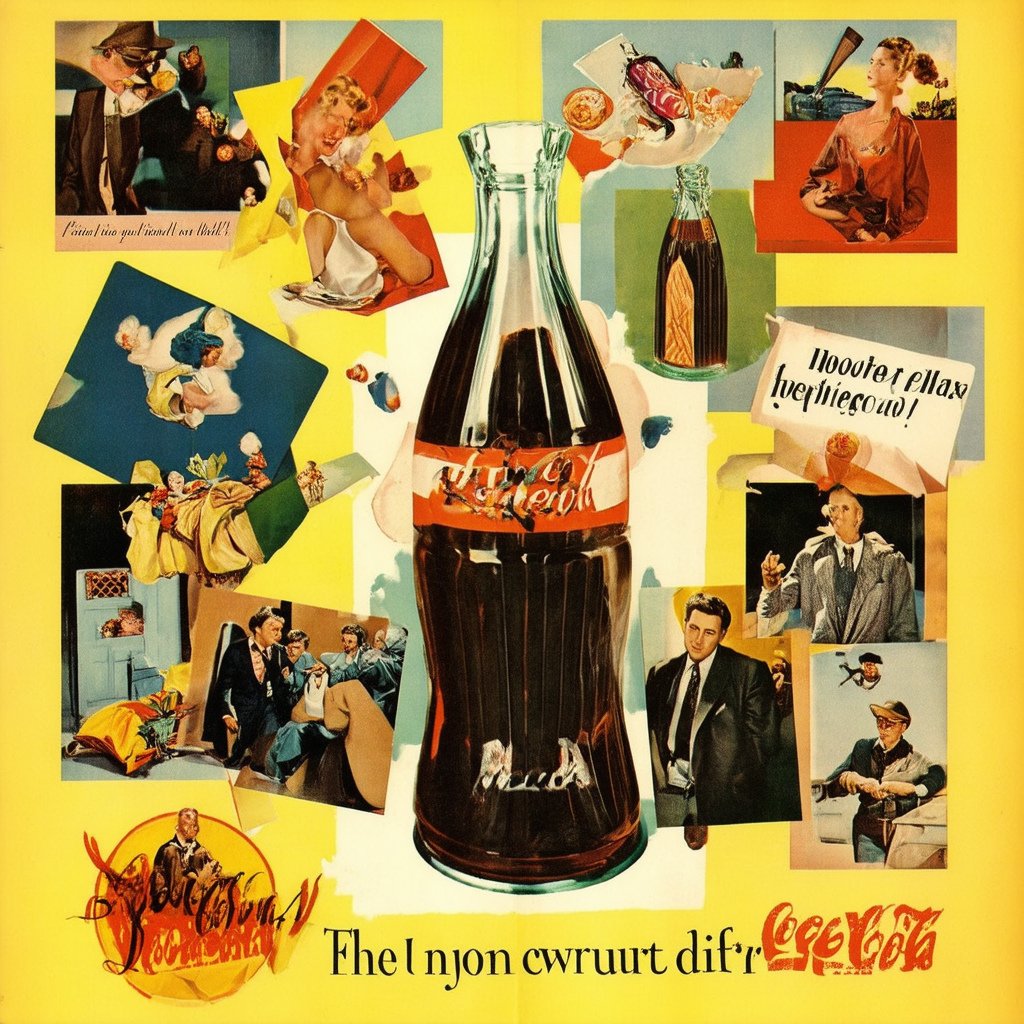} &
\includegraphics[width=0.111\textwidth]{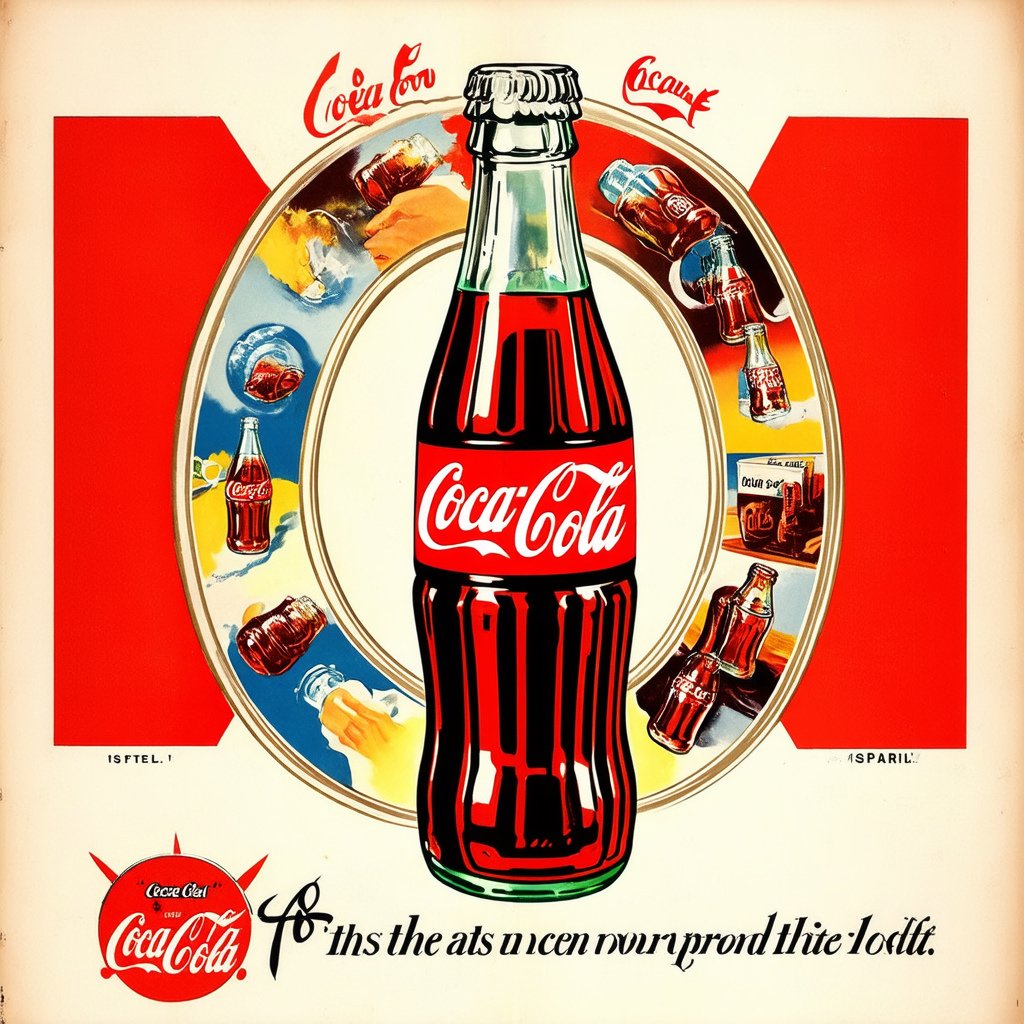} &
\includegraphics[width=0.111\textwidth]{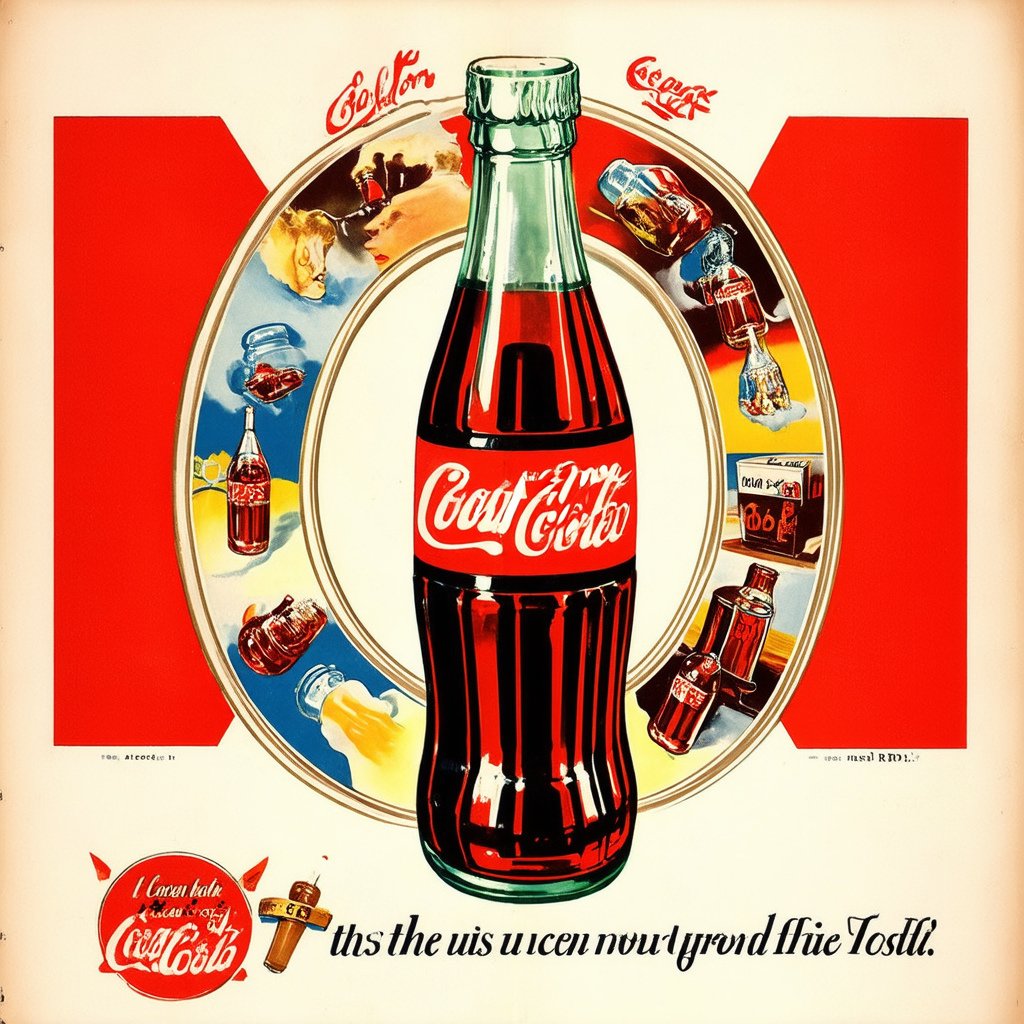} &
\includegraphics[width=0.111\textwidth]{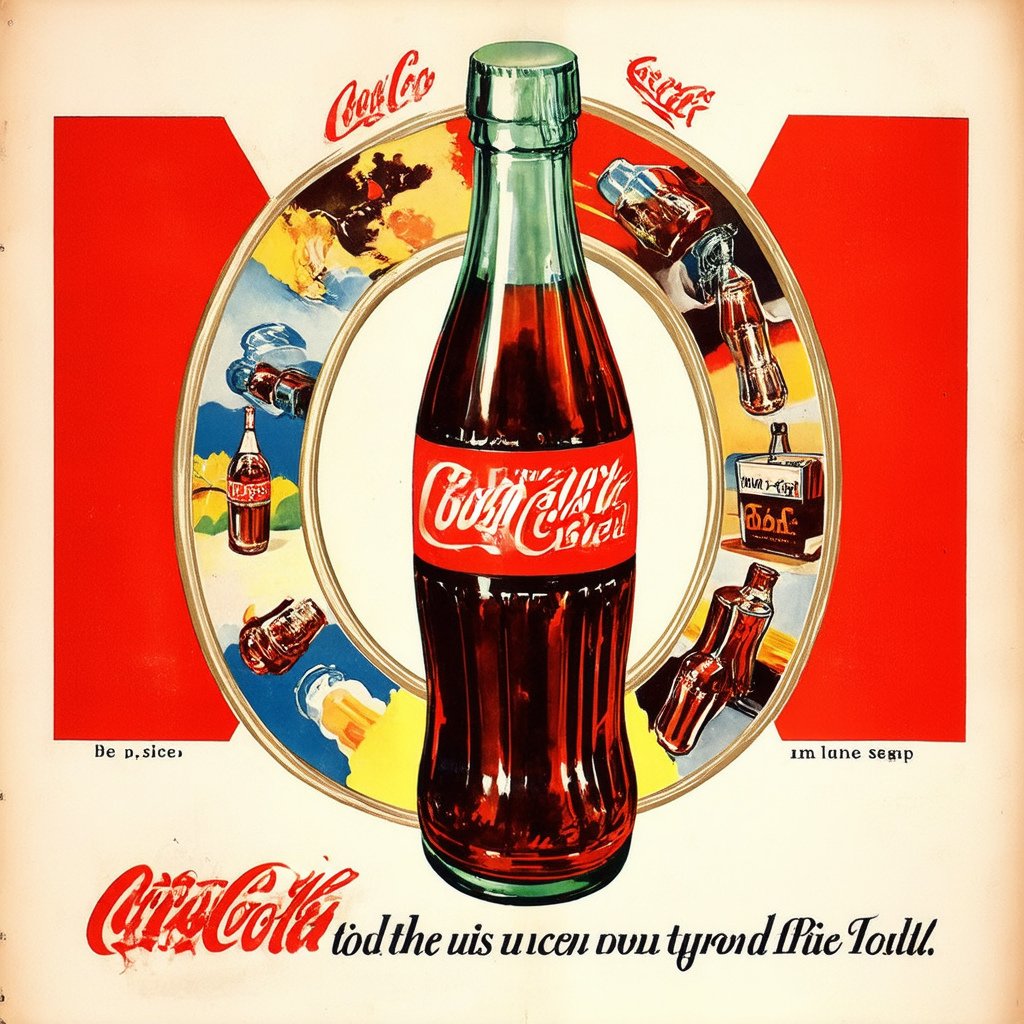} &
\includegraphics[width=0.111\textwidth]{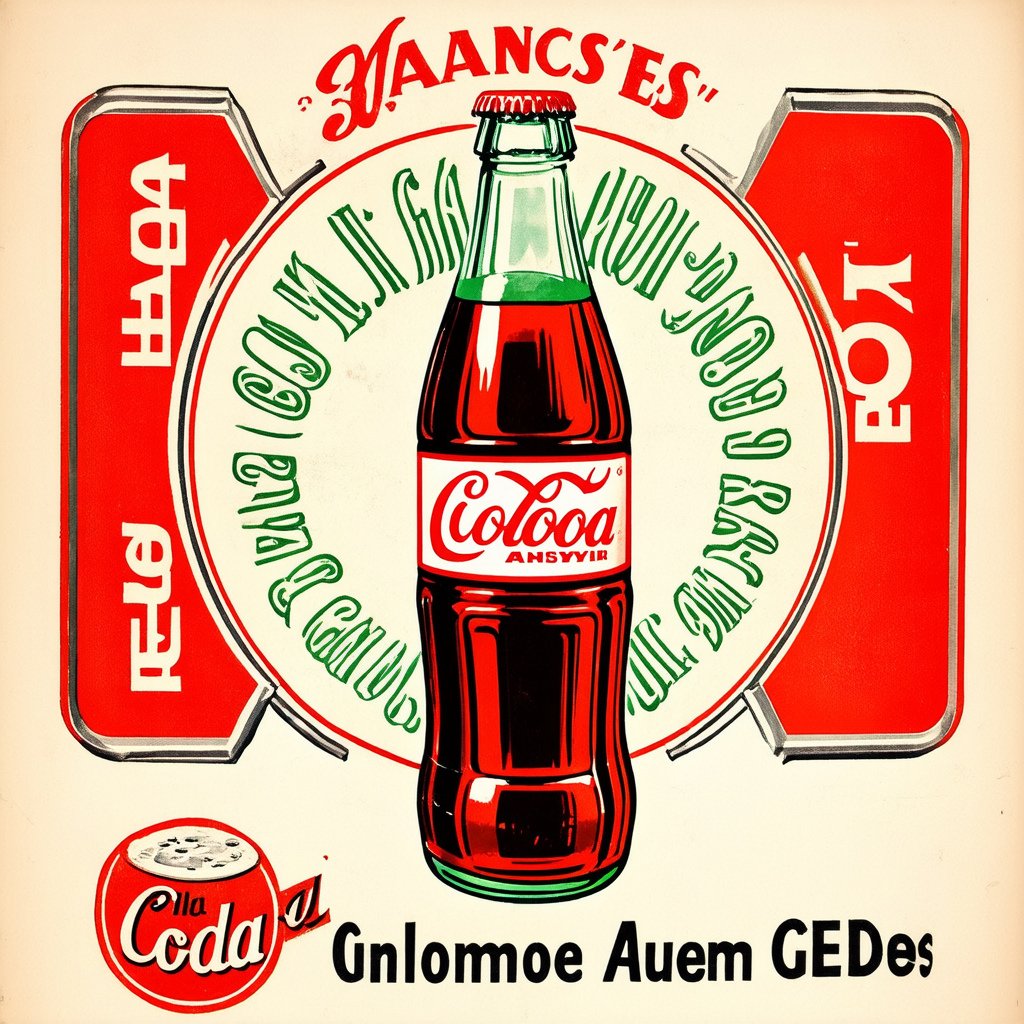} \\

\includegraphics[width=0.111\textwidth]{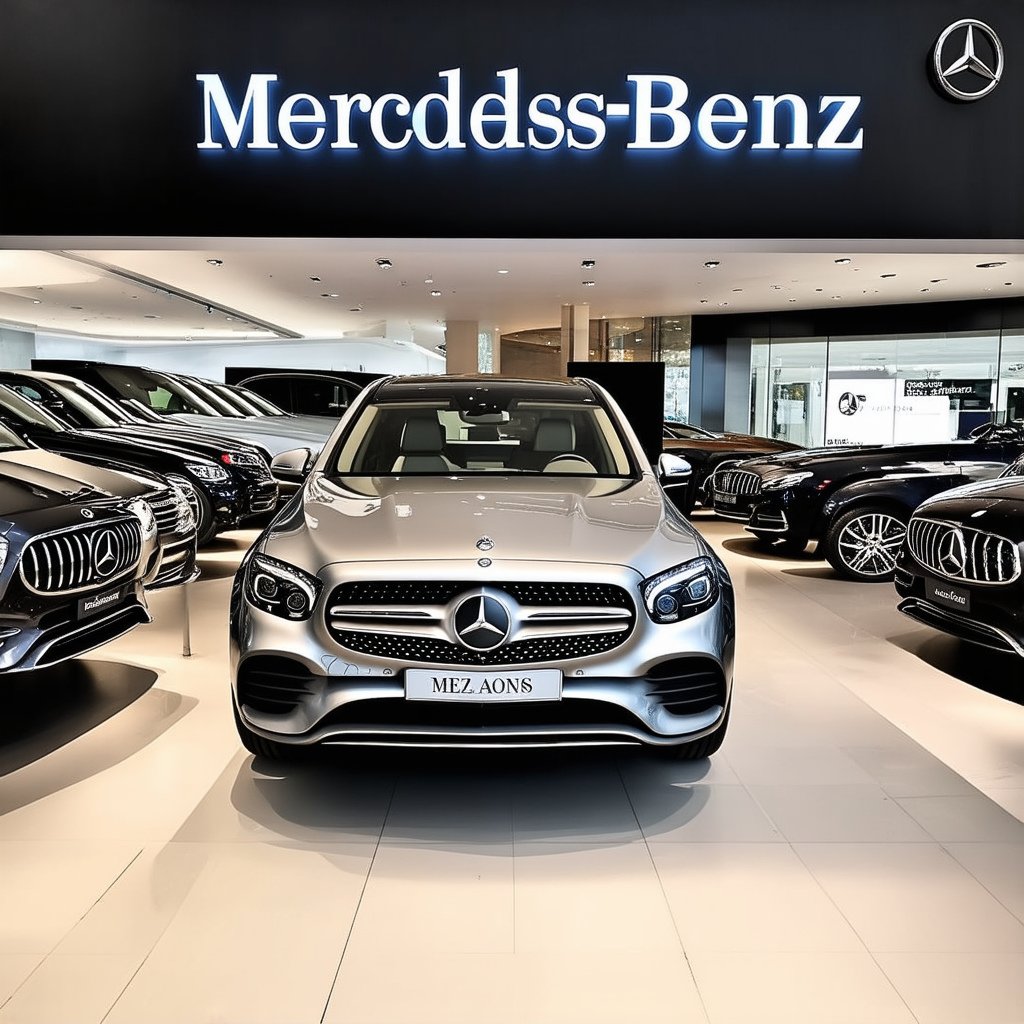} &
\includegraphics[width=0.111\textwidth]{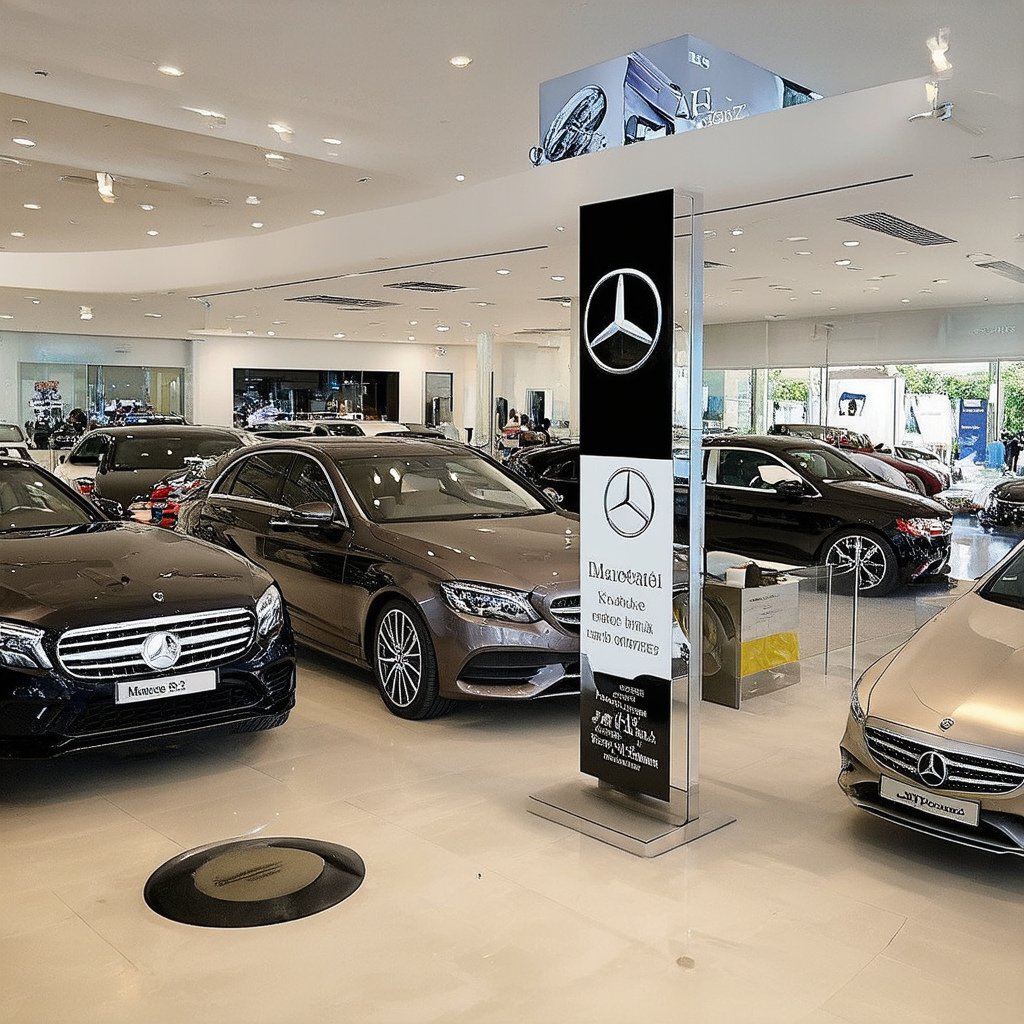} &
\includegraphics[width=0.111\textwidth]{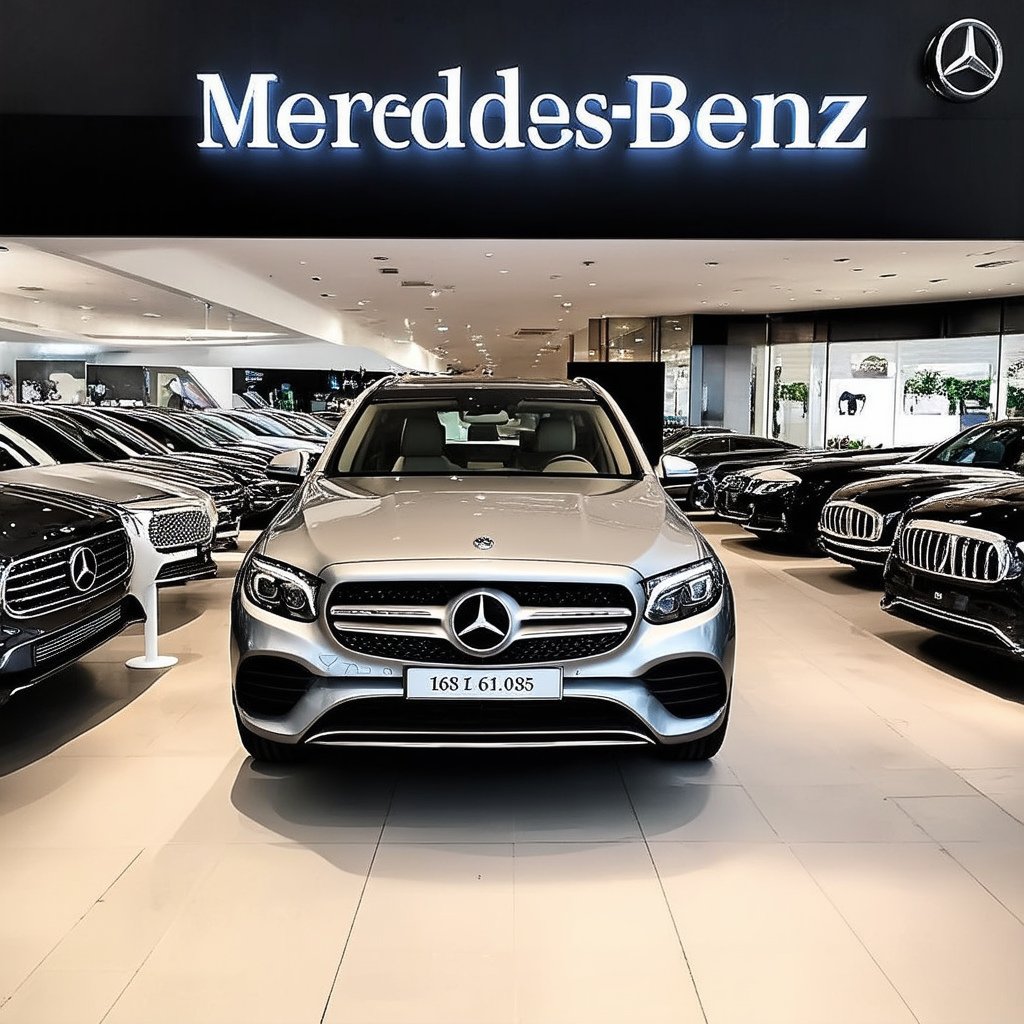} &
\includegraphics[width=0.111\textwidth]{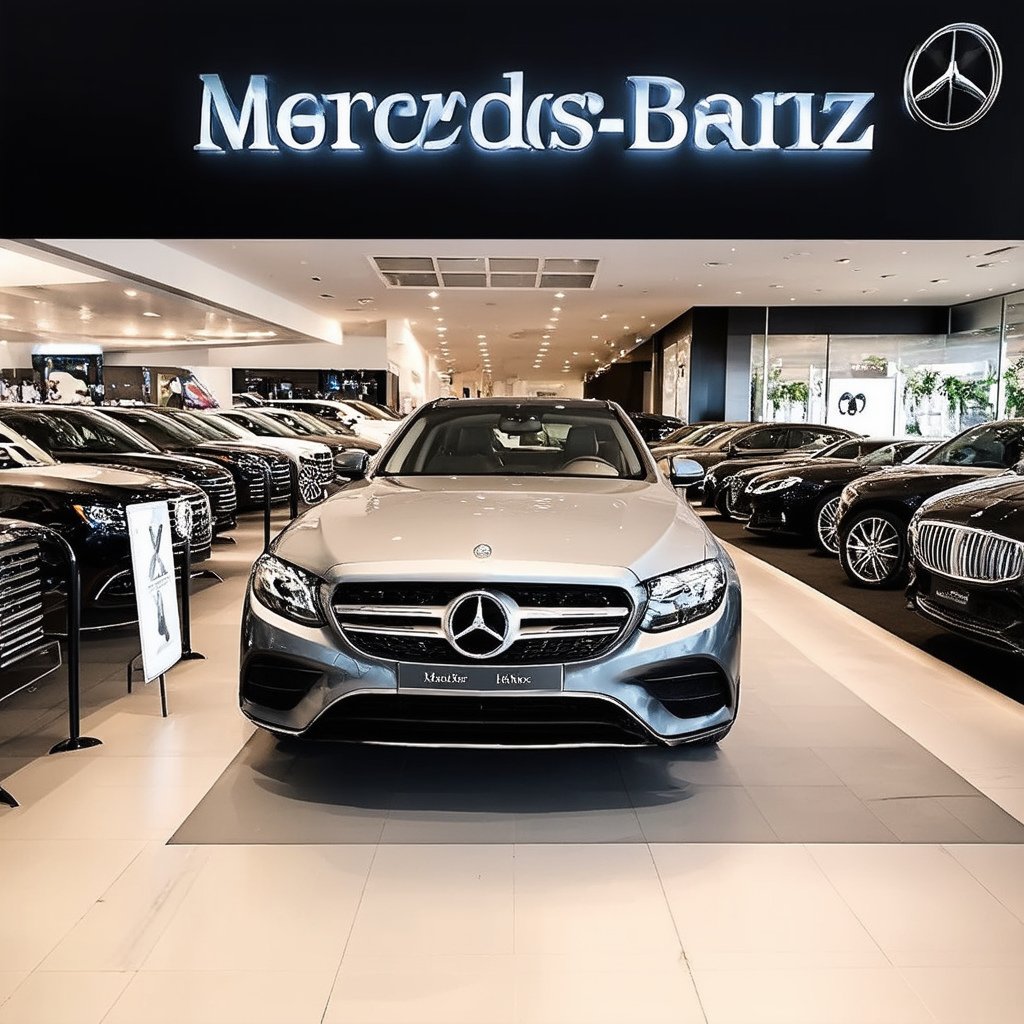} &
\includegraphics[width=0.111\textwidth]{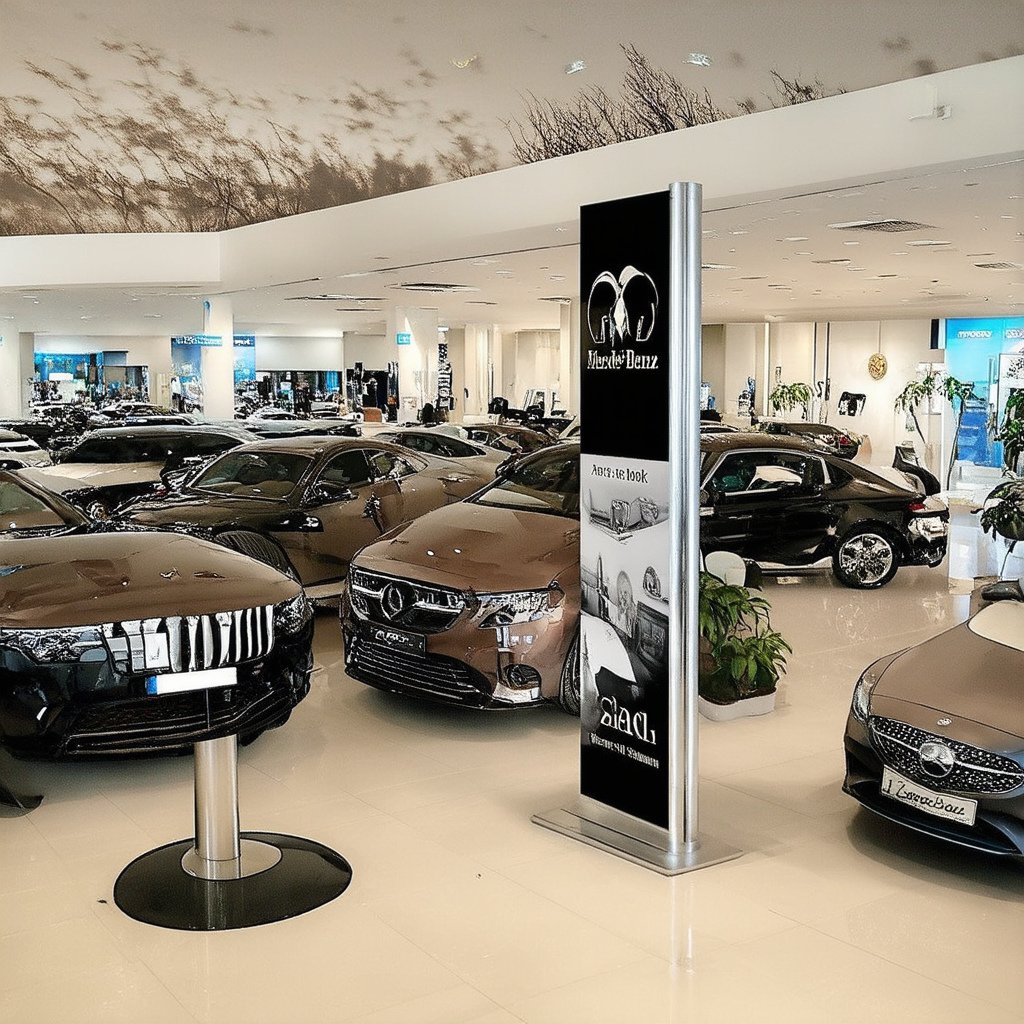} &
\includegraphics[width=0.111\textwidth]{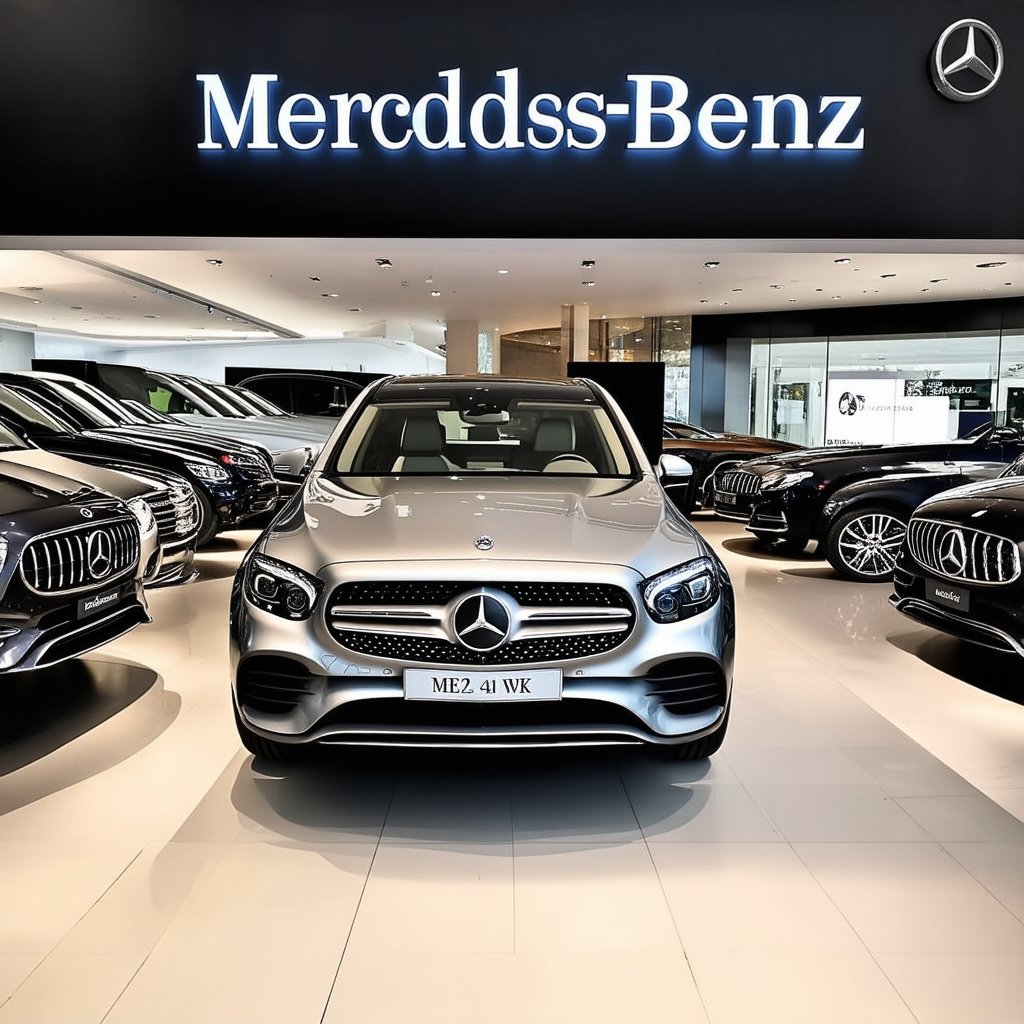} &
\includegraphics[width=0.111\textwidth]{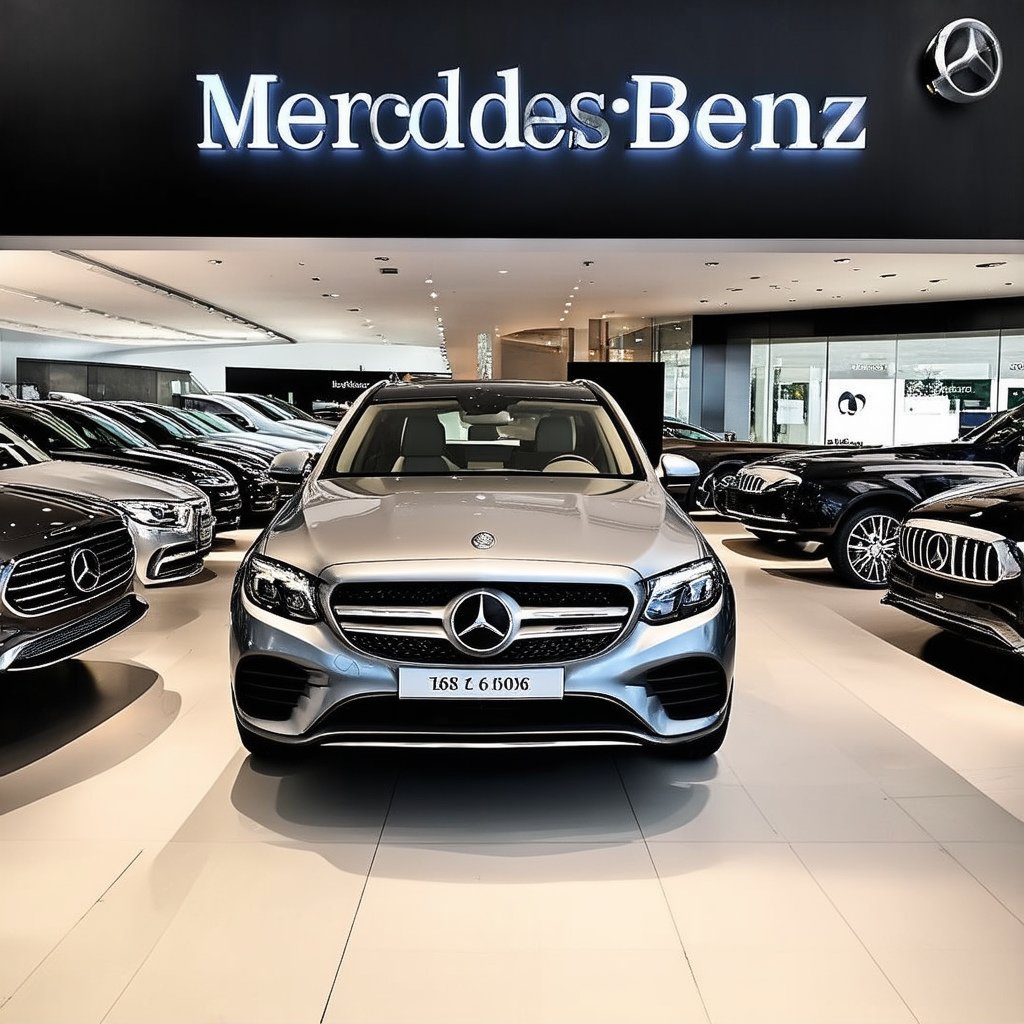} &
\includegraphics[width=0.111\textwidth]{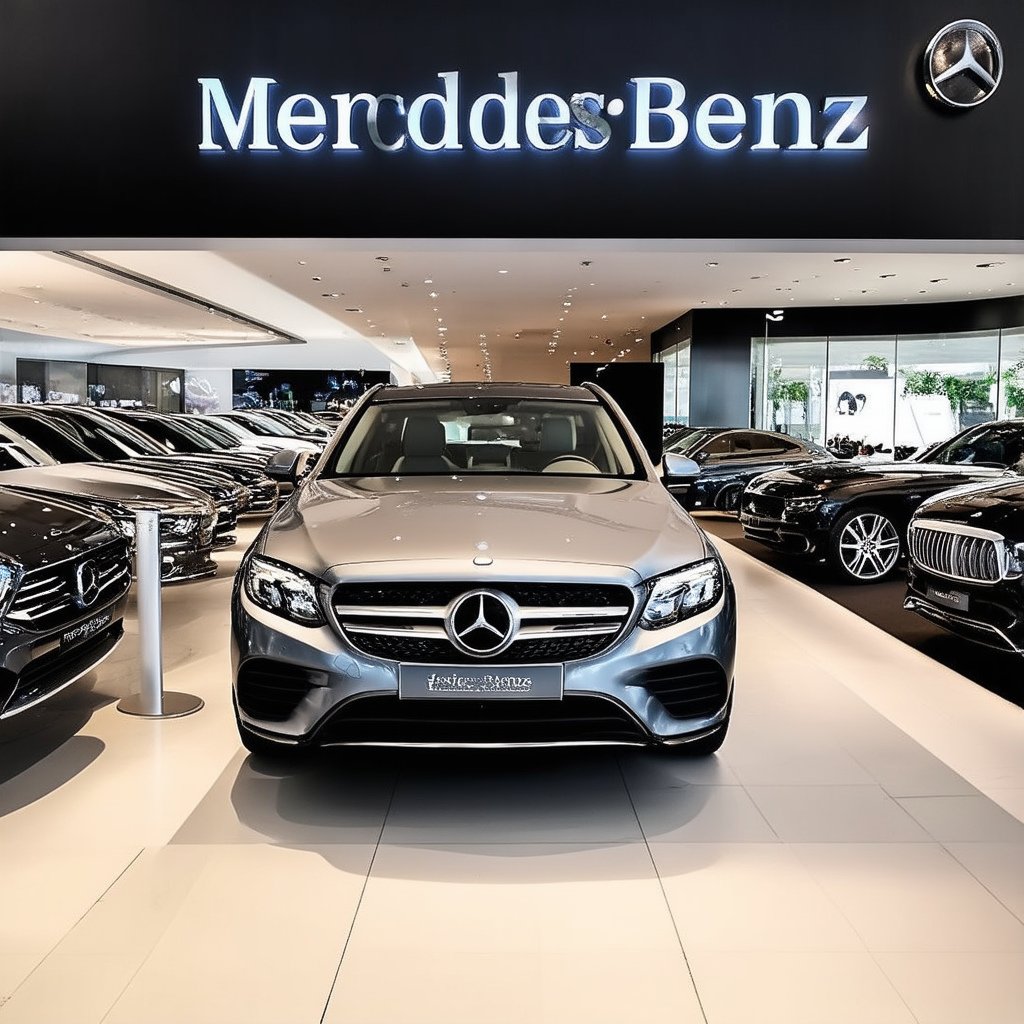} &
\includegraphics[width=0.111\textwidth]{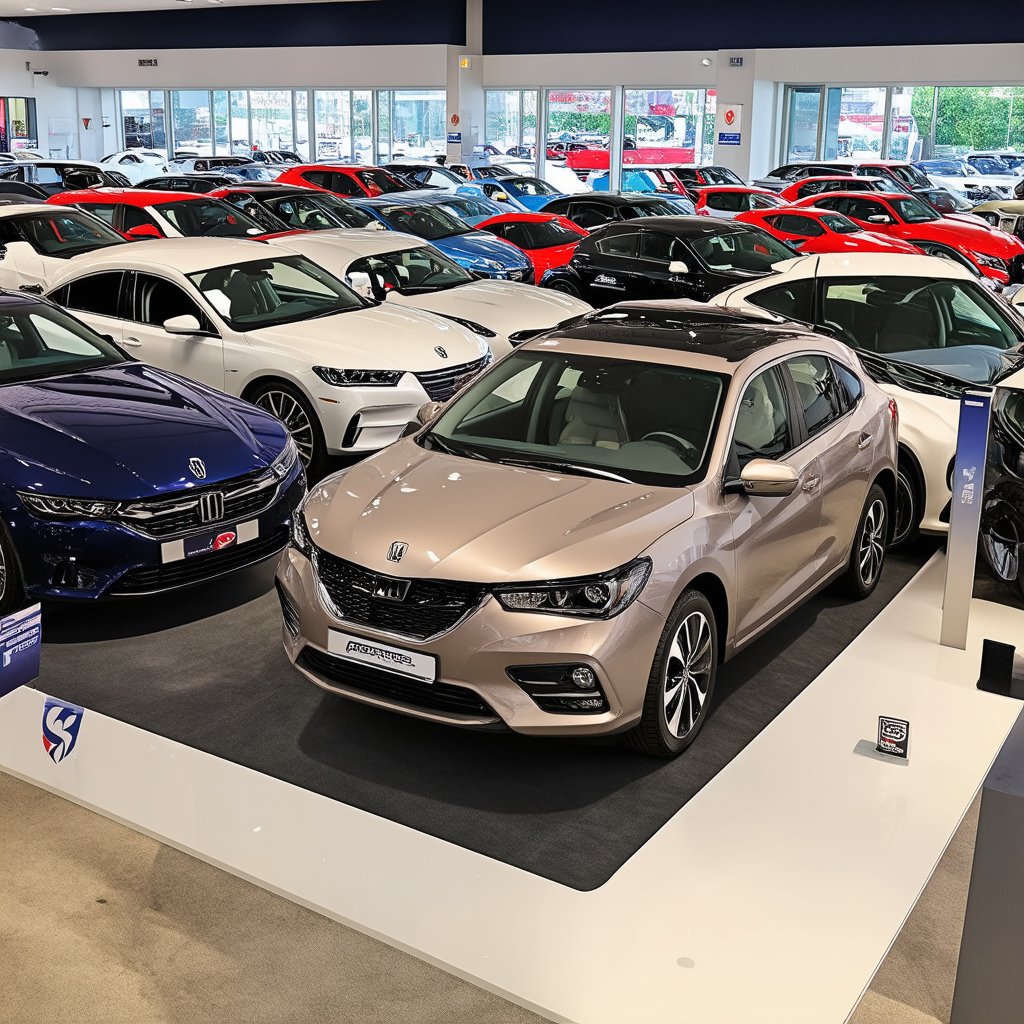} \\

\includegraphics[width=0.111\textwidth]{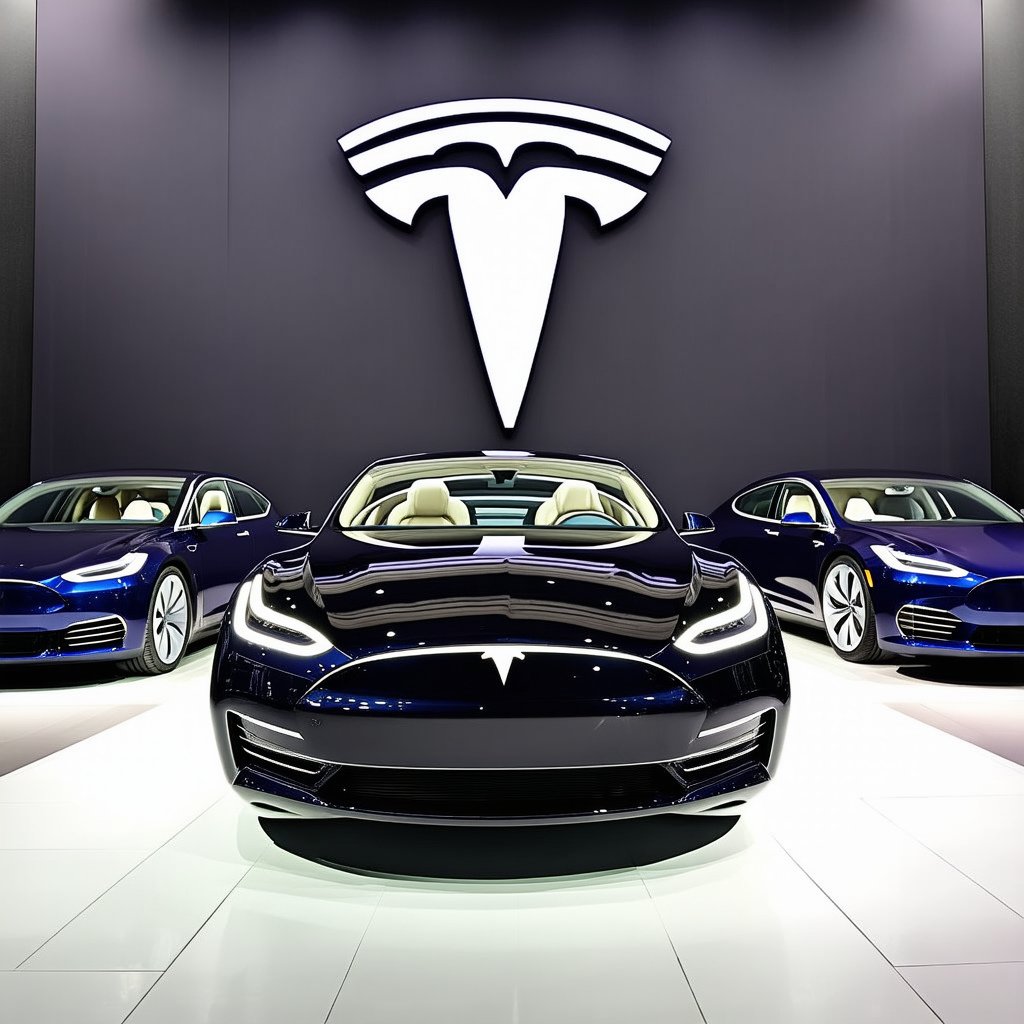} &
\includegraphics[width=0.111\textwidth]{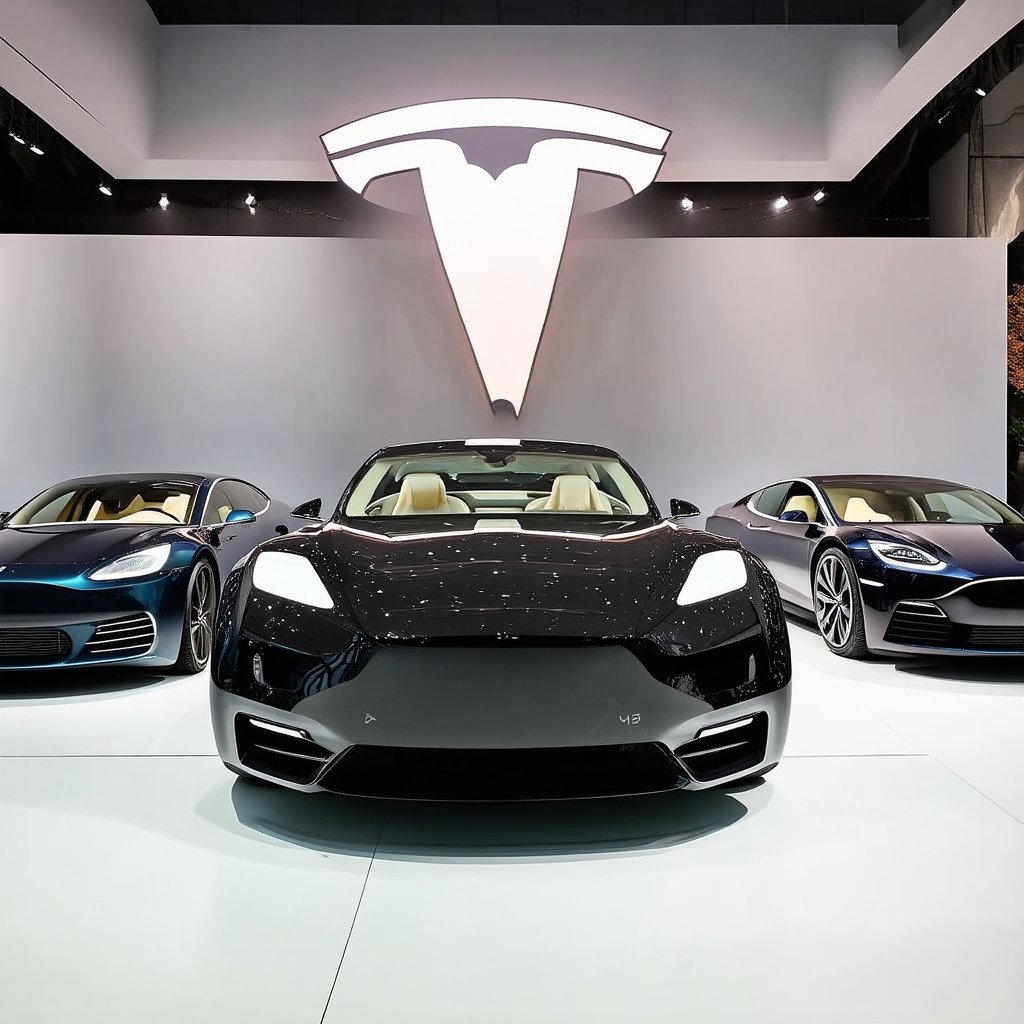} &
\includegraphics[width=0.111\textwidth]{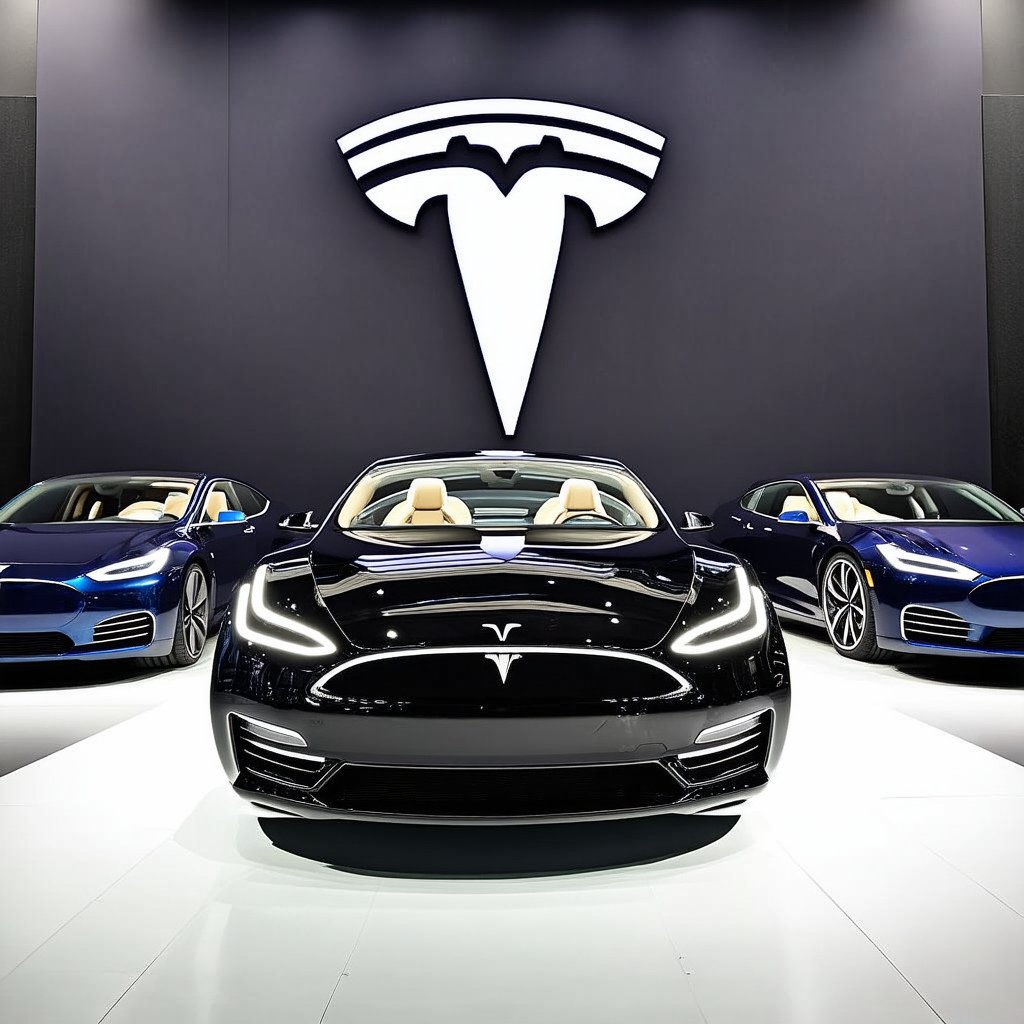} &
\includegraphics[width=0.111\textwidth]{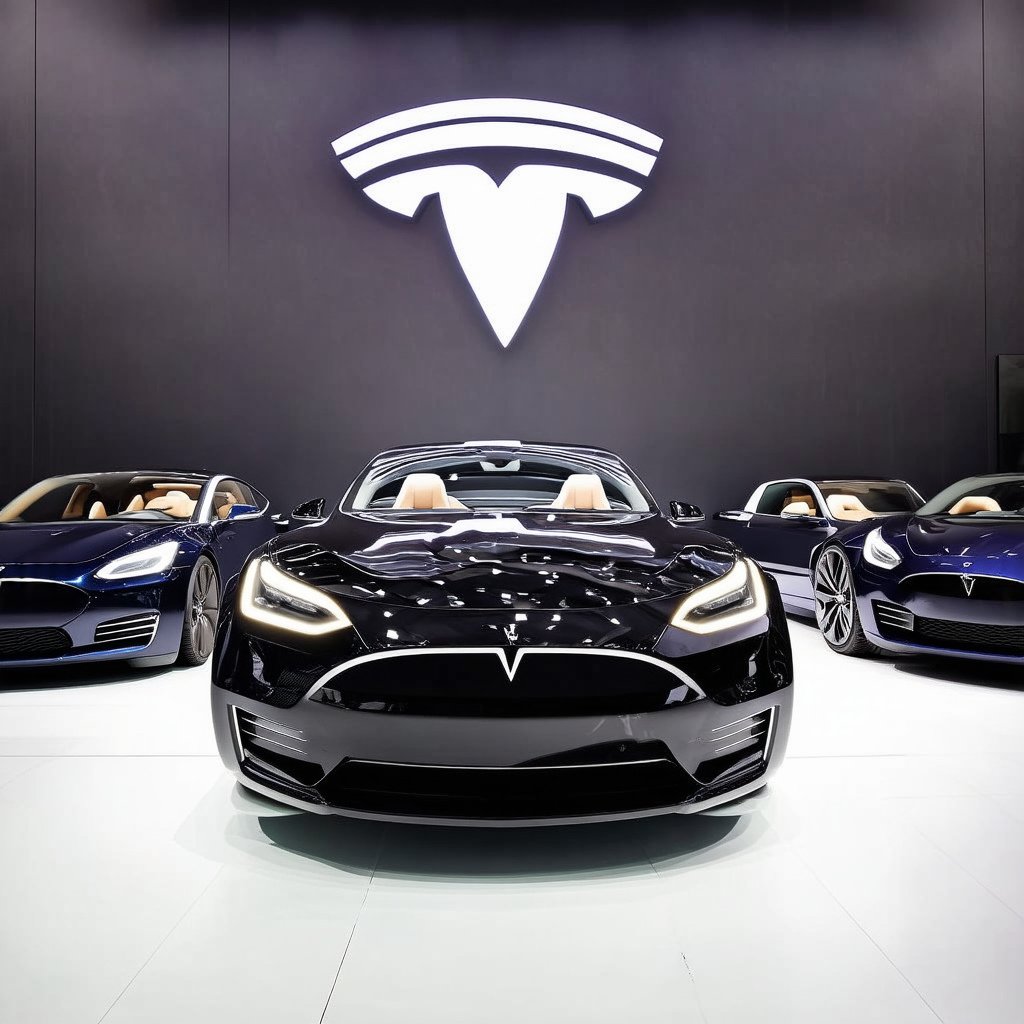} &
\includegraphics[width=0.111\textwidth]{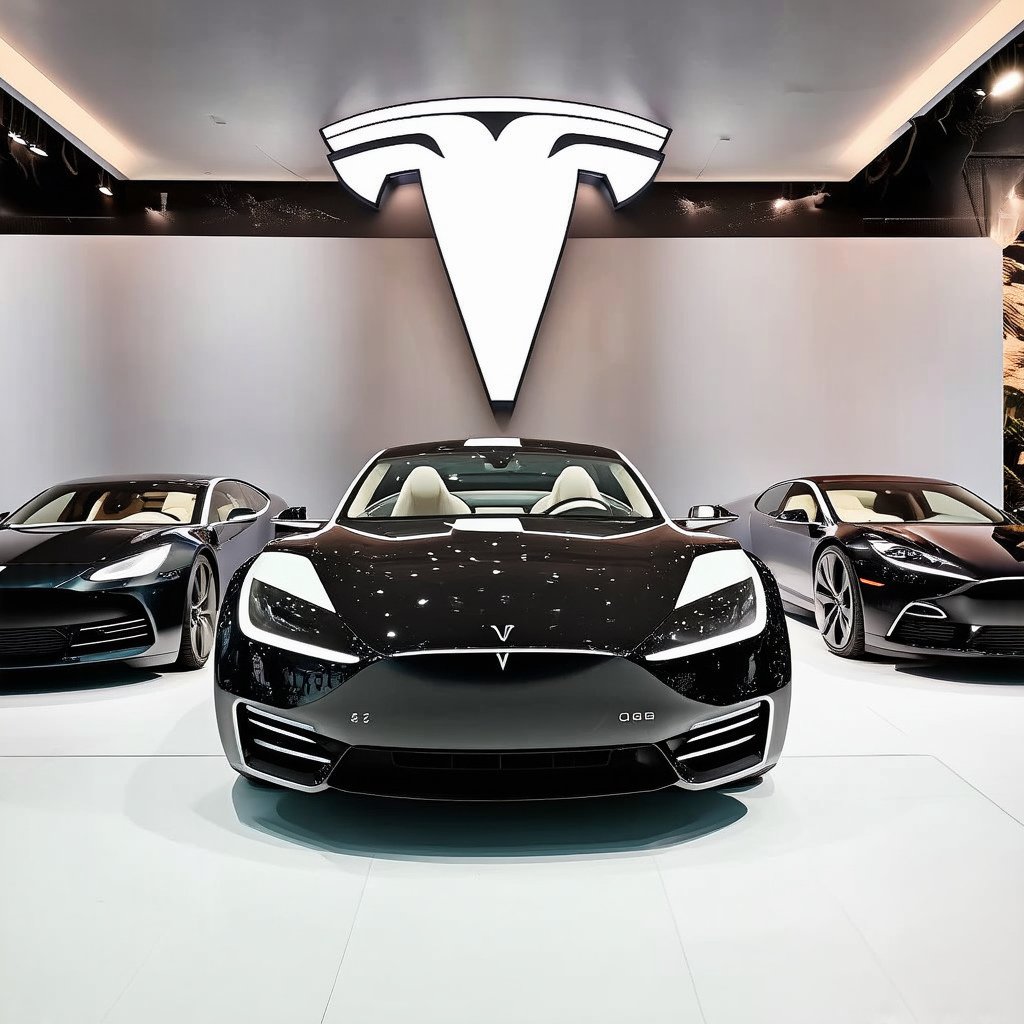} &
\includegraphics[width=0.111\textwidth]{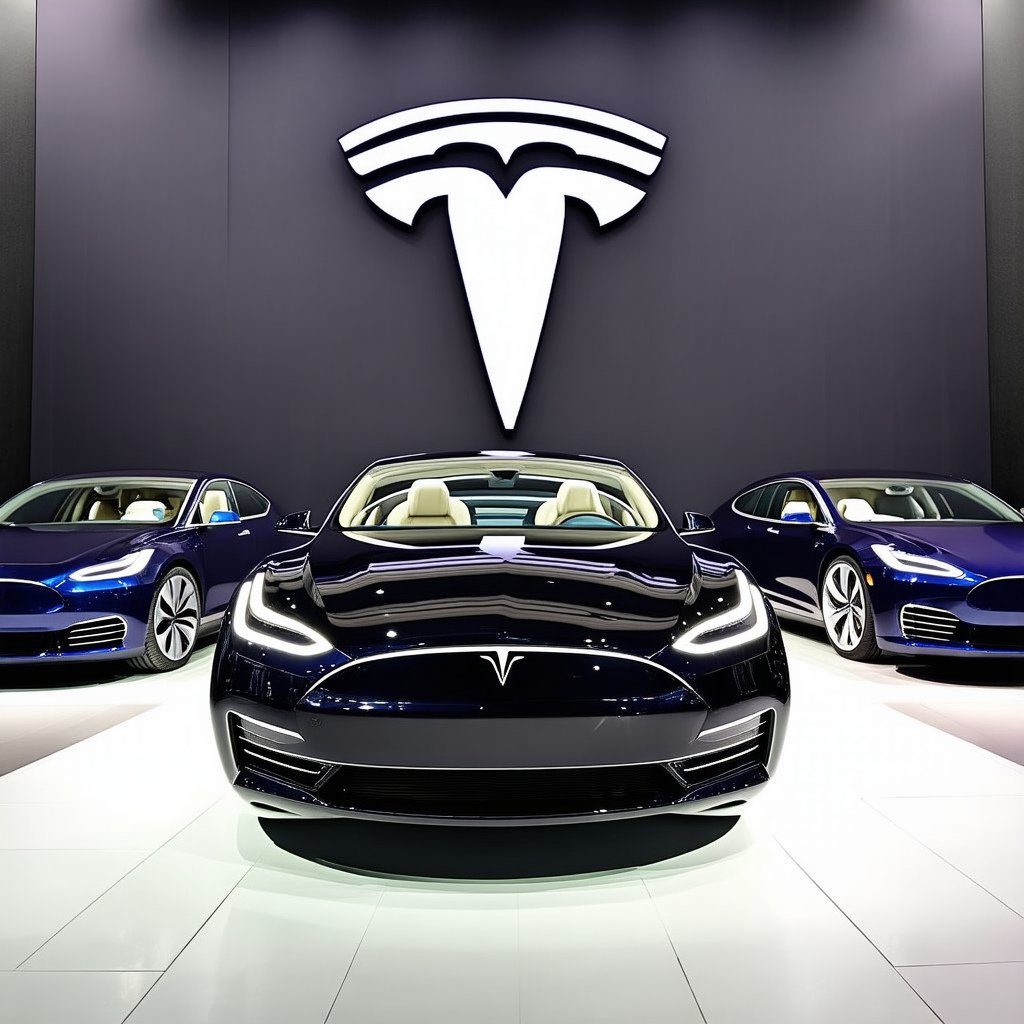} &
\includegraphics[width=0.111\textwidth]{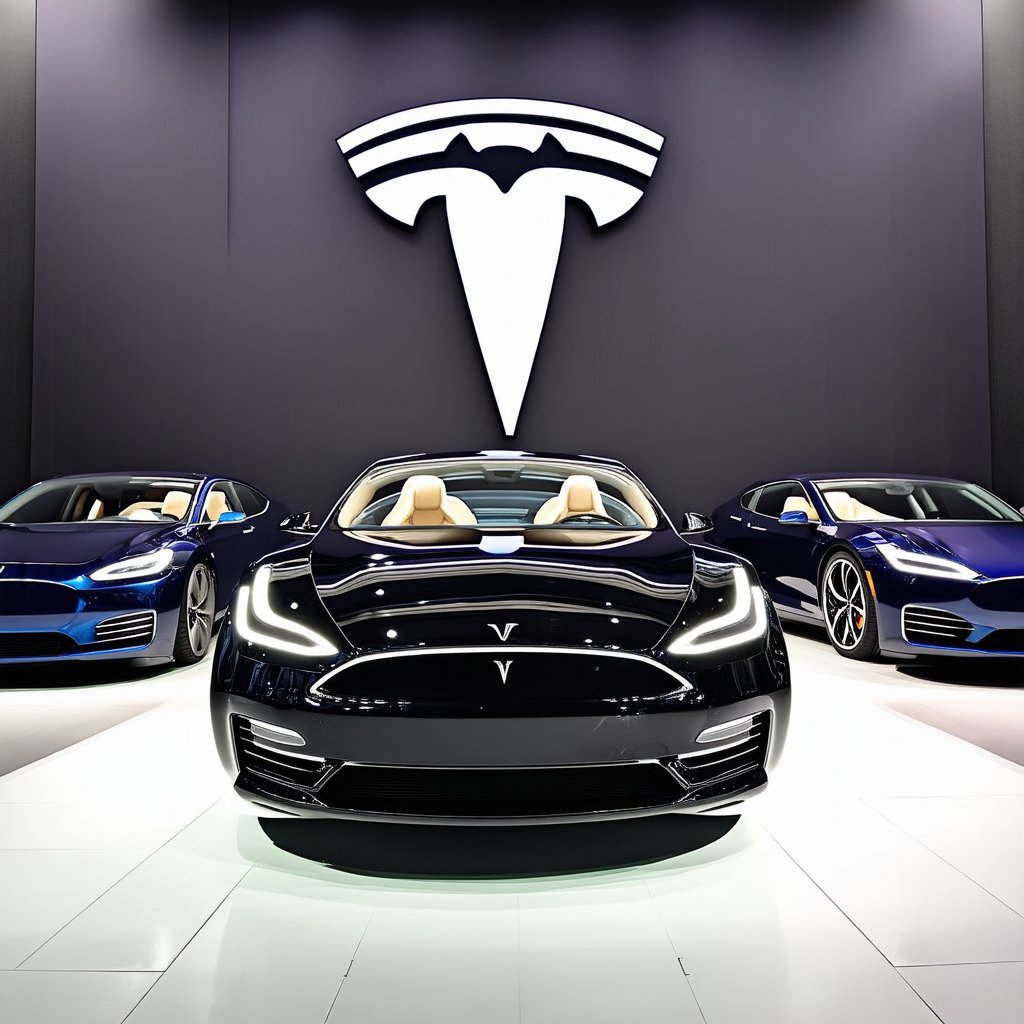} &
\includegraphics[width=0.111\textwidth]{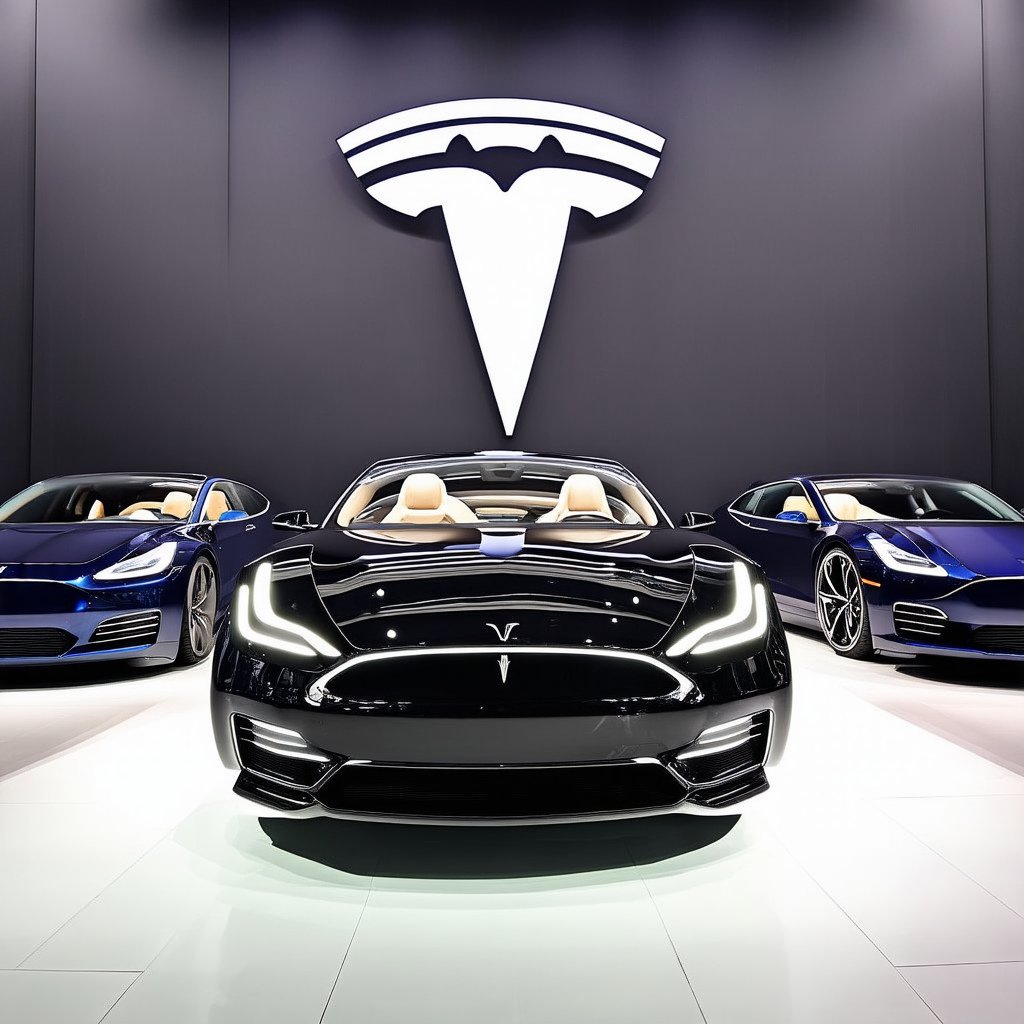} &
\includegraphics[width=0.111\textwidth]{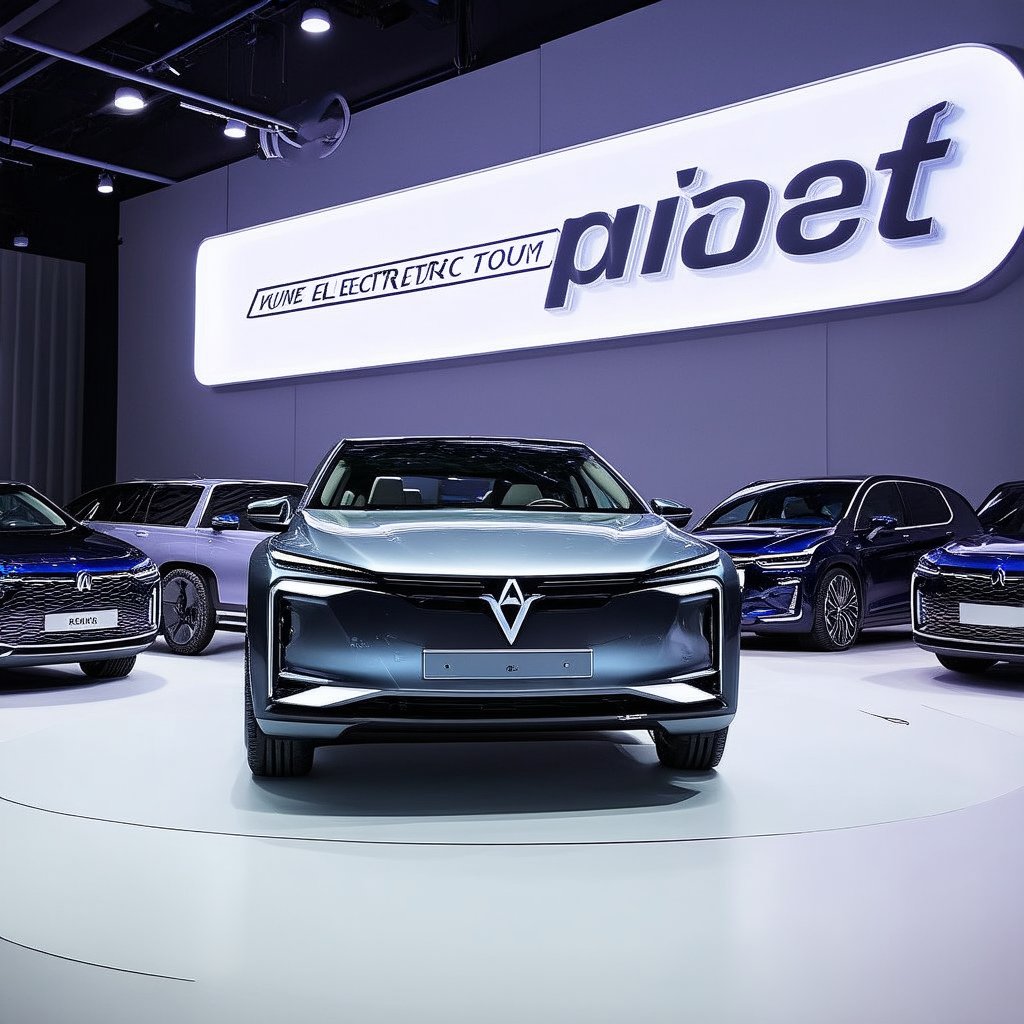} \\

\includegraphics[width=0.111\textwidth]{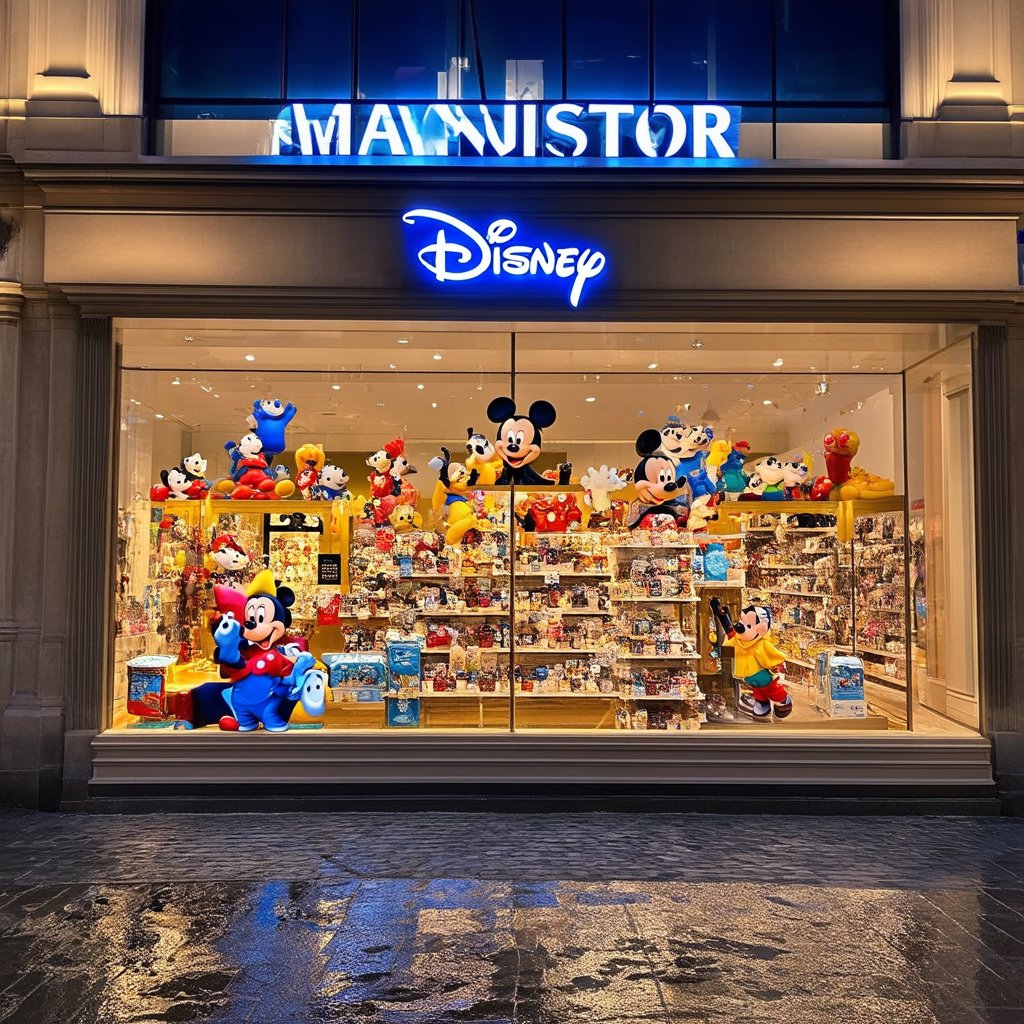} &
\includegraphics[width=0.111\textwidth]{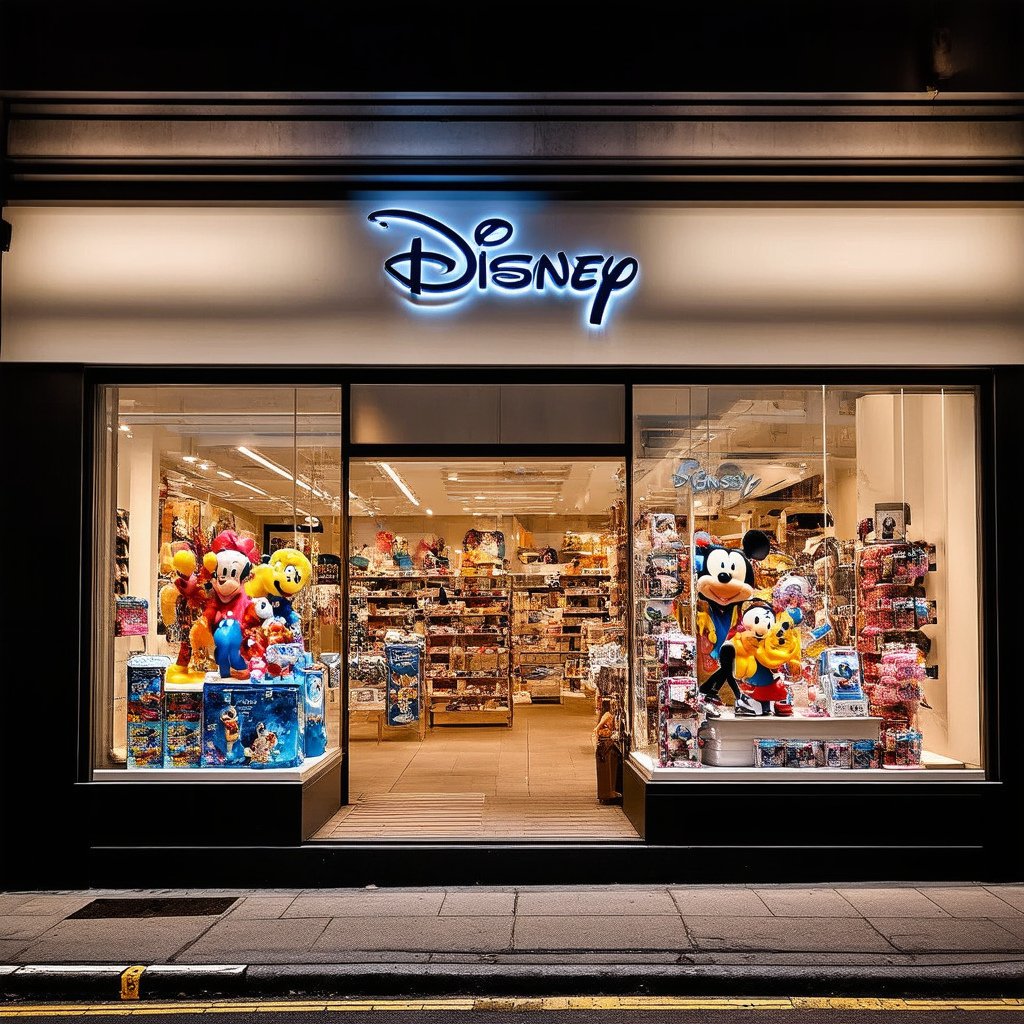} &
\includegraphics[width=0.111\textwidth]{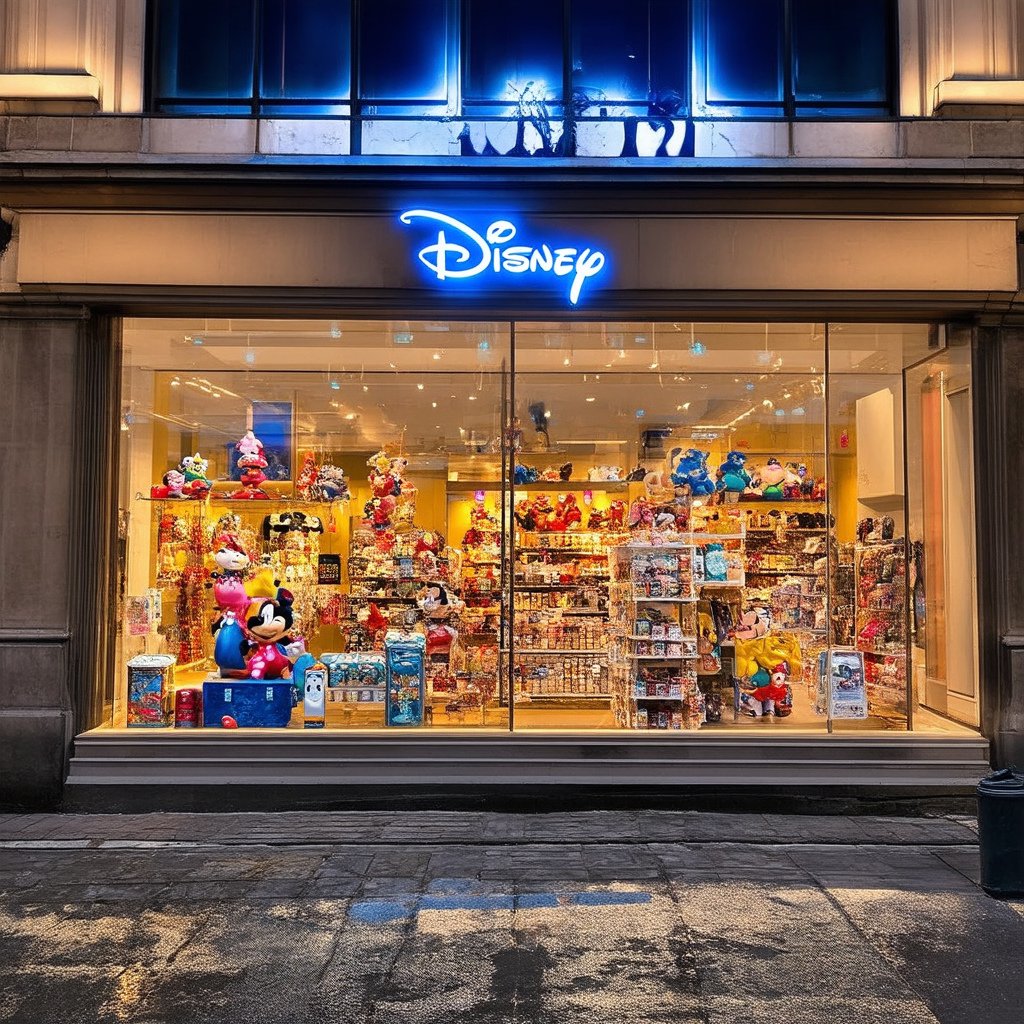} &
\includegraphics[width=0.111\textwidth]{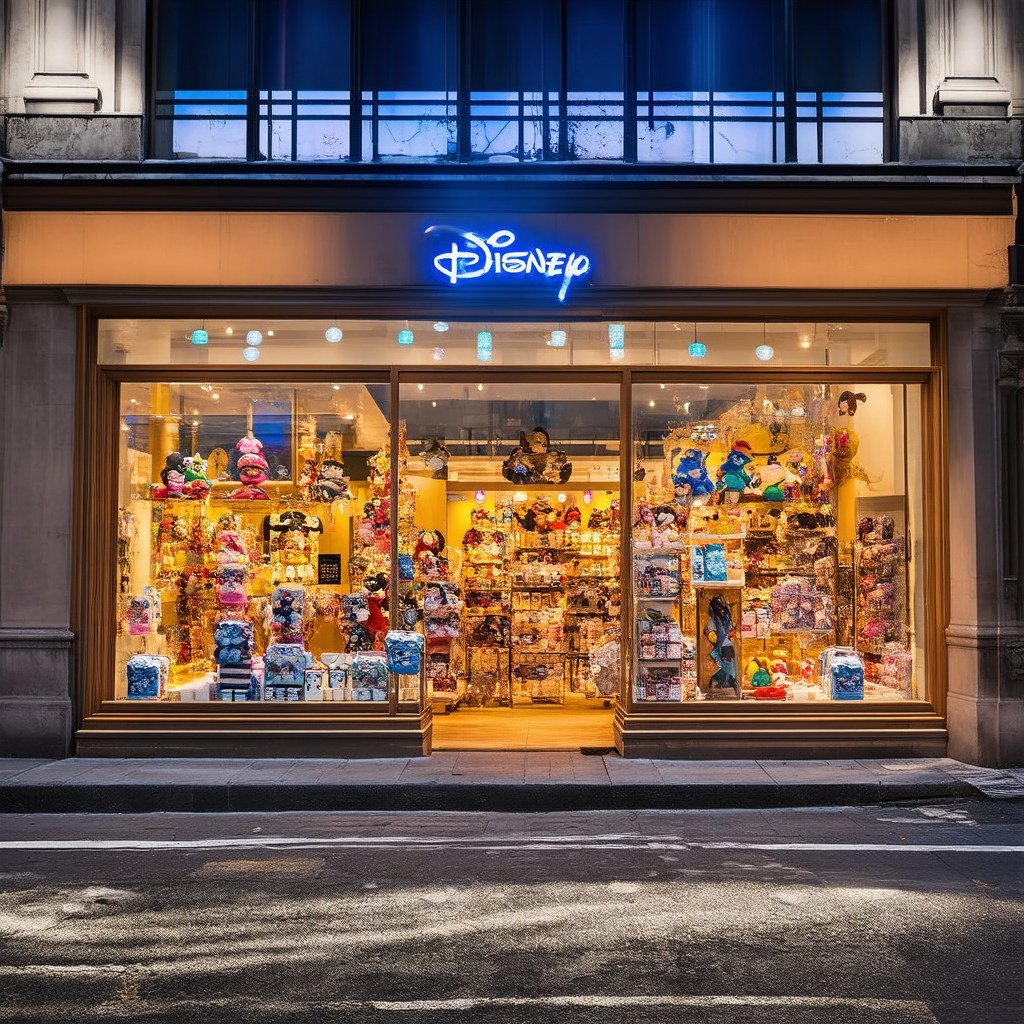} &
\includegraphics[width=0.111\textwidth]{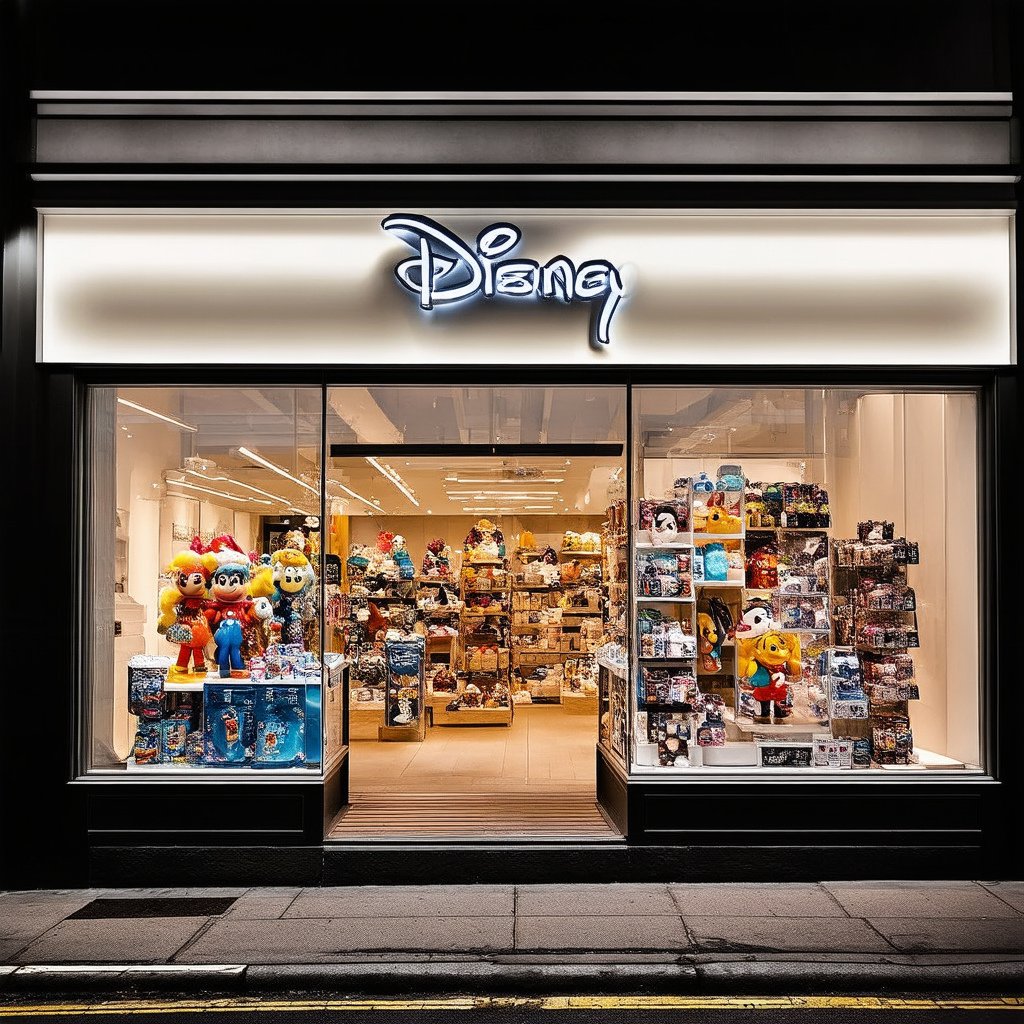} &
\includegraphics[width=0.111\textwidth]{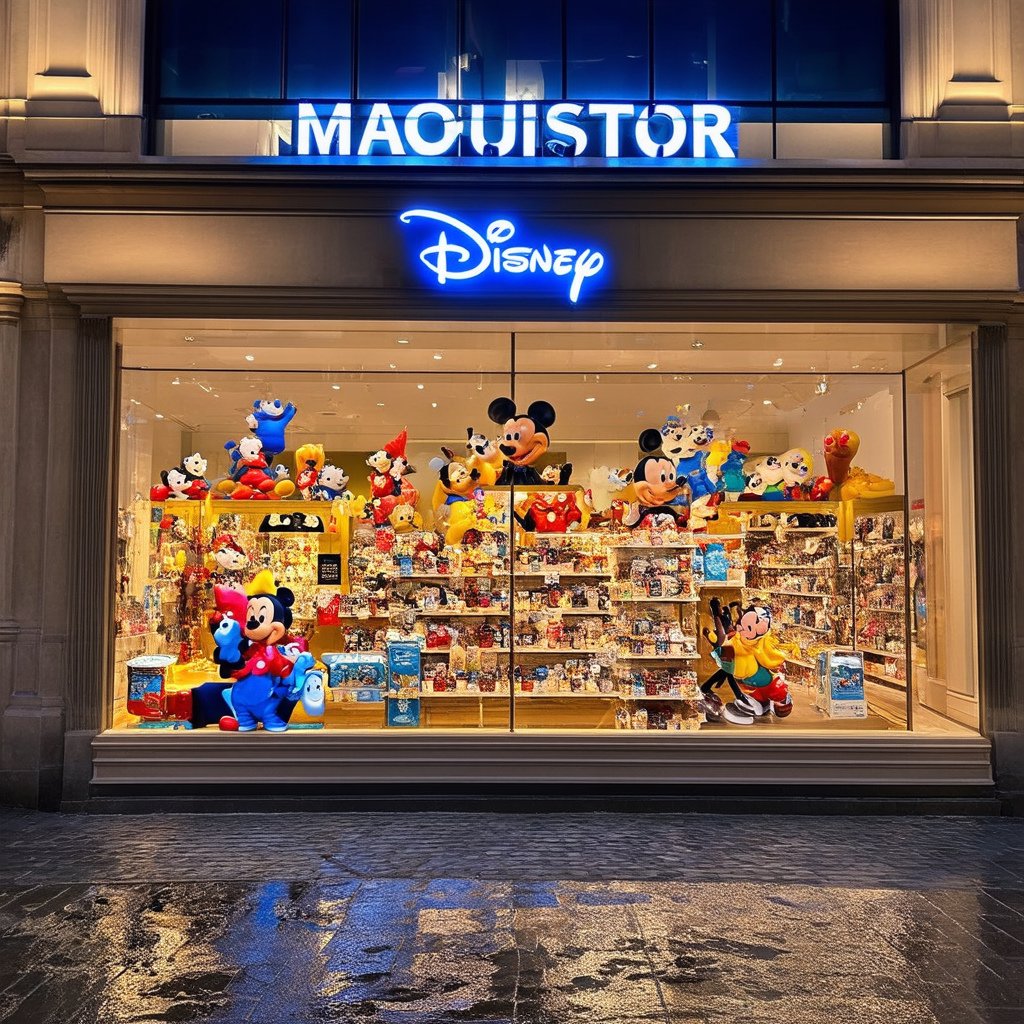} &
\includegraphics[width=0.111\textwidth]{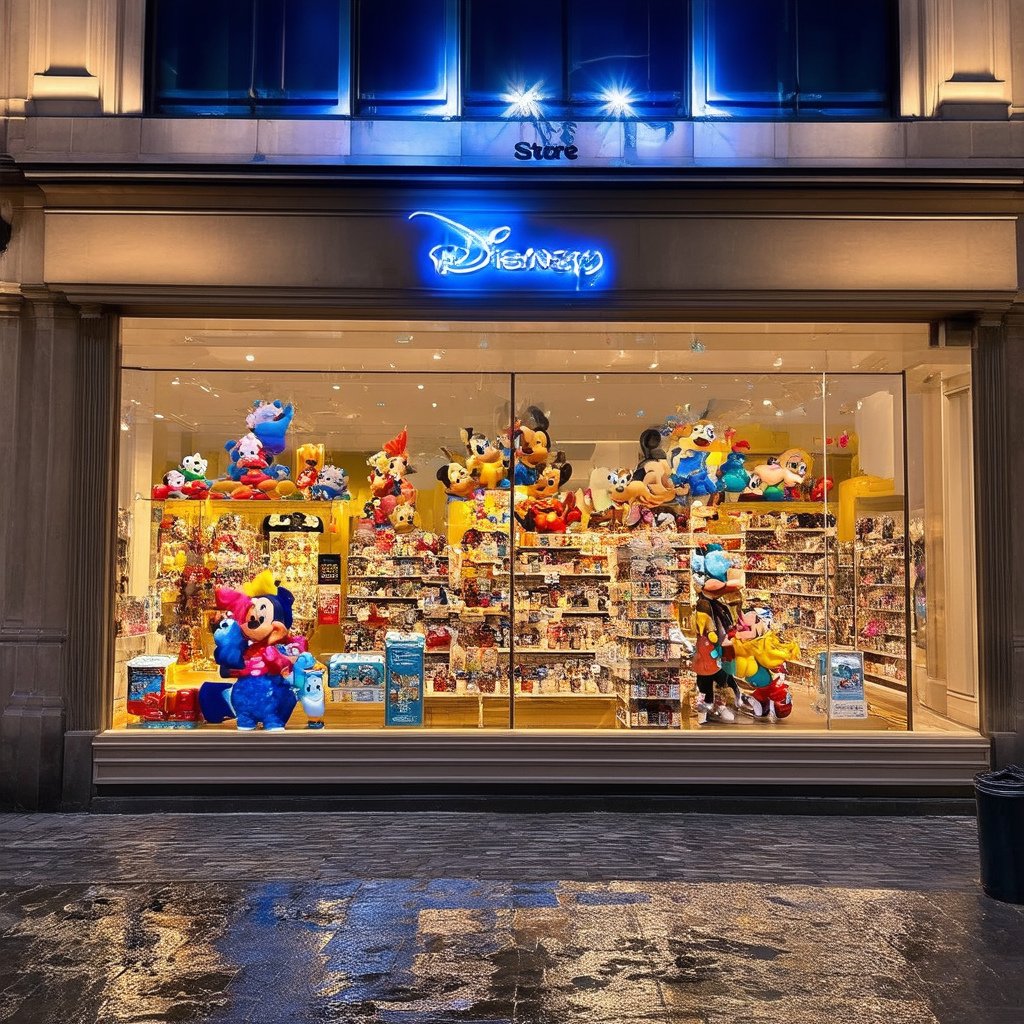} &
\includegraphics[width=0.111\textwidth]{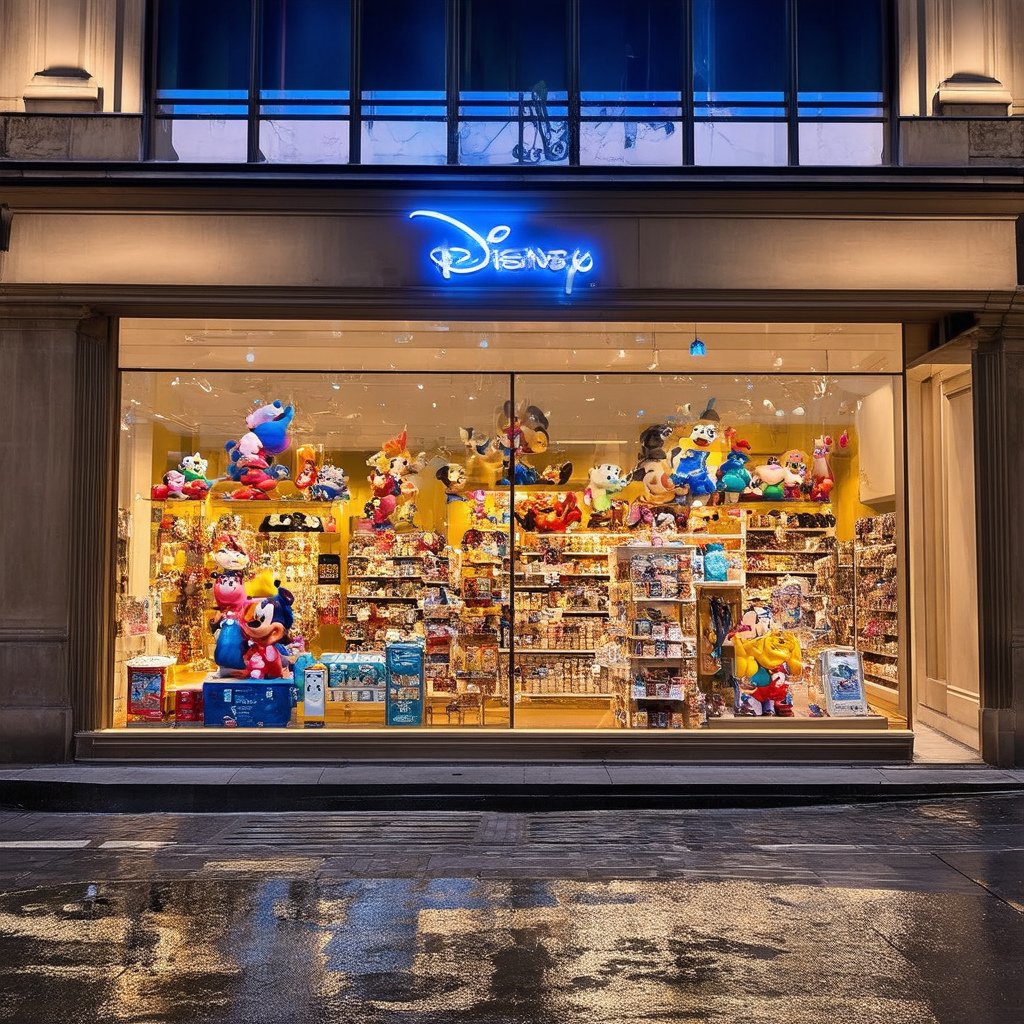} &
\includegraphics[width=0.111\textwidth]{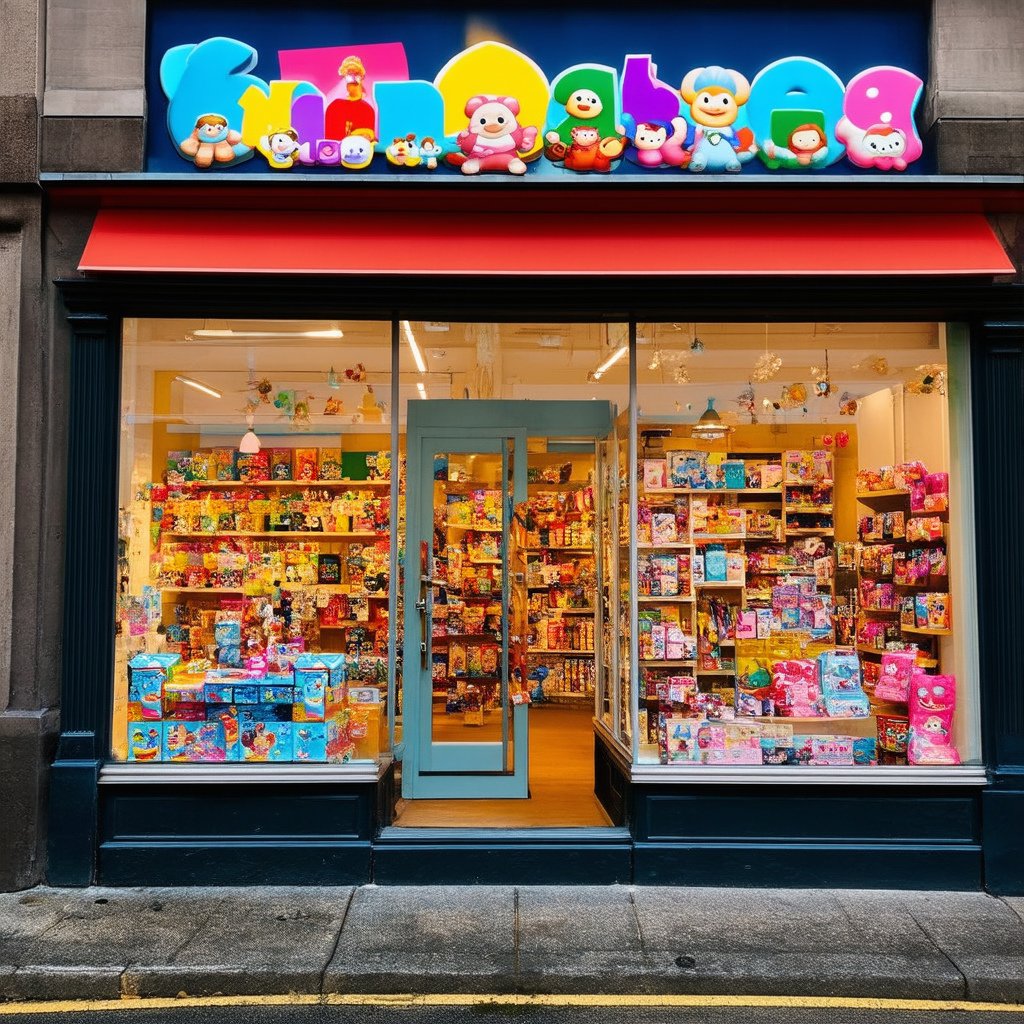} \\

\end{tabular}
\caption{\textbf{Qualitative comparison of inference-time unlearning methods across diverse logo topologies.}
Samples span pure geometric icons (\eg, Apple, Tesla), typography-centric designs (\eg, Coca-Cola, Walt Disney), and complex composites (\eg, Mercedes-Benz).
These results visually corroborate the systemic trade-off in latent-space interventions: scaling safety guidance (left to right for SLD/SEGA) distorts the localized logo but catastrophically degrades global image fidelity. While our semantic baseline (\ourBaseline) achieves superior erasure, it underscores the inescapable challenge of preserving highly entangled backgrounds.
}
\label{tab:results}
\end{figure*}

\subsection{Evaluation}
\label{subsec:method_evaluation}
Logo unlearning cannot be evaluated by target disappearance alone.
A method may suppress the logo by changing the entire image, or it may preserve the scene while leaving a recognizable brand trace.
We therefore evaluate two components: logo detection to identify residual brand regions, and five multi-grained metrics to separately quantify local logo change and global scene preservation.

\subsubsection{Logo Detection}
Logo detection in generated images is nontrivial because the marks may be distorted, partially rendered, or embedded in complex scenes.
Supervised detectors can be tied to fixed training distributions, while template-based matching can be brittle under scale, pose, and typography changes.
We use OWLv2~\citep{minderer2024scaling}, an open-vocabulary detector, to localize candidate logo regions with text queries such as a company name followed by `logo'.
To reduce evaluator noise, we manually verify the automatically produced boxes on the generated corpus and observe 98\% detection accuracy in this verification.
The detector is used only for evaluation, not for the unlearning methods.

\subsubsection{Comprehensive Metrics}
Many inference-time baselines~\citep{schramowski2023safe,brack2023sega,np} are often summarized with binary success rates.
For logos, this is too coarse because the desired edit is spatially small but semantically important.
We define five metrics that jointly capture local erasure and global preservation.
When no logo is detected after unlearning, local metrics are assigned zero, corresponding to no measurable residual logo region under the detector.

\paragraph{\metricone} This metric measures CLIP~\citep{radford2021learning} cosine similarity between the detected logo region and the corresponding company-name query.
Lower scores indicate weaker semantic alignment with the target brand.

\paragraph{\metrictwo and \metricthree} These metrics compare the detected logo region before and after unlearning.
\metrictwo uses CLIP image-feature similarity, while \metricthree uses pixel-level Structural Similarity (SSIM)~\citep{wang2004image}.
Lower values indicate stronger local disruption of the target logo region.

\paragraph{\metricfour and \metricfive} These metrics evaluate whether the non-target scene is preserved by comparing the full image before and after unlearning.
\metricfour measures global CLIP image similarity, and \metricfive measures global SSIM.
Higher values are desirable, indicating that logo removal has not unnecessarily degraded the broader image content.

\begin{table*}[ht]
\centering
\renewcommand{\arraystretch}{1.2}
\caption{\textbf{Quantitative evaluation reveals inference-time unlearning's systemic inefficacy.} Original CLIP-Text scores 0.32 on explicit (Ex) and implicit (Im) tracks. Best and second-best are \textbf{bold} and \underline{underlined}. Stronger safety guidance in SLD~\citep{schramowski2023safe} and SEGA~\citep{brack2023sega} marginally improves local erasure (lower \metricone-\metricthree) but catastrophically degrades global fidelity (plummeting \metricfour, \metricfive). Our baseline (\ourBaseline) achieves deep local ablation but lacks global preservation, proving the inescapable trade-off in disentangling corporate assets.}
\adjustbox{max width=\linewidth}{
\large
\begin{tabular}{ccccccccccc}
    \toprule
    \multirow{2}{*}{Method}   & \multicolumn{2}{c}{\metricone$\downarrow$} & \multicolumn{2}{c}{\metrictwo$\downarrow$} & \multicolumn{2}{c}{\metricthree$\downarrow$} & \multicolumn{2}{c}{\metricfour$\uparrow$} & \multicolumn{2}{c}{\metricfive$\uparrow$}\\
    \cmidrule(r){2-3} \cmidrule(r){4-5} \cmidrule(r){6-7} \cmidrule(r){8-9} \cmidrule(r){10-11}
    & Ex & Im & Ex & Im & Ex & Im & Ex & Im & Ex & Im \\
    \midrule
    NP~\citep{np} & 0.30 & 0.29 & 0.64 & 0.69 & 0.09 & 0.08 & 0.80 & 0.82 & 0.53 & 0.49 \\
    \midrule
    \sldzero    & 0.32 & 0.31 & 0.79 & 0.79 & 0.29 & 0.23 & \underline{0.94} & \underline{0.94} & 0.86 & 0.83 \\
    \sldhalf & 0.29 & 0.28 & 0.65 & 0.69 & 0.14 & 0.11 & 0.82 & 0.85 & 0.73 & 0.68 \\
    \sldone    & \underline{0.28} & \underline{0.28} & \underline{0.60} & \underline{0.66} & \underline{0.08} & \underline{0.07} & 0.75 & 0.79 & 0.50 & 0.47 \\
    \midrule
    \segazero  & 0.32 & 0.31 & 0.84 & 0.85 & 0.42 & 0.37 & \textbf{0.98} & \textbf{0.98} & \textbf{0.97} & \textbf{0.97} \\
    \segahalf & 0.30 & 0.30 & 0.75 & 0.76 & 0.24 & 0.19 & 0.91 & 0.92 & \underline{0.90} & \underline{0.87} \\
    \segaone & 0.30 & 0.29 & 0.70 & 0.73 & 0.19 & 0.14 & 0.87 & 0.89 & 0.83 & 0.79 \\
    \midrule
    \ourBaseline & \textbf{0.26} & \textbf{0.26} & \textbf{0.52} & \textbf{0.57} & \textbf{0.07} & \textbf{0.05} & 0.71 & 0.77 & 0.55 & 0.53 \\
    \bottomrule
\end{tabular}
}
\label{tab:metric}
\end{table*}

\section{Experiment}
\label{sec:exp}

\subsection{Experimental Setting}
\label{subsec:setting}
We use \ourSD~\citep{esser2024scaling,sd3} as the main T2I backbone for \ourBench.
NP~\citep{np} is directly supported.
SLD~\citep{schramowski2023safe} and SEGA~\citep{brack2023sega} were originally developed for stable-diffusion-v1-5~\citep{sd1.5}; we adapt their inference logic to the \ourSD pipeline following Algorithm 1 in Appendix H of~\citep{schramowski2023safe} and Algorithm 1 in Appendix A of~\citep{brack2023sega}.
All three are inference-time interventions and do not require model retraining.
Implementation details are provided in the released code.

For SLD and SEGA, we evaluate the three hyperparameter settings in \Cref{tab:hyper}.
For SLD, \sldzero follows Hyp-Strong and \sldhalf follows Hyp-Max from the original configuration~\citep{schramowski2023safe}; \sldone probes a stronger guidance regime.
For SEGA, \segazero, \segahalf, and \segaone follow progressively different editing strengths based on~\citep{brack2023sega}.
We fix the random seed across paired generations to reduce sampling variance.

\paragraph{Evaluation Metrics Recap}
\label{subsec:exp_metrics}
As defined in \Cref{sec:method}, \metricone measures semantic alignment between the detected logo and the target company name.
\metrictwo and \metricthree measure local logo-region similarity in CLIP feature space and pixel space.
\metricfour and \metricfive measure global image preservation using full-image CLIP similarity and SSIM~\citep{wang2004image}.

\subsection{Performance of Inference-Time Methods}
\label{subsec:exp_performance}

\Cref{tab:metric} shows a consistent local-global trade-off for the evaluated inference-time methods.
NP~\citep{np}, SLD~\citep{schramowski2023safe}, and SEGA~\citep{brack2023sega} reduce local logo metrics only partially in many settings, and \metricone remains well above the ideal value of zero.
The pattern appears in both explicit and implicit tracks, indicating that logo leakage is not fully addressed by suppressing direct logo keywords.

Increasing safety or editing guidance often improves \metricone, \metrictwo, and \metricthree, but the same settings reduce \metricfour and \metricfive.
In other words, stronger guidance can disrupt the target region, but it also changes global content, texture, layout, or object identity.
This is precisely the failure mode that \ourBench is designed to reveal: for logos, a useful intervention must be local enough to preserve the image while still removing the protected mark.

\ourBaseline provides a complementary reference point.
Because it operates by rewriting prompts, it can remove many logo-inducing semantic cues and achieves the strongest local erasure scores in \Cref{tab:metric}.
However, the improvement is obtained by changing the input description, so non-target scene content can also change.
This result is useful diagnostically: prompt-space filtering can mitigate some explicit and contextual triggers, but it does not solve model-level logo unlearning.

\begin{table}
\centering
\renewcommand{\arraystretch}{1.2}
\caption{\textbf{Calibration of safety guidance hyperparameters for inference-time unlearning.}
To evaluate the trade-off between local concept erasure and global fidelity, we establish a calibrated spectrum of intervention intensities for SLD and SEGA. Generation parameters remain strictly constant across all cohorts (\eg, 28 inference steps, guidance scale 7.0, $1024 \times 1024$ resolution).}
\adjustbox{max width=0.8\linewidth}{
\large
\begin{tabular}{c c c c c c}
\toprule
& \makecell{Warmup Steps} & \makecell{Safety Guidance} & Threshold & \makecell{Momentum Scale} & \makecell{Momentum Beta}\\
\midrule
\sldzero & 7 & 2000 & 0.025 & 0.5 & 0.7\\
\sldhalf & 4 & 3000 & 0.500 & 0.5 & 0.7\\
\sldone & 0 & 5000 & 1.000 & 0.5 & 0.7\\
\midrule
\segazero & 10 & 4 & 0.990 & 0.3 & 0.6\\
\segahalf & 7 & 5 & 0.950 & 0.3 & 0.6\\
\segaone & 5 & 5 & 0.900 & 0.3 & 0.6\\
\bottomrule
\end{tabular}
}
\label{tab:hyper}
\end{table}

\Cref{tab:results} gives qualitative examples of these trends.
For simple geometric logos such as Apple, latent baselines often blur, resize, or deform the mark rather than cleanly remove the brand cue.
For text-heavy logos such as Coca-Cola or Disney, interventions can produce illegible text-like artifacts.
In scenes with multiple logo instances, such as Tesla, some methods affect only the most salient mark.
Under strong guidance, images may acquire unrelated artifacts or background shifts, reinforcing the need to evaluate local erasure together with global fidelity.

\begin{table*}[ht]
\centering
\renewcommand{\arraystretch}{1.2}
\caption{\textbf{Evaluating fine-tuning unlearning architectures on LUim-500.} Best results are \textbf{bold}. For fairness, our baseline (\ourBaseline) uses each method's legacy checkpoint. \ourBaseline outperforms ESD and Forget-Me-Not in local erasure (lower \metricone-\metricthree) without expensive weight updates. Conversely, Forget-Me-Not resists local unlearning but preserves higher global fidelity (\metricfour, \metricfive), confirming the inescapable systemic trade-off.}
\adjustbox{max width=\linewidth}{
\large
\begin{tabular}{lcccccc}
\toprule
T2I Architecture & Method& \metricone$\downarrow$ & \metrictwo$\downarrow$ & \metricthree$\downarrow$ & \metricfour$\uparrow$ & \metricfive$\uparrow$\\
\midrule
SD-v1-4 & ESD~\citep{gandikota2023erasing} & 0.24 & 0.89 & 0.15 & 0.72 & 0.38\\
SD-v1-4 & \ourBaseline & \textbf{0.23} & \textbf{0.89} & \textbf{0.03} & \textbf{0.78} & \textbf{0.45}\\
\midrule
SD-v2-1-base & Forget-Me-Not~\citep{zhang2024forget} & 0.28 & 0.80 & 0.13 & \textbf{0.80} & \textbf{0.24}\\
SD-v2-1-base & \ourBaseline & \textbf{0.25} & \textbf{0.65} & \textbf{0.11} & 0.75 & 0.22\\
\bottomrule
\end{tabular}
}
\label{tab:ft}
\end{table*}

\subsection{Evaluations of Fine-Tuning-Based Approaches}

We also evaluate two fine-tuning-based methods, ESD~\citep{gandikota2023erasing} and Forget-Me-Not~\citep{zhang2024forget}, on the LUim-500 subset.
This comparison is constrained by implementation compatibility: ESD is evaluated on SD-v1-4 and Forget-Me-Not on SD-v2-1-base, following their official settings.
\ourBaseline is evaluated separately within each corresponding backbone as a prompt-space reference.

\Cref{tab:ft} shows that these compatible fine-tuning settings also do not consistently achieve both strong local logo removal and high global preservation.
We therefore treat the result as scoped evidence rather than a universal statement about all weight-editing methods or modern backbones.
Still, the same tension appears: more disruptive behavior can reduce logo evidence but may also lower background fidelity, while conservative behavior preserves the image at the risk of leaving brand traces.
Detailed fine-tuning hyperparameters are provided in the supplementary materials.

\begin{table}[h!]
\centering
\renewcommand{\arraystretch}{1.2}
\caption{\textbf{Quantitative ablation of the \ourBaseline framework on \ourBenchHard.} Measured via \metricone, we isolate the exact contribution of each agent. The primary semantic ablator (Remover) drives over 80\% of the unlearning efficacy. The semantic stabilizer (Reflector) provides critical refinement, contributing $\sim$17\%. While the deterministic safeguard (Checker) yields a 0\% direct metric delta, it remains architecturally essential to guarantee the zero-tolerance robustness of the revision loop.}
\adjustbox{max width=\linewidth}{
\begin{tabular}{lccccc}
\toprule
& Before & \ourBaseline & w/o Remover & w/o Reflector & w/o Checker \\
\midrule
\metricone$\downarrow$  & 0.3156 & 0.2627 & 0.3059 & 0.2678 & 0.2627 \\ 
\metricone Reduction $\uparrow$  & 0\% & 16.76\%  & 3.07\%  & 15.15\%  & 16.76\%  \\
Relative Contribution $\uparrow$ & N/A & 100\%  & \textbf{83.2\%}  & \underline{16.8\%}  & 0\%  \\
\bottomrule
\end{tabular}
}
\label{tab:ablation}
\end{table}

\subsection{Diagnosing Failures: A Correlation Analysis}
\label{subsec:correlation}

To better understand when logo unlearning is difficult, we analyze whether local visual attributes correlate with unlearning behavior.
For detected logo regions, we measure Area, Location, Edge Density~\citep{canny1986computational}, Shape Count, Texture Complexity~\citep{gebejes2013texture}, and Fractal Dimension~\citep{lin1991divergence}.
We then compute Pearson correlation coefficients~\citep{pearson1896vii} between these attributes and the local metrics \metricone, \metrictwo, and \metricthree.

\begin{figure*}[t]
\centering
\includegraphics[width=\linewidth]{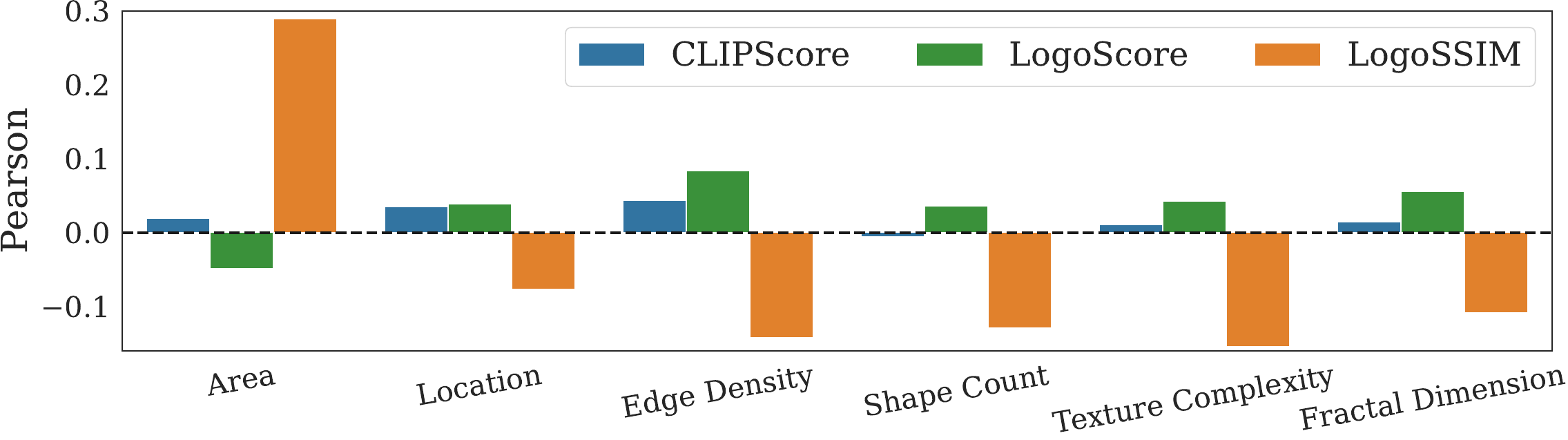}
\caption{
\textbf{Diagnostic correlation between visual attributes and SEGA~\citep{brack2023sega} unlearning efficacy.}
We compute Pearson correlations across intrinsic geometric and textural properties (\eg, area, texture complexity). While most factors exhibit negligible correlation, \metricthree (local SSIM) reveals a notable relationship. This suggests that incorporating structural heuristics (\eg, SSIM-guided spatial controls) is vital for future latent-space unlearning architectures.
}
\label{fig:pearson}
\end{figure*}

\Cref{fig:pearson} shows that most attributes have weak correlations with semantic scores, which is reasonable because logo leakage depends on both visual form and prompt context.
The clearest signal appears in \metricthree: larger logo regions tend to be harder to structurally alter without visible changes, while high-frequency or visually complex regions can be easier to disrupt.
These findings are not a causal explanation of all failures, but they support the case for explicit spatial control, such as SSIM-guided constraints that target logo regions while protecting surrounding content.

\subsection{Ablation of the Exploratory Baseline}

We ablate \ourBaseline on \ourBenchHard to clarify the role of each agent.
\Cref{tab:ablation} shows that the Remover is the main contributor: without it, the \metricone reduction drops from 16.76\% to 3.07\%.
The Reflector provides additional stabilization by revising prompts that are either under-edited or over-edited.
The Checker has little direct effect on the aggregate metric in this experiment, but it provides a final textual guard against obvious residual logo references.
Overall, the ablation confirms that \ourBaseline's gains mainly come from semantic rewriting, reinforcing its role as a diagnostic prompt-space baseline rather than a weight-level unlearning method.

\section{Discussion}
\label{sec:con}

\paragraph{Conclusion} 
We introduced \ourBench, a benchmark of nearly 10,000 text-image pairs for studying logo unlearning on Fortune Global 500 companies.
The benchmark frames logos as localized but semantically entangled concepts: removing them requires changing a small protected mark while preserving the rest of the generated scene.
Across the evaluated inference-time and compatible fine-tuning-based methods, our multi-grained metrics reveal a persistent local-global trade-off that single success rates would obscure.
\ourBaseline further shows that prompt-space rewriting can reduce logo triggers, but also clarifies that prompt filtering is not the same as localized model-level unlearning.

\paragraph{Limitations and Future Work} 
Our empirical conclusions are scoped to the evaluated methods, compatible backbones, and Fortune Global 500 coverage.
\ourBench does not yet cover long-tail regional brands, multilingual logo variants, or full adaptive attack protocols, and the fine-tuning comparisons remain constrained by each method's official implementation.
Future work should extend the benchmark along these axes, add stronger and more diverse weight-editing baselines where compatible, and report broader variance analyses.

\newpage

\bibliographystyle{unsrtnat}
\bibliography{main}


\newpage



\appendix

\section{Comprehensive Evaluation Metrics}

We establish a comprehensive, multi-grained evaluation framework comprising five distinct metrics, each with a stringently defined scope to capture the complex trade-offs between local concept erasure and global image preservation, as outlined in \Cref{tab:metric_category}. 

\metricone calculates the CLIP similarity between the textual company name query and the isolated logo region extracted from the generated image. A lower score indicates a highly effective semantic decoupling, demonstrating that the generated logo no longer aligns with the targeted brand identity.

\metrictwo evaluates the latent CLIP feature similarity between the localized logo regions extracted before and after the unlearning intervention. A lower value implies a profound feature-level disruption, marking a successful localized unlearning process.

\metricthree utilizes the Structural Similarity Index Measure (SSIM) to quantify the pixel-level structural alterations within the localized logo bounding box before and after unlearning. A diminished SSIM value confirms that the geometric and textural integrity of the logo has been thoroughly dismantled.

\metricfour computes the global CLIP similarity across the entire image composition before and after the unlearning process. A higher score is imperative here; it serves as a critical counterbalance, signifying that the broad contextual semantics and background elements remain fundamentally undisturbed by the localized unlearning.

\metricfive measures the global SSIM for the complete visual scene before and after unlearning. A higher value dictates rigorous structural preservation of the non-target background, which is the hallmark of a non-destructive, highly precise unlearning methodology.

\begin{figure*}[ht]
    \centering
    \includegraphics[width=0.8\linewidth]{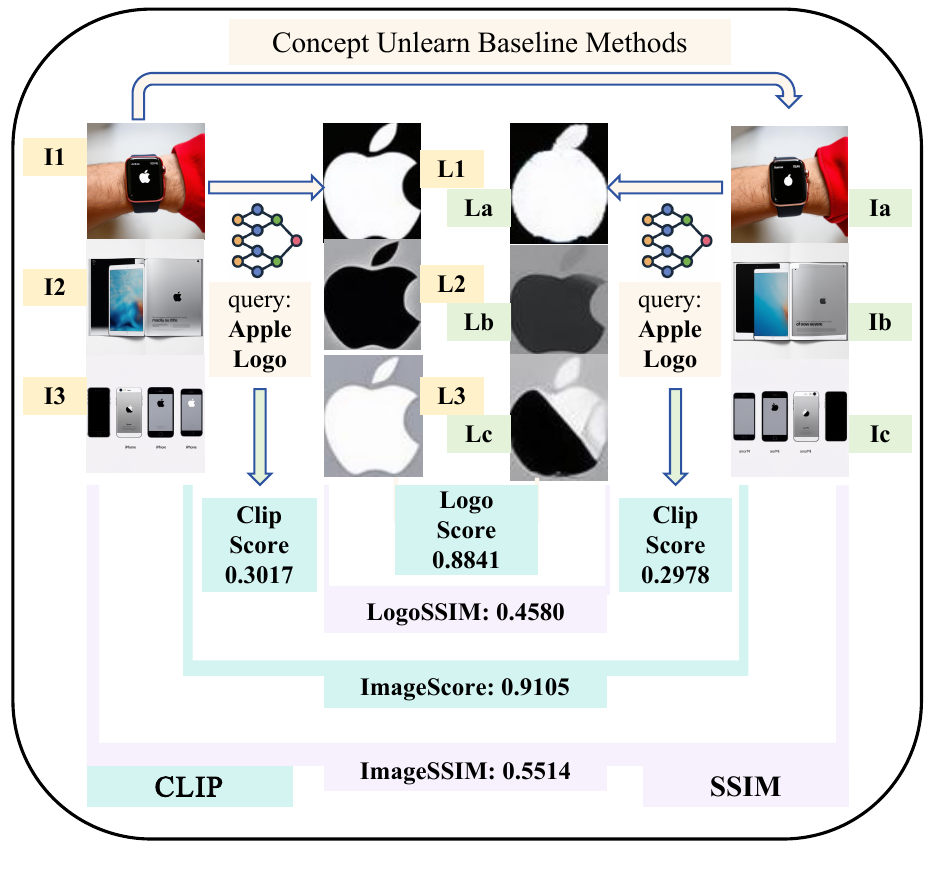}
    \caption{Original images and their unlearned counterparts are analyzed using metrics focused on logos and overall image similarity. \metricone calculates the CLIP similarity between detected logos and the text query "apple logo." \metrictwo measures the CLIP similarity between logos before and after unlearning. \metricthree evaluates the SSIM similarity between logos before and after unlearning. \metricfour assesses the CLIP similarity for the entire image before and after unlearning, while \metricfive evaluates the SSIM similarity for the entire image. Light purple represents the SSIM score, while light blue represents the CLIP score.}
    \label{fig:evaluation}
\end{figure*}

These metrics constitute the rigorous backbone of our evaluation methodology, providing a mathematically structured approach to assessing unlearning performance. Furthermore, a comprehensive visual schema of our evaluation pipeline is depicted in \Cref{fig:evaluation} to ensure absolute methodological transparency. 

Formally, let $I_1, I_2$, and $I_3$ denote the original synthesized images, while their corresponding counterparts post-unlearning are denoted as $I_a, I_b$, and $I_c$. Following the application of our open-vocabulary logo detection, the localized bounding box regions yielding the highest confidence are extracted and denoted as $L_1, L_2, L_3$, and $L_a, L_b, L_c$ respectively. \metricone calculates the CLIP cosine similarity between the localized sets $\{L_1, L_2, L_3, L_a, L_b, L_c\}$ and the explicit text query (e.g., "apple logo"). \metrictwo computes the pairwise CLIP feature similarity between the pre- and post-unlearning local regions: $(L_1, L_a), (L_2, L_b)$, and $(L_3, L_c)$. \metricthree measures the pixel-wise SSIM over the same localized pairs: $(L_1, L_a), (L_2, L_b)$, and $(L_3, L_c)$. To capture global fidelity, \metricfour calculates the holistic CLIP similarity between the full image pairs: $(I_1, I_a), (I_2, I_b)$, and $(I_3, I_c)$. Finally, \metricfive computes the global SSIM similarity for $(I_1, I_a), (I_2, I_b)$, and $(I_3, I_c)$.

The algorithmic formulation for this evaluation framework is meticulously detailed in \Cref{alg:metric}. Both CLIP and SSIM metrics are initially computed per individual prompt by contrasting the pre- and post-intervention states, followed by rigorous aggregation at both the prompt and company levels to ensure statistical robustness.

\section{Implementation Details for Fine-Tuning Baselines}

\textbf{Forget-Me-Not}~
To guarantee a fair evaluation environment, we strictly adhere to the authors' original architectural constraints, utilizing the "stabilityai/stable-diffusion-2-1-base" text-to-image model. During the baseline generation phase, the inference parameters are fixed: num\_inference\_steps is set to 50, guidance\_scale to 7, and num\_images\_per\_prompt to 1. For the fine-tuning phase, the hyperparameter configuration includes a train\_batch\_size of 1, a learning\_rate of 2.0e-06, and a max\_train\_steps limit of 35. We employ the AdamW optimizer with adam\_beta1 at 0.9, adam\_beta2 at 0.999, adam\_weight\_decay of 0.01, adam\_epsilon of 1.0e-08, and a max\_grad\_norm bounded at 1.

\textbf{ESD}~
In strict accordance with the foundational ESD framework, we deploy the "CompVis/stable-diffusion-v1-4" model. For the image synthesis phase, structural parameters are set to img\_size of 512, n\_steps of 50, n\_imgs of 1, and a guidance\_scale of 7.5. Throughout the fine-tuning phase, we apply the targeted cross-attention (xattn) modification methodology coupled with a learning\_rate of 1e-5.

\section{Empirical Ranking of Logo Entanglement}
\label{sec:metric_rank}
Beyond benchmarking unlearning algorithms, a pivotal secondary contribution of \ourBench is its capacity to empirically quantify and rank the deep-seated entanglement—or unlearning difficulty—of specific corporate logos within the model's latent space. We formulate this difficulty ranking across the Fortune 500 companies by analyzing the localized disruption metric (\metrictwo) under the adversarial conditions of \ourBenchHard utilizing our exploratory \ourBaseline.

Given that a minimized \metrictwo score denotes optimal semantic disruption, an elevated \metrictwo score indicates profound resistance to unlearning (i.e., extreme latent entanglement). Sorted in descending order of unlearning resistance, the five most intractably entangled company logos are \textit{HYUNDAI MOTOR}, \textit{CHINA AEROSPACE SCIENCE \& INDUSTRY}, \textit{PANASONIC HOLDINGS}, \textit{RTX}, and \textit{TATA MOTORS}, registering extreme \metrictwo retention values of $0.834, 0.8299, 0.8142, 0.8122$, and $0.8051$ respectively. Conversely, the five most highly malleable logos, demonstrating the least resistance to semantic erasure, are \textit{ORANGE}, \textit{WELLS FARGO}, \textit{AMAZON.COM}, \textit{TARGET}, and \textit{WORLD KINECT}, yielding disrupted \metrictwo scores of $0.0907, 0.1838, 0.2119, 0.2159$, and $0.2206$.

\begin{algorithm}
\caption{The CLIP and SSIM similarities are first calculated for each individual prompt by comparing the images before and after unlearning. Then, the results are averaged at both the prompt level and the company level.}
\label{alg:metric}
\begin{algorithmic}[1]
\REQUIRE Datasets $D_i$, $i=1,2$; Unlearning methods $M_{il}$; Companies $C_{ilj}$; Prompts $P_{iljk}$
\ENSURE Metrics $ME_{il}$ for evaluating unlearning methods

\FOR{each dataset $D_i$}
    \FOR{each unlearning method $M_{il}$}
        \FOR{each company $C_{ilj}$, $j = 1$ to $500$}
            \FOR{each prompt $P_{iljk}$, $k = 1$ to $10$}
                \STATE \COMMENT{Generate images before and after unlearning}
                \STATE Generate image $I\_ori_{iljk}$ using Stable Diffusion before unlearning
                \STATE Apply unlearning method $M_{il}$ to generate $I\_un_{iljk}$
                \STATE \COMMENT{Metric 1: Text-Logo Alignment (Local View)}
                \STATE Use OWLv2 with $C_{ilj}$ logo as text query on $I\_ori_{iljk}$ and $I\_un_{iljk}$
                \STATE Extract top confidence scores corresponding boxes containing logos as $L\_ori_{iljk}$ and $L\_un_{iljk}$
                \STATE Extract CLIP logo features $F\_L\_ori_{iljk}$ and $F\_L\_un_{iljk}$
                \STATE Extract CLIP text features $T_{ilj}$ with text query $C_{ilj}$ logo
                \STATE Compute cosine similarity $s_0, s_1$ between $F\_L\_ori_{iljk}$, $F\_L\_un_{iljk}$ and $T_{ilj}$
                \STATE $ME1_{iljk} = s_0 \text{ before unlearn or } = s_1 \text{ after unlearn}$
                \STATE \COMMENT{Metric 2: Logo-Logo Alignment (Local View)}
                \STATE Extract CLIP features $F\_L\_ori_{iljk}$ and $F\_L\_un_{iljk}$
                \STATE Compute cosine similarity $ME2_{iljk}$ between $F\_L\_ori_{iljk}$ and $F\_L\_un_{iljk}$
                \STATE \COMMENT{Metric 3: Logo-Logo Alignment (Local View)}
                \STATE Compute SSIM score $ME3_{iljk}$ between $L\_ori_{iljk}$ and $L\_un_{iljk}$
                \STATE \COMMENT{Metric 4: Image-Image Alignment (Global View)}
                \STATE Extract CLIP features $F\_I\_ori_{iljk}$ and $F\_I\_un_{iljk}$
                \STATE Compute cosine similarity $ME4_{iljk}$ between $F\_I\_ori_{iljk}$ and $F\_I\_un_{iljk}$
                \STATE \COMMENT{Metric 5: Image-Image Alignment (Global View)}
                \STATE Compute SSIM score $ME5_{iljk}$ between $I\_ori_{iljk}$ and $I\_un_{iljk}$
            \ENDFOR
            \STATE \COMMENT{Average metrics over $k$}
            \STATE Compute $ME1_{ilj} = \frac{1}{10} \sum_{k=1}^{10} ME1_{iljk}$
            \STATE Compute $ME2_{ilj} = \frac{1}{10} \sum_{k=1}^{10} ME2_{iljk}$
            \STATE Compute $ME3_{ilj} = \frac{1}{10} \sum_{k=1}^{10} ME3_{iljk}$
            \STATE Compute $ME4_{ilj} = \frac{1}{10} \sum_{k=1}^{10} ME4_{iljk}$
            \STATE Compute $ME5_{ilj} = \frac{1}{10} \sum_{k=1}^{10} ME5_{iljk}$
        \ENDFOR
        \STATE \COMMENT{Average metrics over $j$}
        \STATE Compute $ME1_{il} = \frac{1}{500} \sum_{j=1}^{500} ME1_{ilj}$
        \STATE Compute $ME2_{il} = \frac{1}{500} \sum_{j=1}^{500} ME2_{ilj}$
        \STATE Compute $ME3_{il} = \frac{1}{500} \sum_{j=1}^{500} ME3_{ilj}$
        \STATE Compute $ME4_{il} = \frac{1}{500} \sum_{j=1}^{500} ME4_{ilj}$
        \STATE Compute $ME5_{il} = \frac{1}{500} \sum_{j=1}^{500} ME5_{ilj}$
    \ENDFOR
\ENDFOR
\RETURN{$ME_{il}$ for all methods $M_{il}$}
\end{algorithmic}
\end{algorithm}

\begin{table*}[h]
    \centering
        \caption{\metricone, \metrictwo and \metricthree focus on the perspective of logos extracted from local regions, with \metricone considering the relationship between text and image, and \metrictwo and \metricthree focusing on the relationship between images. \metricfour and \metricfive, on the other hand, evaluate the overall background. \metricone, \metrictwo and \metricfour calculate CLIP similarity, while \metricthree and \metricfive measure SSIM.}
    \begin{tabular}{ccccccc}
    \toprule
         Metric Name& Text-Image & Image-Image & CLIP &  SSIM& Local & Global\\
         \midrule
        \metricone & \checkmark &  & \checkmark &  & \checkmark & \\
        \midrule
        \metrictwo &  & \checkmark & \checkmark &  & \checkmark & \\
        \midrule
        \metricthree &  & \checkmark &  & \checkmark & \checkmark & \\
        \midrule
        \metricfour &  & \checkmark & \checkmark &  &  & \checkmark\\
        \midrule
        \metricfive &  & \checkmark &  & \checkmark &  & \checkmark\\ \bottomrule
    \end{tabular}

    \label{tab:metric_category}
\end{table*}

\section{Prompt Curation and Multi-Agent Pipeline}

\begin{figure*}[ht]
    \centering
    \includegraphics[width=1.0\linewidth]{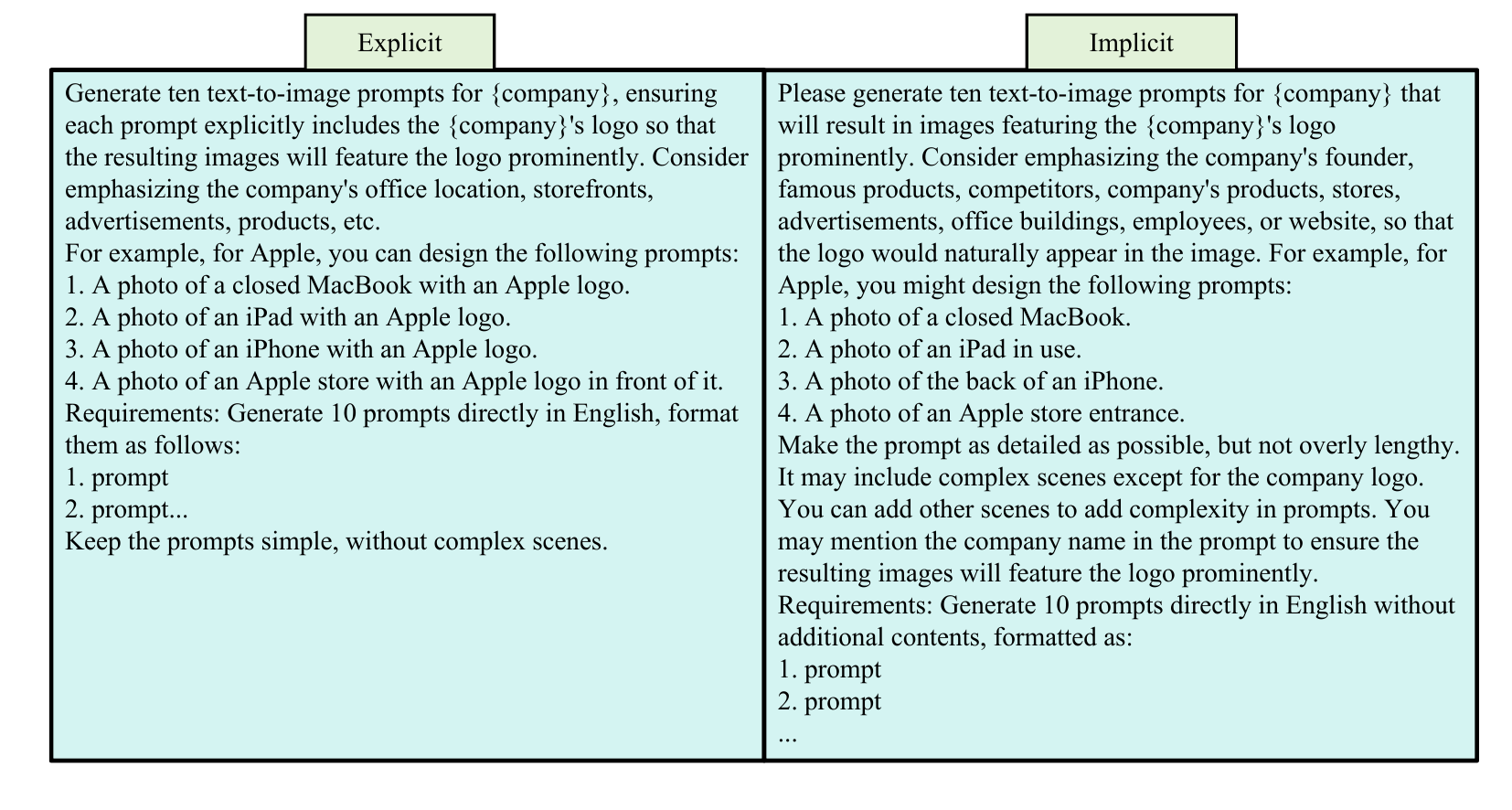}
    \caption{The agent prompts for generating \ourBench was crafted using OpenAI's GPT-4o model.}
    \label{fig:agent_prompts}
\end{figure*}

\begin{figure*}[ht]
    \centering
    \includegraphics[width=1.0\linewidth]{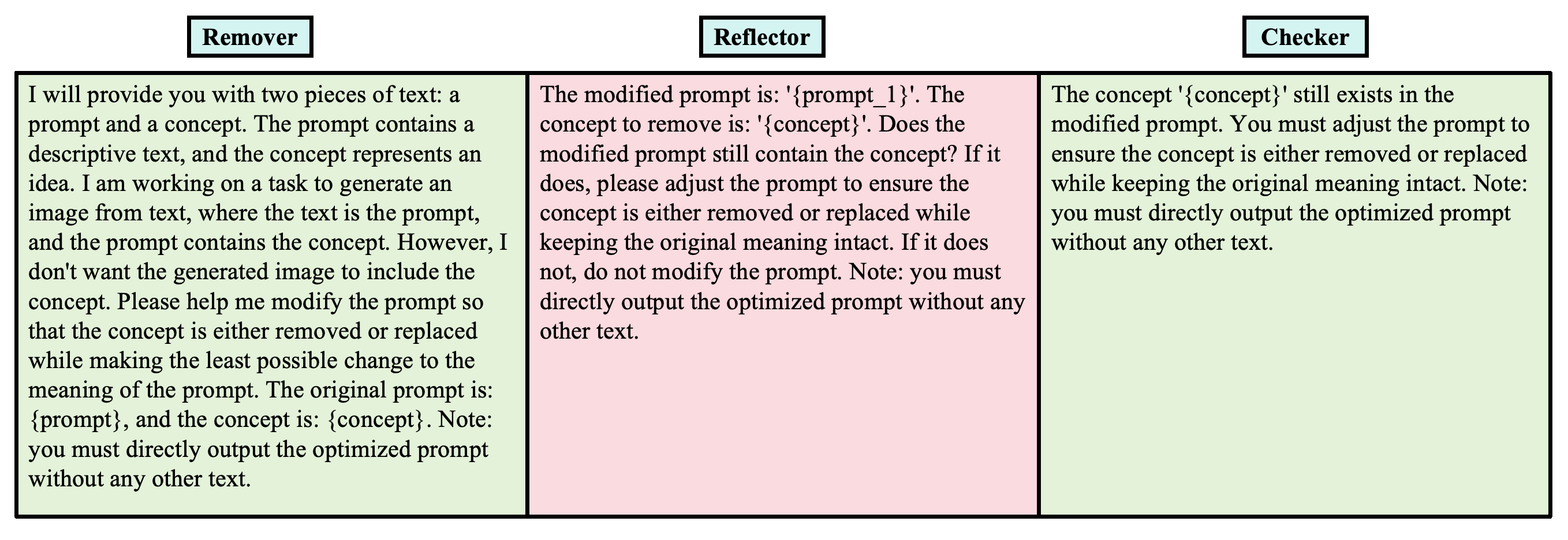}
    \caption{The Remover, Reflector, and Checker prompts in \ourBaseline. The Remover is used to eliminate elements related to the company logo from the original prompt while keeping other parts as consistent as possible. The Reflector evaluates whether the Remover has successfully completed its task and provides further optimized prompts. The Checker performs a final review to ensure that the final prompt does not contain any company logo; if any logo-related elements remain, they are directly removed.}
    \label{fig:ours_prompts}
\end{figure*}

To ensure absolute methodological transparency in the construction of \ourBench, we disclose the exact meta-prompts utilized to synthesize the dataset. Capitalizing on the advanced reasoning capabilities of state-of-the-art language models, these were generated leveraging OpenAI's GPT-5, as thoroughly documented in \Cref{fig:agent_prompts}. Furthermore, we transparently detail the architectural prompt logic underpinning our exploratory multi-agent baseline (\ourBaseline), which orchestrates the collaborative functionalities of the Remover, Reflector, and Checker agents, visually formalized in \Cref{fig:ours_prompts}. 

Within this semantic intervention pipeline: The \textit{Remover} acts as the primary semantic ablator, engineered to surgically excise any latent textual triggers linked to the target logo while meticulously preserving the surrounding contextual phrasing. The \textit{Reflector} functions as a semantic stabilizer; it critically evaluates the Remover's output against the original user intent, iteratively optimizing the prompt to correct over-edits or hallucinated contexts. The \textit{Checker} provides a deterministic final safeguard, executing a rigorous review to certify the absolute absence of implicit or explicit brand references. Should residual triggers be detected, the prompt is immediately cycled back for aggressive, direct ablation.


\section{Extended Qualitative Evaluation}

\begin{figure*}[ht]
    \centering
    \renewcommand{\arraystretch}{1.2} 
    \setlength{\tabcolsep}{0pt} 
    \begin{tabular}{ccccccccc}

    \multicolumn{1}{c}{\tiny \textbf{Before}} &
        \multicolumn{1}{c}{\tiny \textbf{NP}} &
        \multicolumn{1}{c}{\tiny \textbf{\sldzero}} &
        \multicolumn{1}{c}{\tiny \textbf{\sldhalf}} &
        \multicolumn{1}{c}{\tiny \textbf{\sldone}} &
        \multicolumn{1}{c}{\tiny \textbf{\segazero}} &
        \multicolumn{1}{c}{\tiny \textbf{\segahalf}} &
        \multicolumn{1}{c}{\tiny \textbf{\segaone}} &
        \multicolumn{1}{c}{\tiny \textbf{\ourBaseline}} \\

        \includegraphics[width=0.1\textwidth]{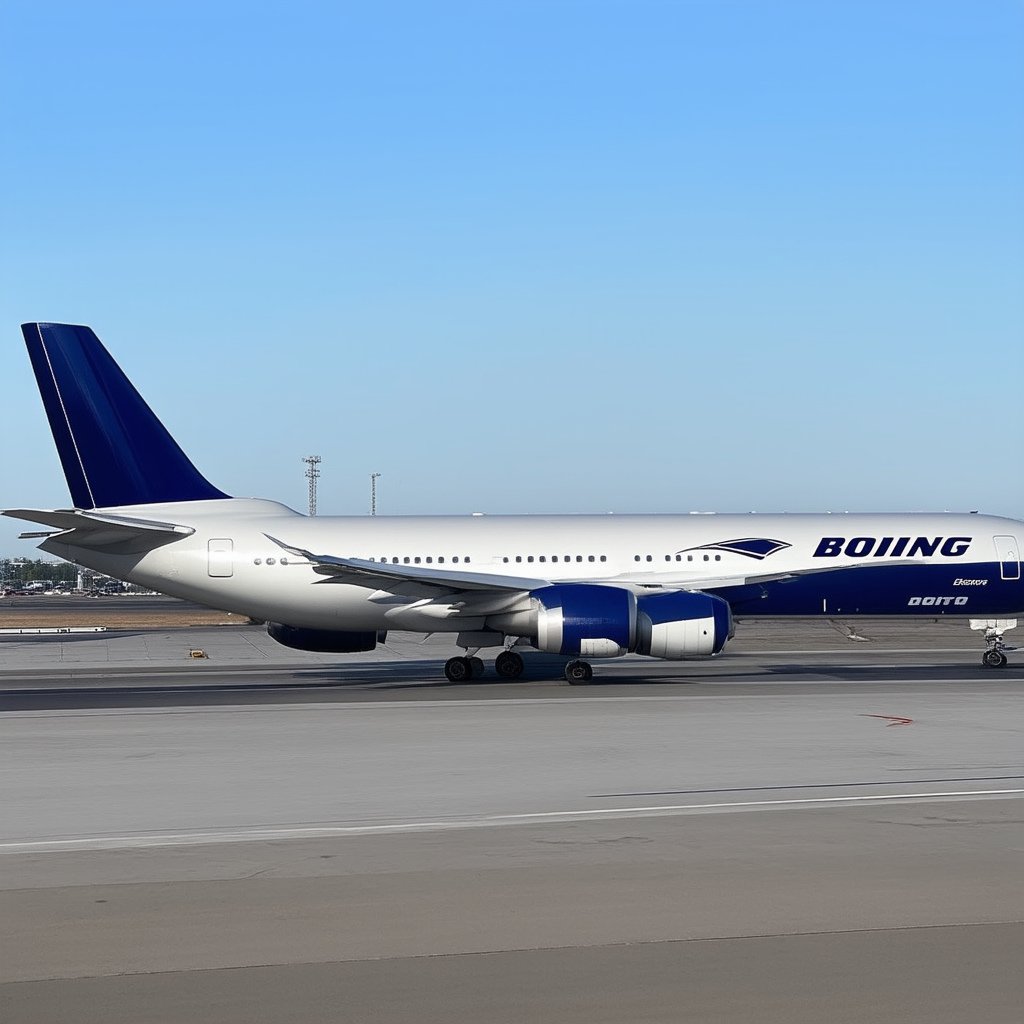} &
        \includegraphics[width=0.1\textwidth]{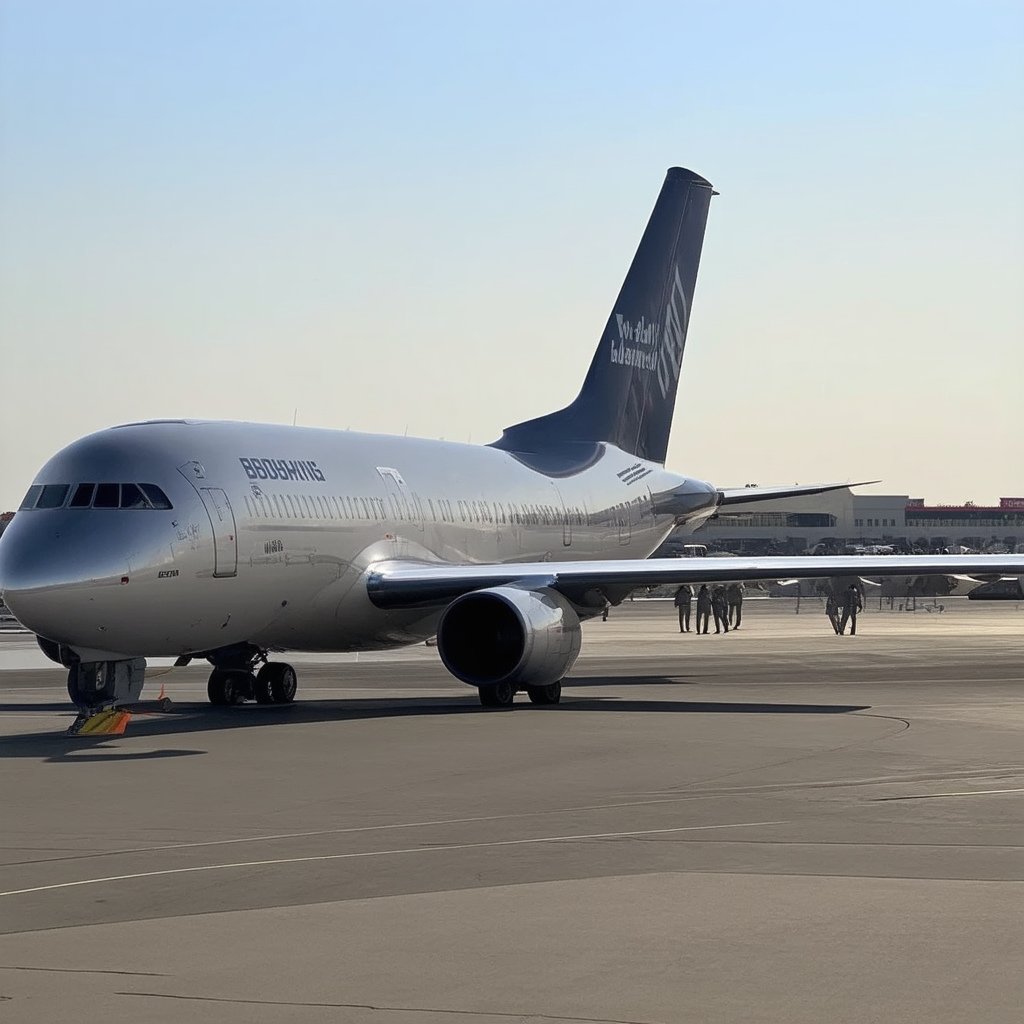} &
        \includegraphics[width=0.1\textwidth]{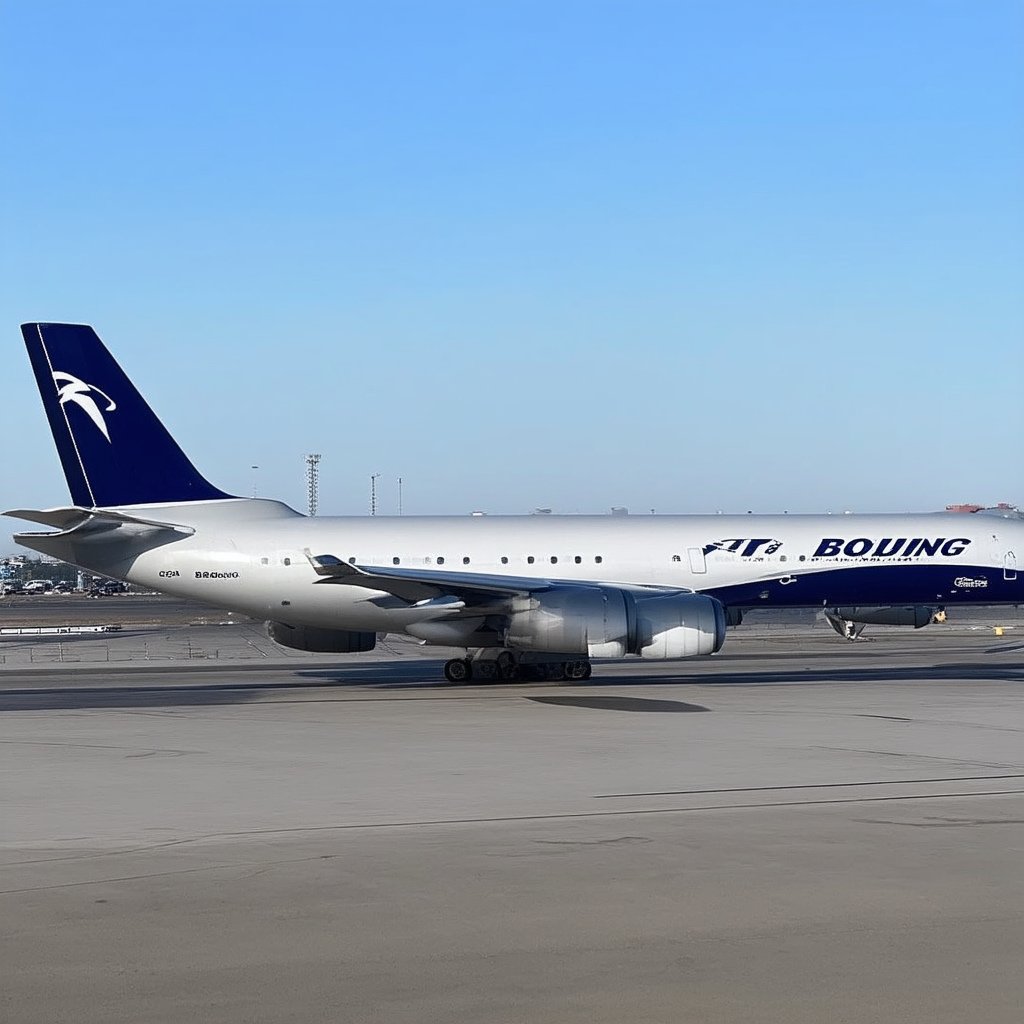} &
                \includegraphics[width=0.1\textwidth]{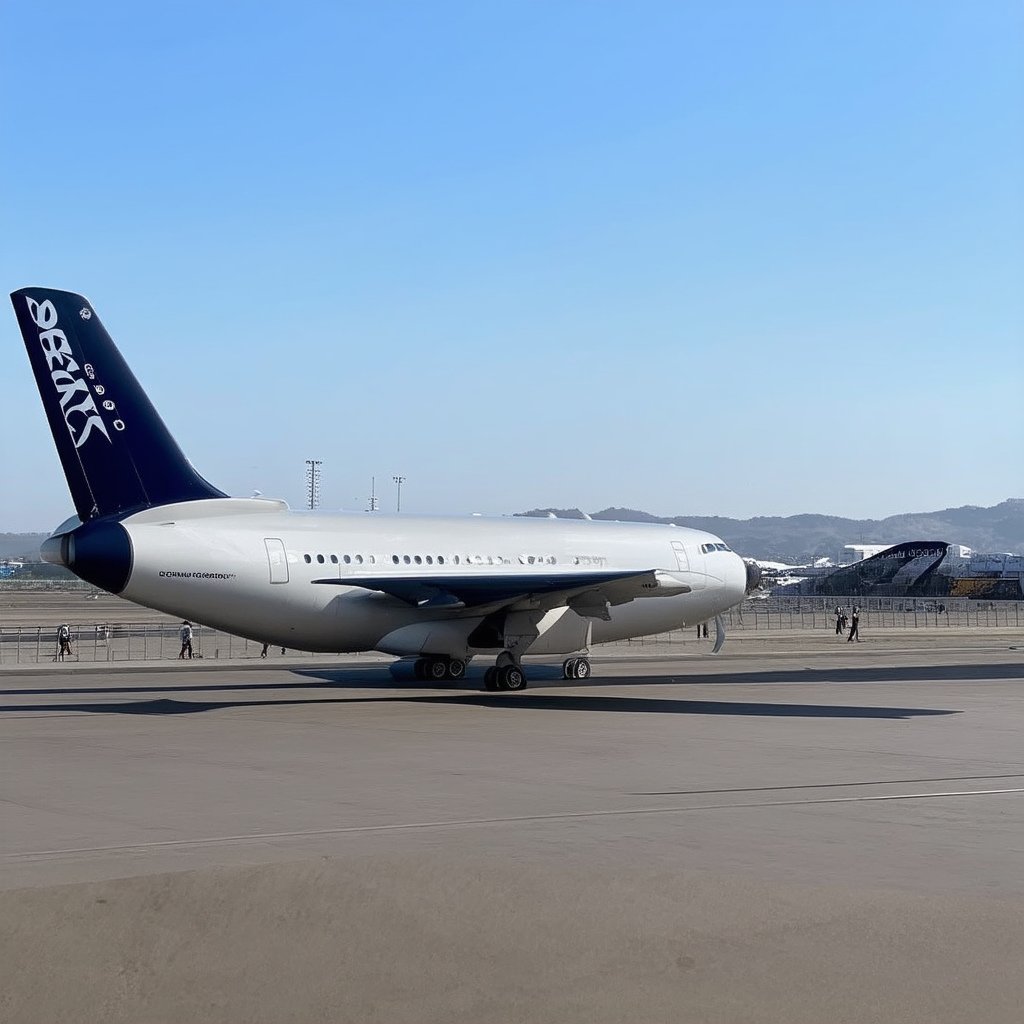} &
                        \includegraphics[width=0.1\textwidth]{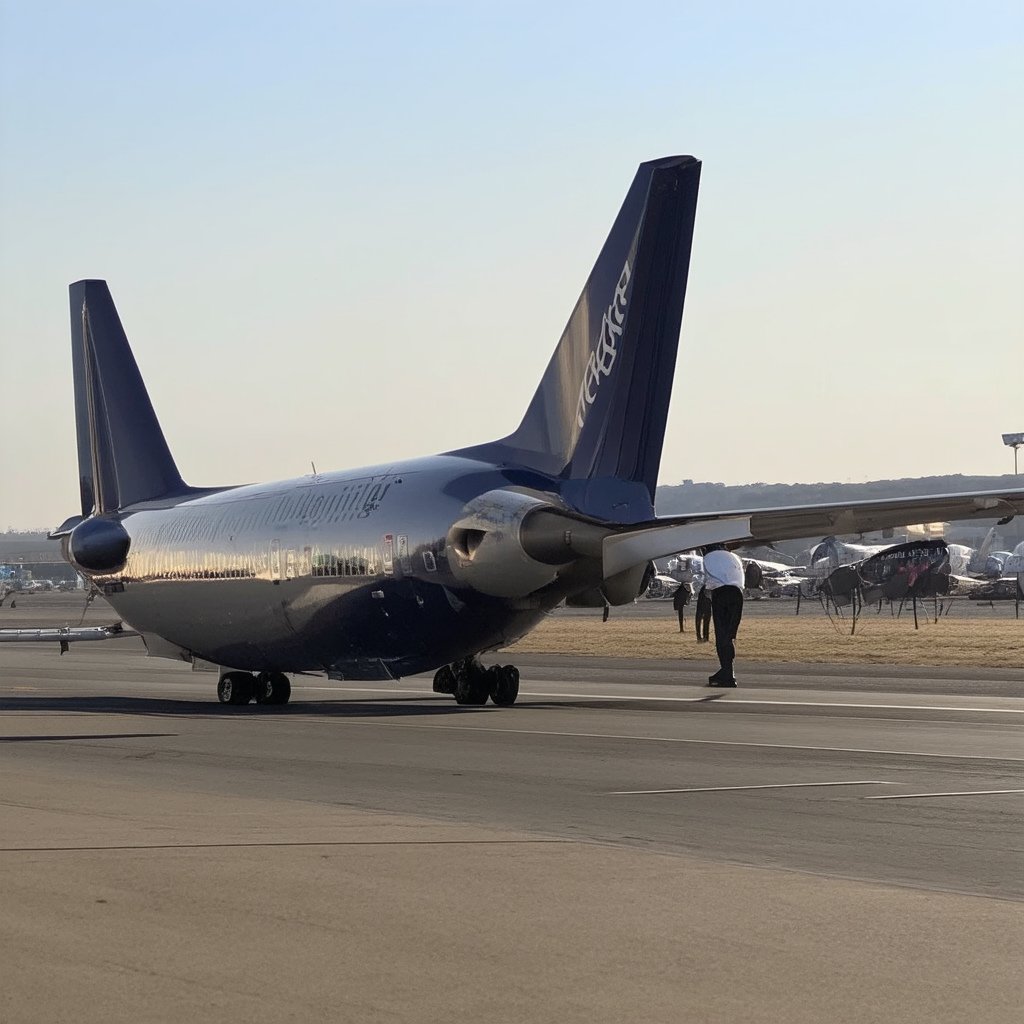} &
        \includegraphics[width=0.1\textwidth]{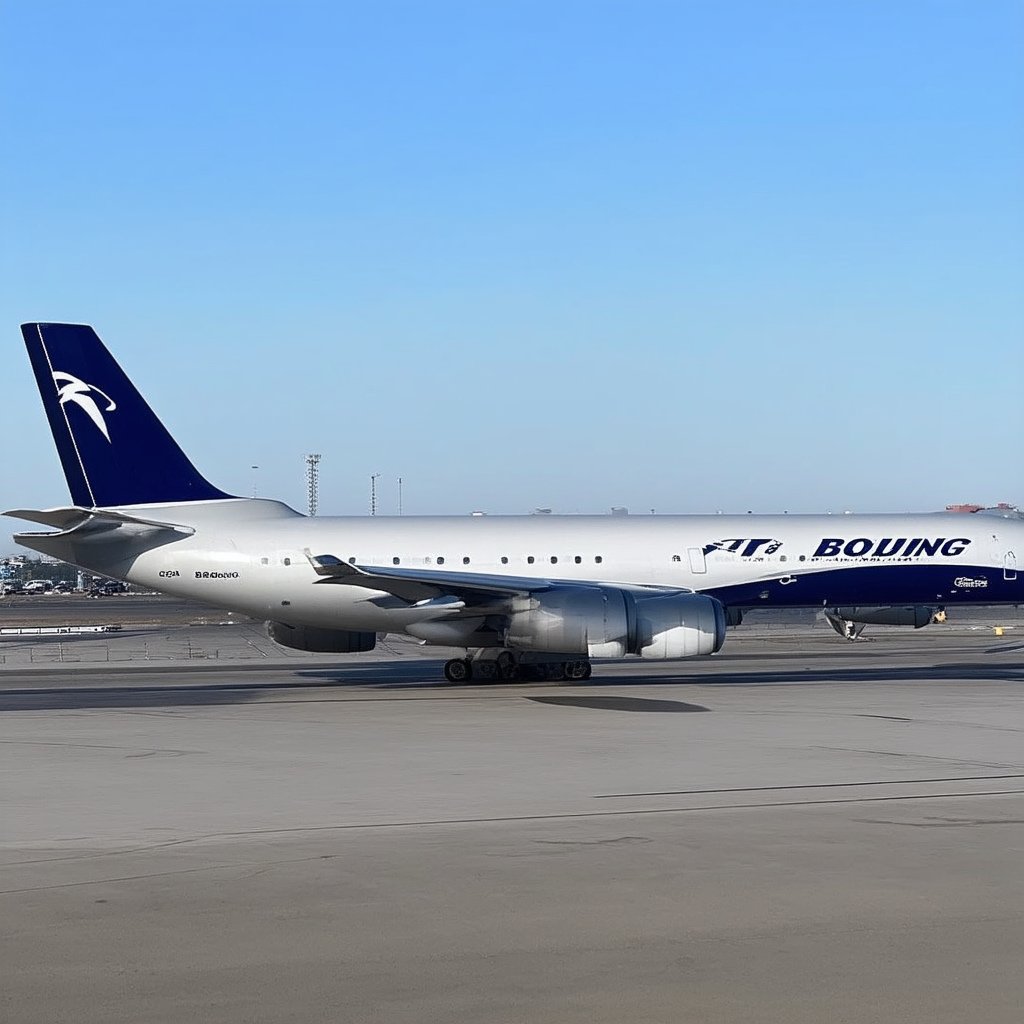} &
                \includegraphics[width=0.1\textwidth]{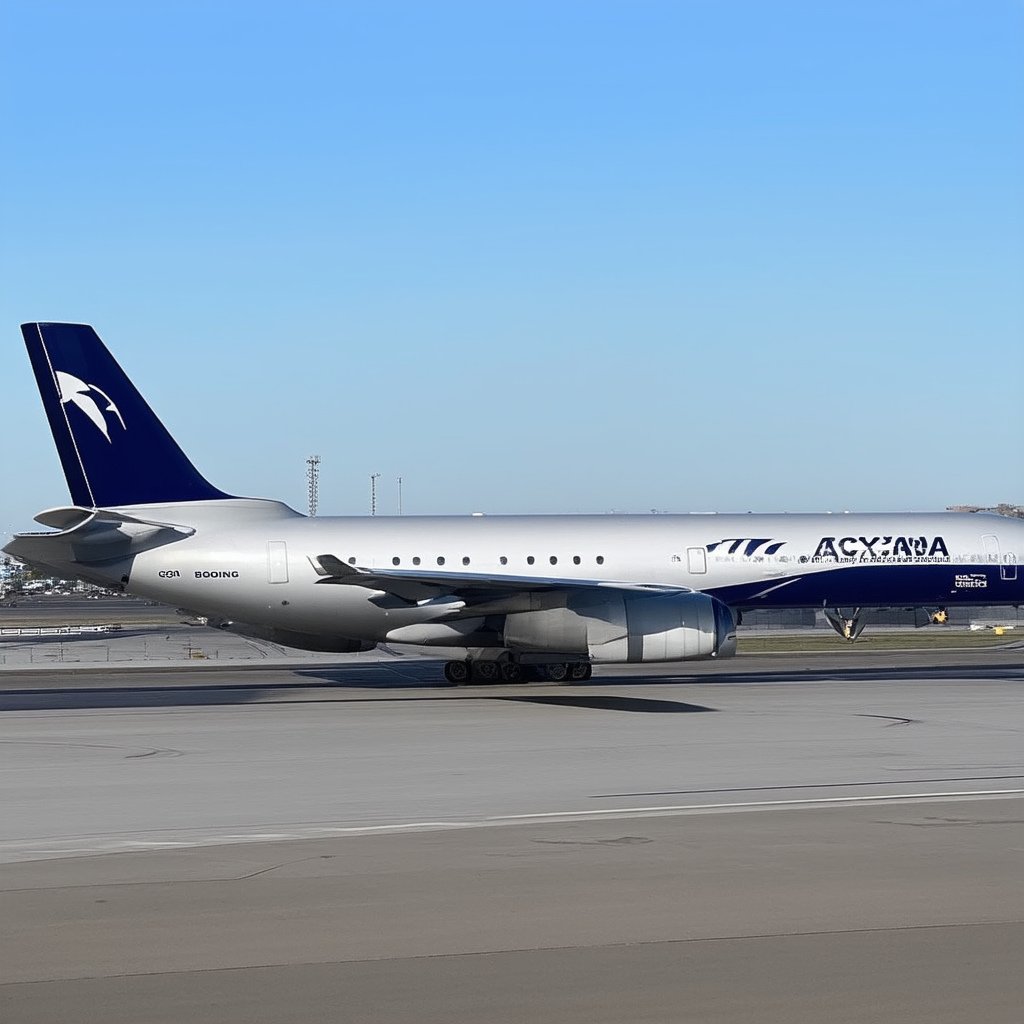} &
                        \includegraphics[width=0.1\textwidth]{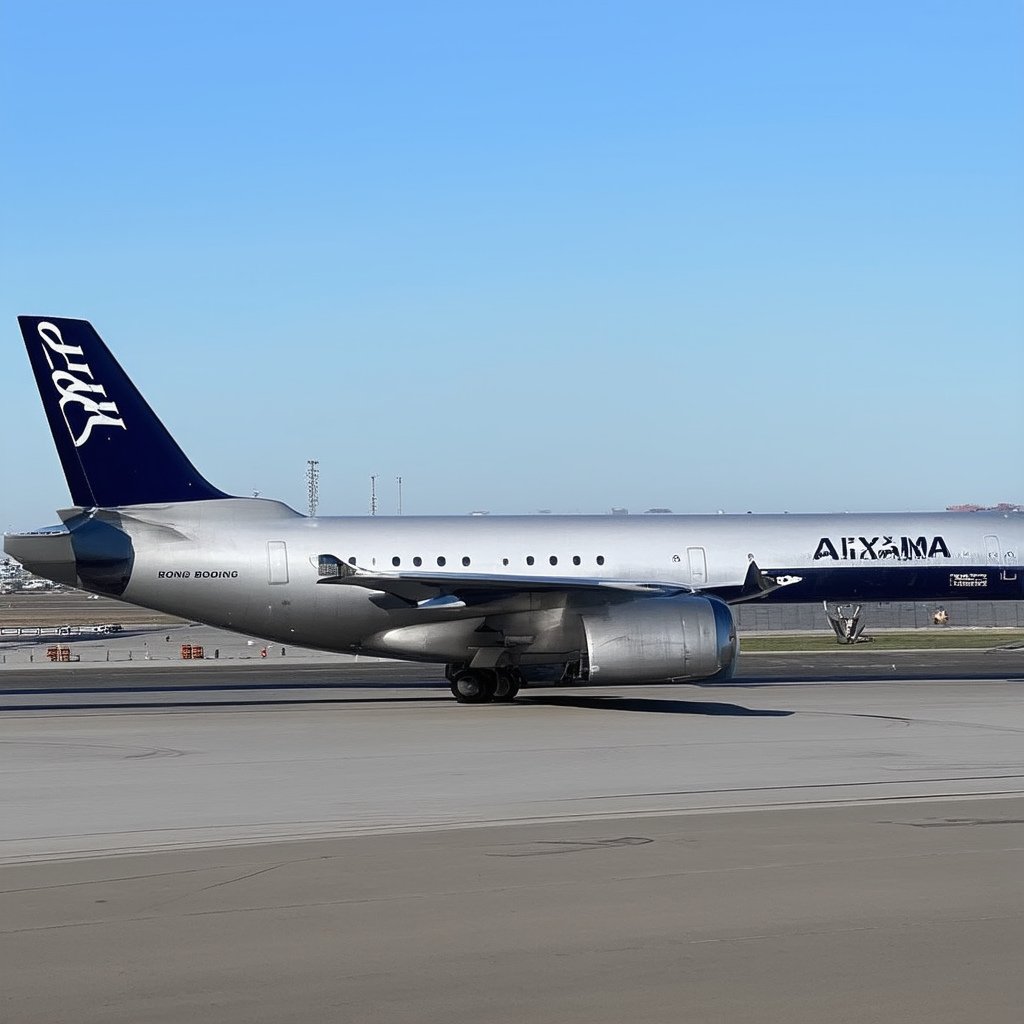} &
        \includegraphics[width=0.1\textwidth]{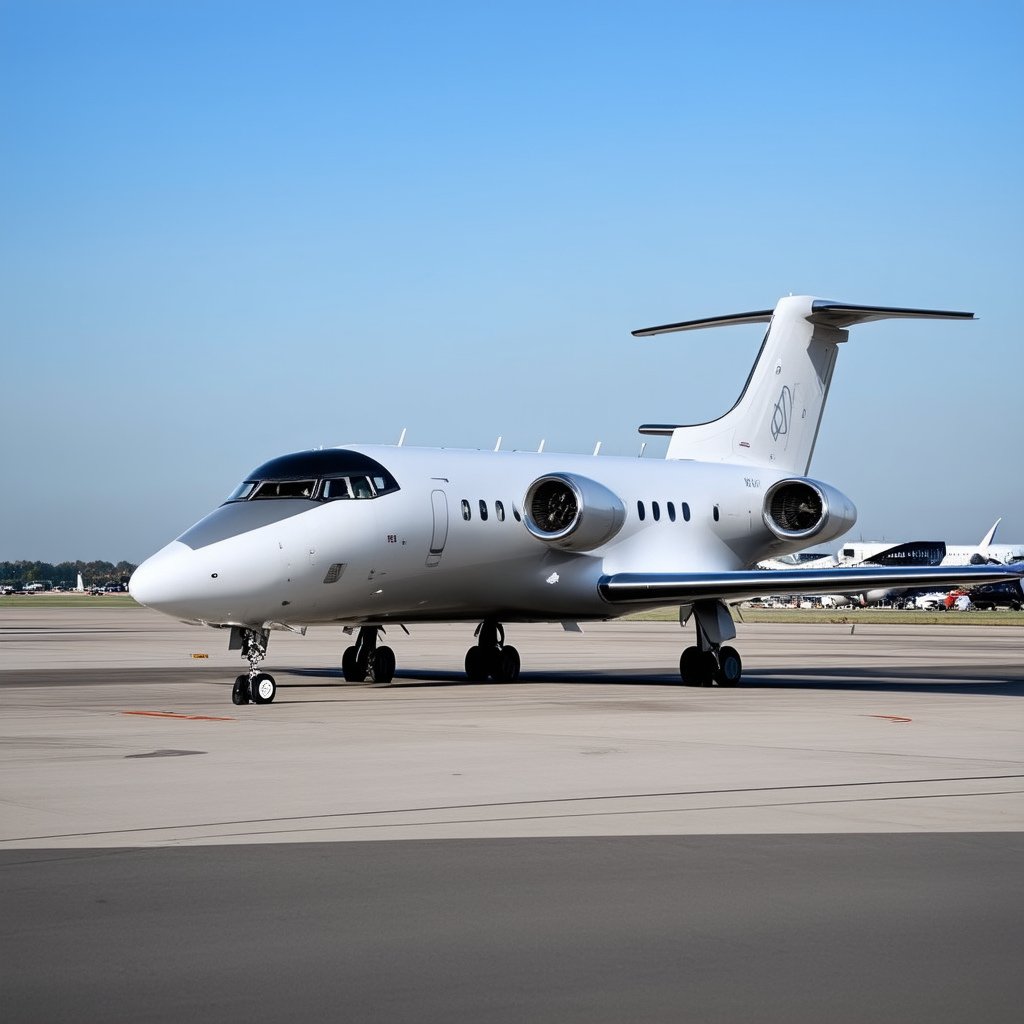} \\
    
        \includegraphics[width=0.1\textwidth]{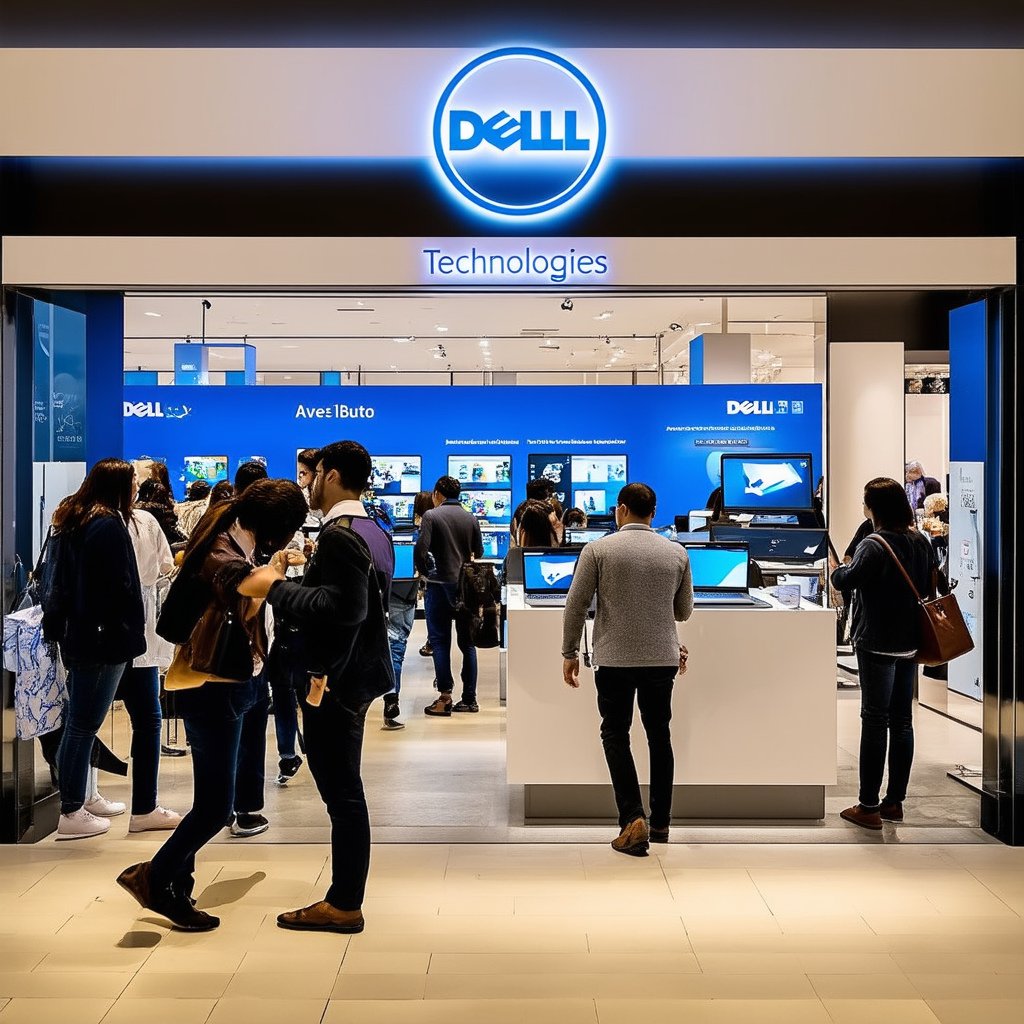} &
        \includegraphics[width=0.1\textwidth]{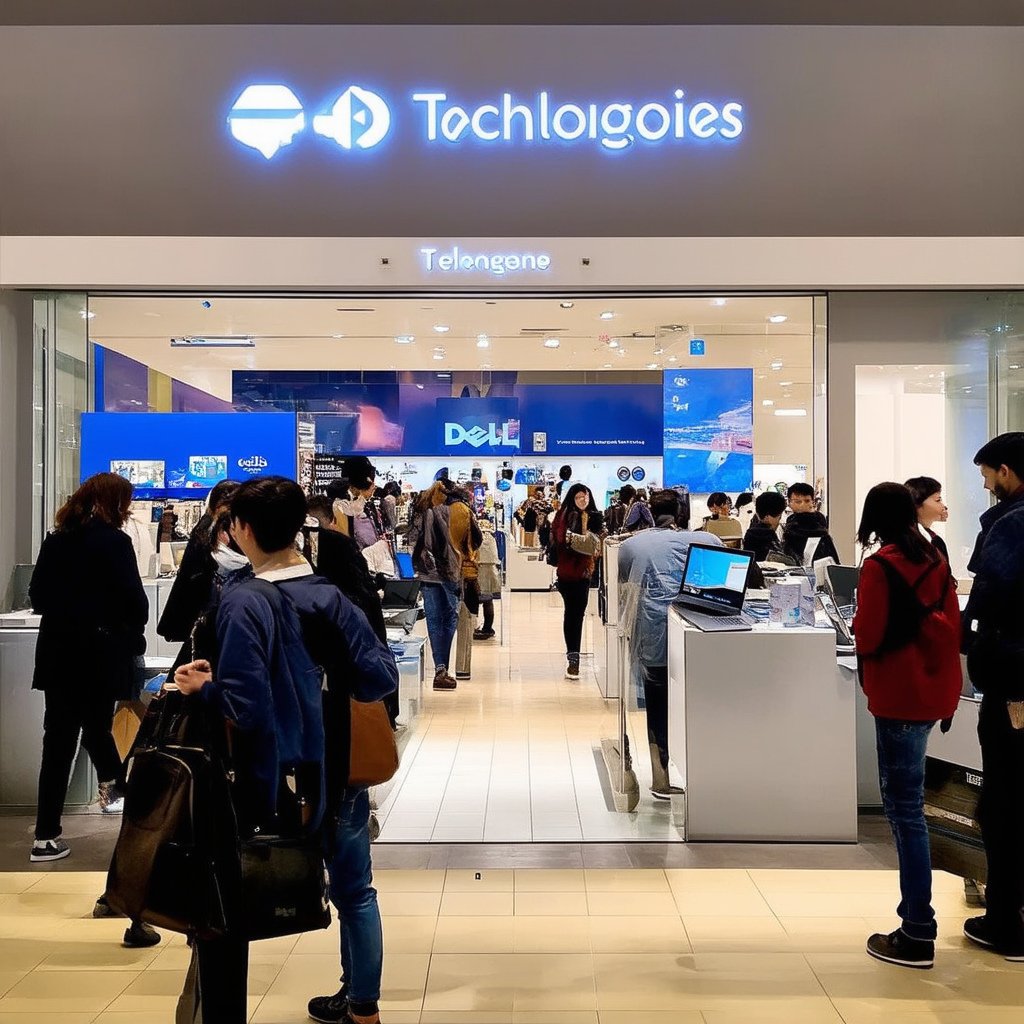} &
        \includegraphics[width=0.1\textwidth]{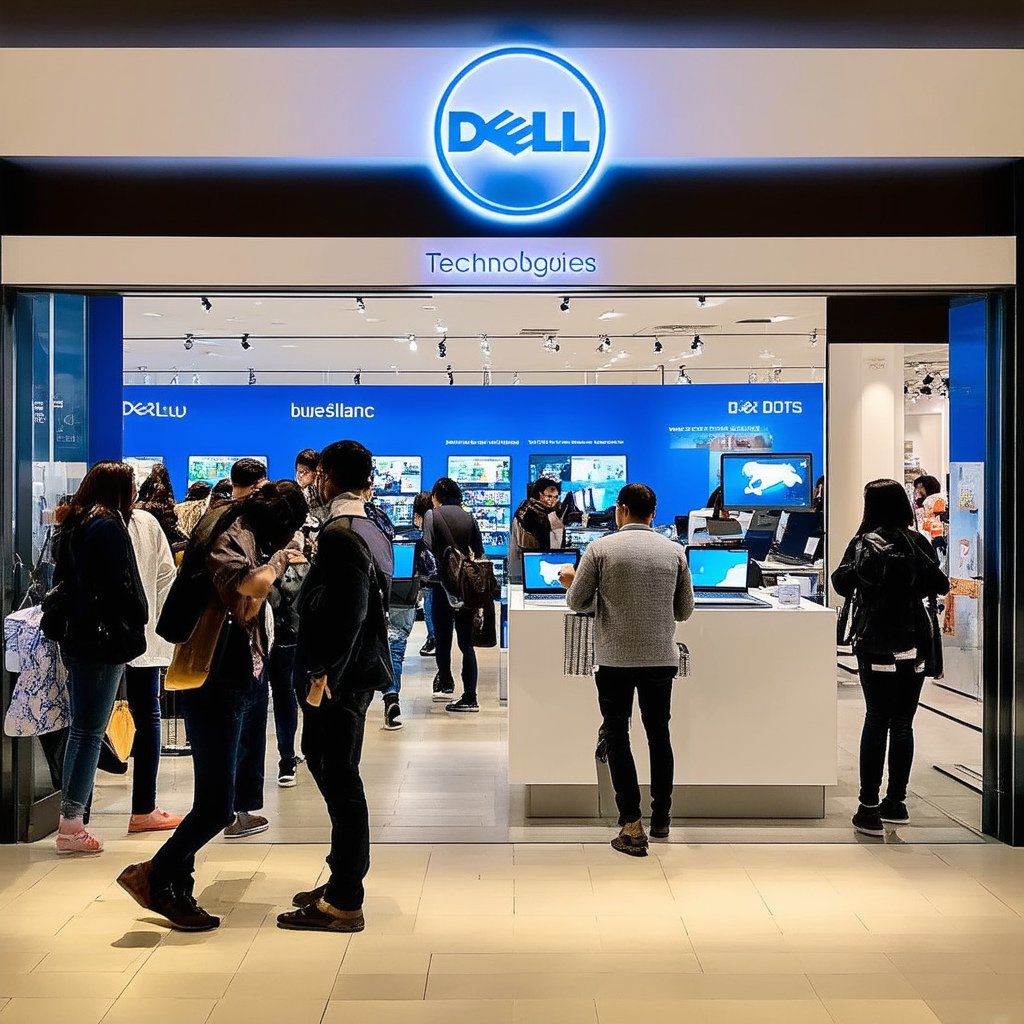} &
                \includegraphics[width=0.1\textwidth]{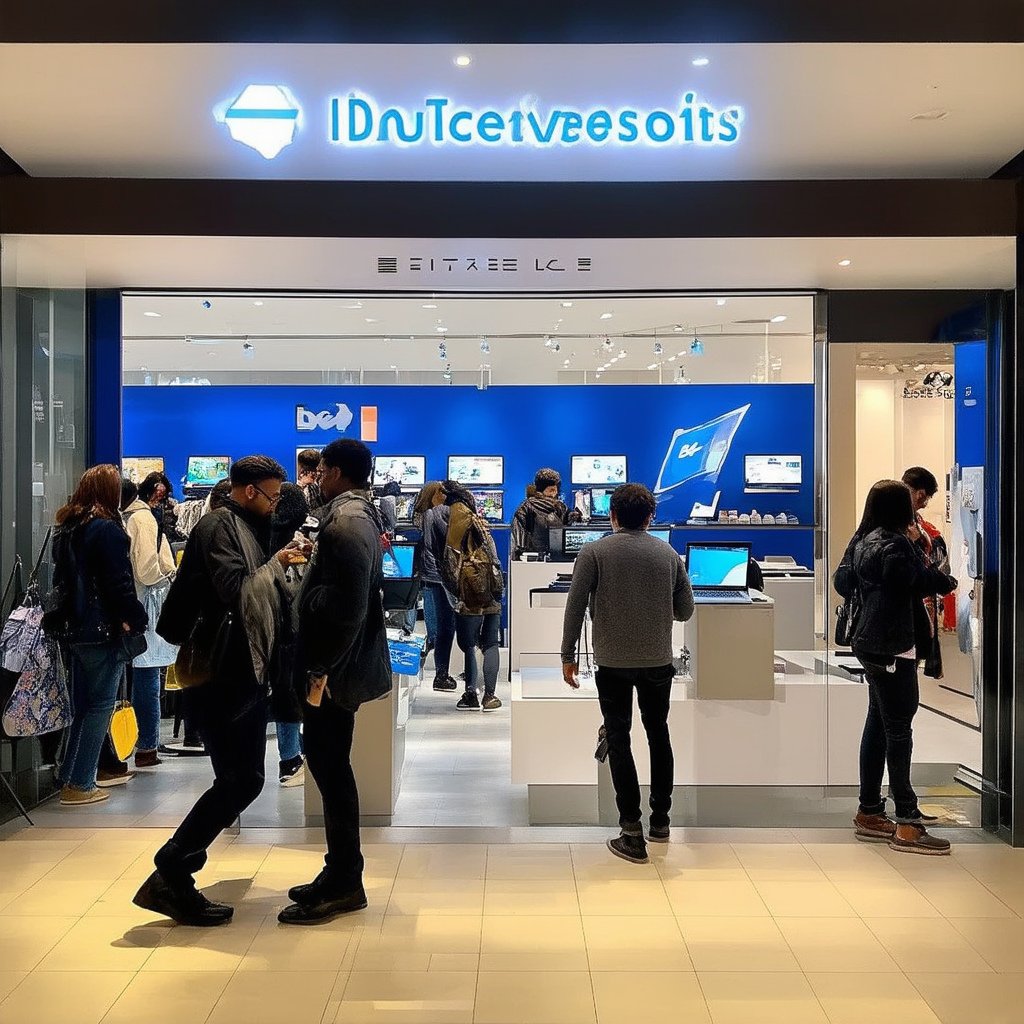} &
                        \includegraphics[width=0.1\textwidth]{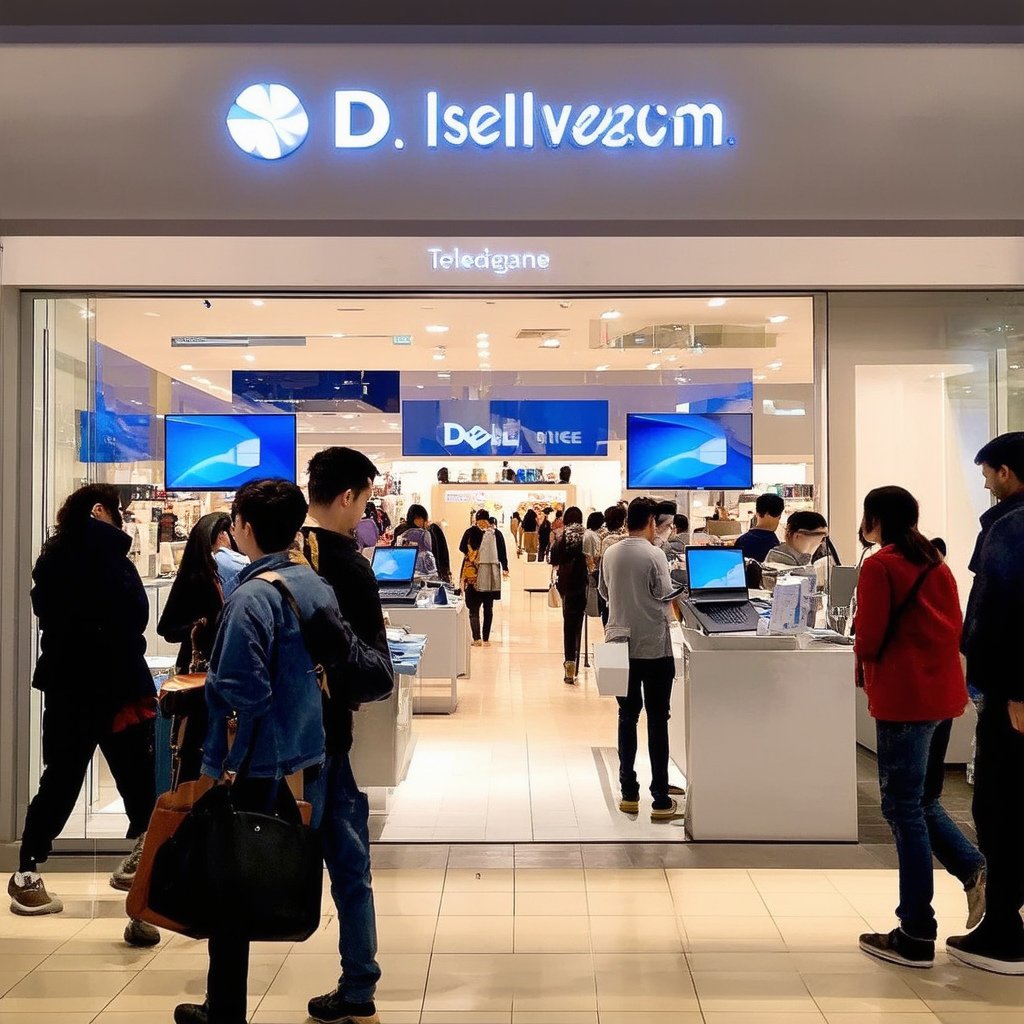} &
        \includegraphics[width=0.1\textwidth]{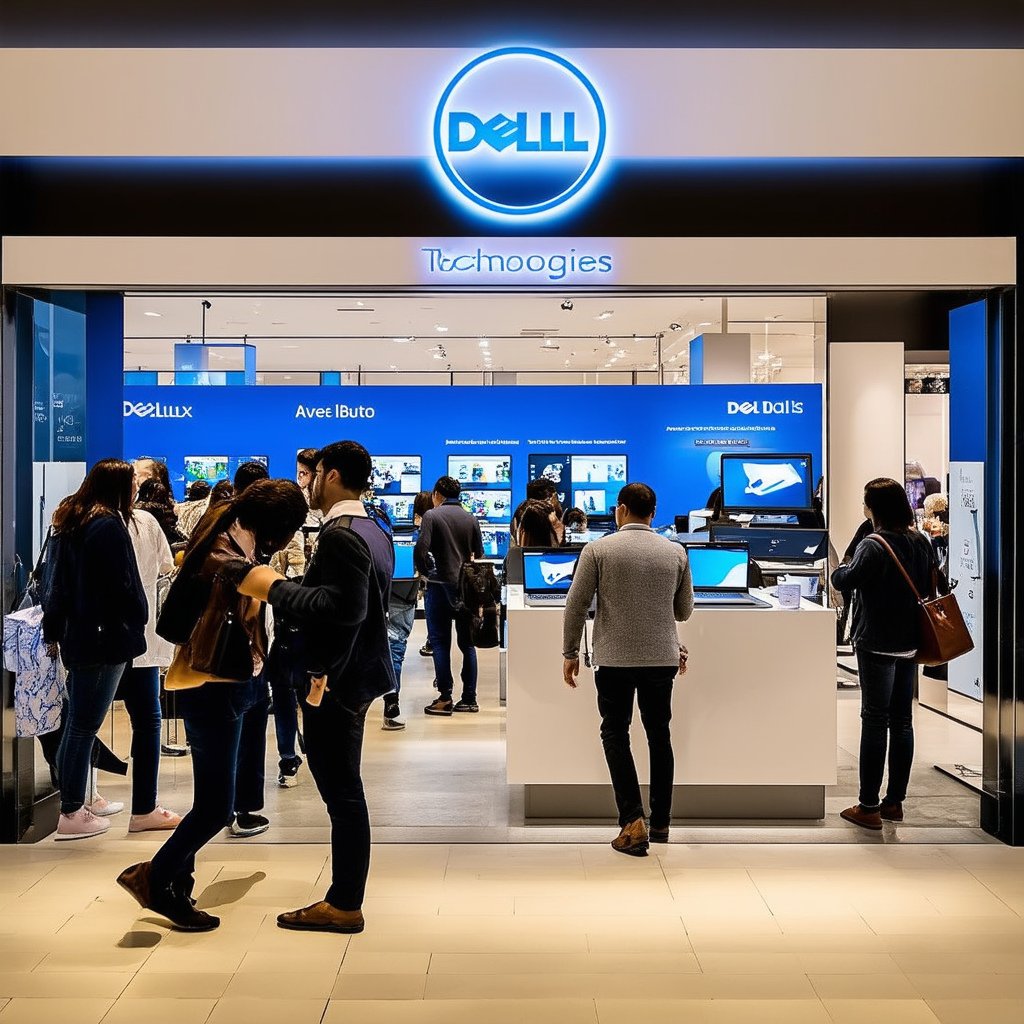} &
                \includegraphics[width=0.1\textwidth]{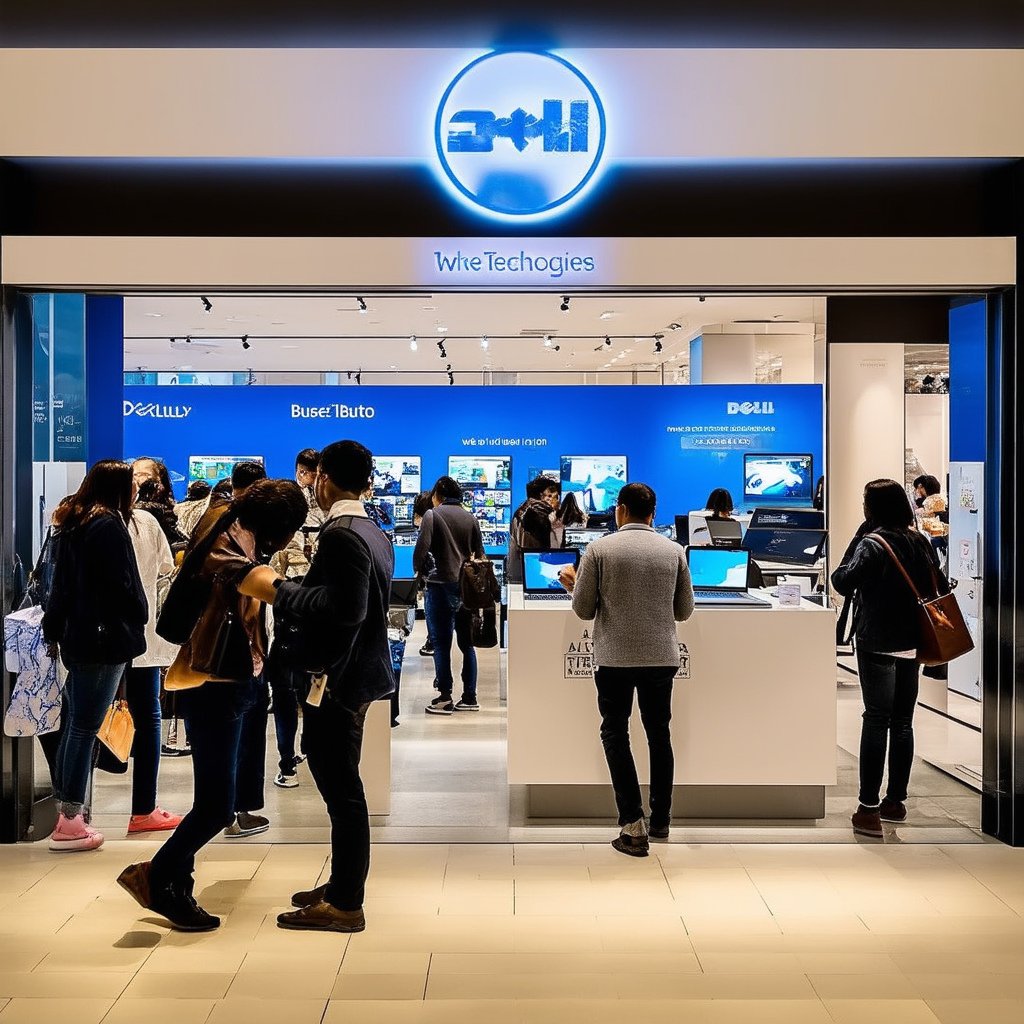} &
                        \includegraphics[width=0.1\textwidth]{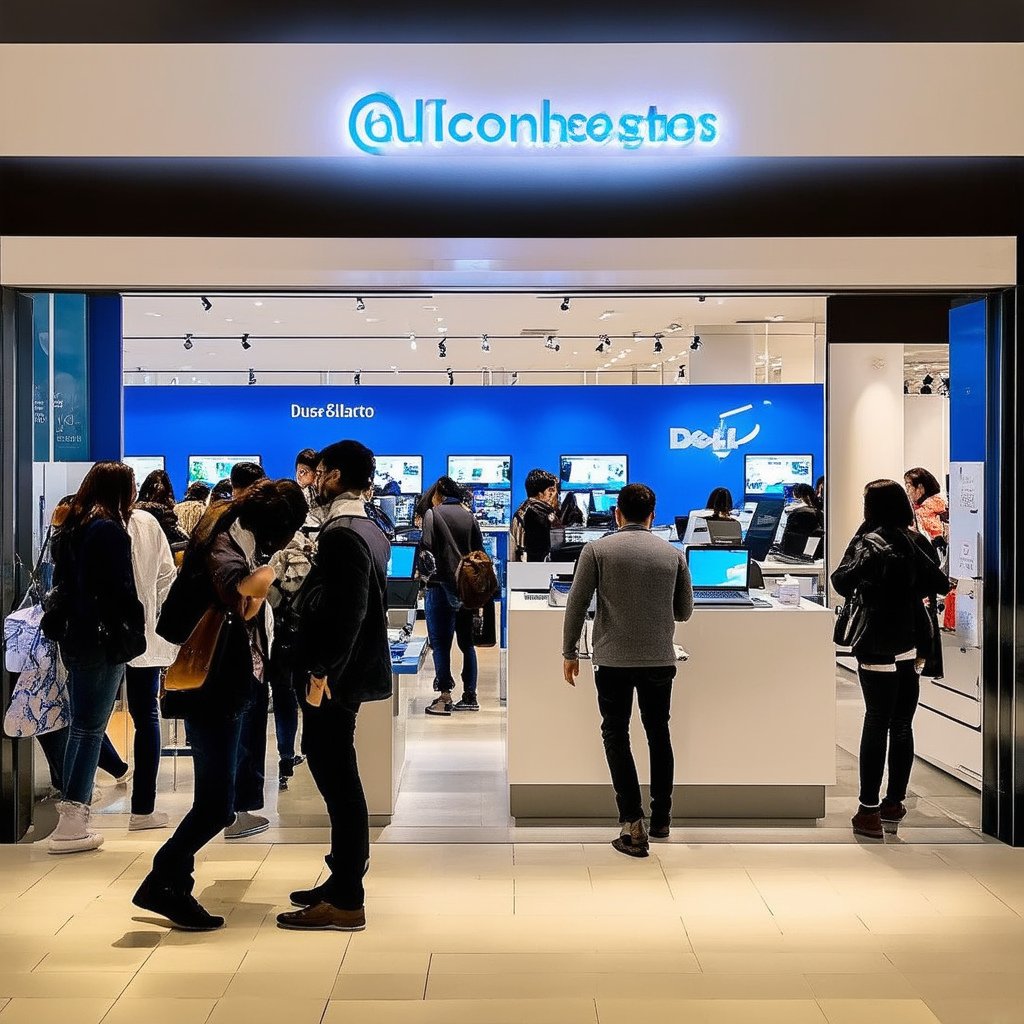} &
        \includegraphics[width=0.1\textwidth]{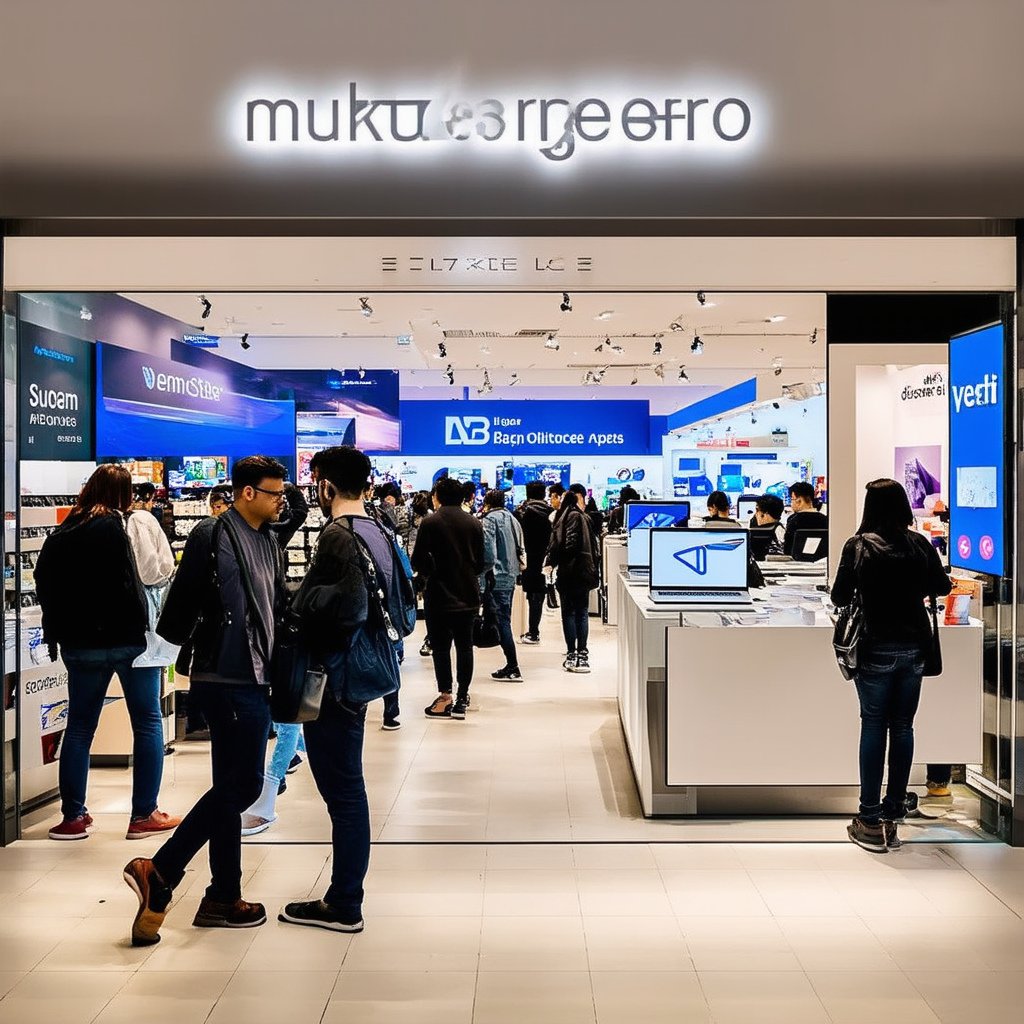} \\

        \includegraphics[width=0.1\textwidth]{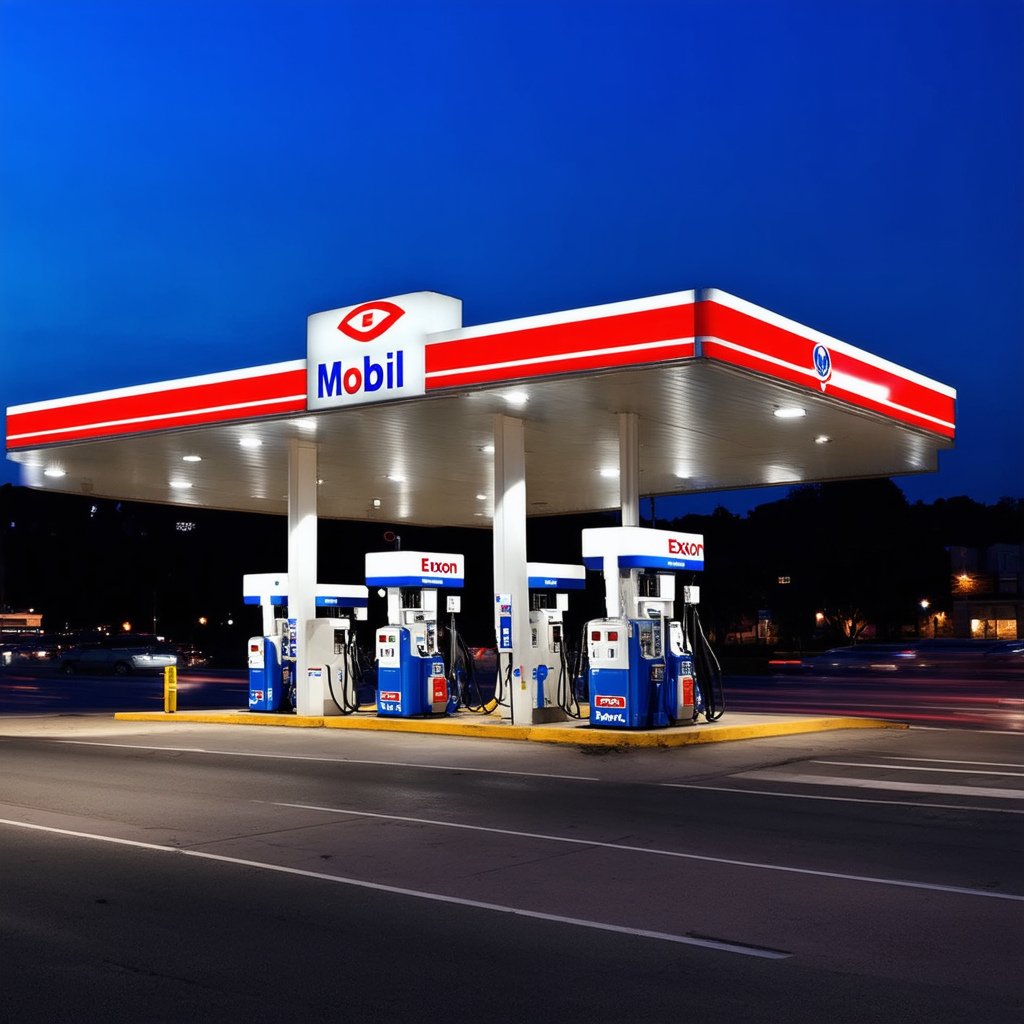} &
        \includegraphics[width=0.1\textwidth]{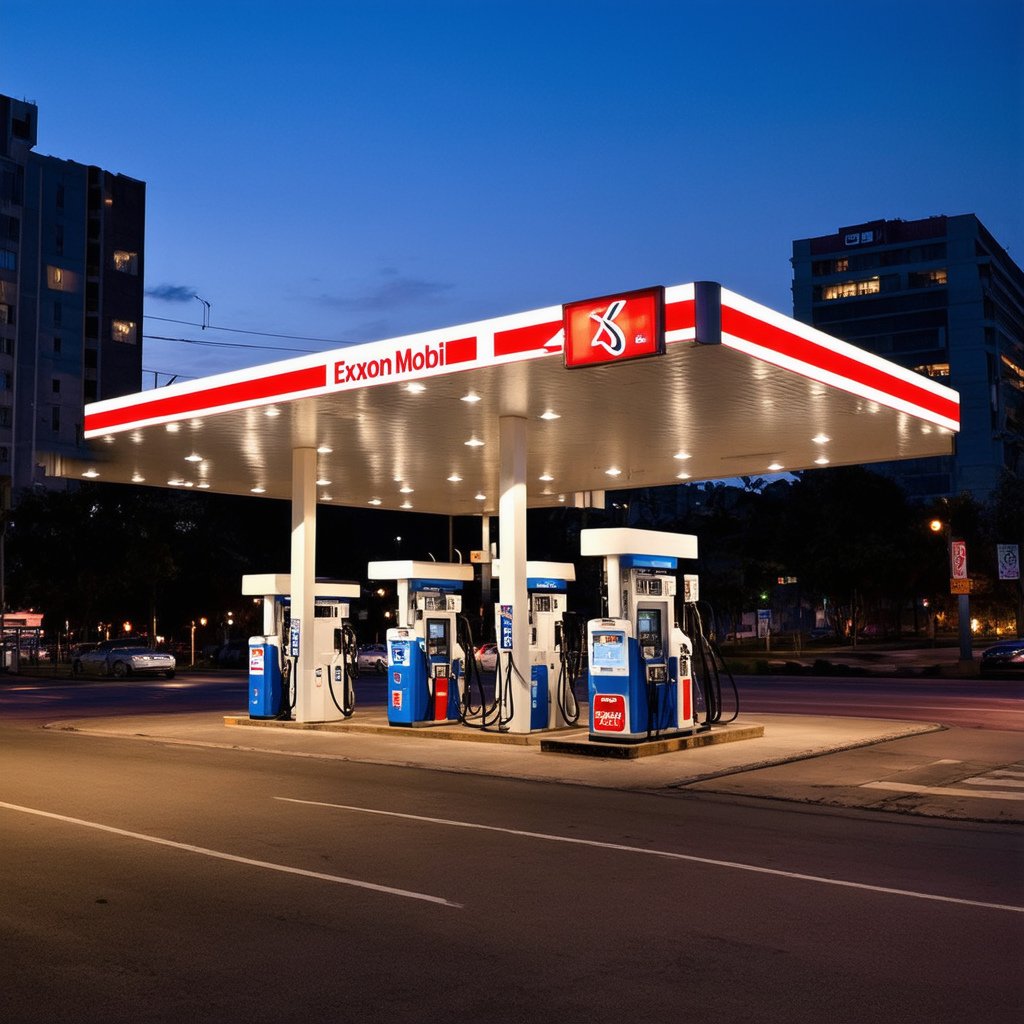} &
        \includegraphics[width=0.1\textwidth]{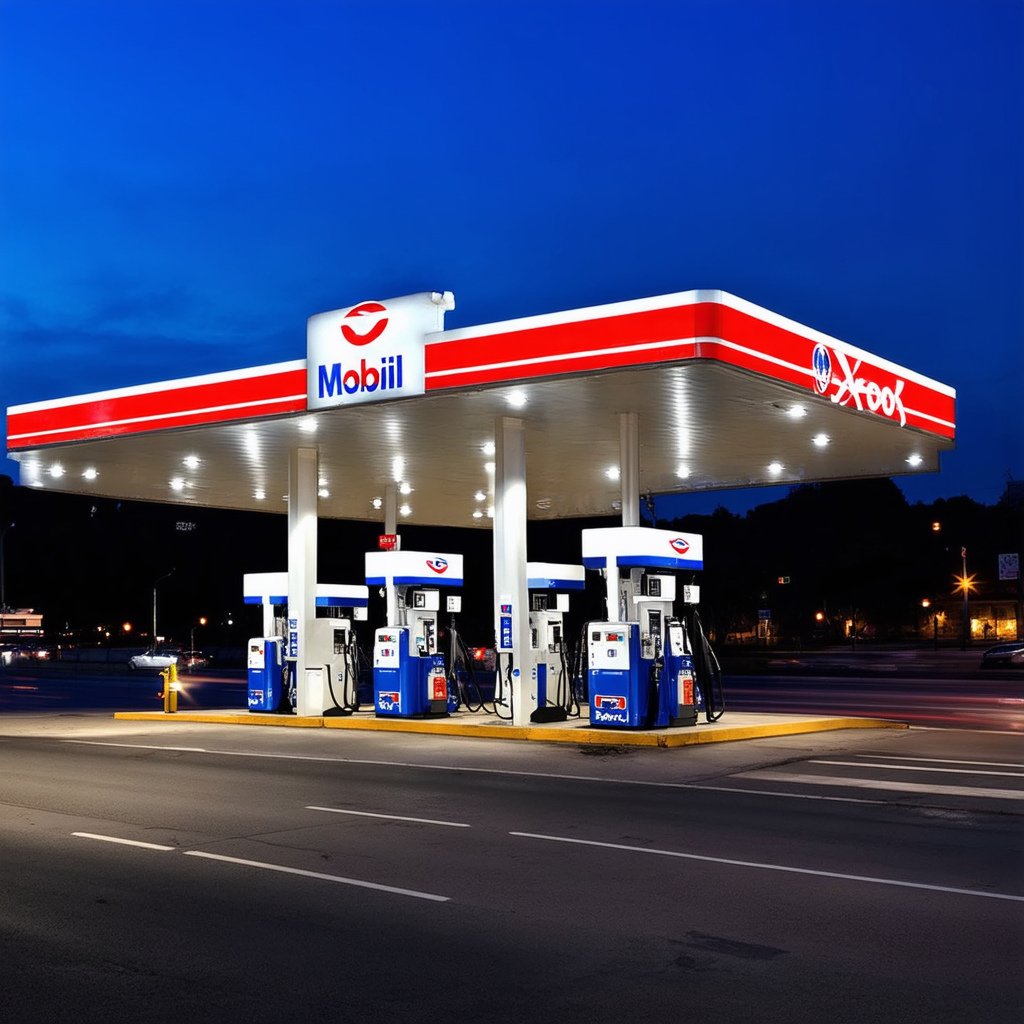} &
                \includegraphics[width=0.1\textwidth]{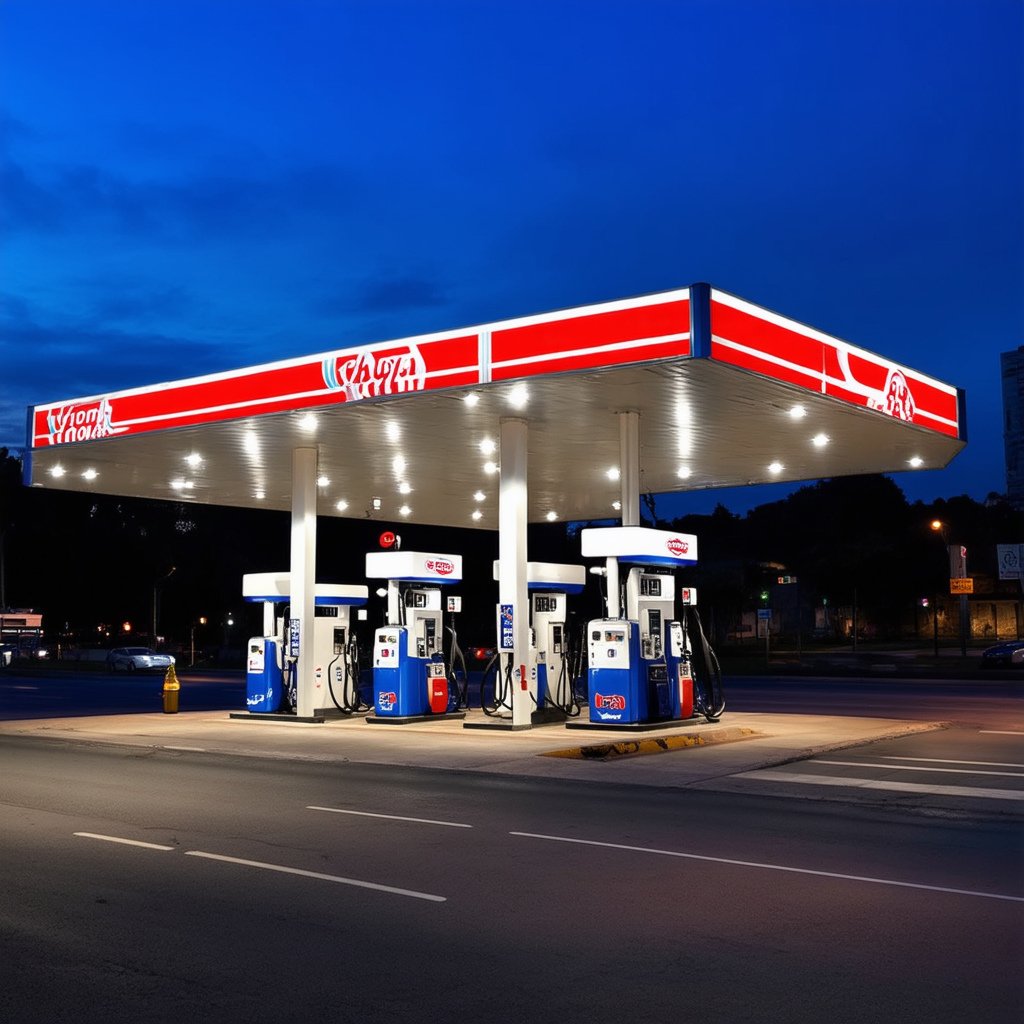} &
                        \includegraphics[width=0.1\textwidth]{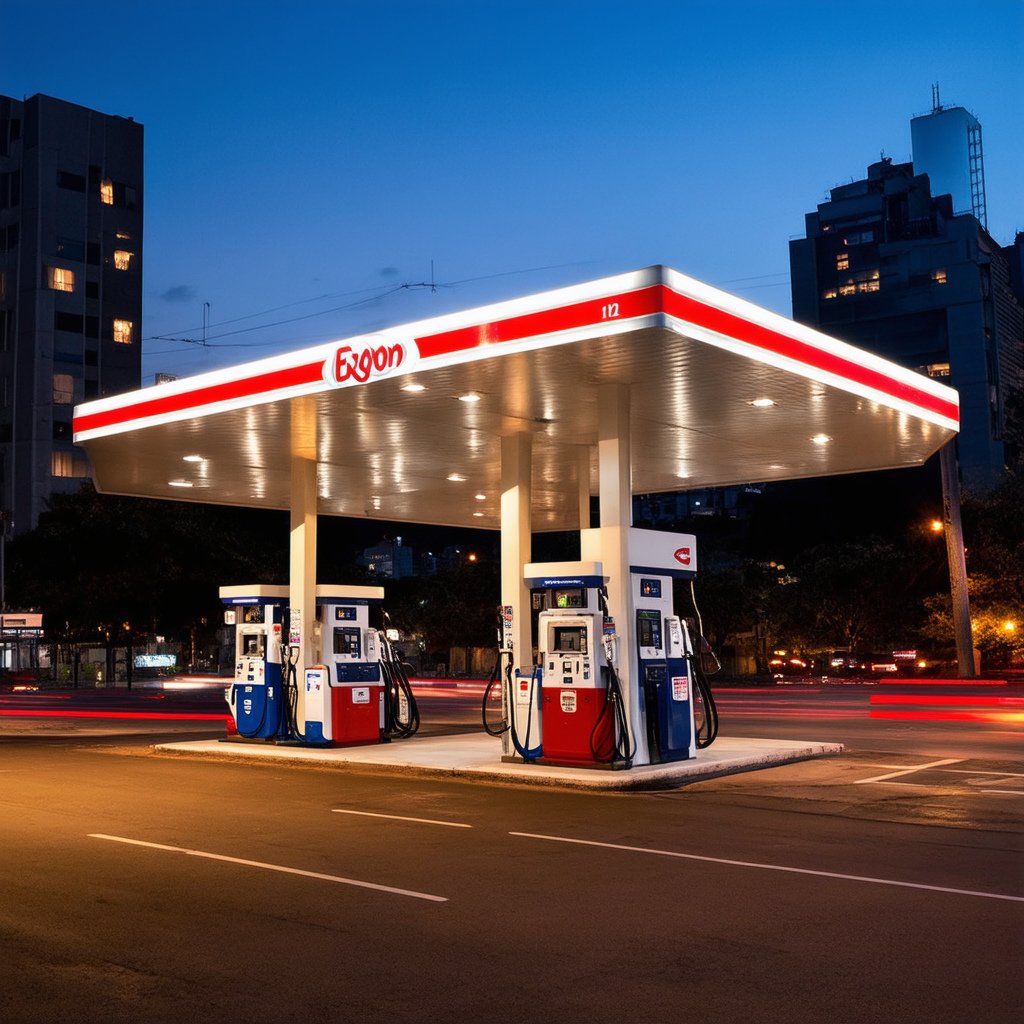} &
        \includegraphics[width=0.1\textwidth]{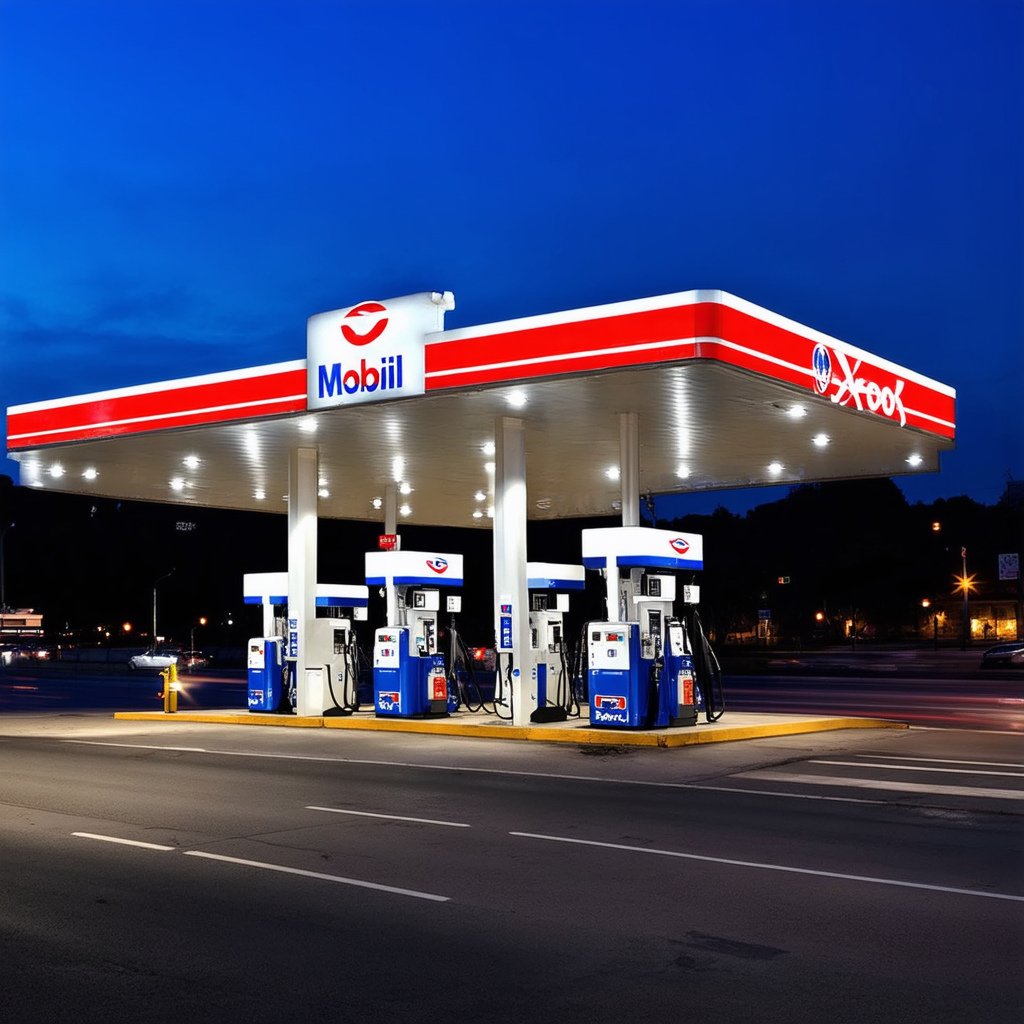} &
                \includegraphics[width=0.1\textwidth]{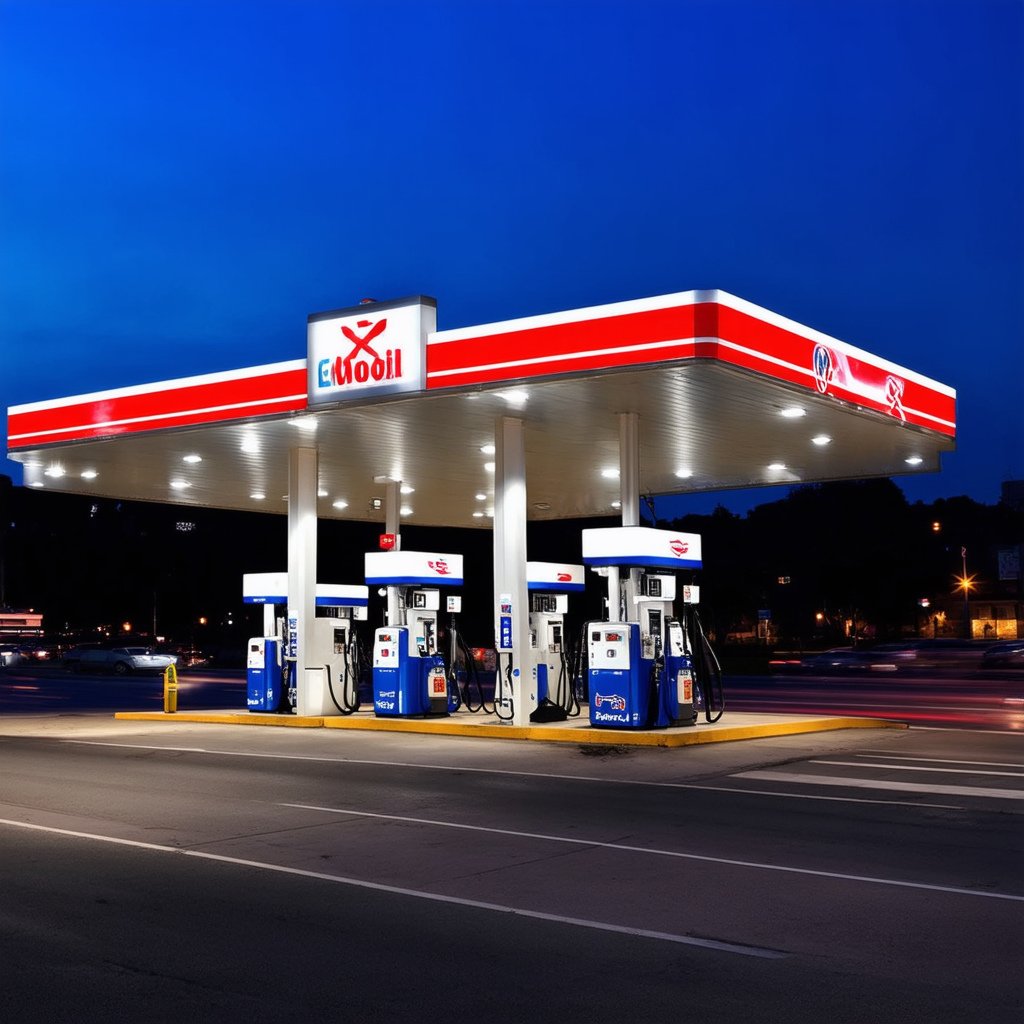} &
                        \includegraphics[width=0.1\textwidth]{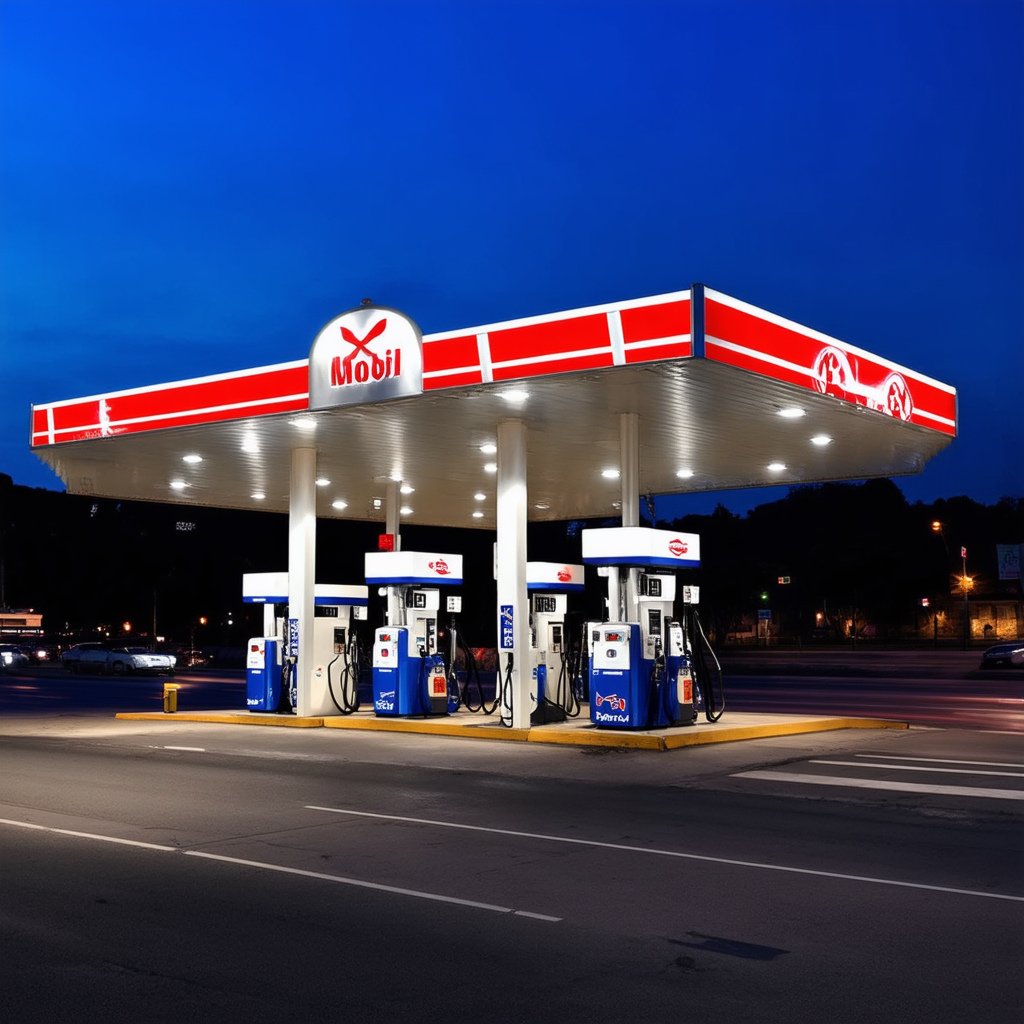} &
        \includegraphics[width=0.1\textwidth]{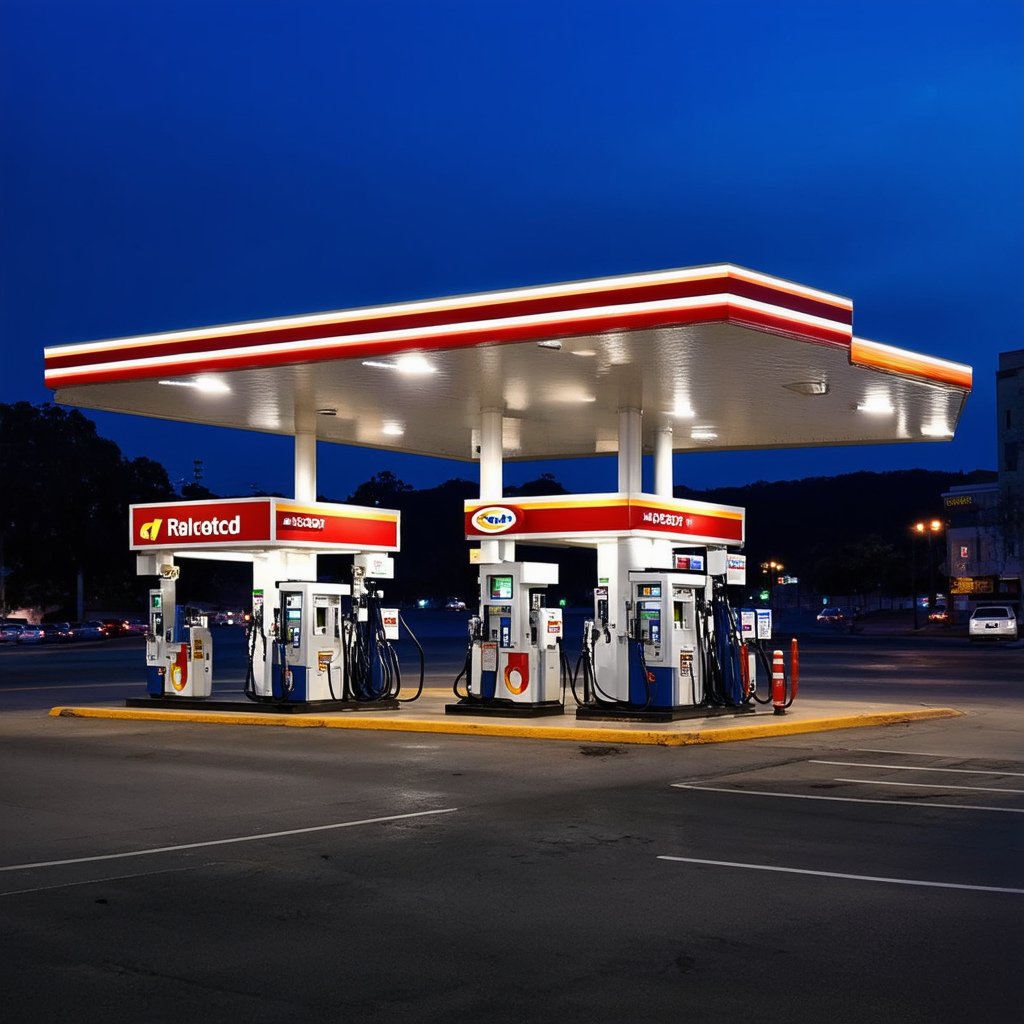} \\

        \includegraphics[width=0.1\textwidth]{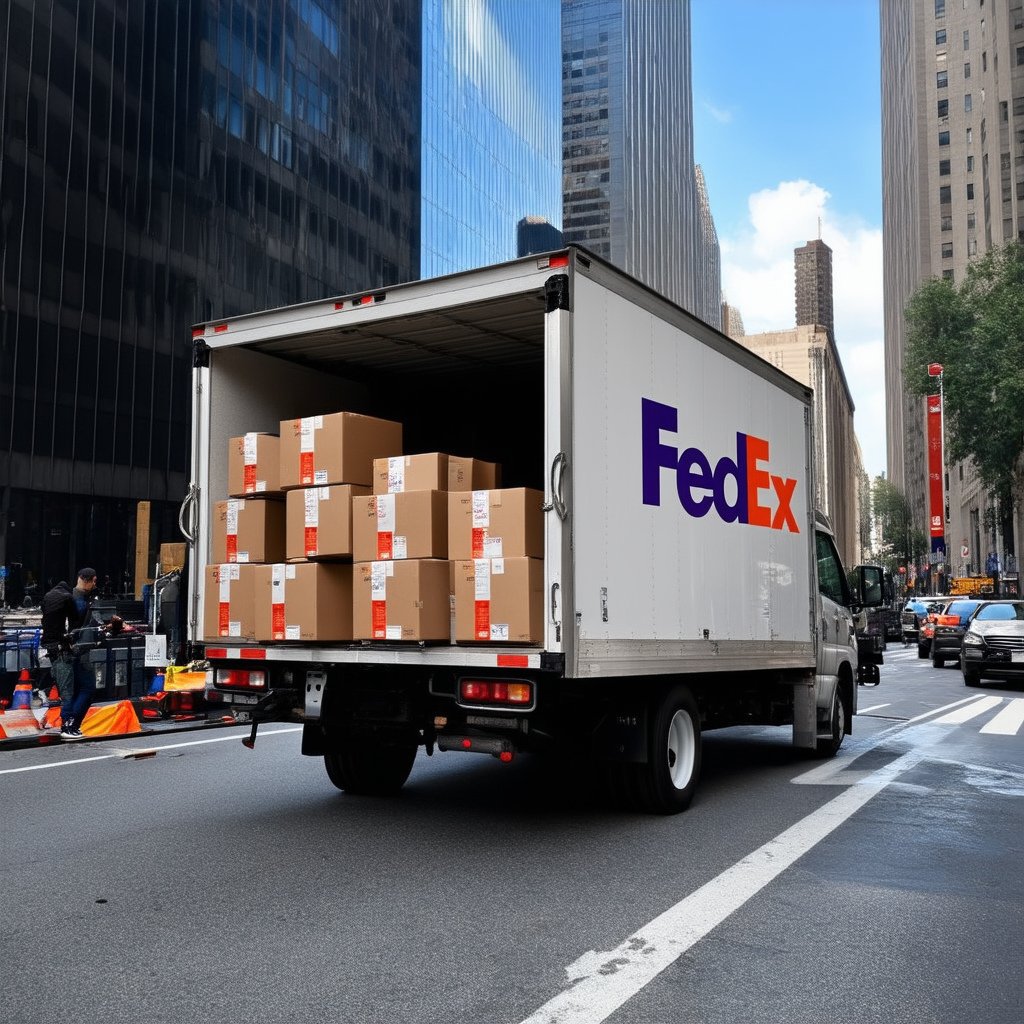} &
        \includegraphics[width=0.1\textwidth]{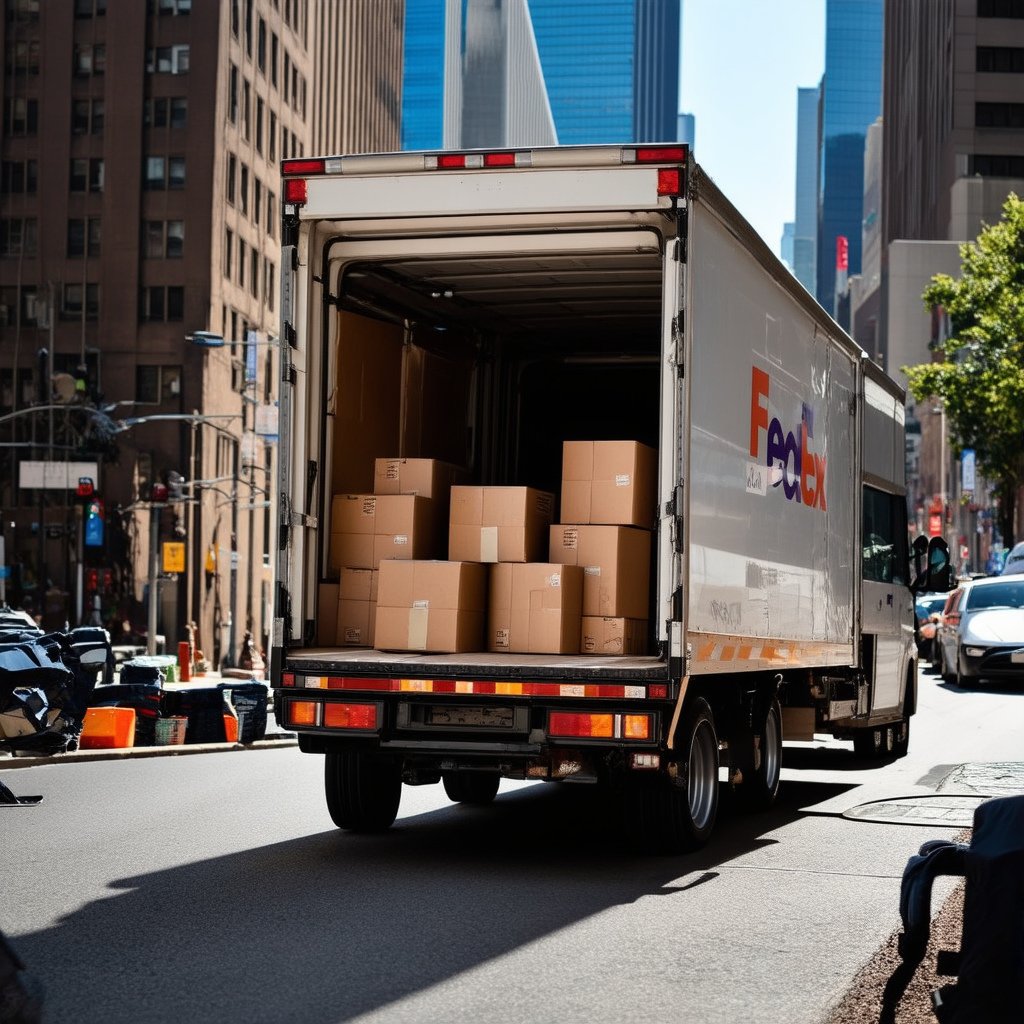} &
        \includegraphics[width=0.1\textwidth]{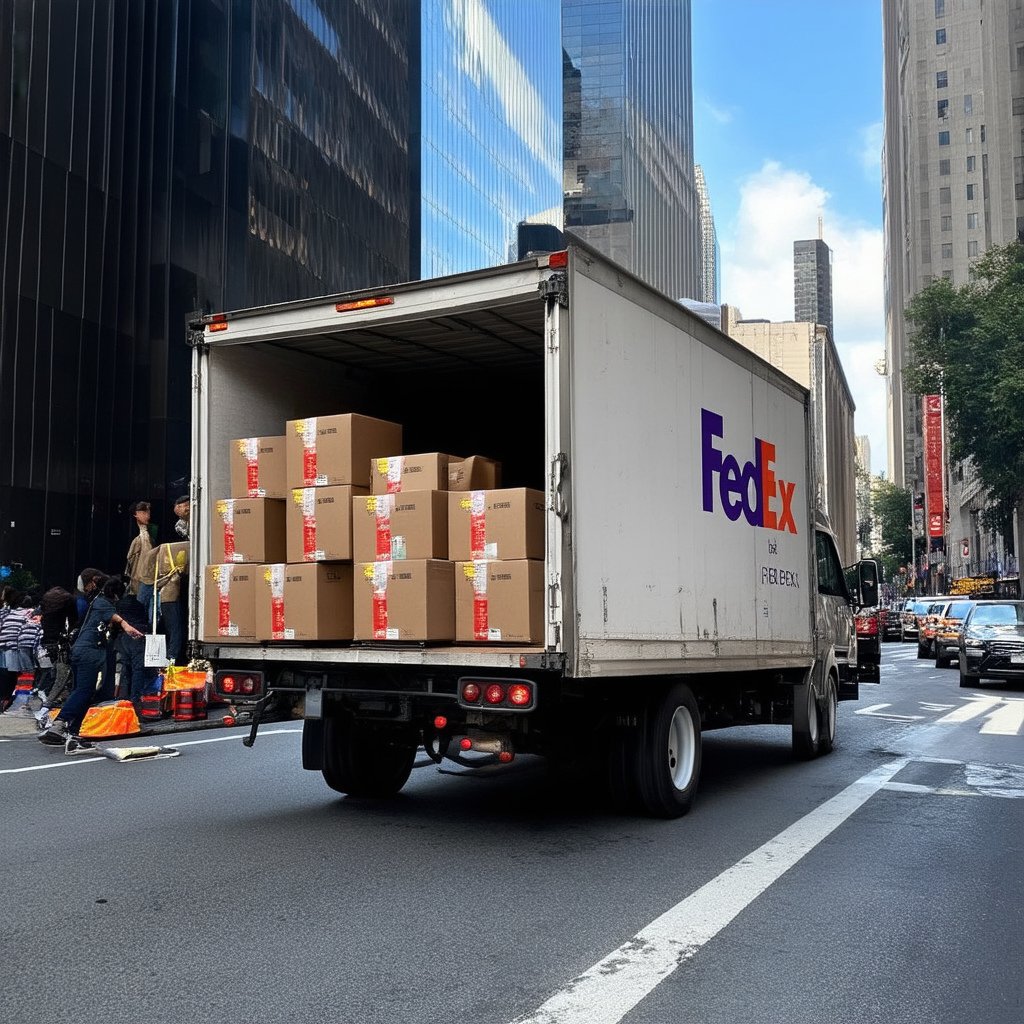} &
                \includegraphics[width=0.1\textwidth]{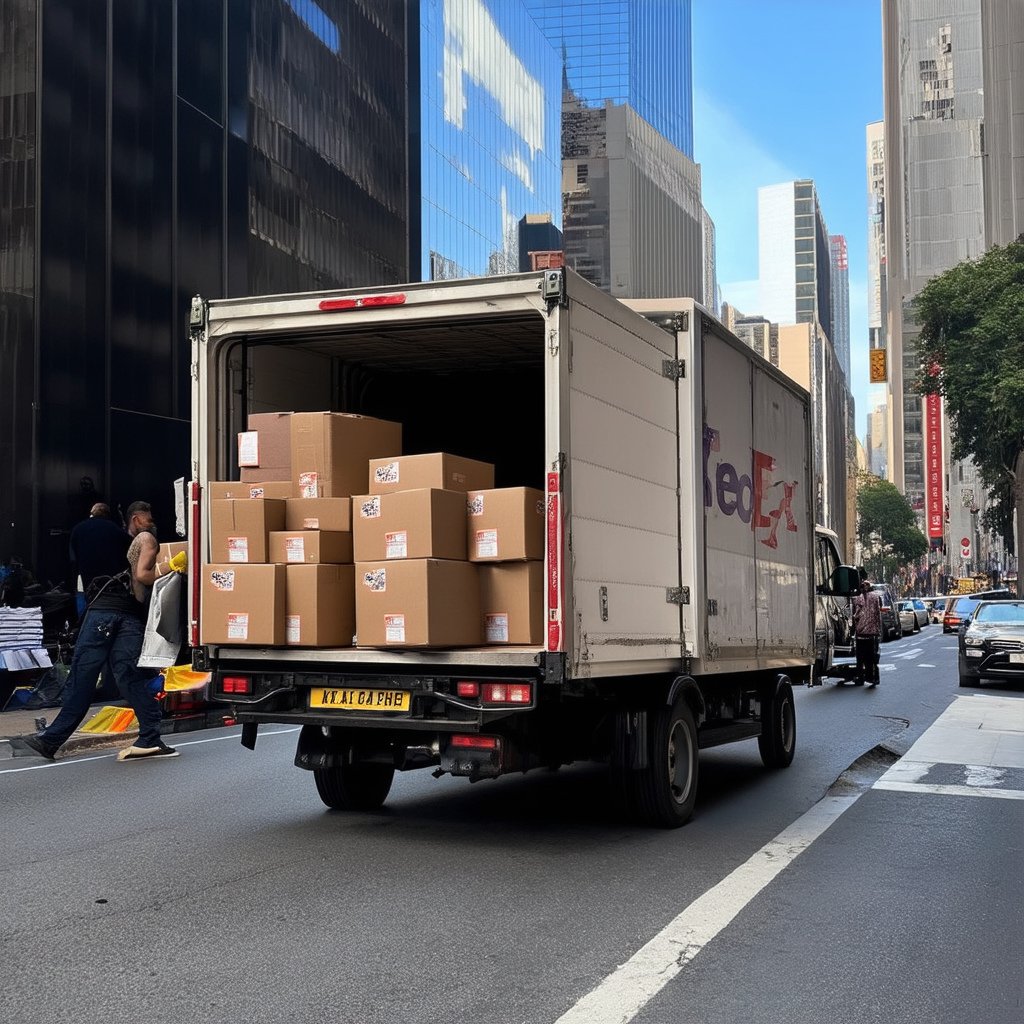} &
                        \includegraphics[width=0.1\textwidth]{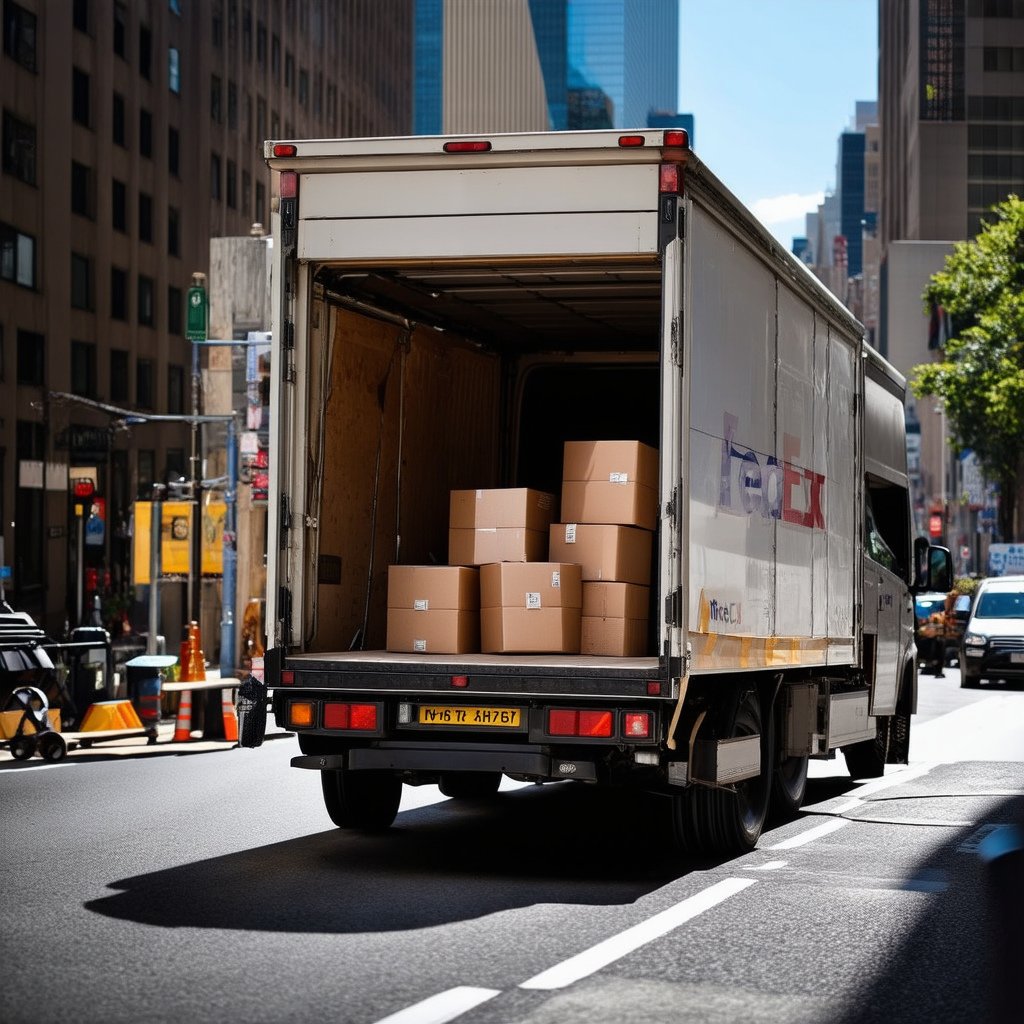} &
        \includegraphics[width=0.1\textwidth]{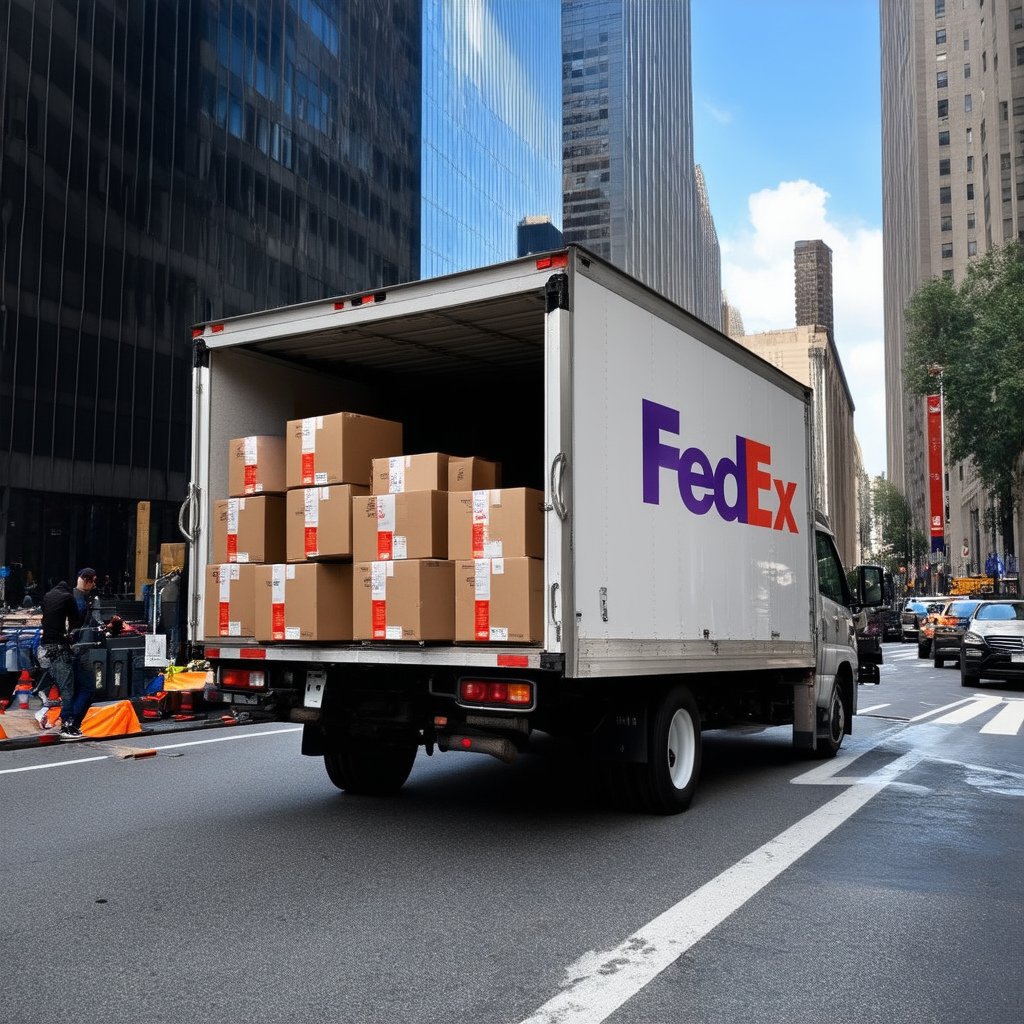} &
                \includegraphics[width=0.1\textwidth]{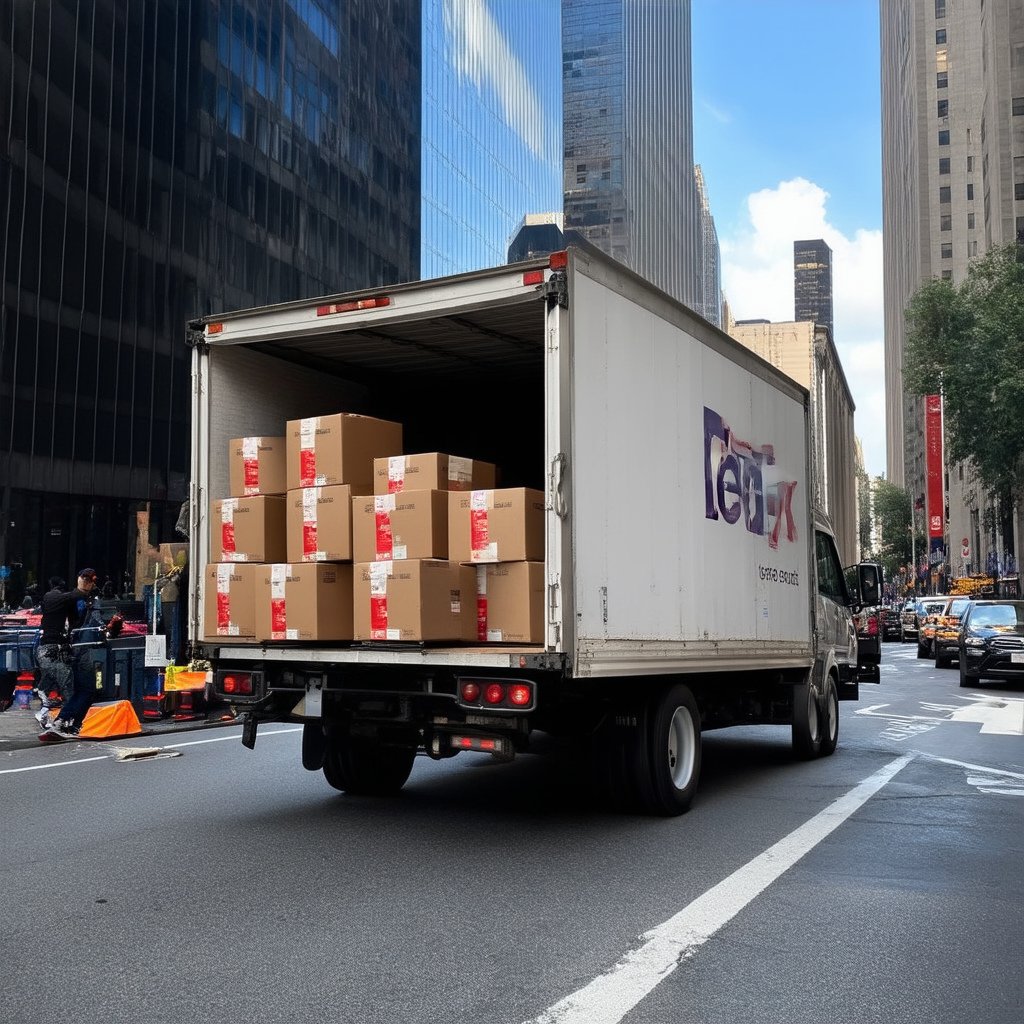} &
                        \includegraphics[width=0.1\textwidth]{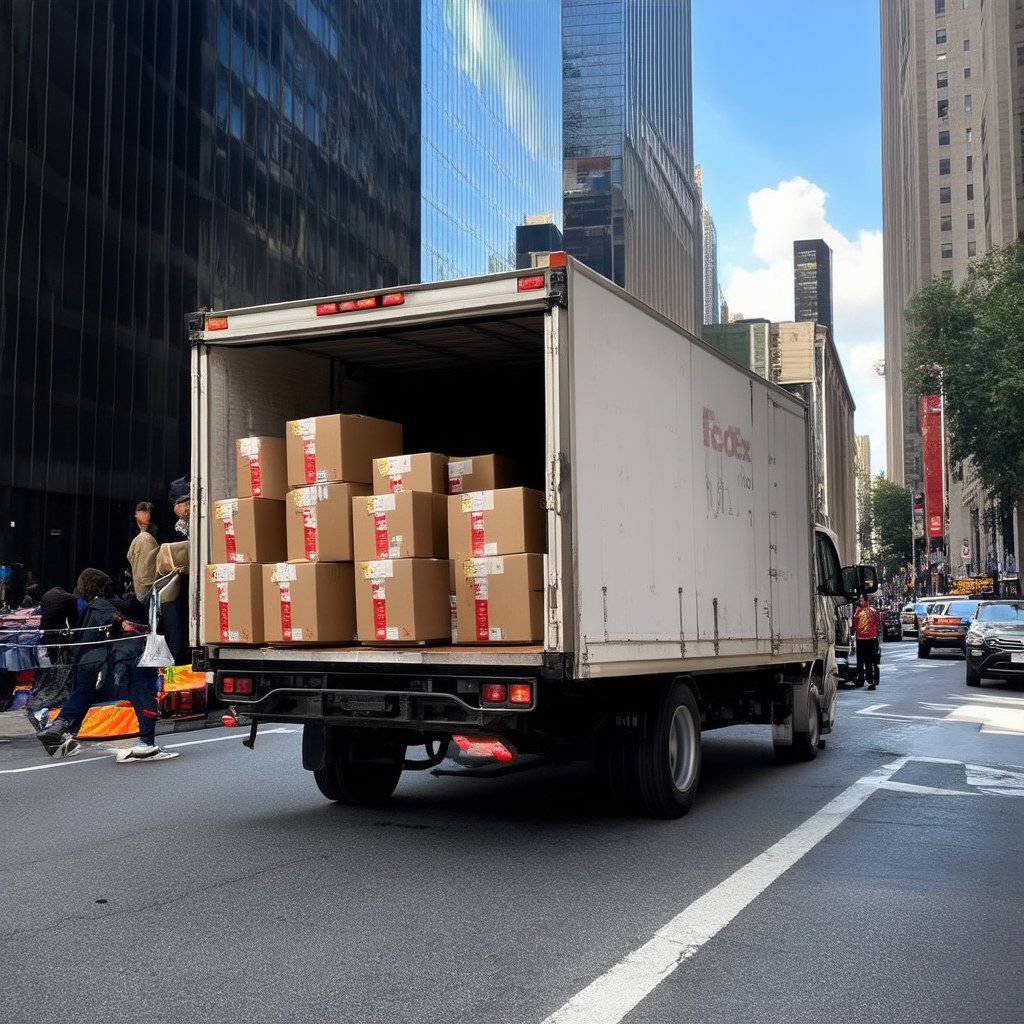} &
        \includegraphics[width=0.1\textwidth]{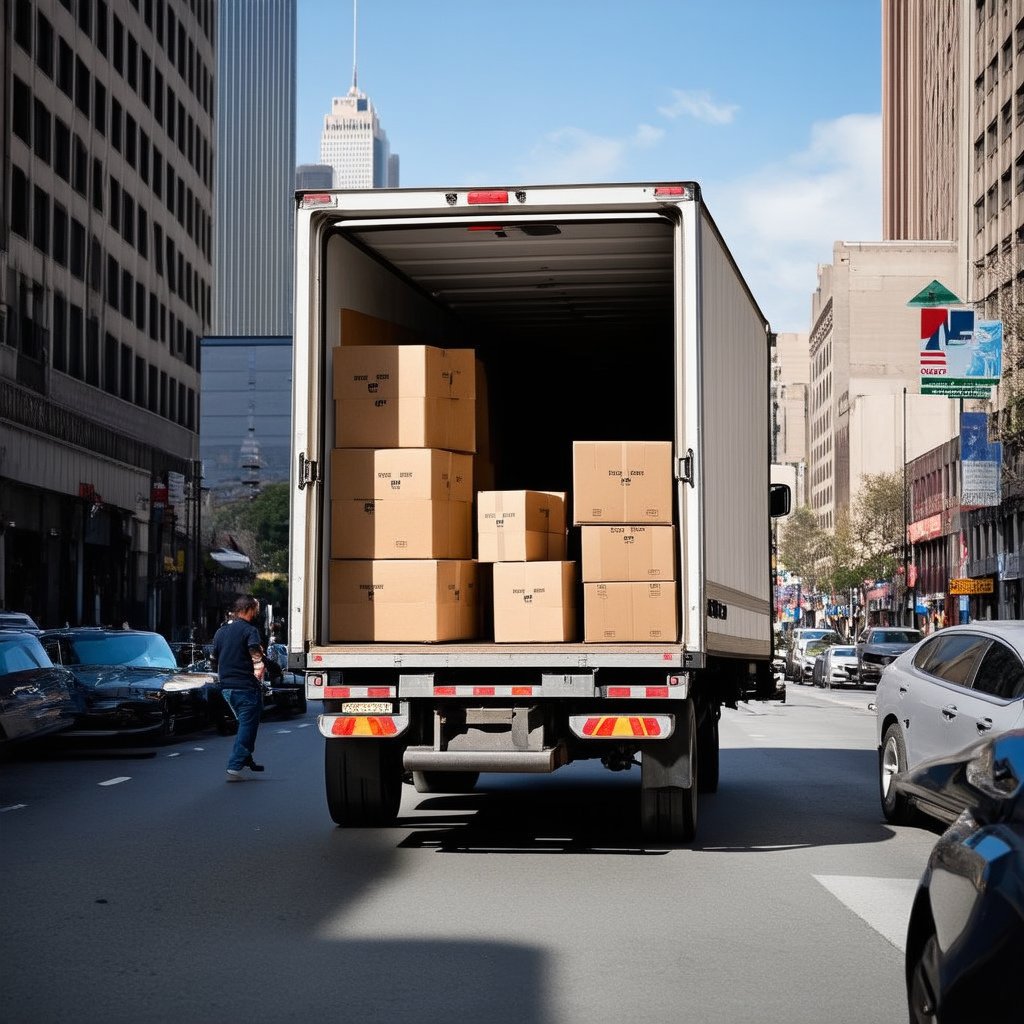} \\

        \includegraphics[width=0.1\textwidth]{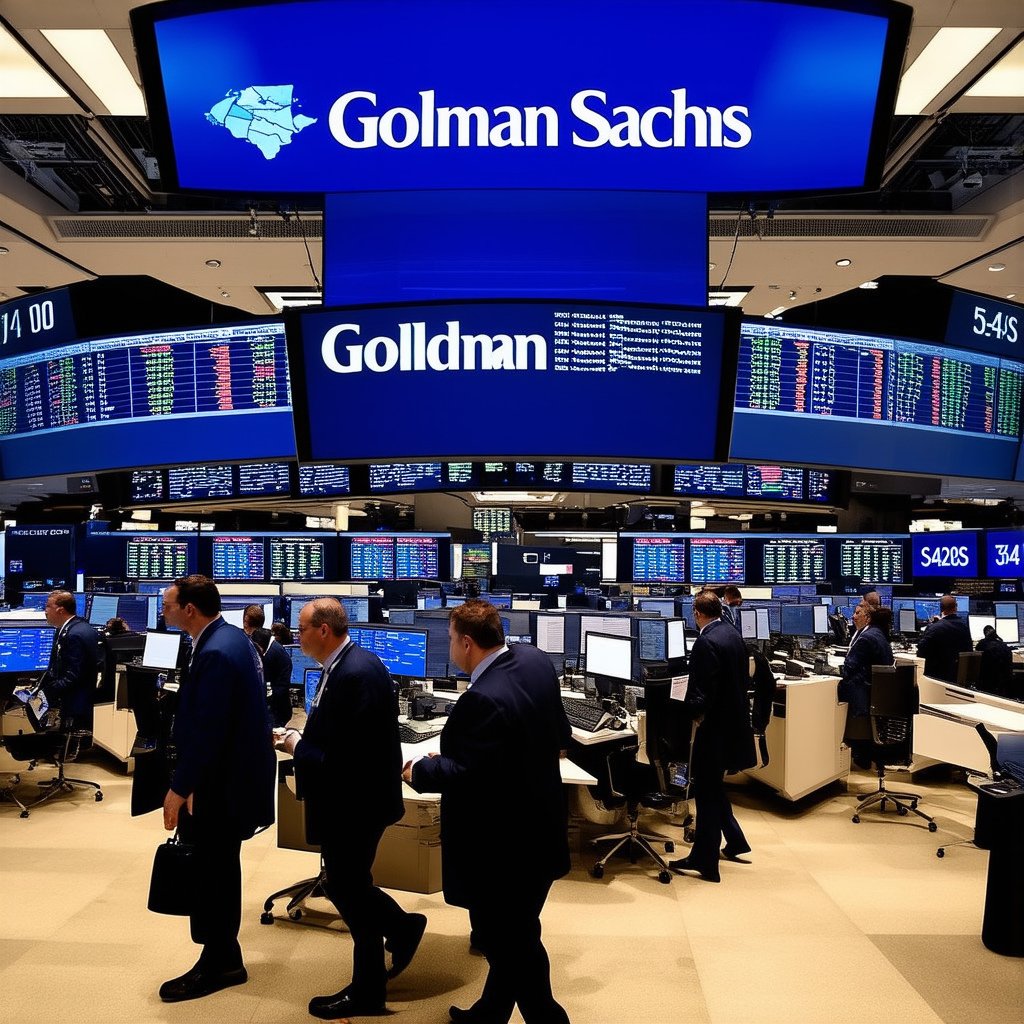} &
        \includegraphics[width=0.1\textwidth]{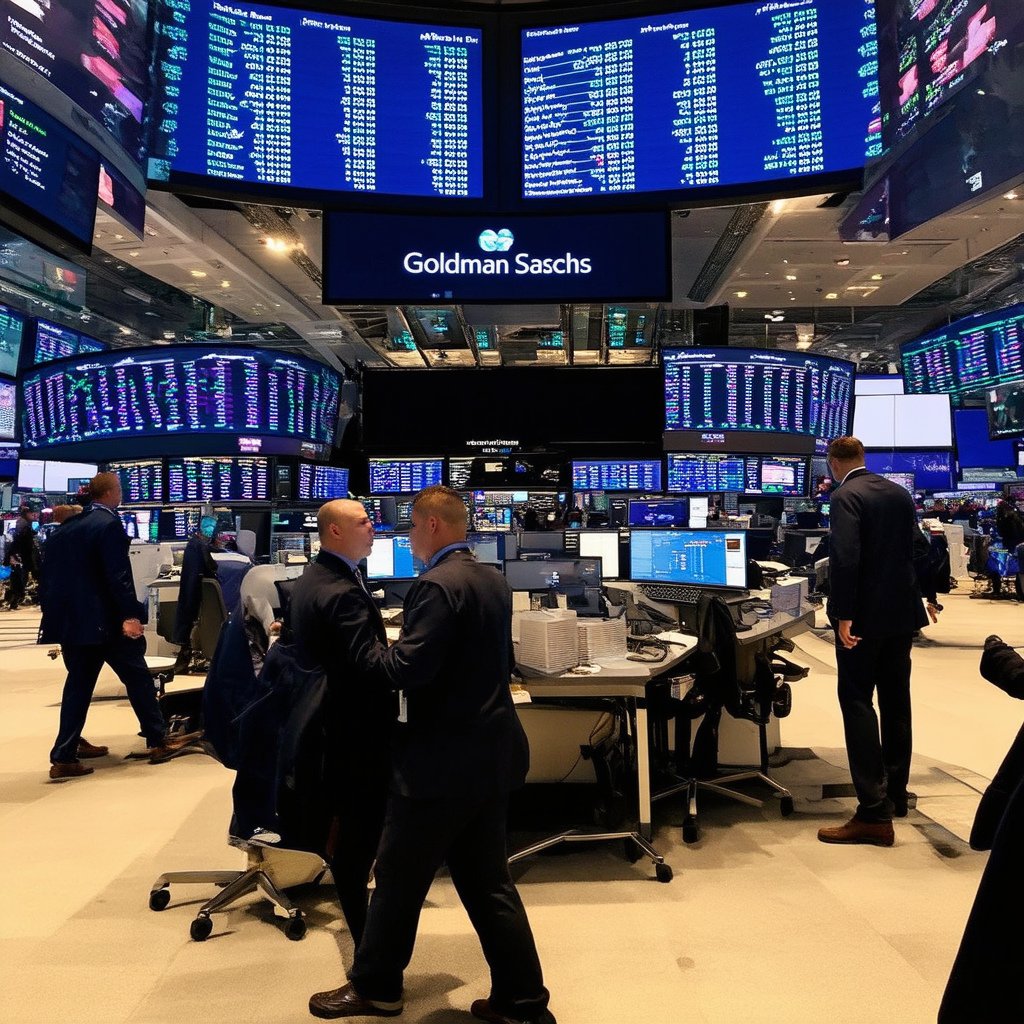} &
        \includegraphics[width=0.1\textwidth]{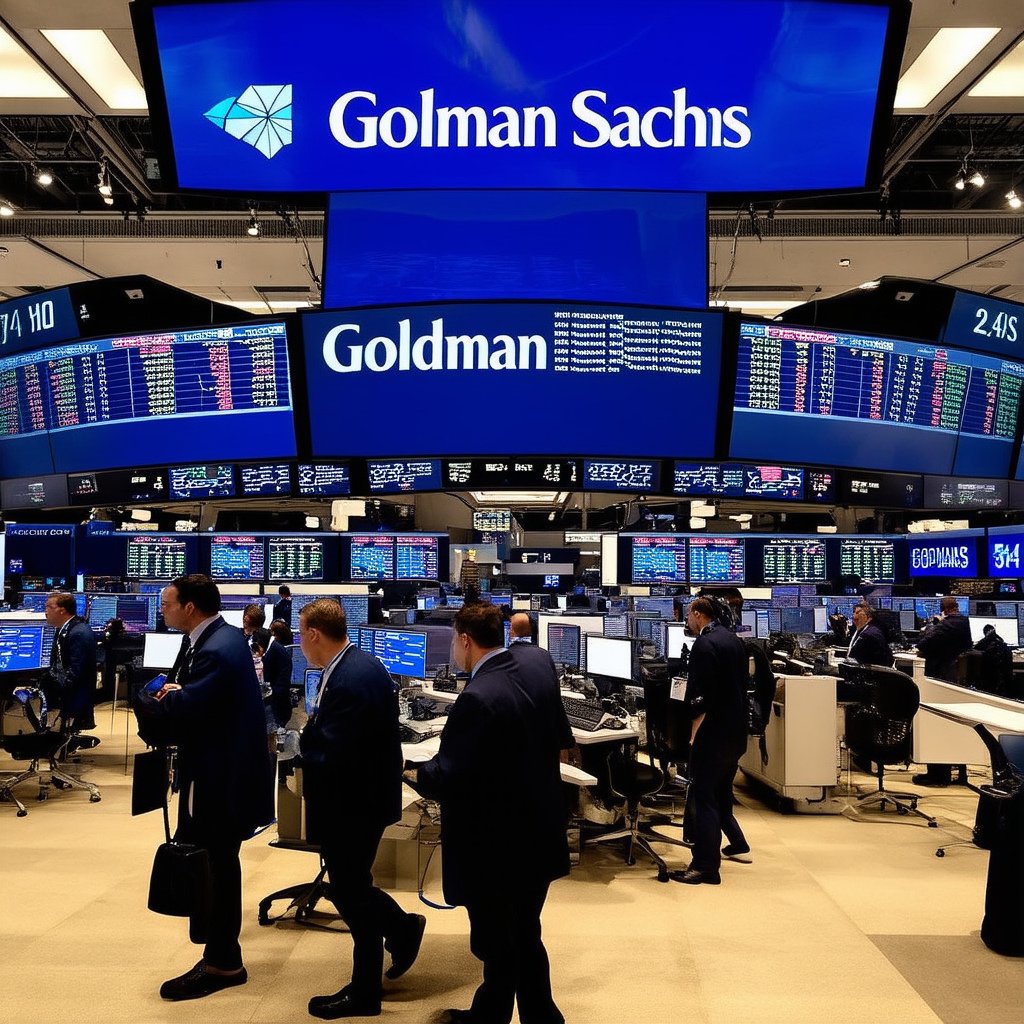} &
                \includegraphics[width=0.1\textwidth]{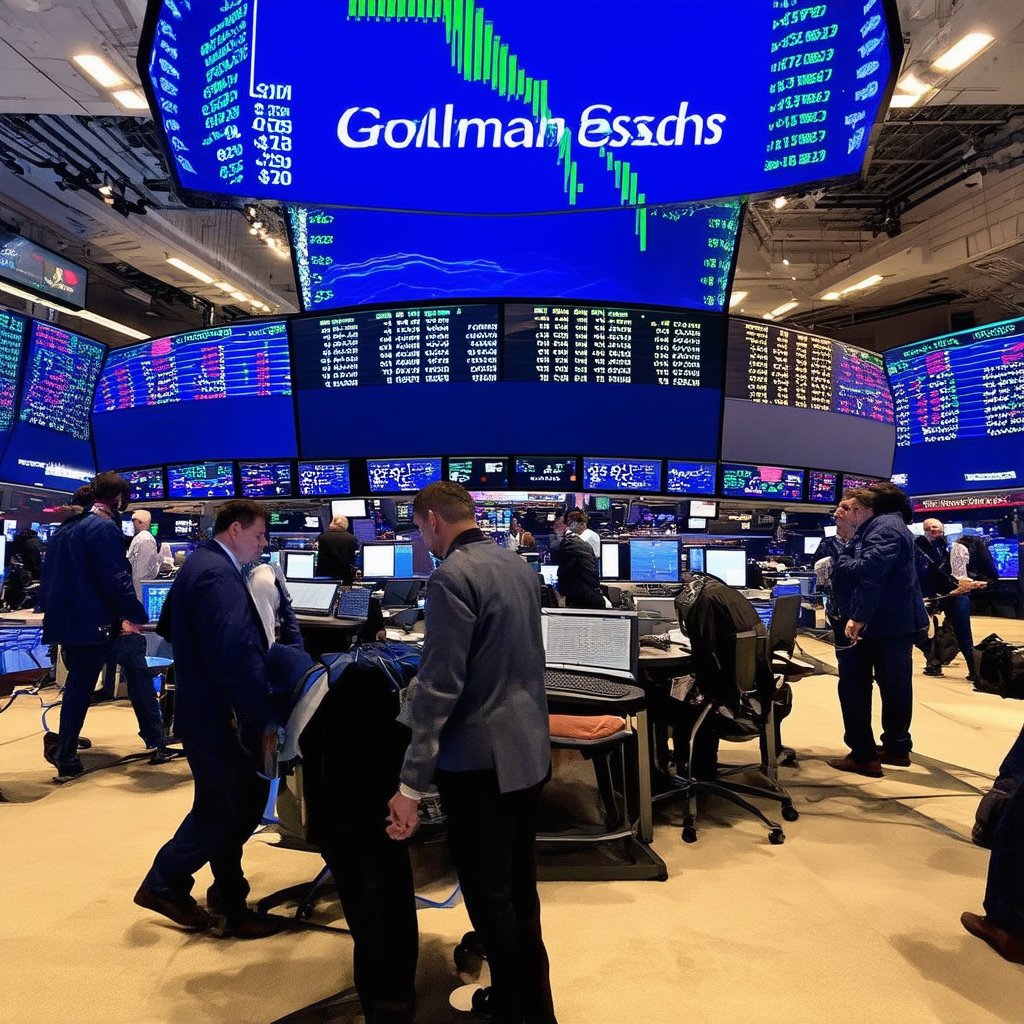} &
                        \includegraphics[width=0.1\textwidth]{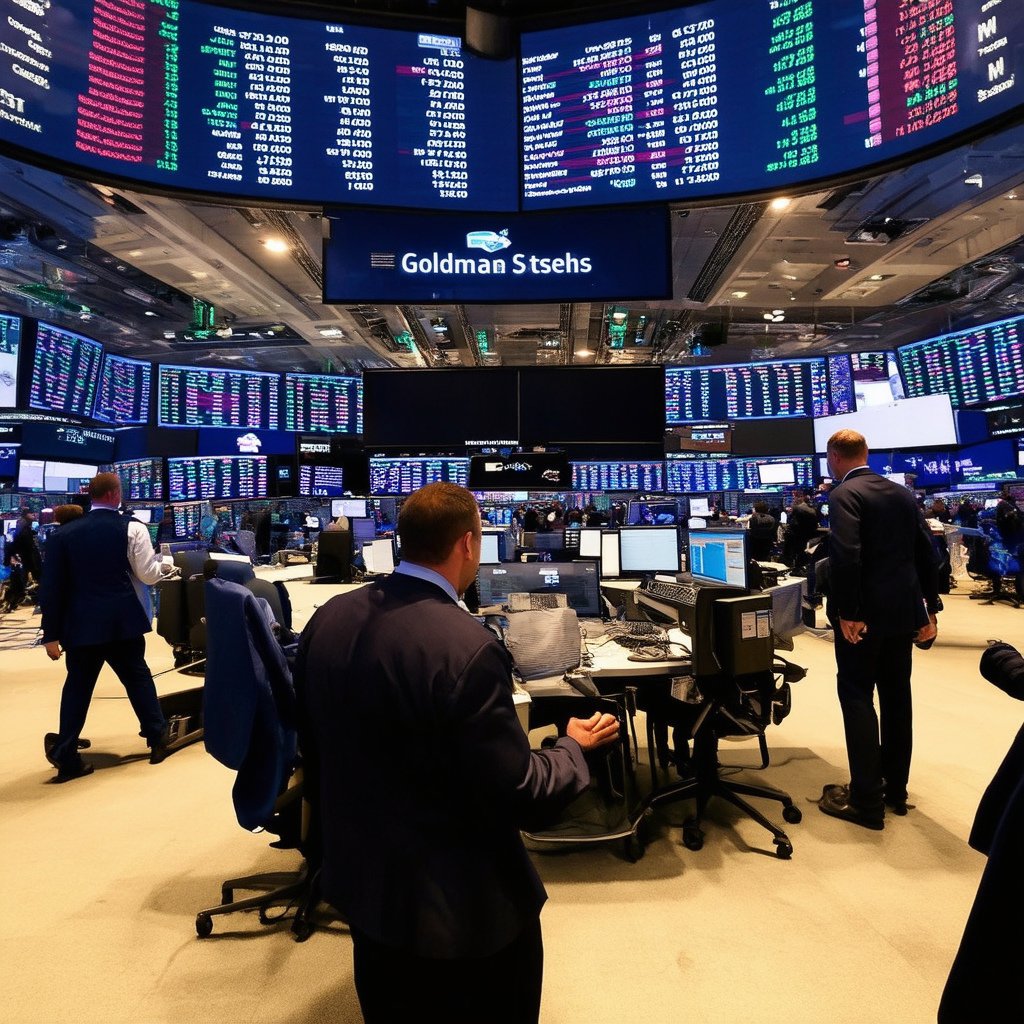} &
        \includegraphics[width=0.1\textwidth]{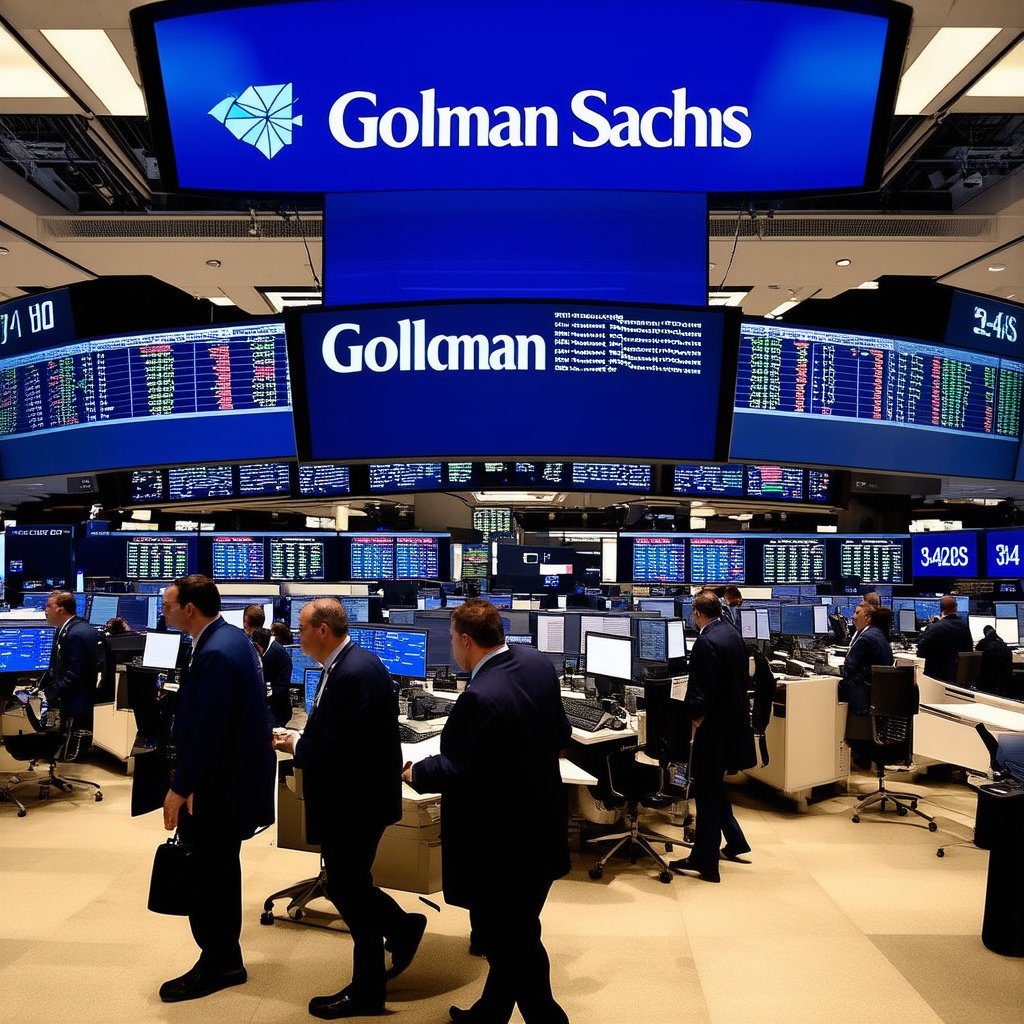} &
                \includegraphics[width=0.1\textwidth]{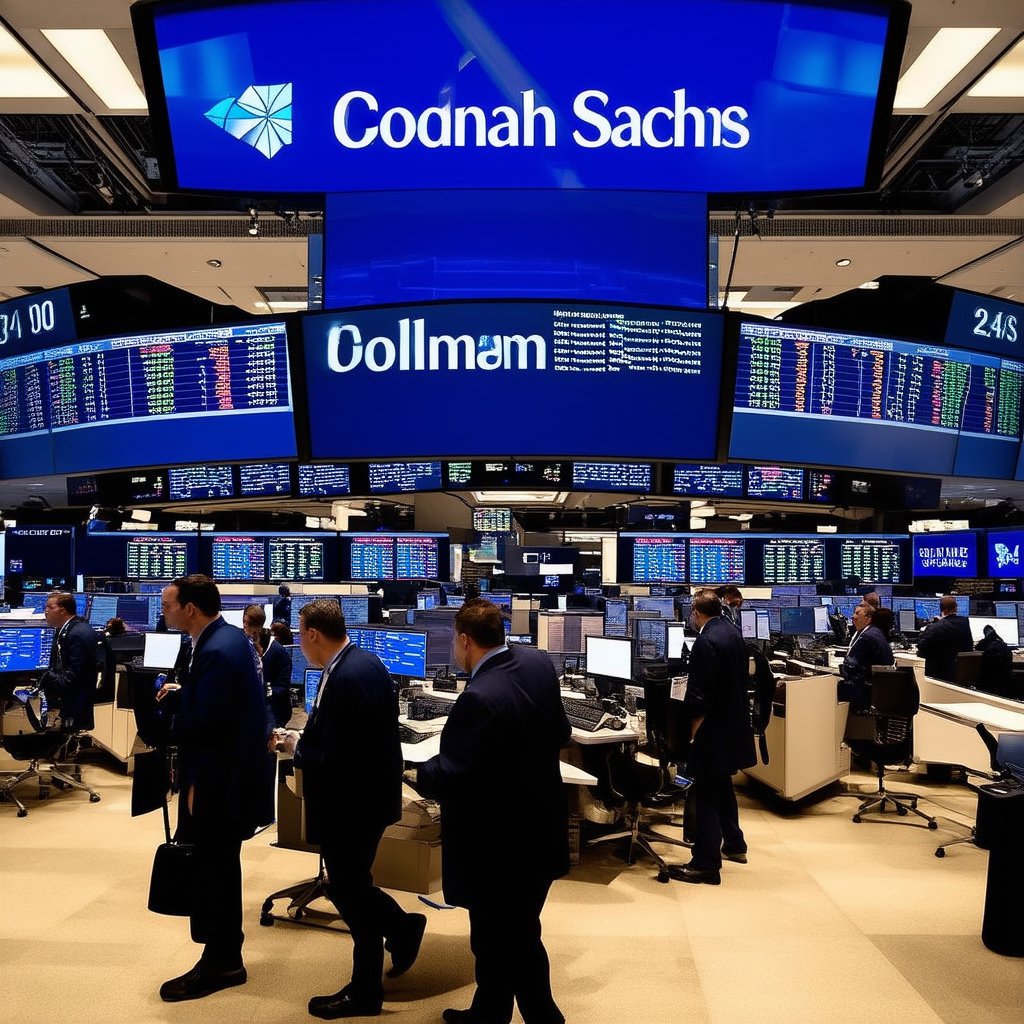} &
                        \includegraphics[width=0.1\textwidth]{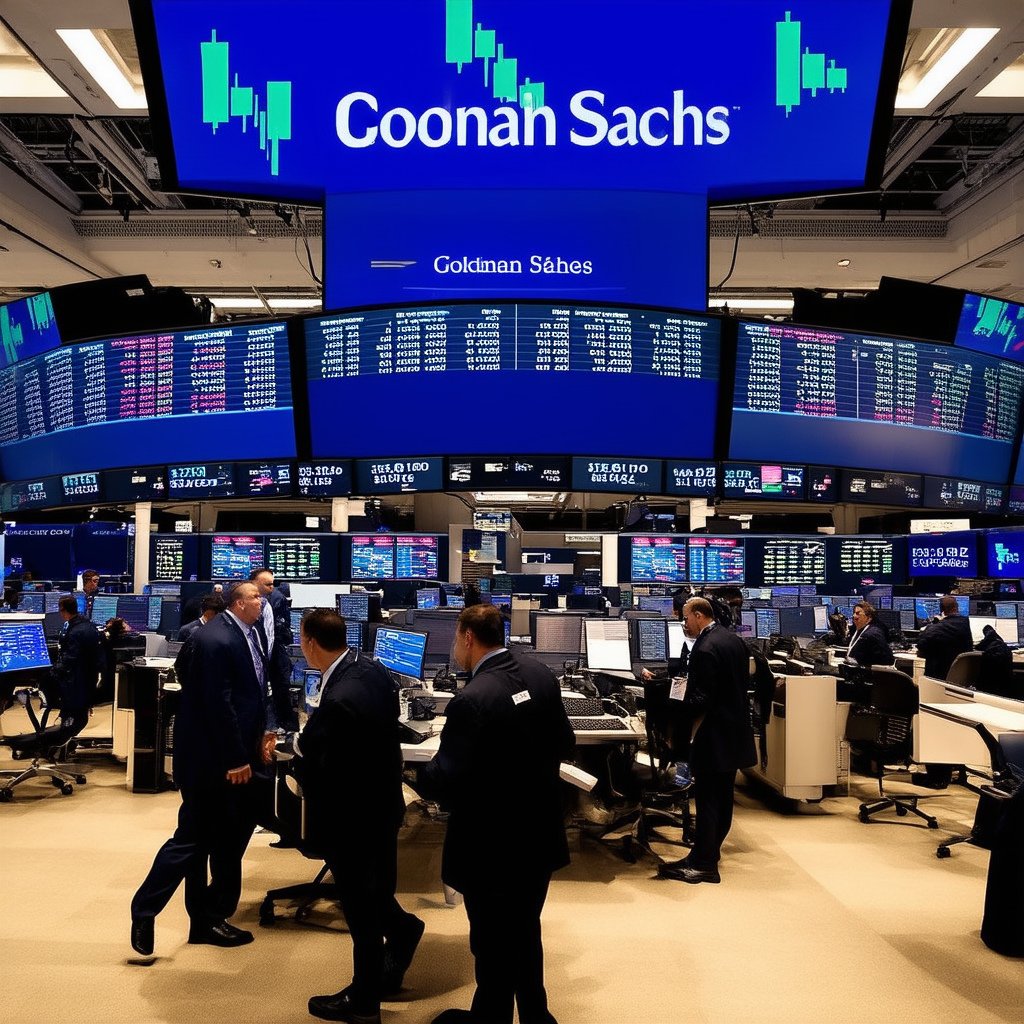} &
        \includegraphics[width=0.1\textwidth]{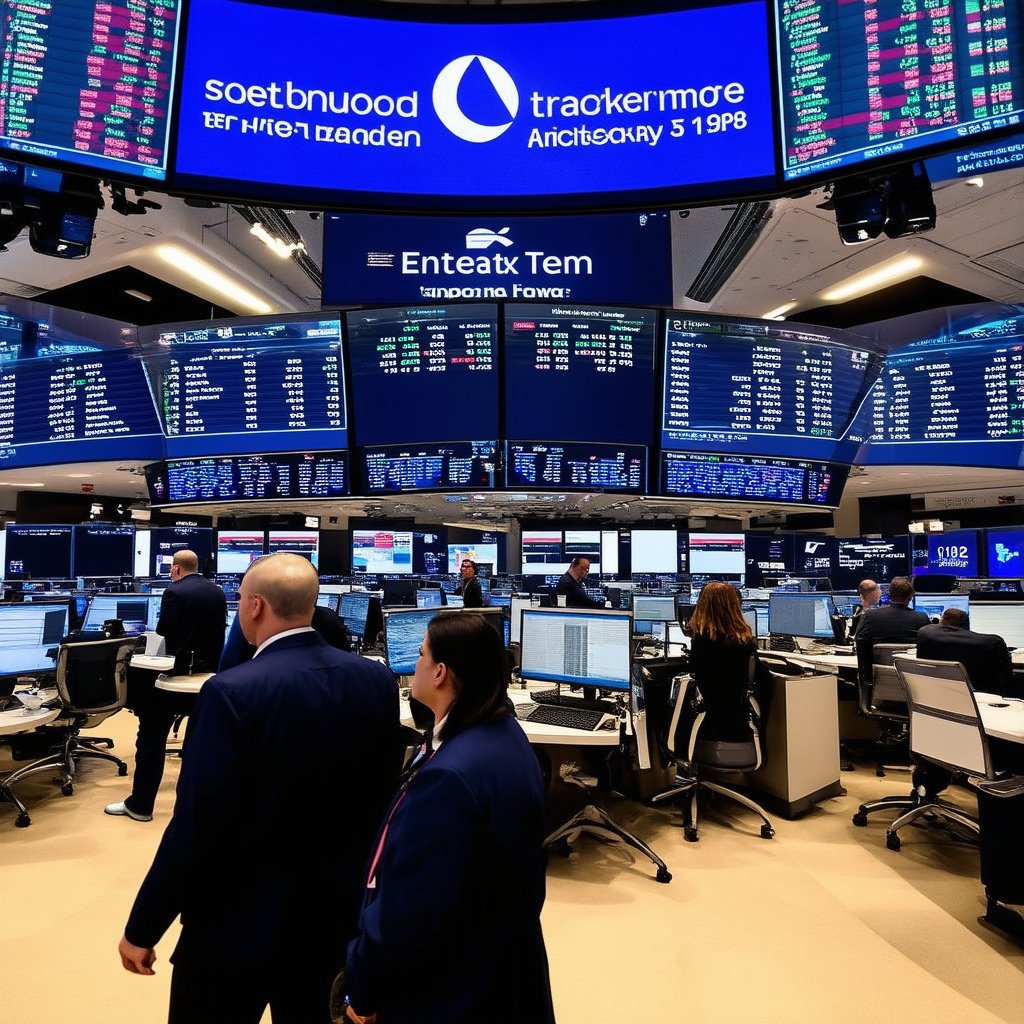} \\

        \includegraphics[width=0.1\textwidth]{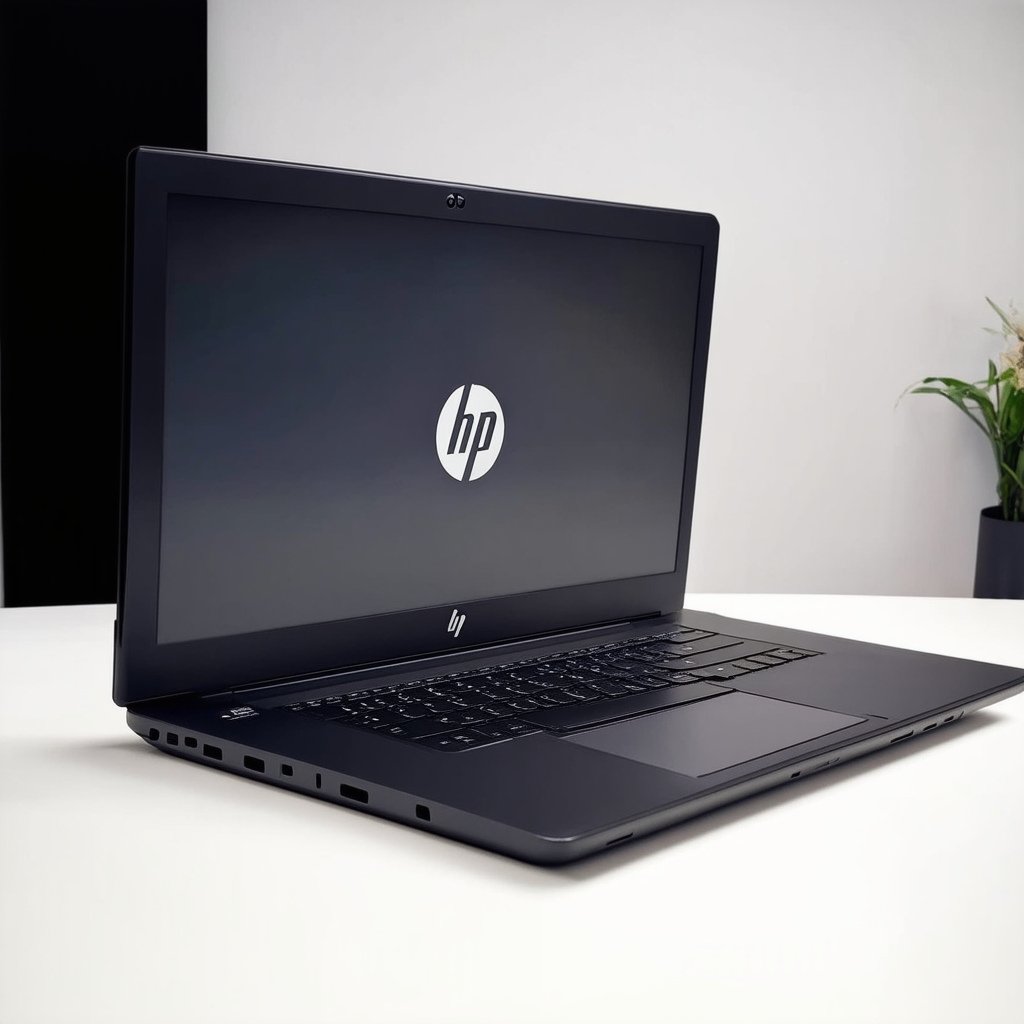} &
        \includegraphics[width=0.1\textwidth]{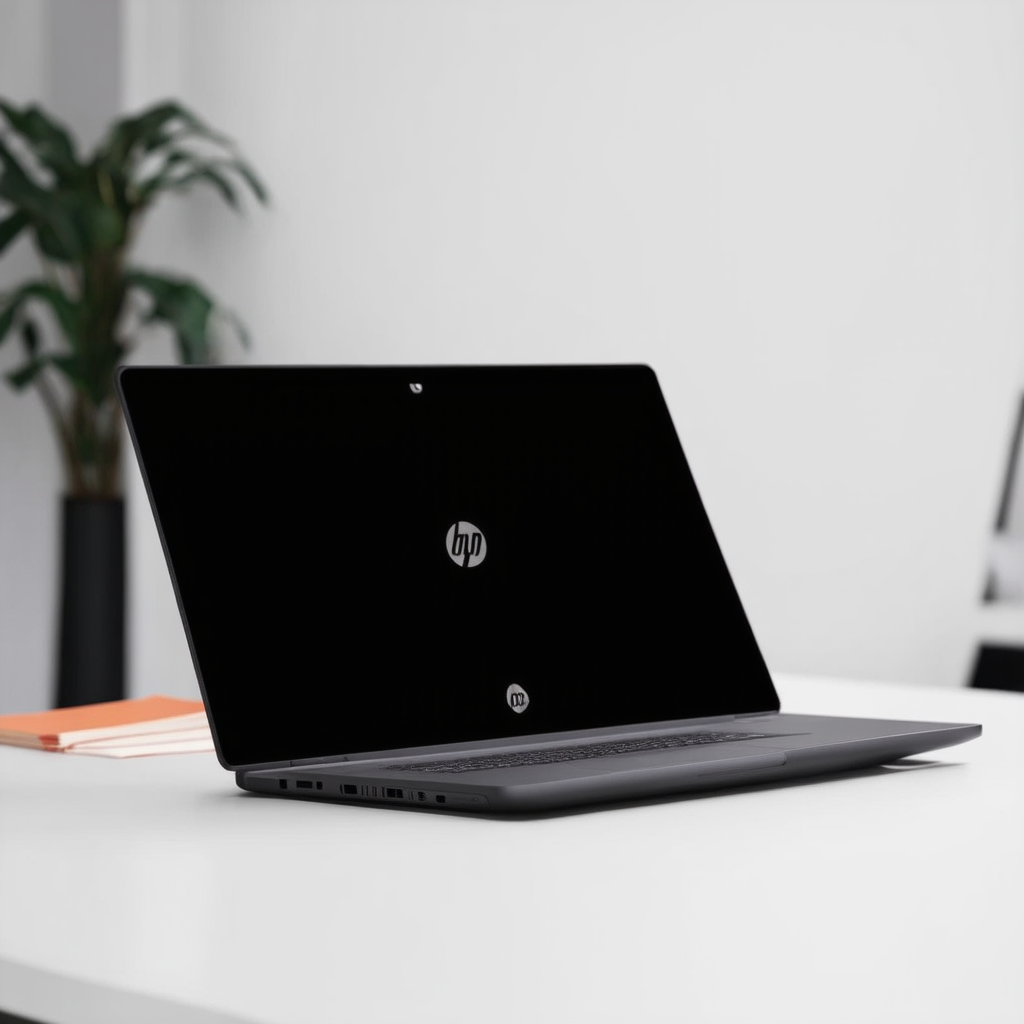} &
        \includegraphics[width=0.1\textwidth]{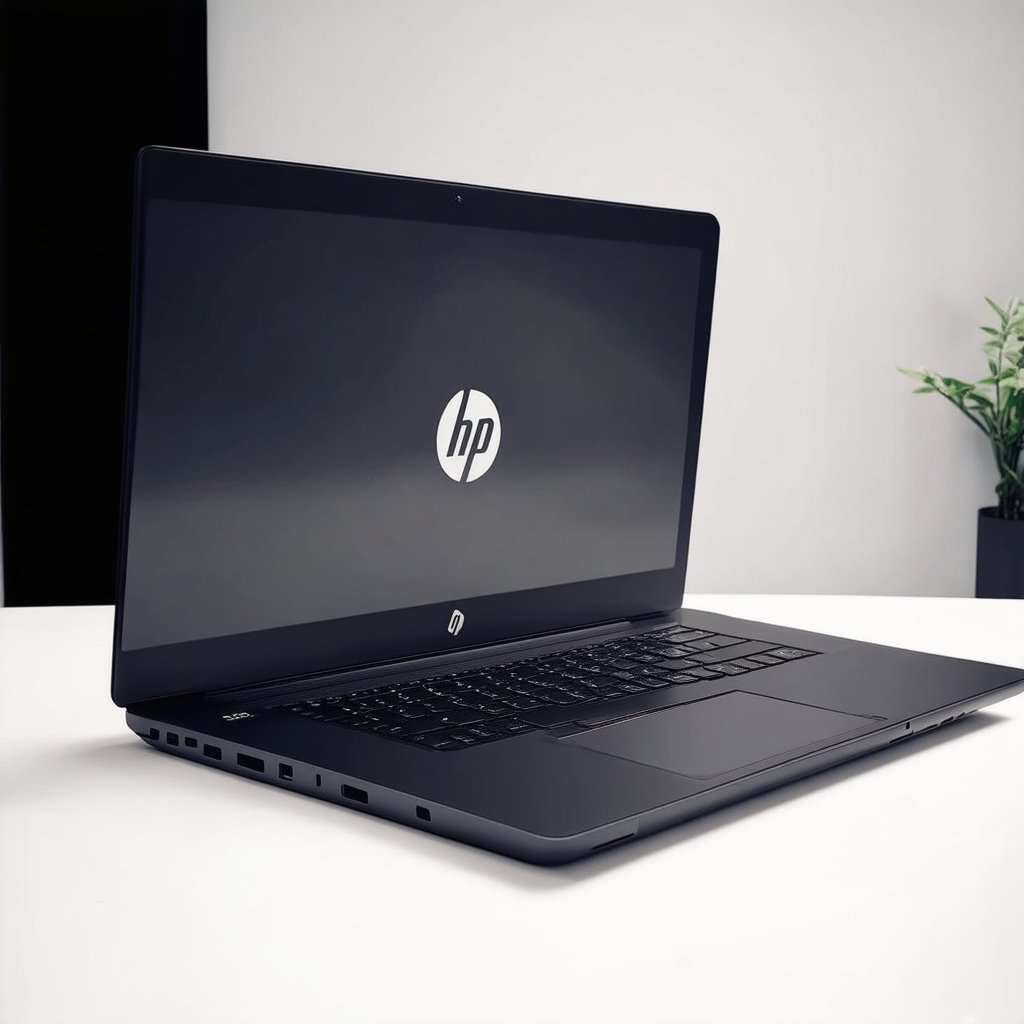} &
                \includegraphics[width=0.1\textwidth]{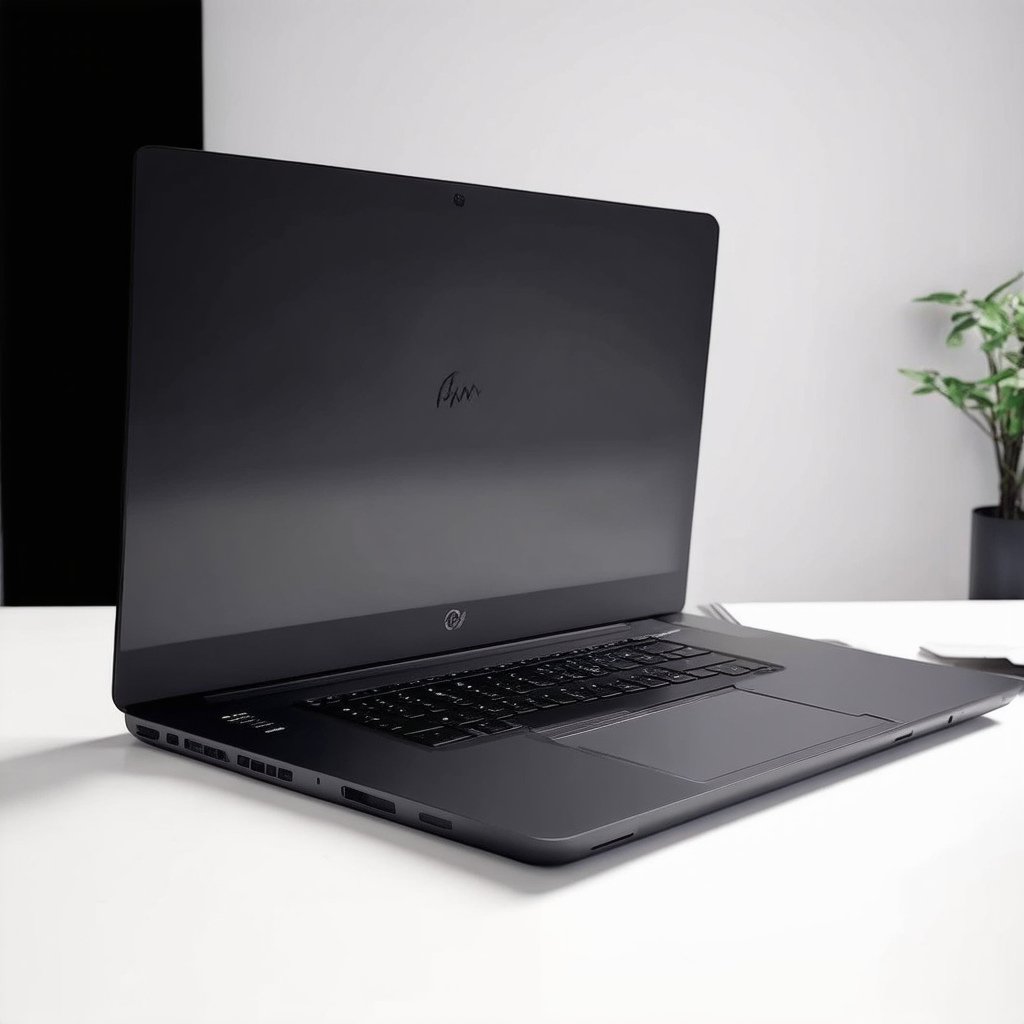} &
                        \includegraphics[width=0.1\textwidth]{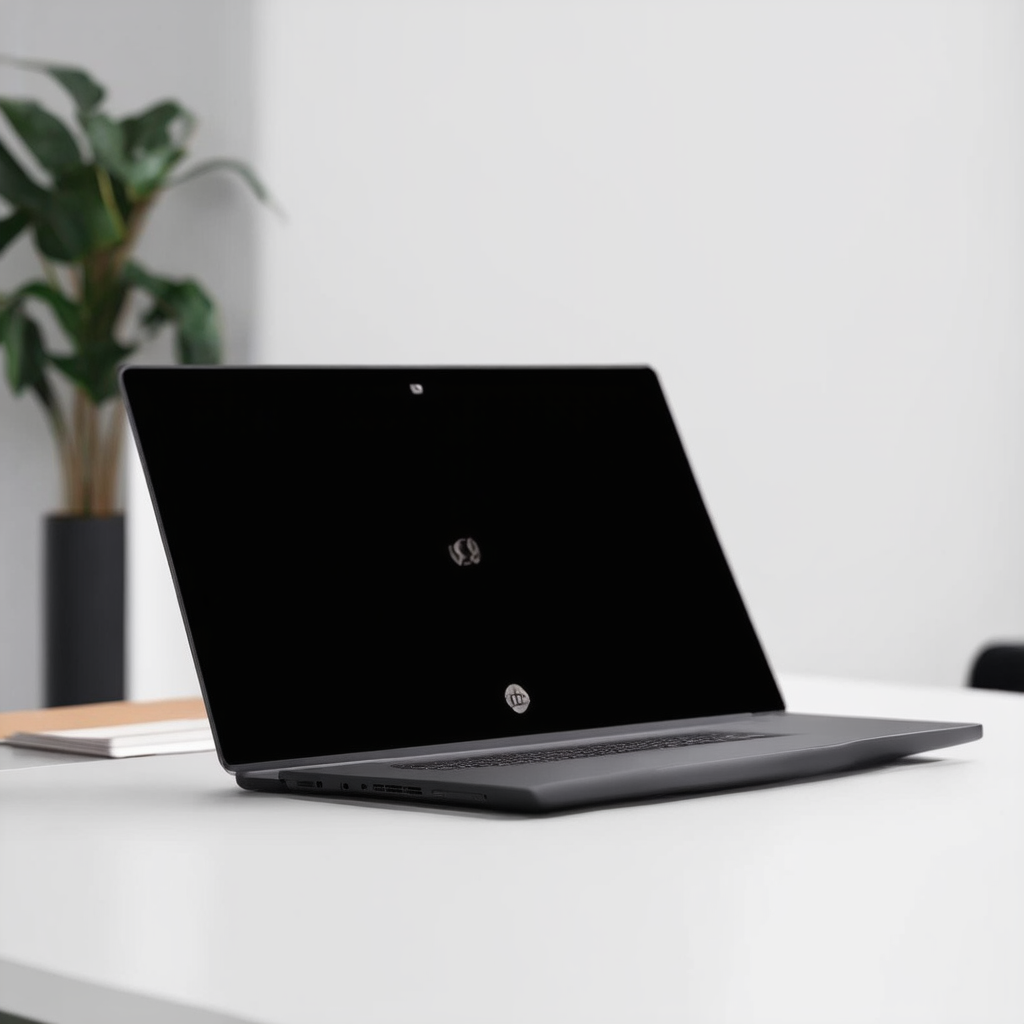} &
        \includegraphics[width=0.1\textwidth]{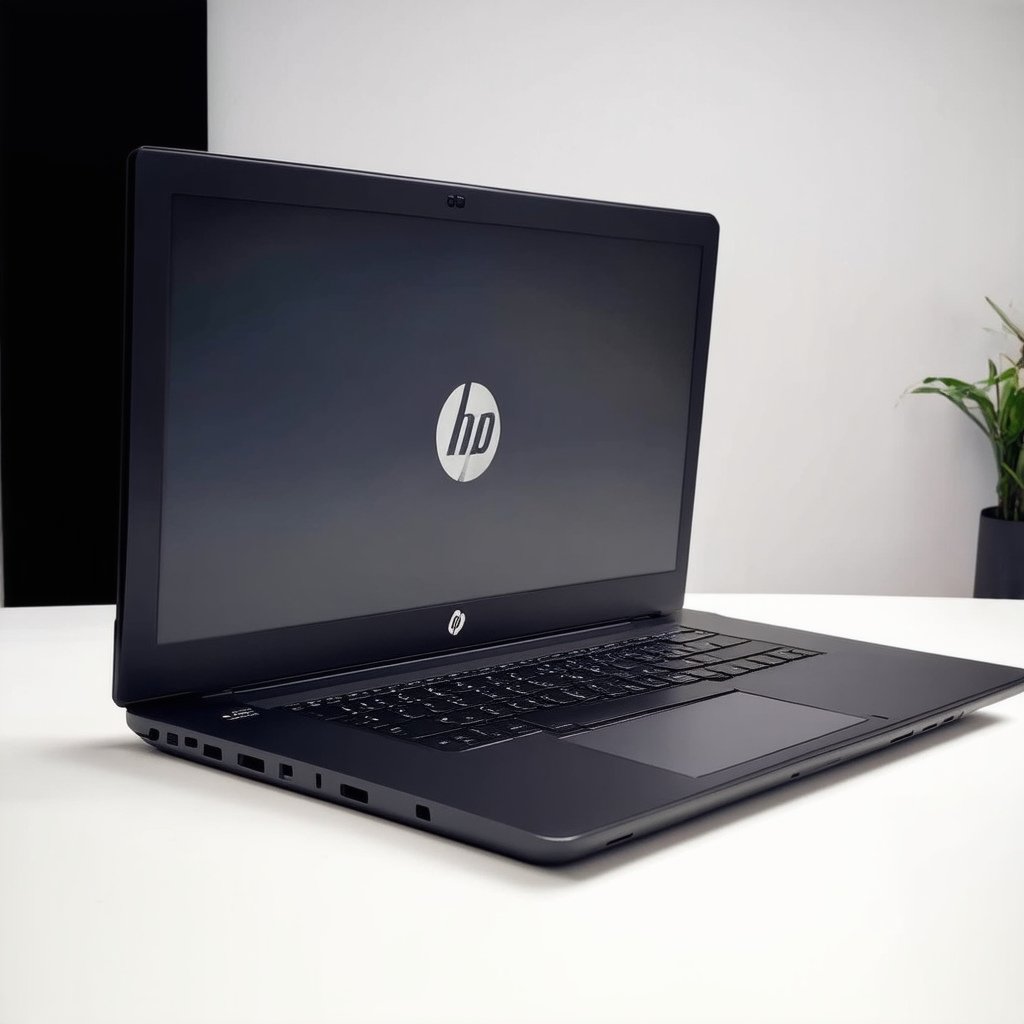} &
                \includegraphics[width=0.1\textwidth]{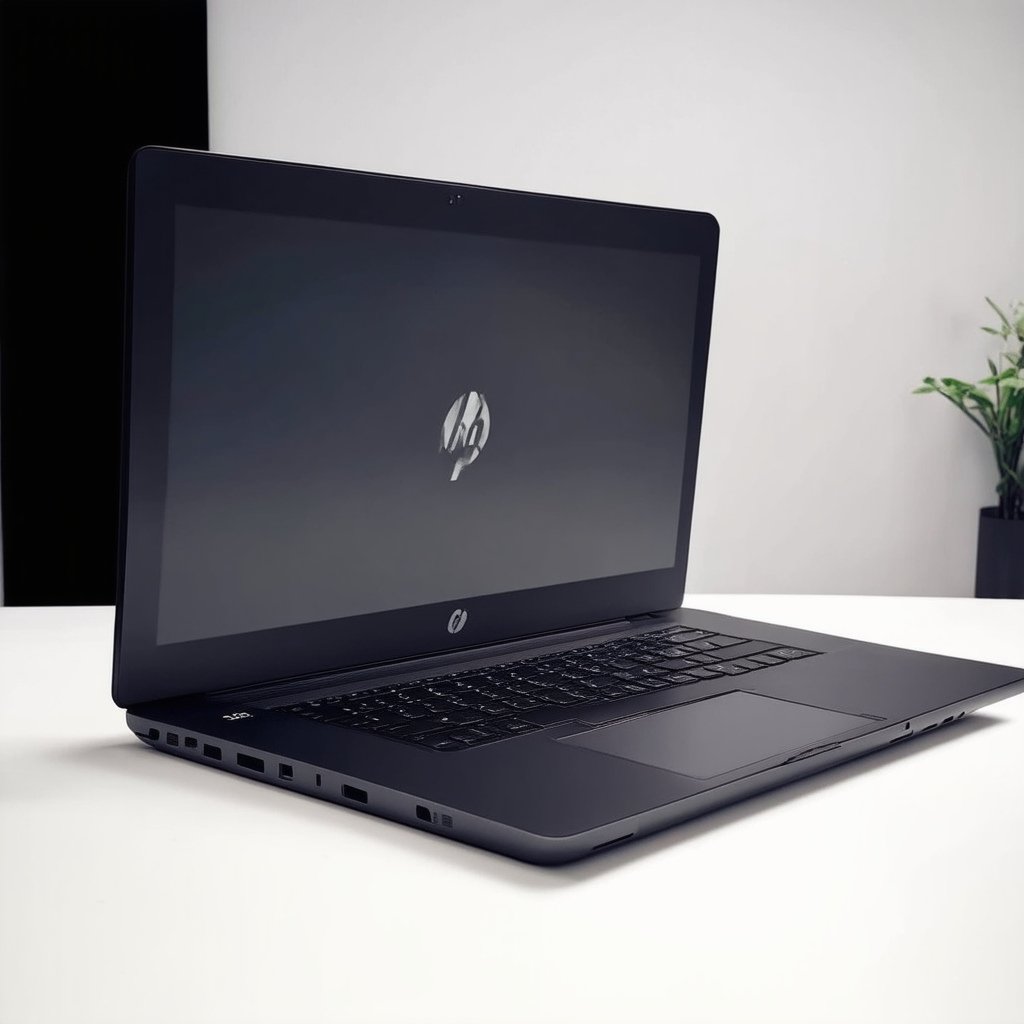} &
                        \includegraphics[width=0.1\textwidth]{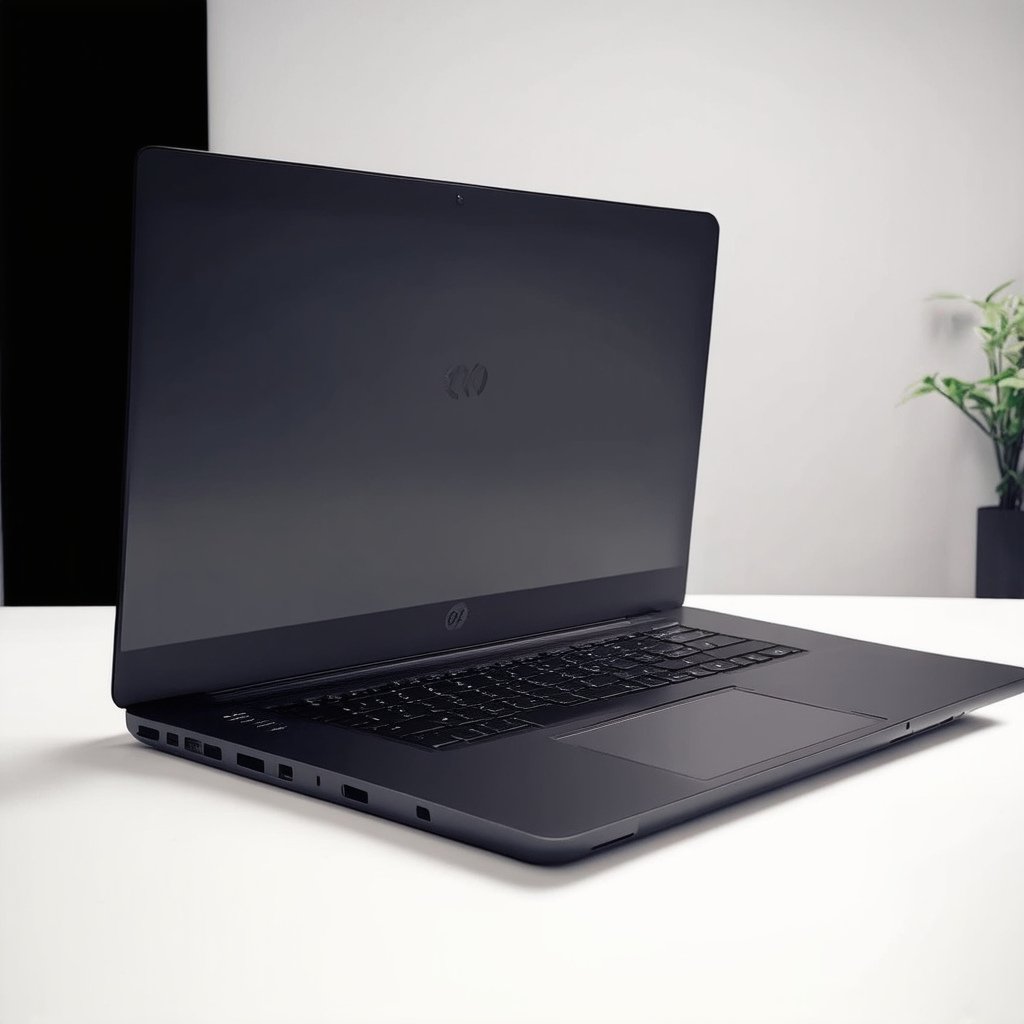} &
        \includegraphics[width=0.1\textwidth]{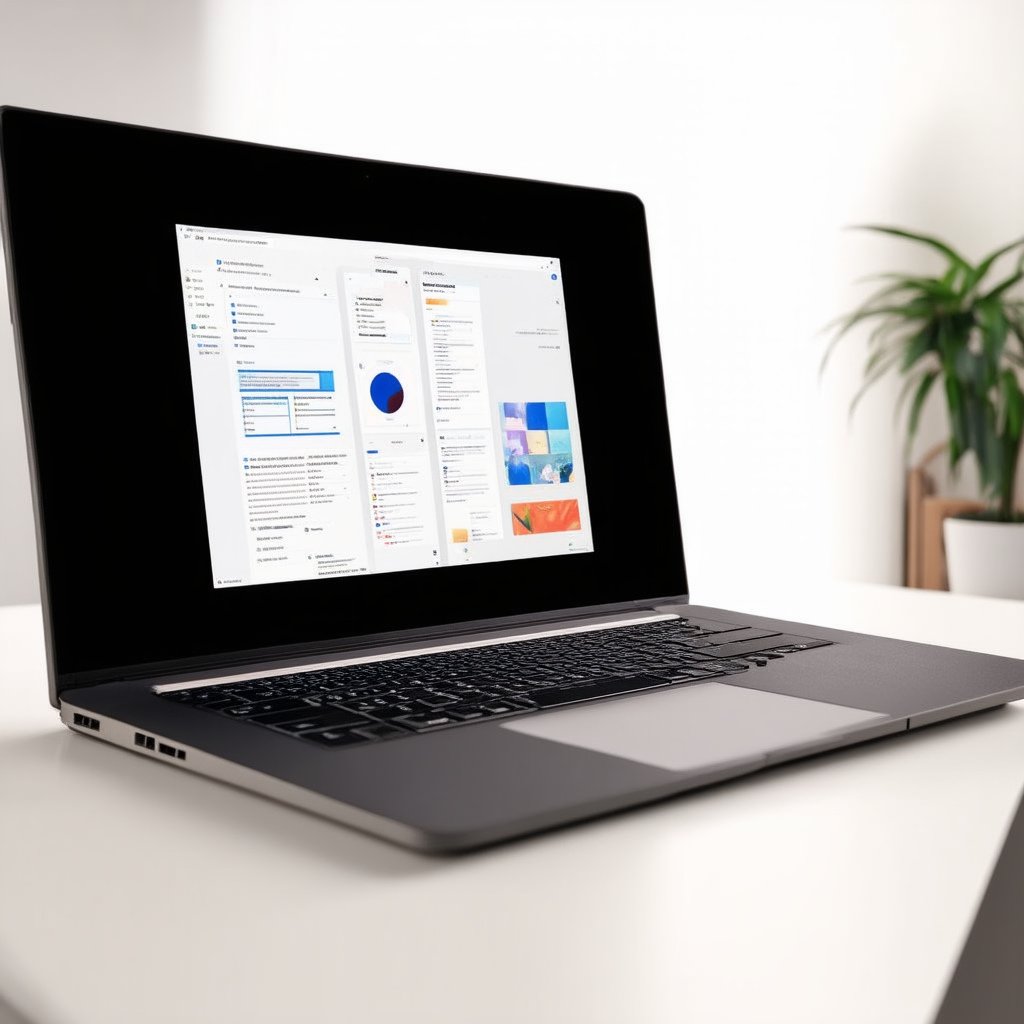} \\

        \includegraphics[width=0.1\textwidth]{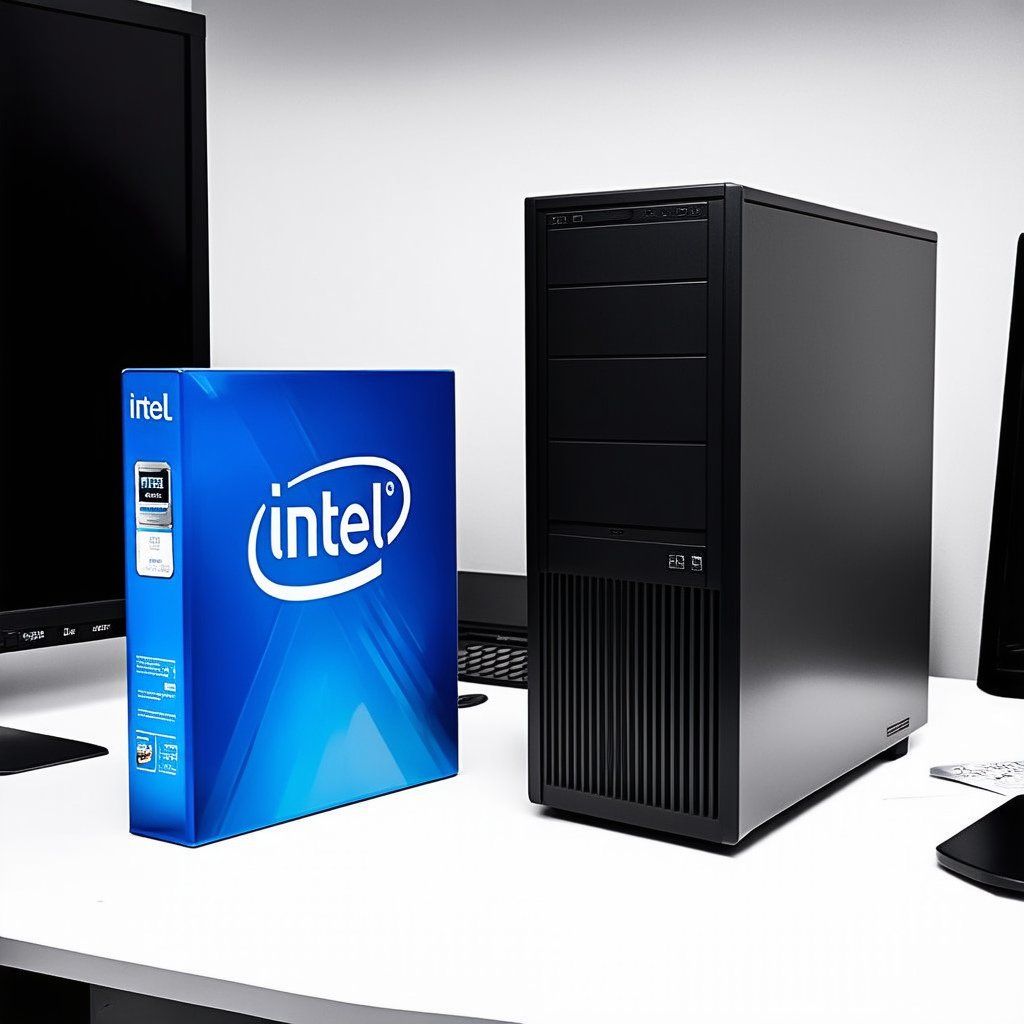} &
        \includegraphics[width=0.1\textwidth]{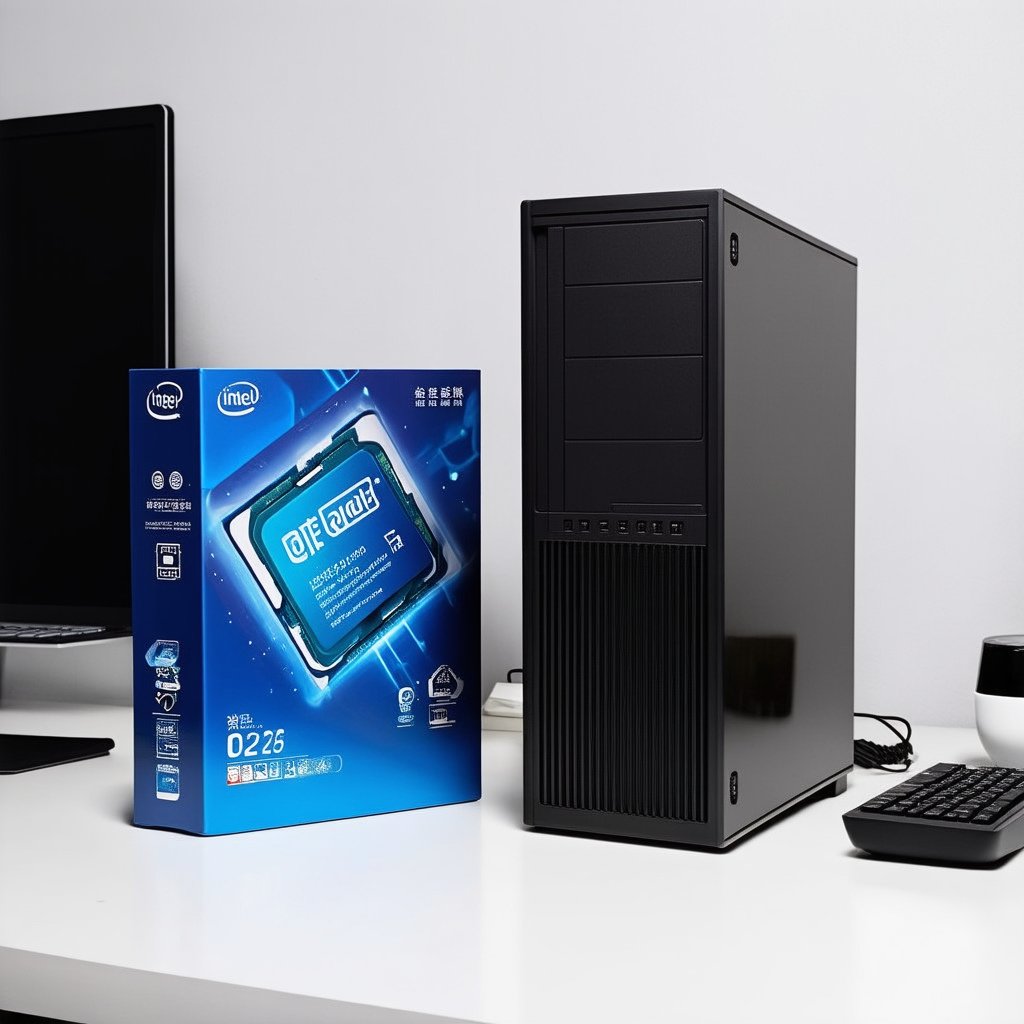} &
        \includegraphics[width=0.1\textwidth]{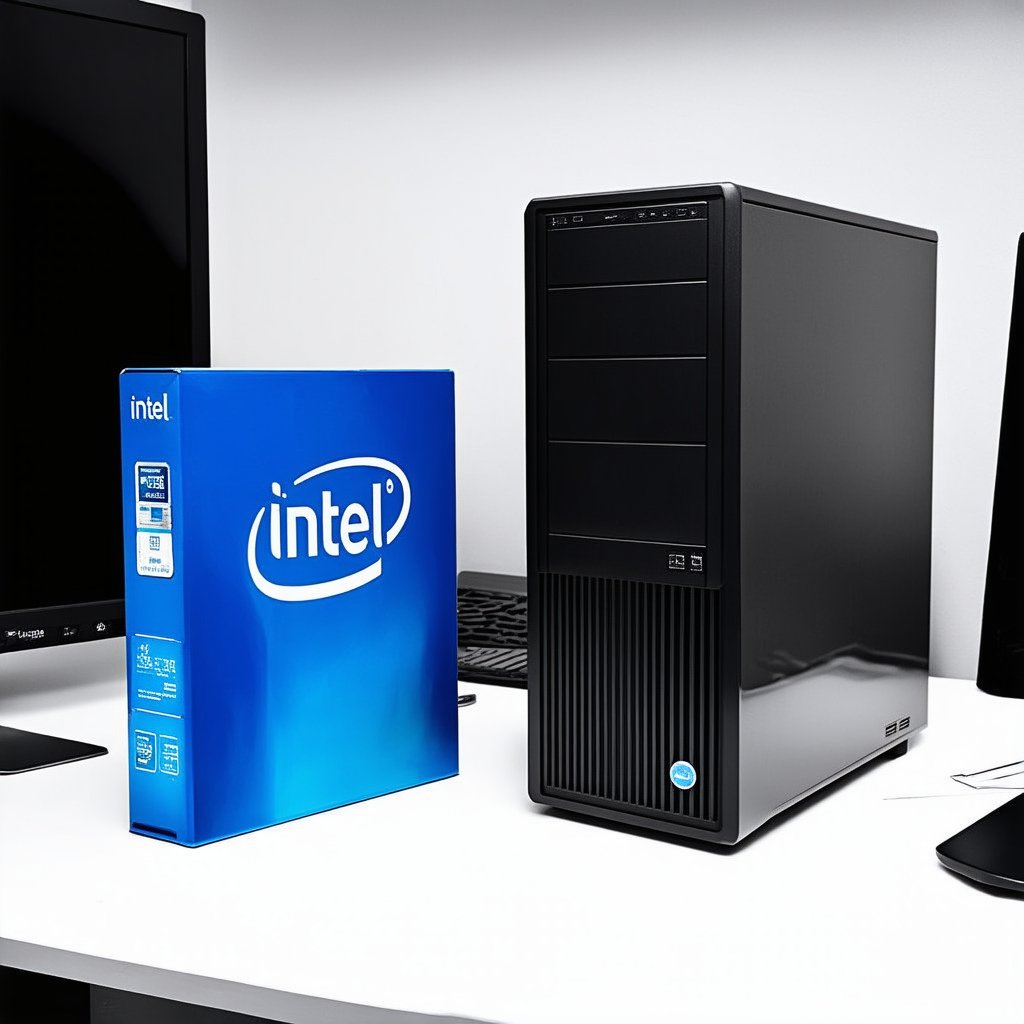} &
                \includegraphics[width=0.1\textwidth]{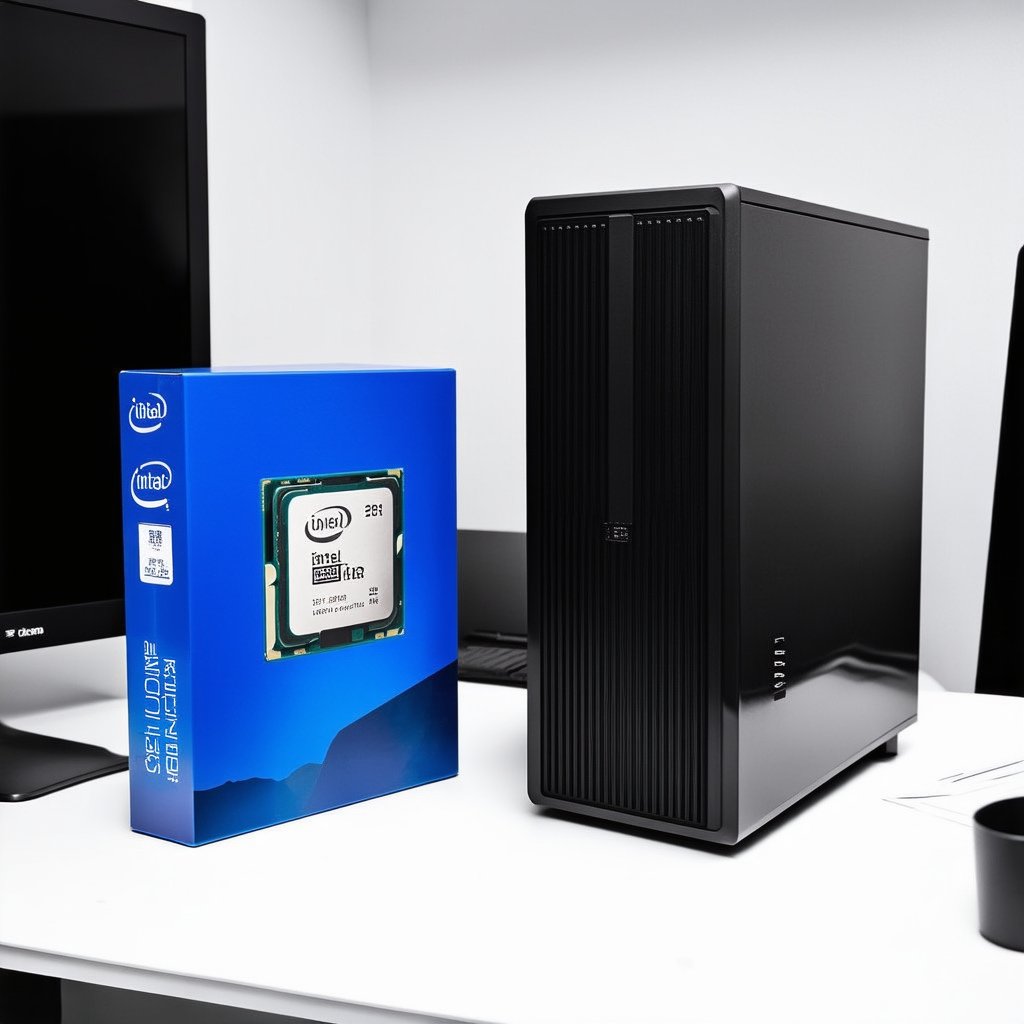} &
                        \includegraphics[width=0.1\textwidth]{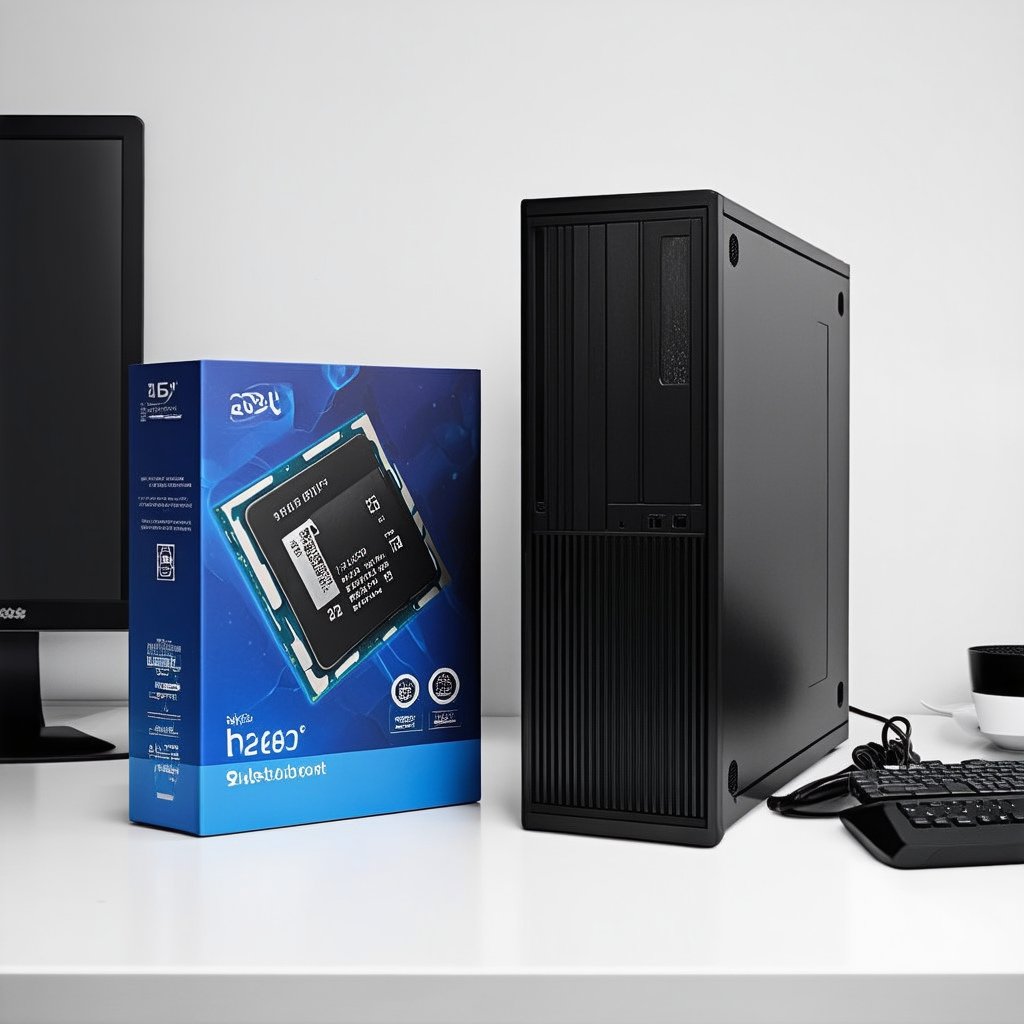} &
        \includegraphics[width=0.1\textwidth]{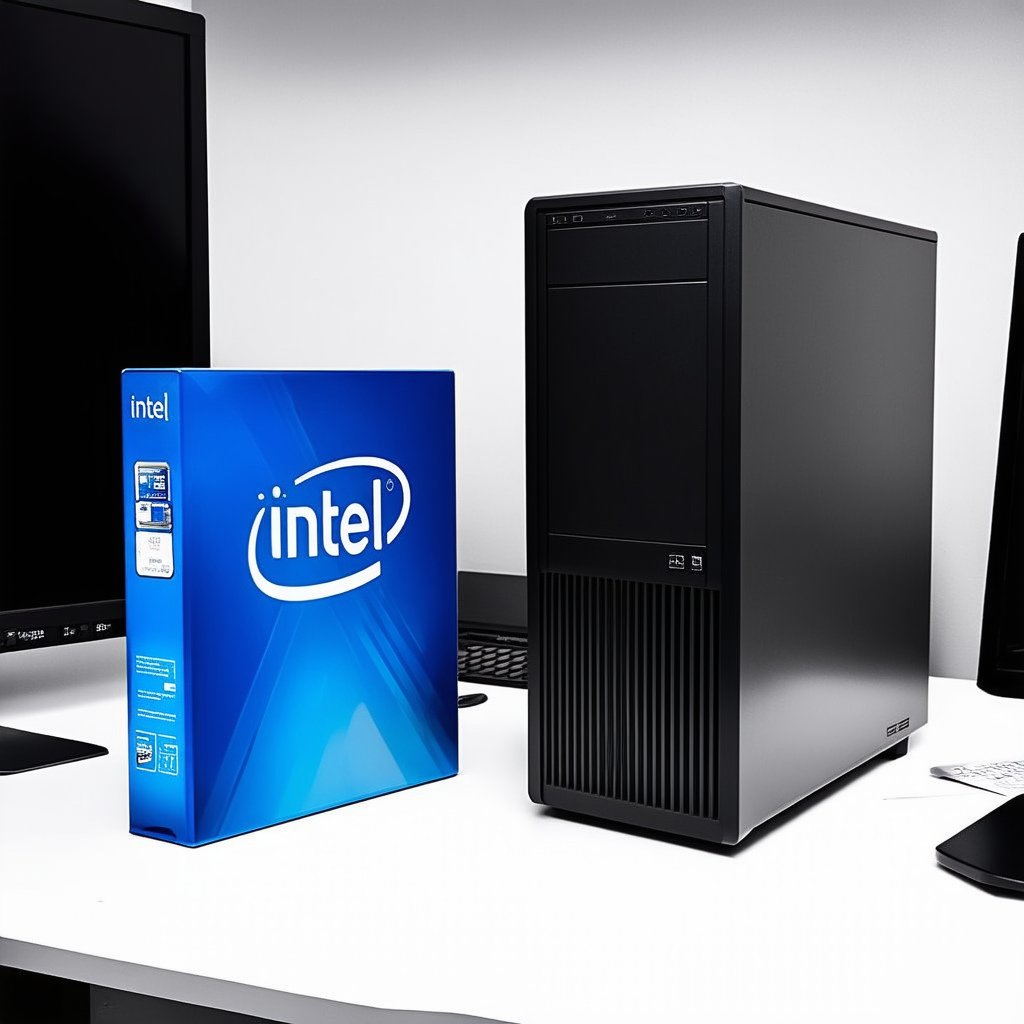} &
                \includegraphics[width=0.1\textwidth]{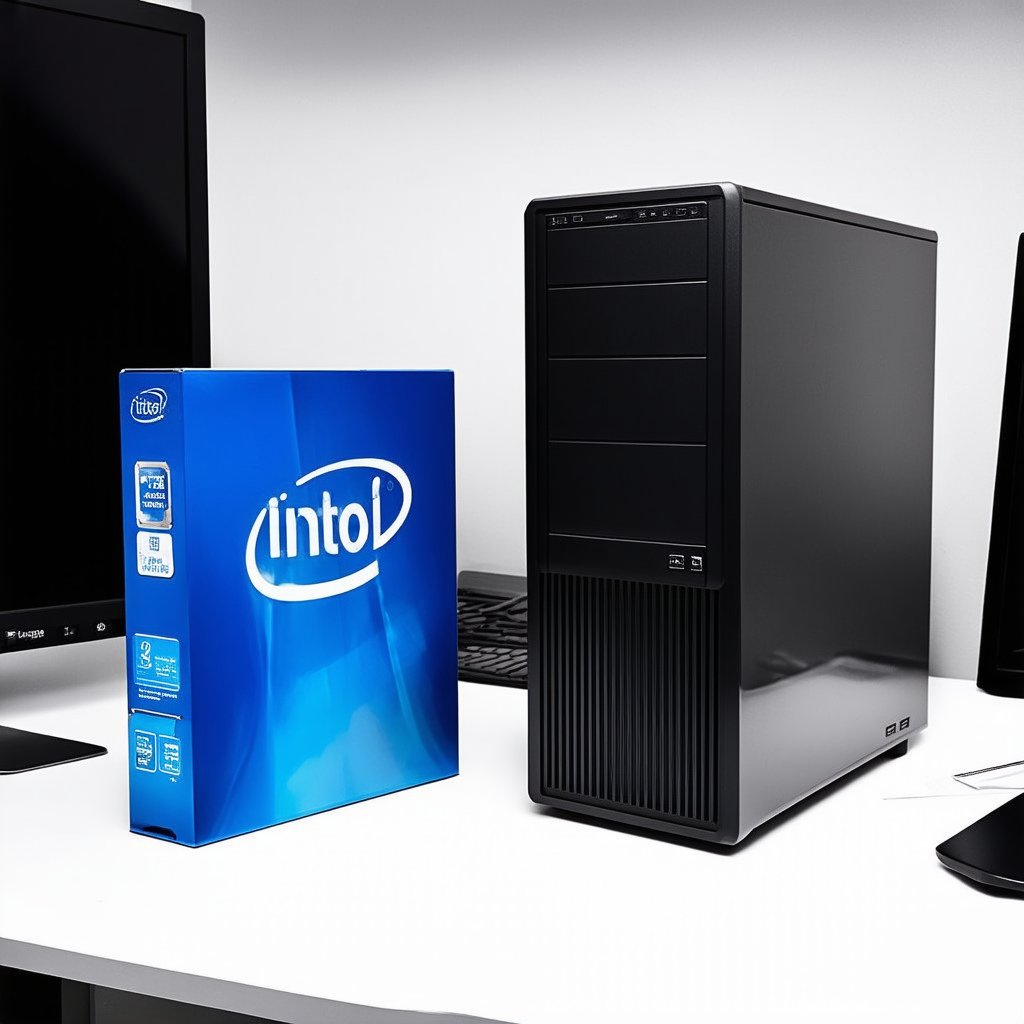} &
                        \includegraphics[width=0.1\textwidth]{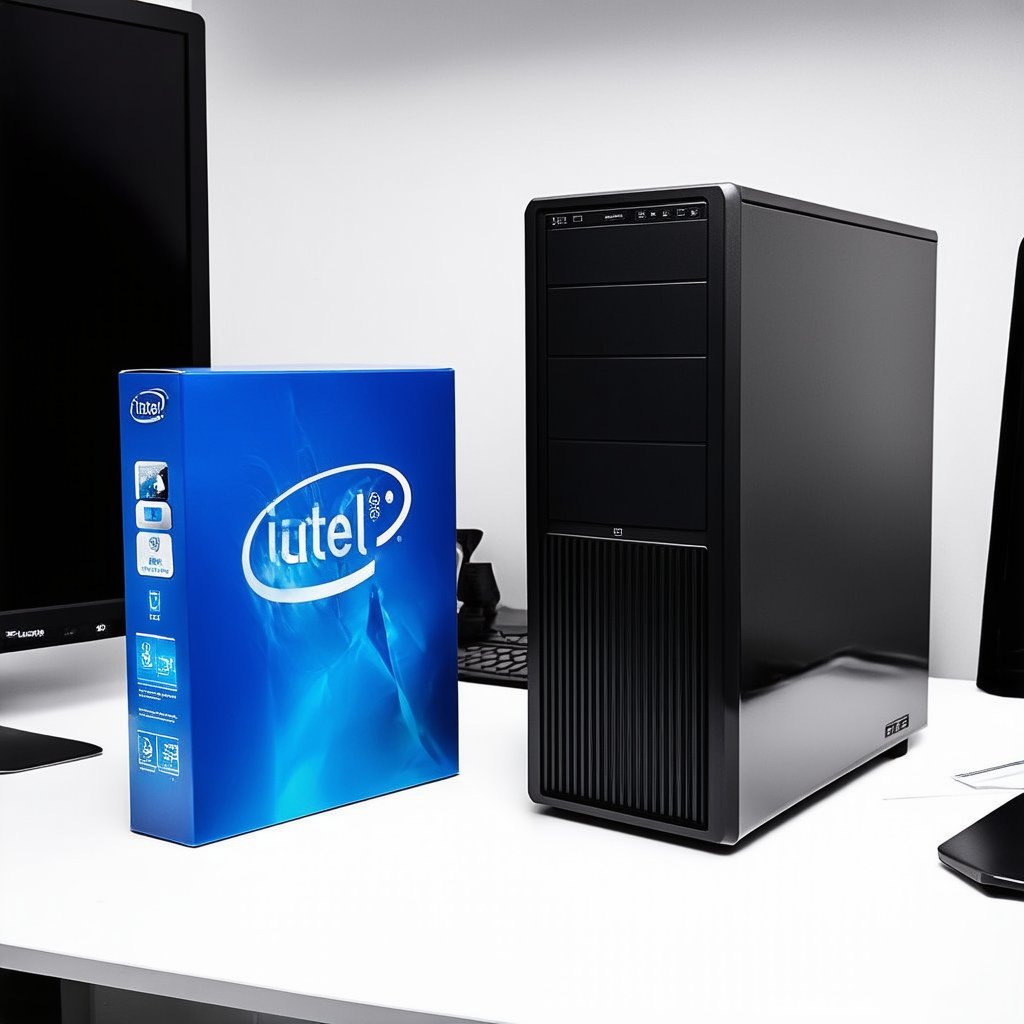} &
        \includegraphics[width=0.1\textwidth]{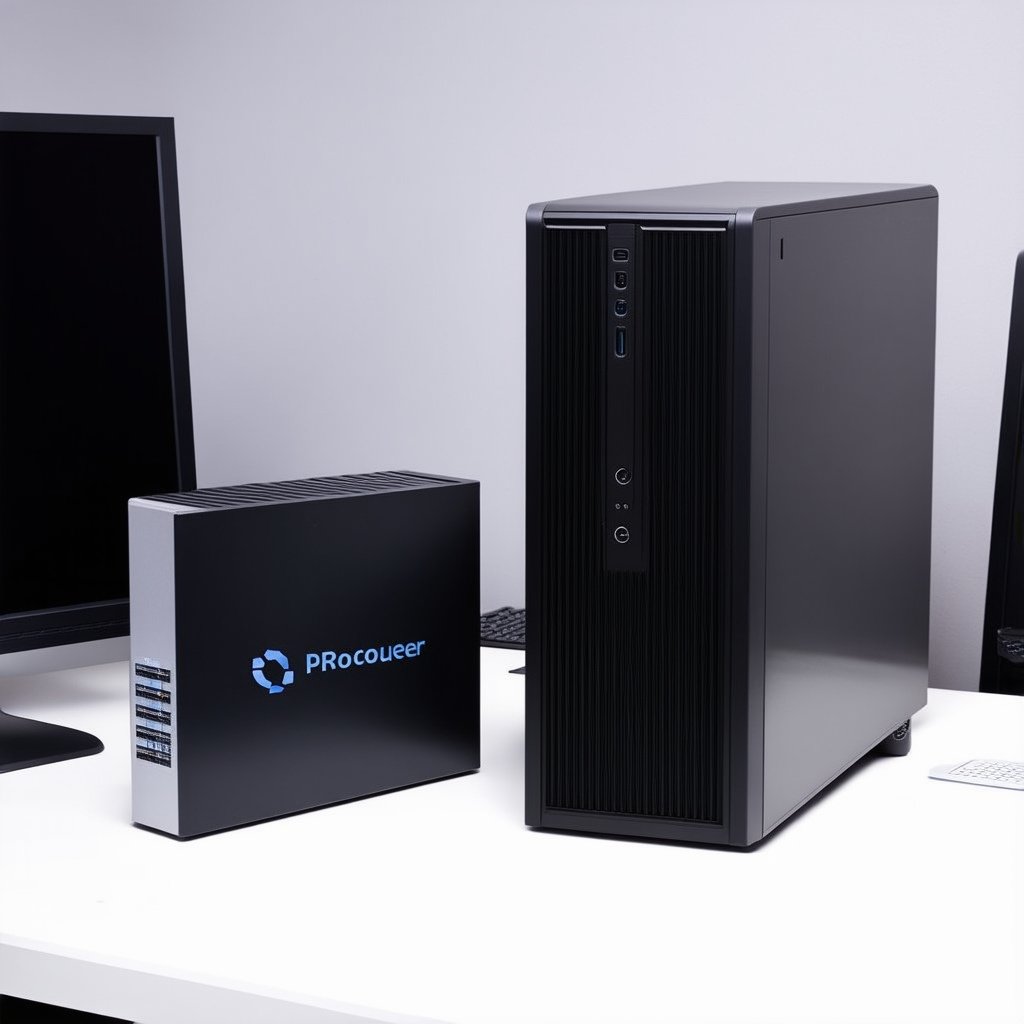} \\

        \includegraphics[width=0.1\textwidth]{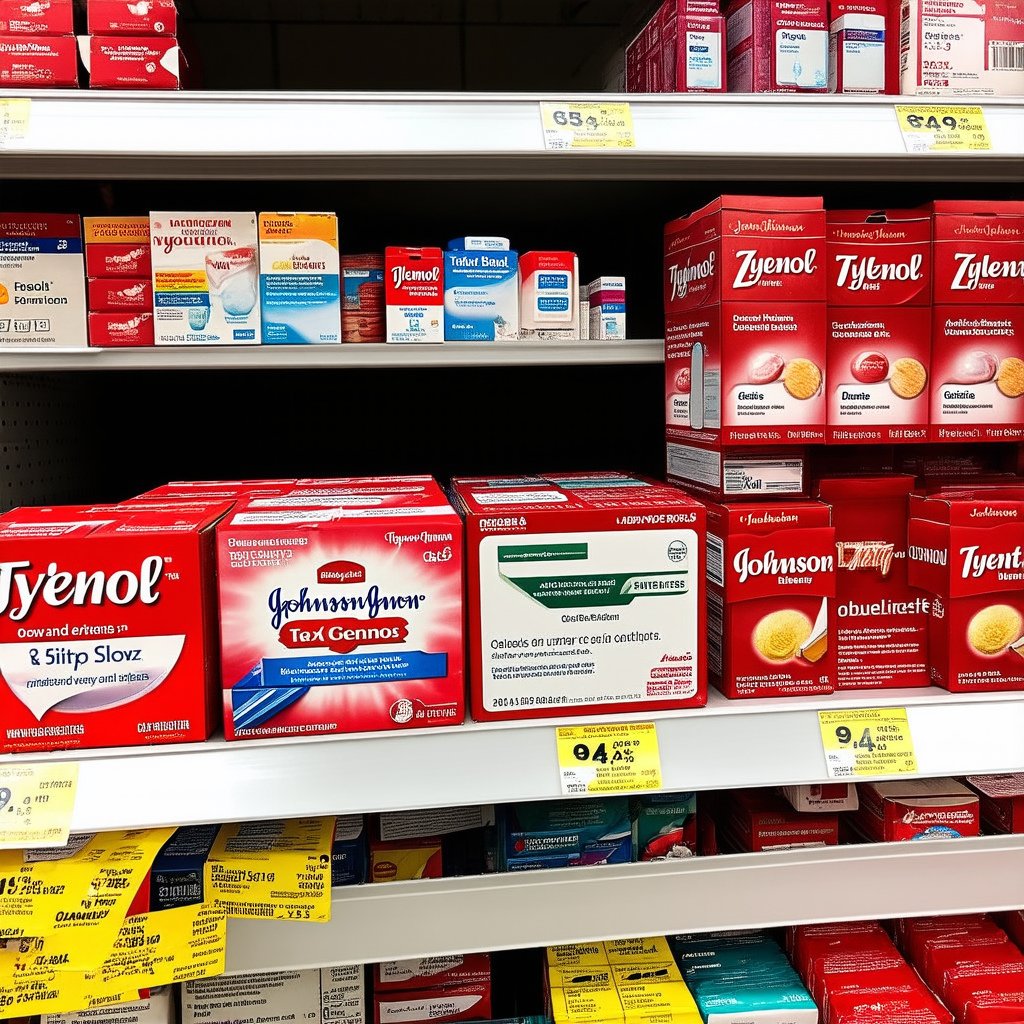} &
        \includegraphics[width=0.1\textwidth]{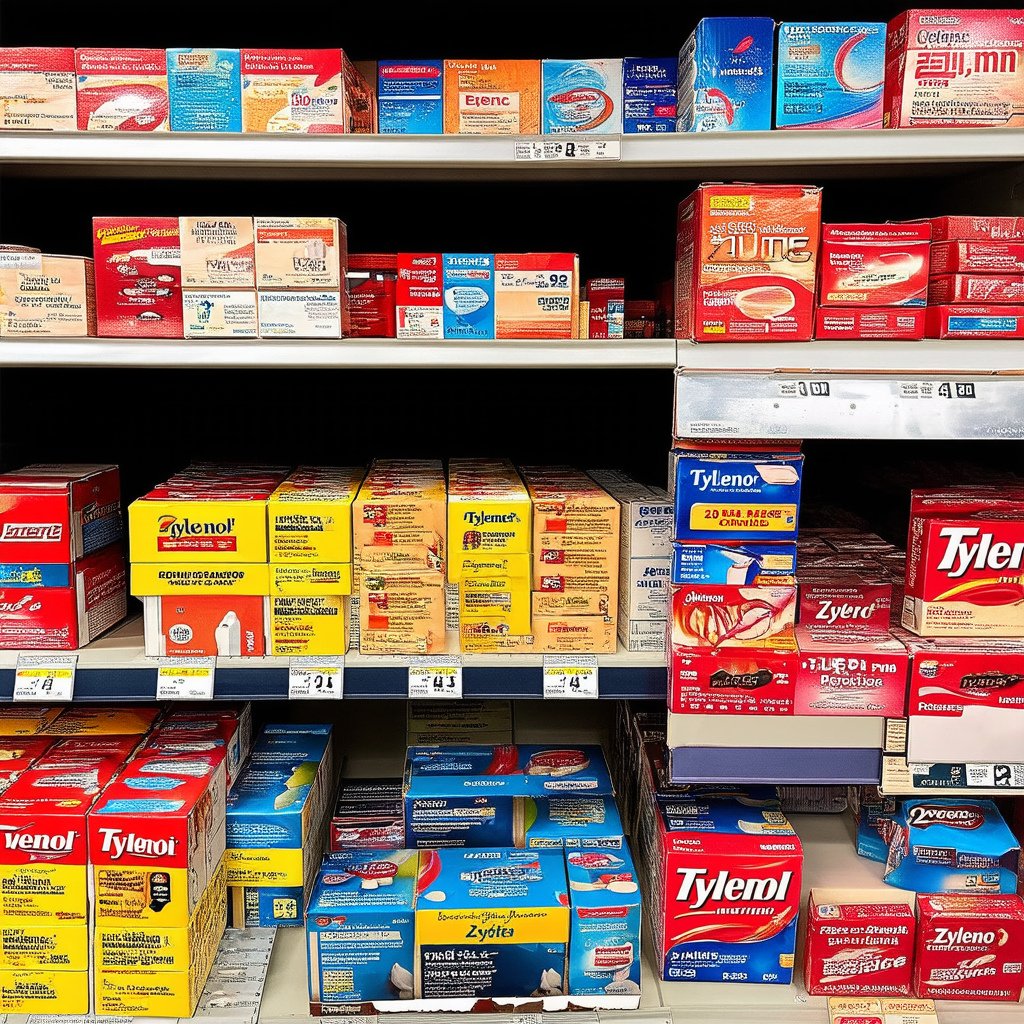} &
        \includegraphics[width=0.1\textwidth]{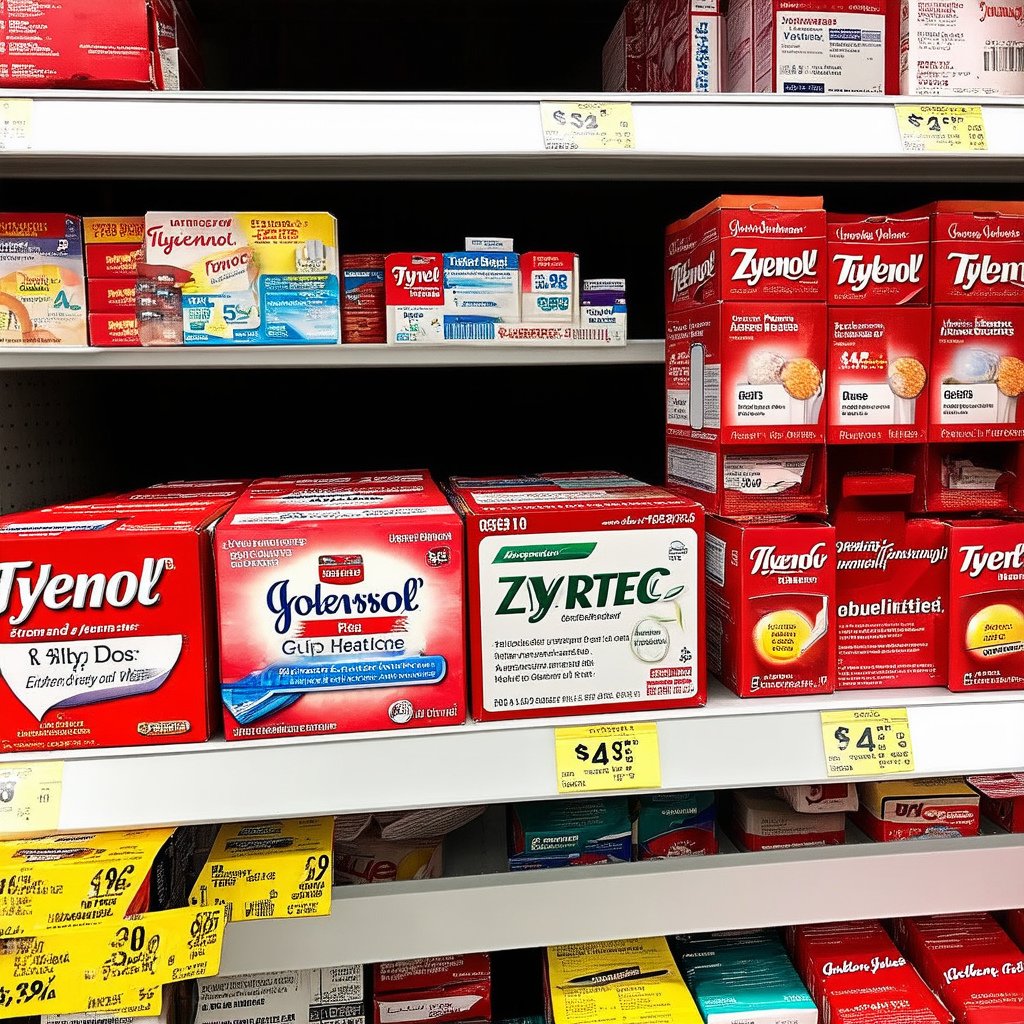} &
                \includegraphics[width=0.1\textwidth]{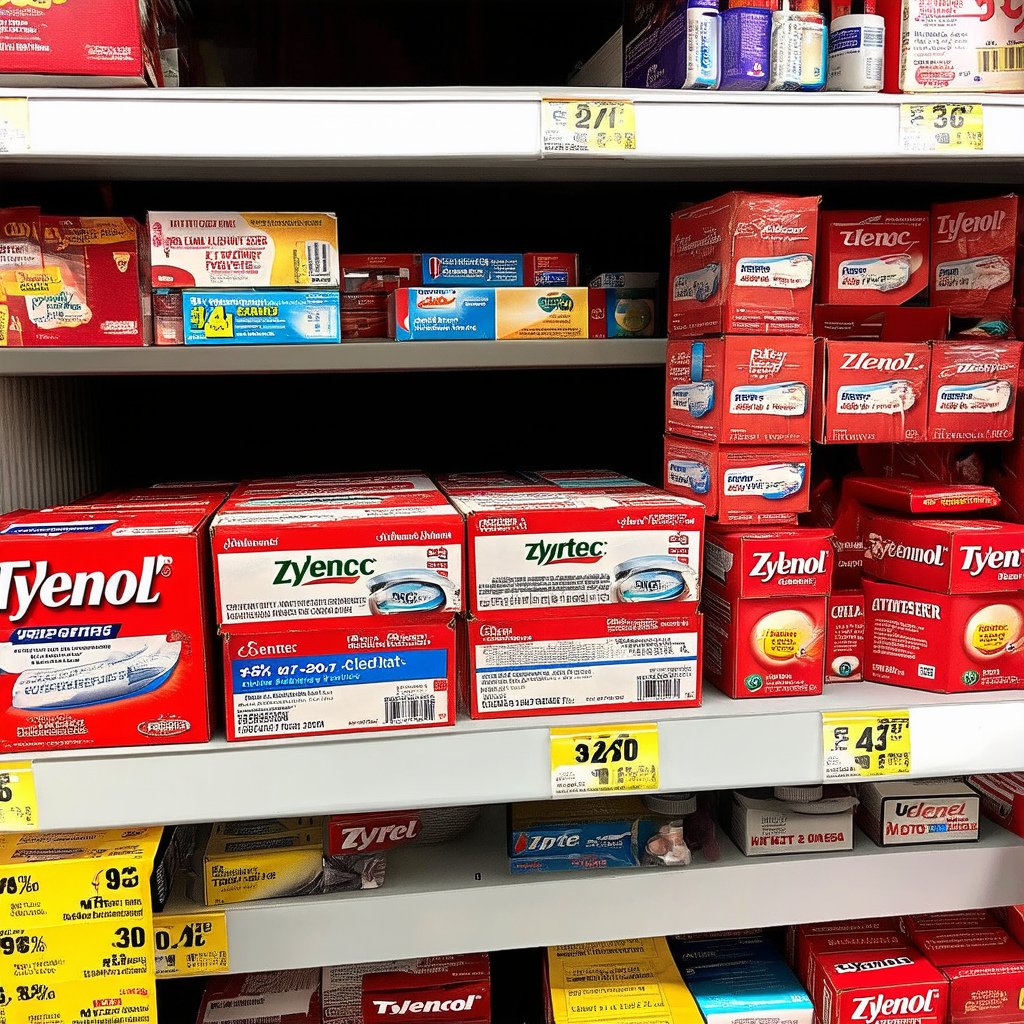} &
                        \includegraphics[width=0.1\textwidth]{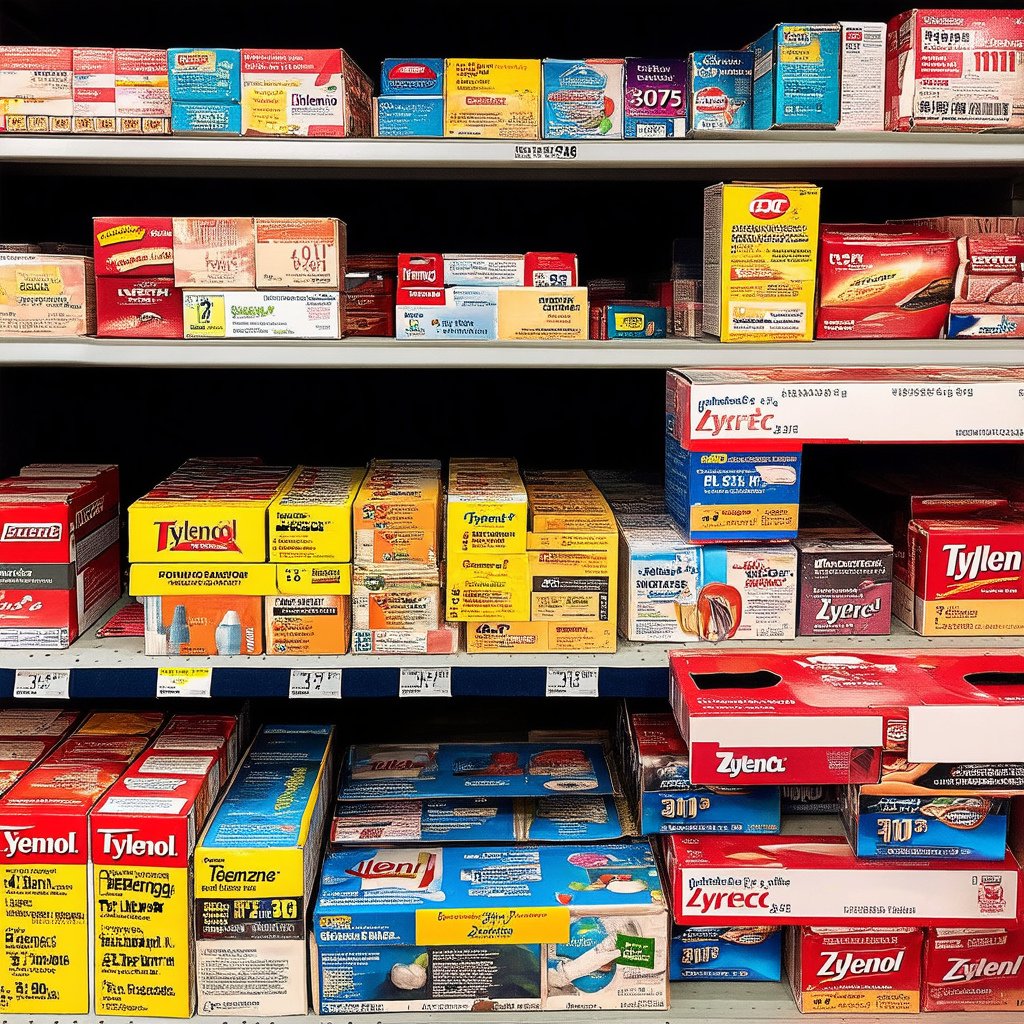} &
        \includegraphics[width=0.1\textwidth]{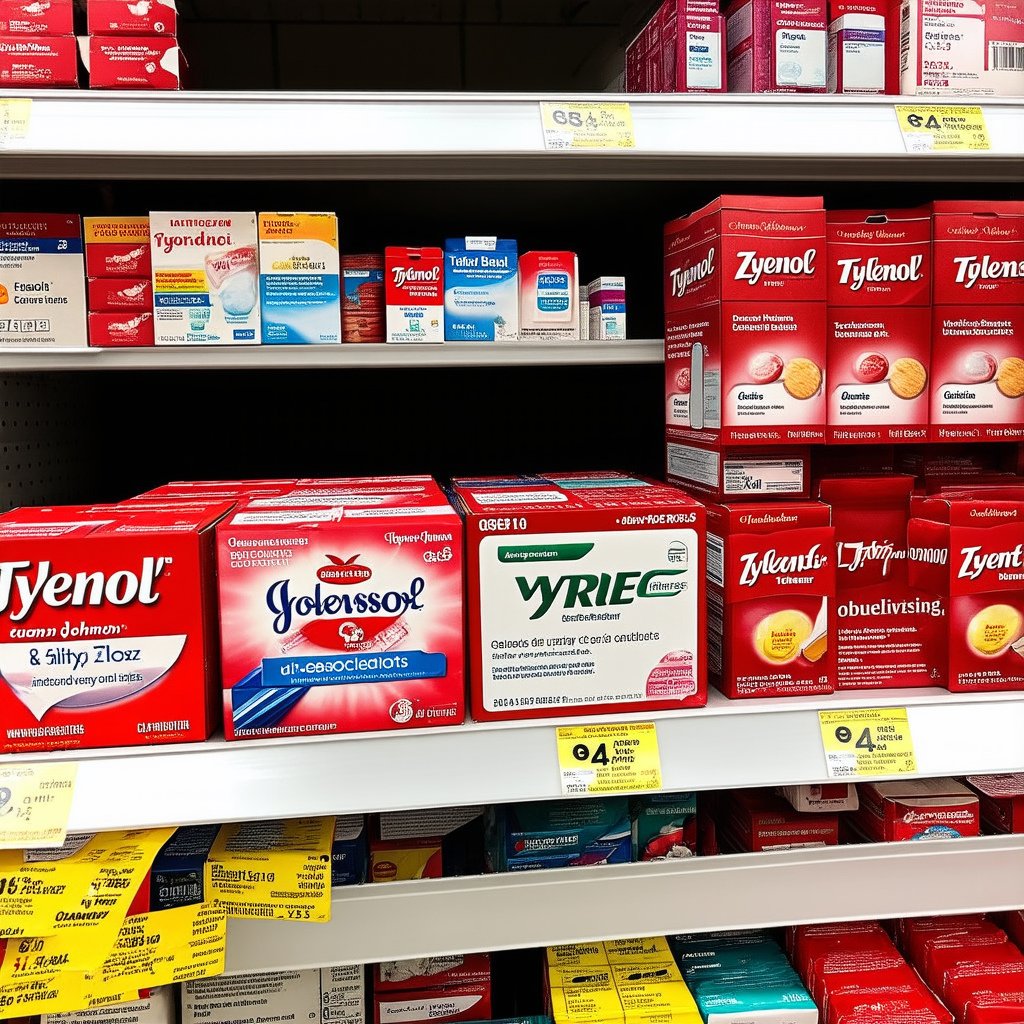} &
                \includegraphics[width=0.1\textwidth]{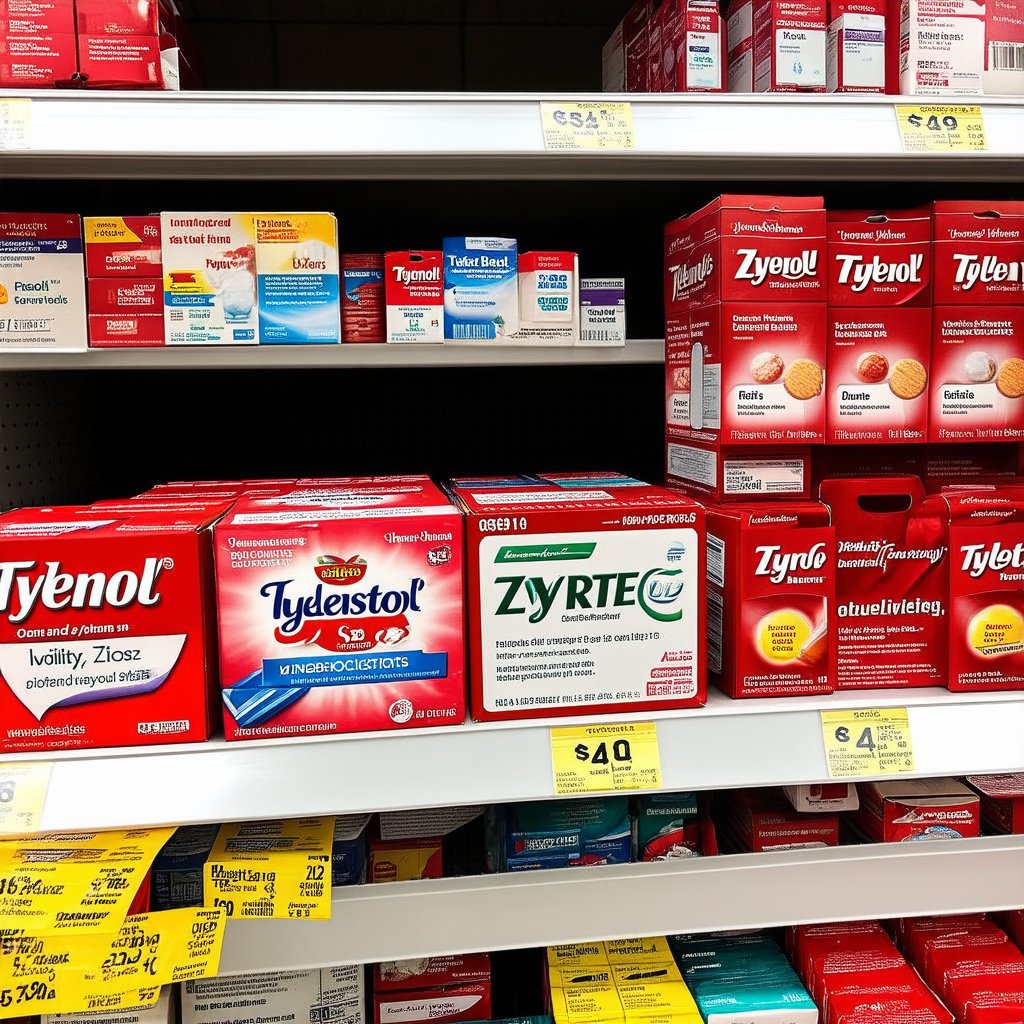} &
                        \includegraphics[width=0.1\textwidth]{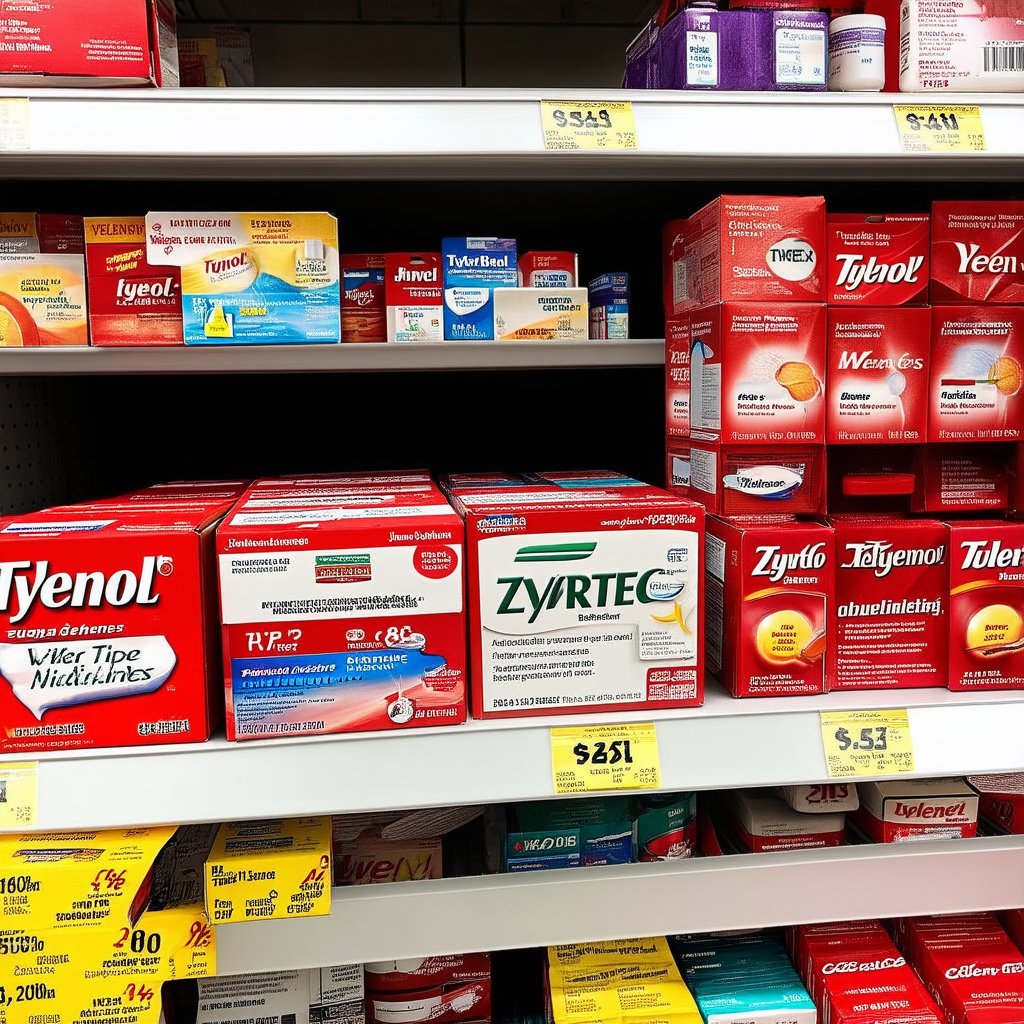} &
        \includegraphics[width=0.1\textwidth]{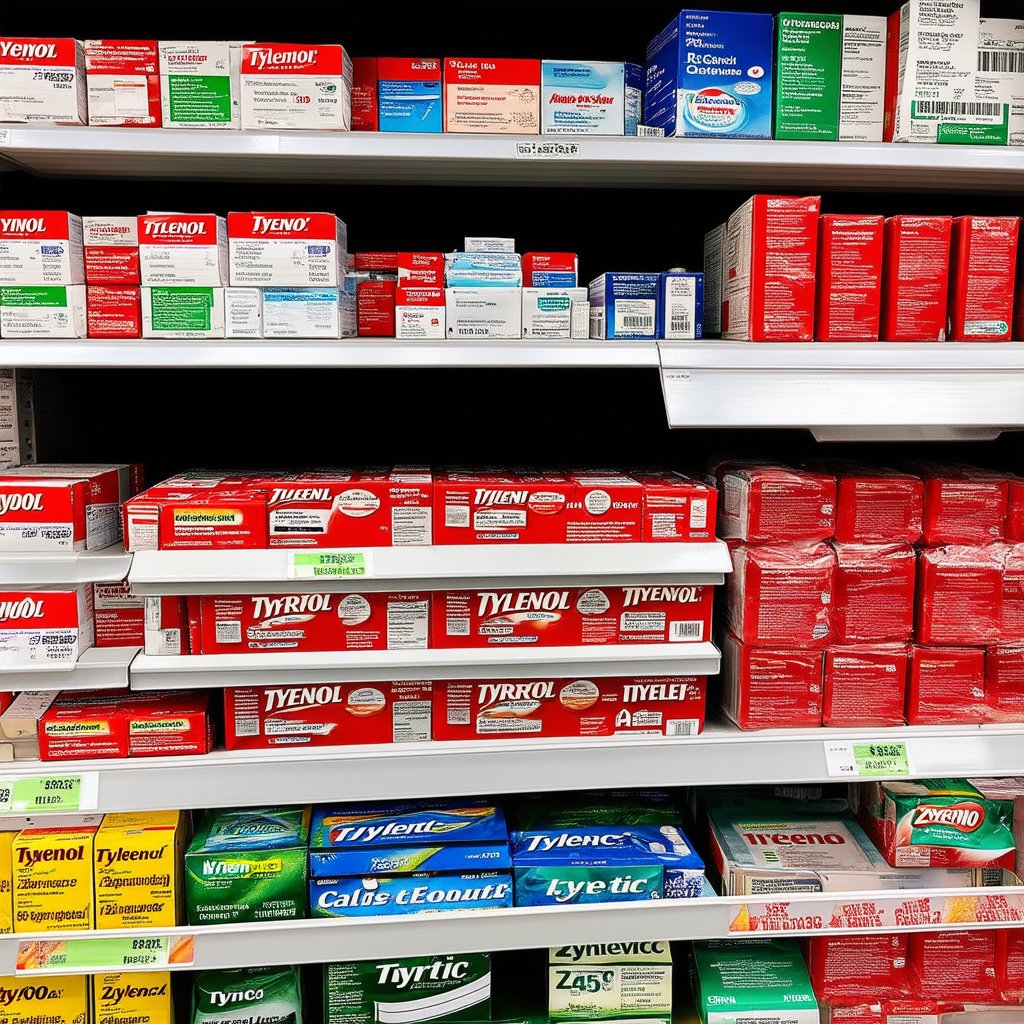} \\

        \includegraphics[width=0.1\textwidth]{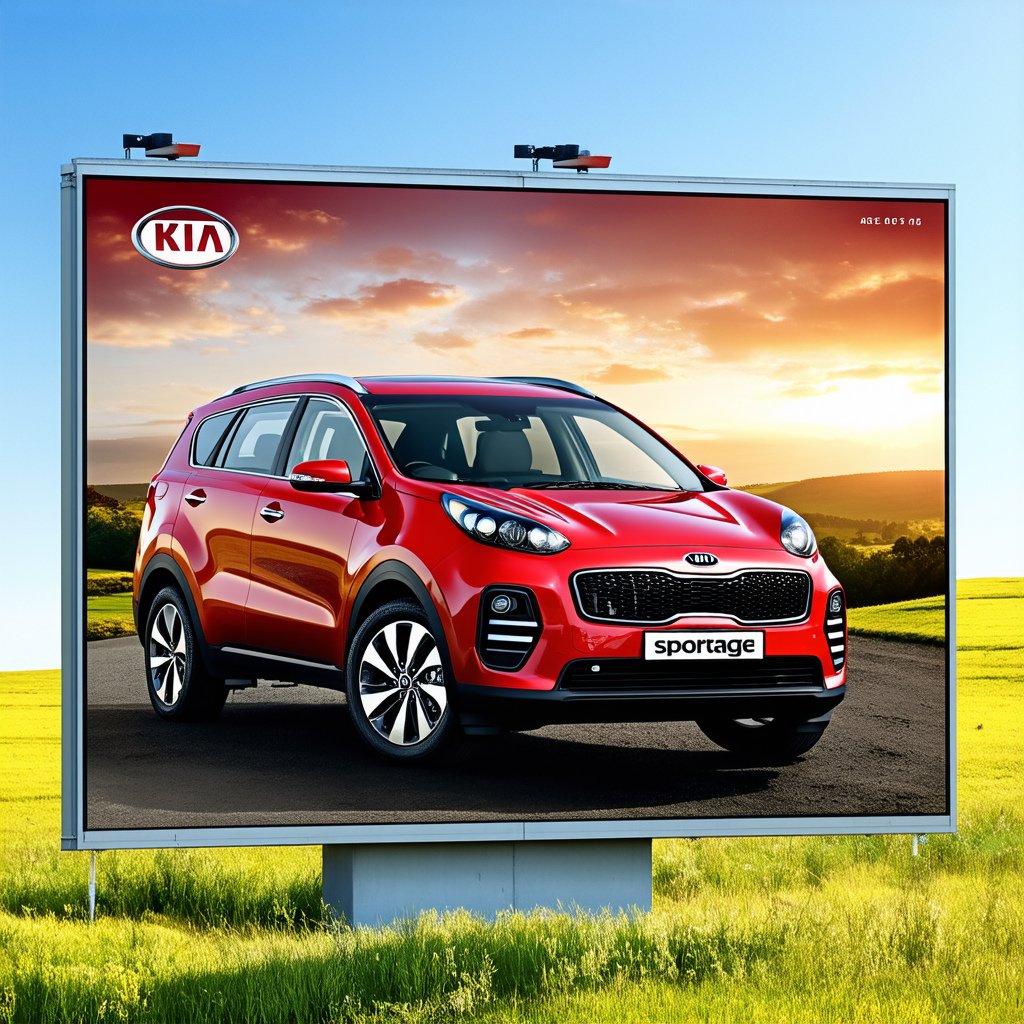} &
        \includegraphics[width=0.1\textwidth]{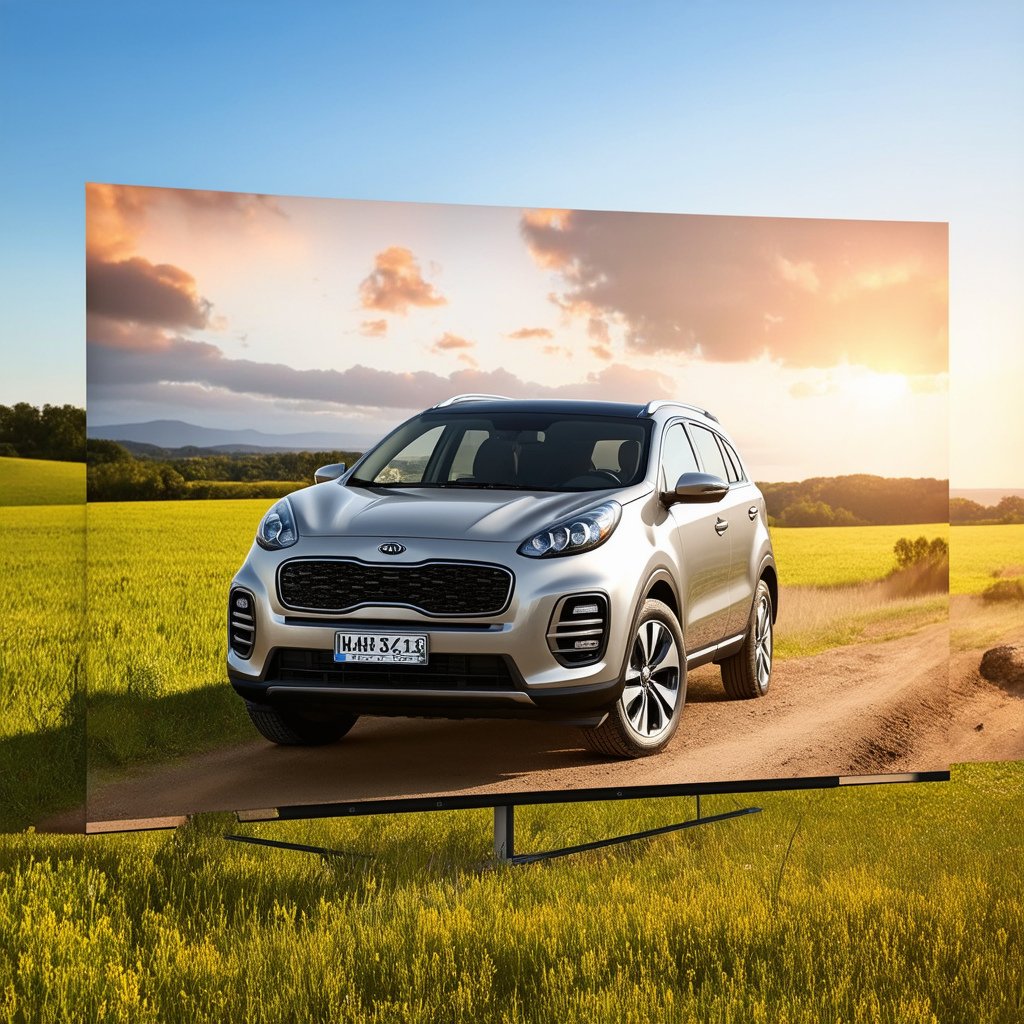} &
        \includegraphics[width=0.1\textwidth]{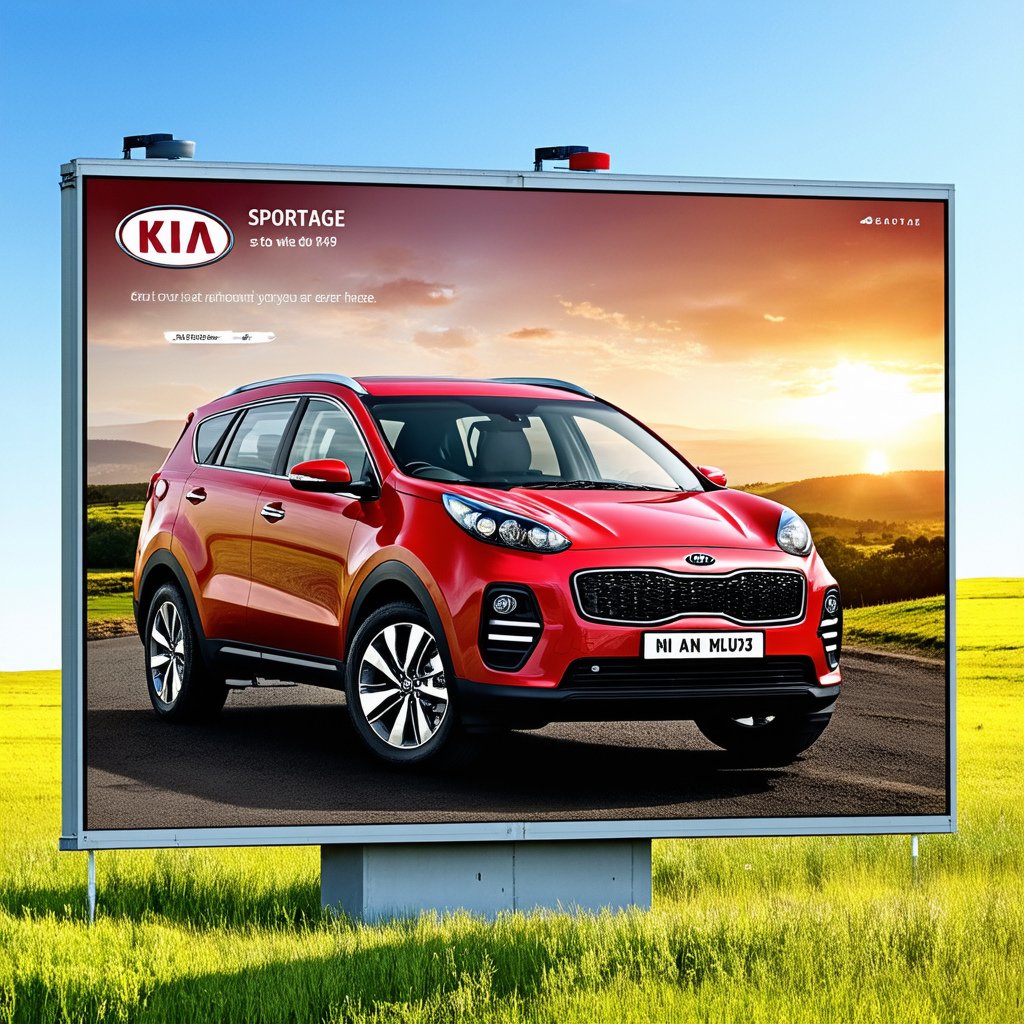} &
                \includegraphics[width=0.1\textwidth]{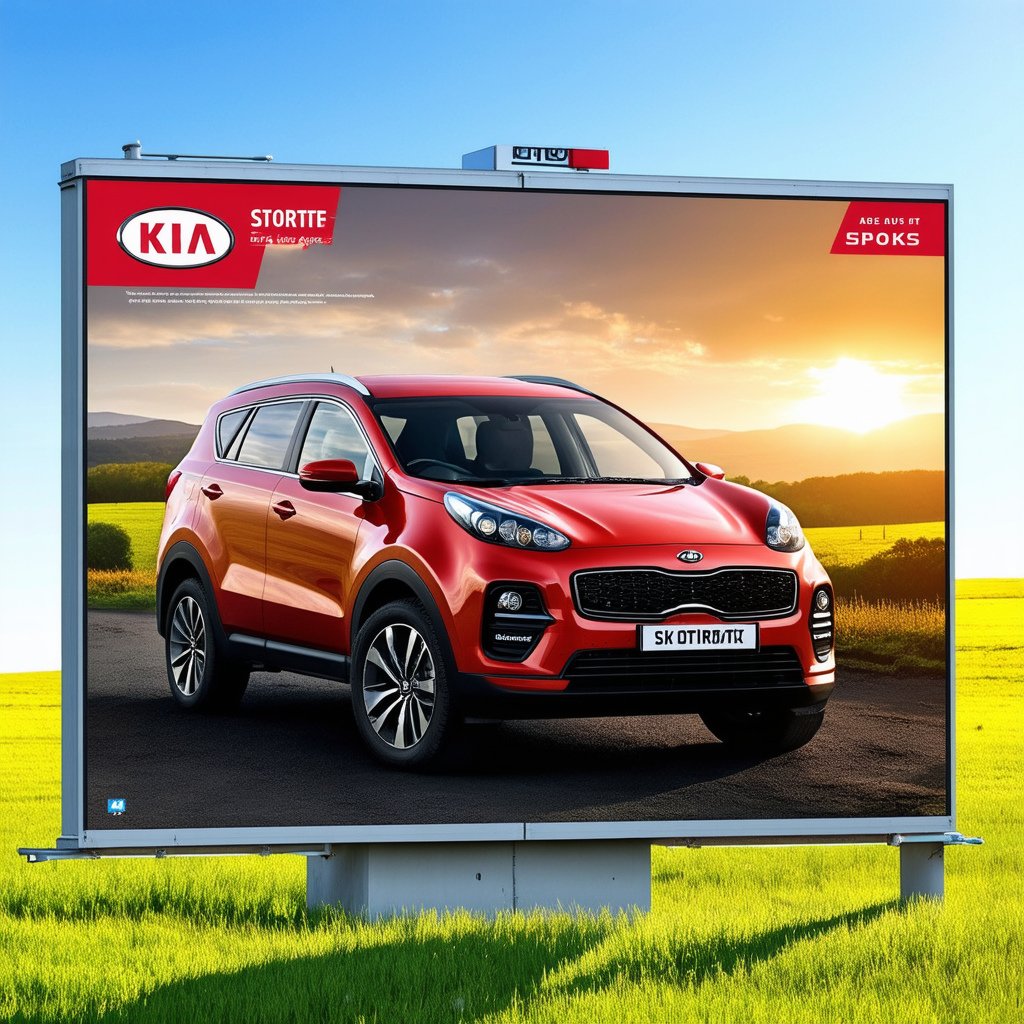} &
                        \includegraphics[width=0.1\textwidth]{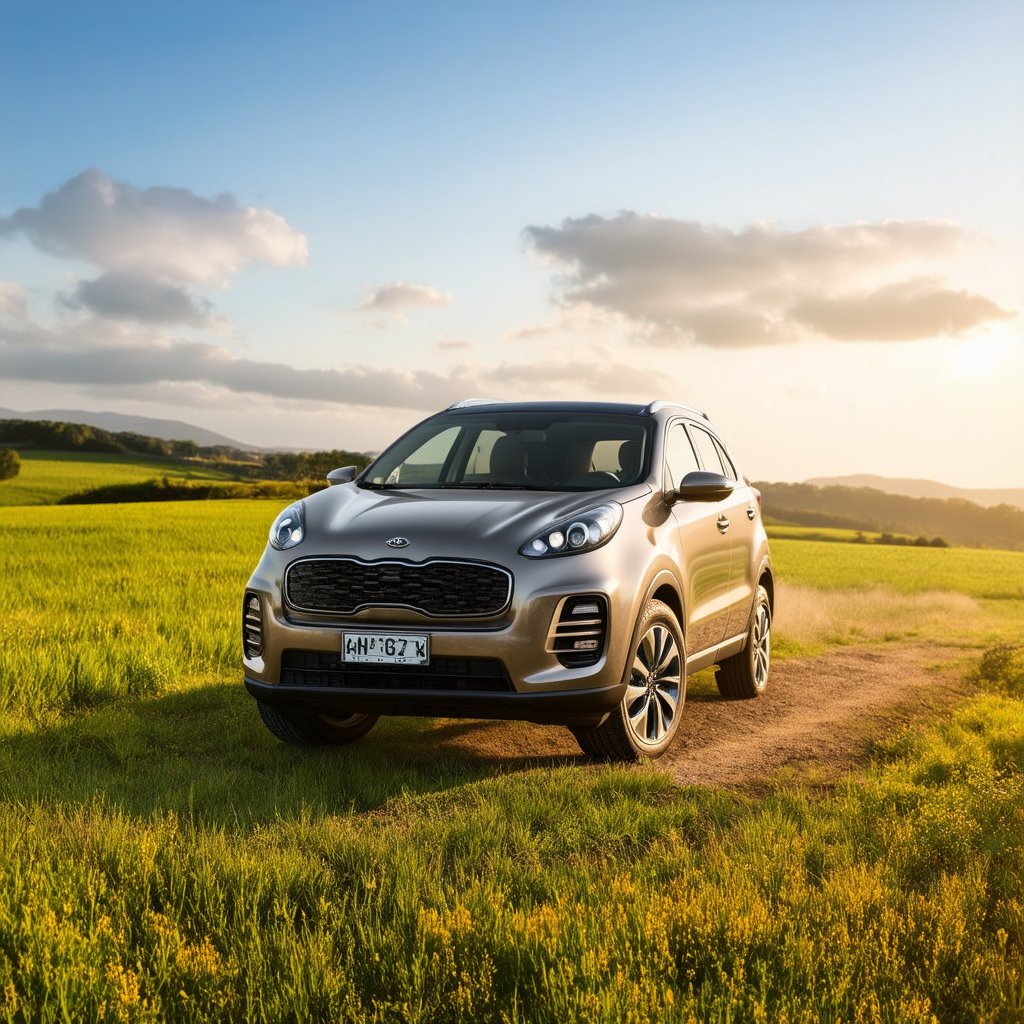} &
        \includegraphics[width=0.1\textwidth]{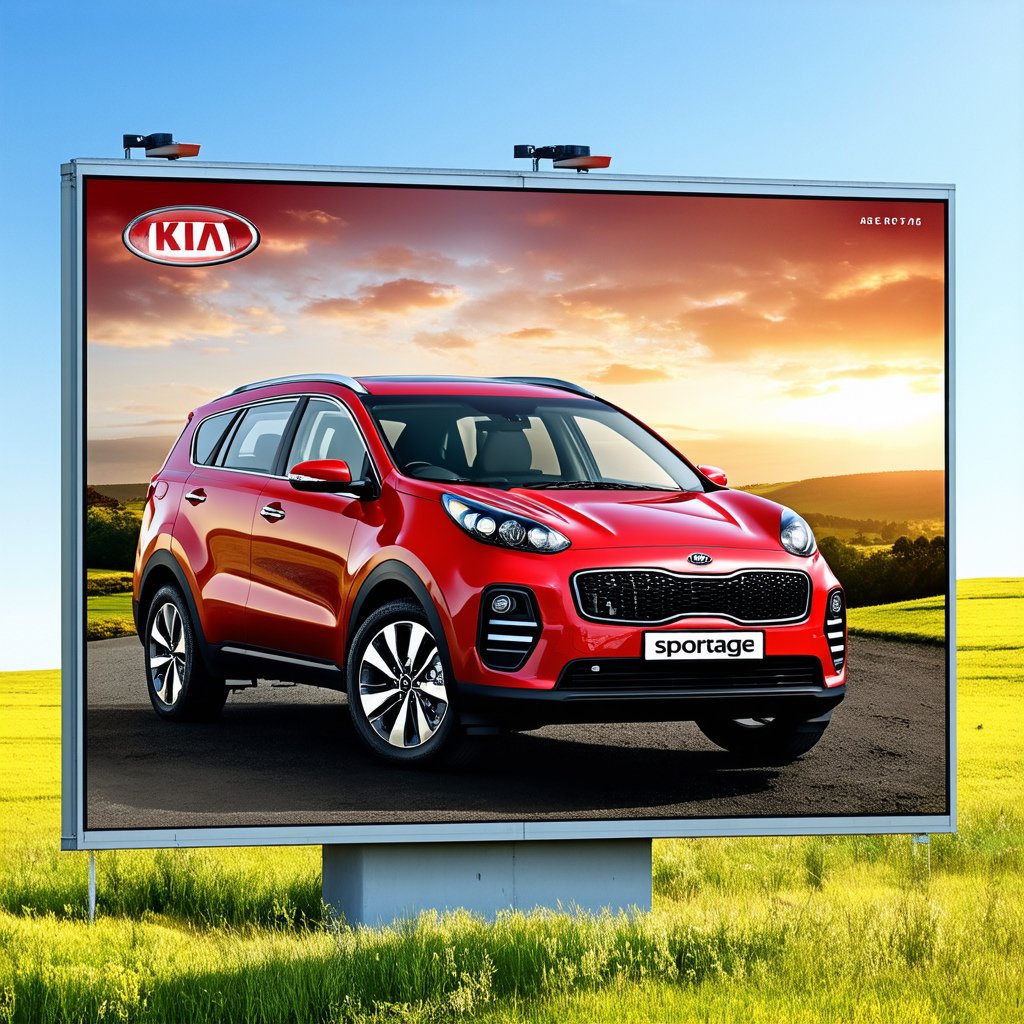} &
                \includegraphics[width=0.1\textwidth]{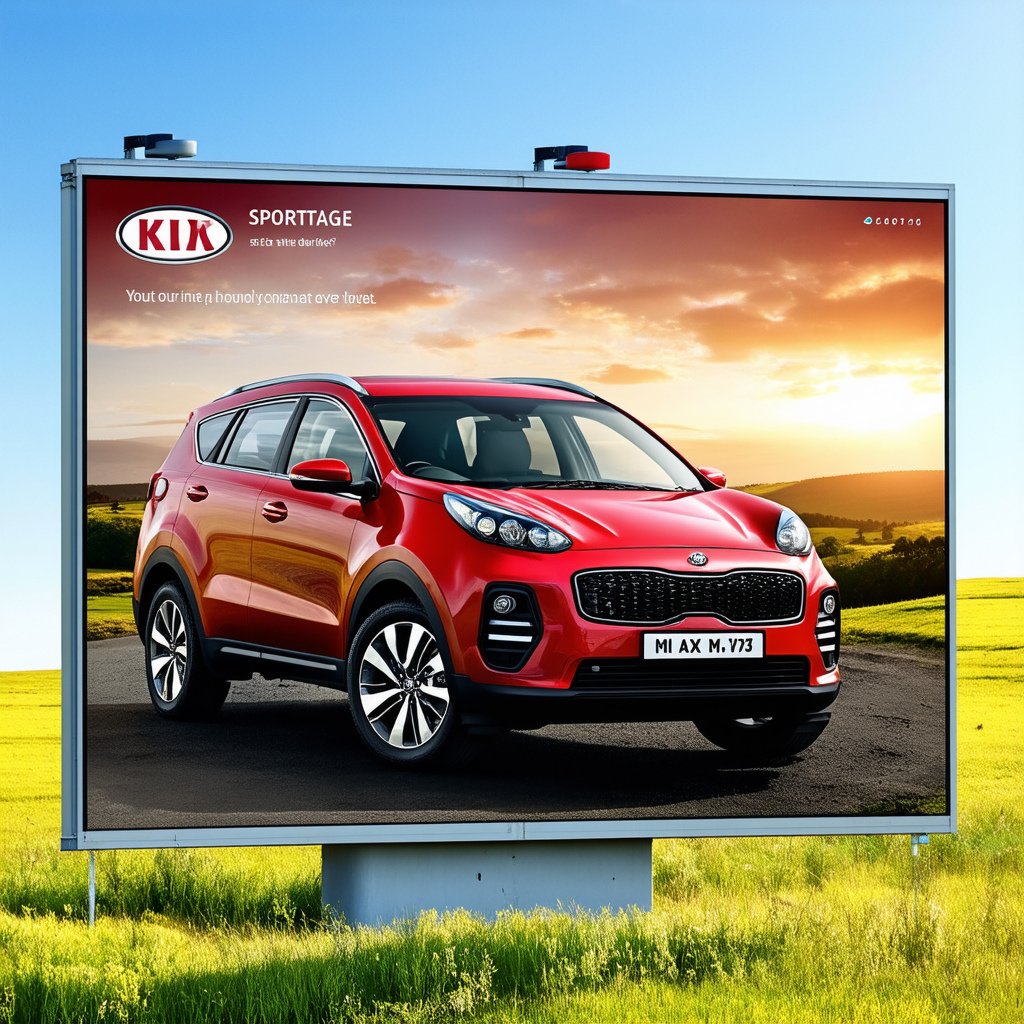} &
                        \includegraphics[width=0.1\textwidth]{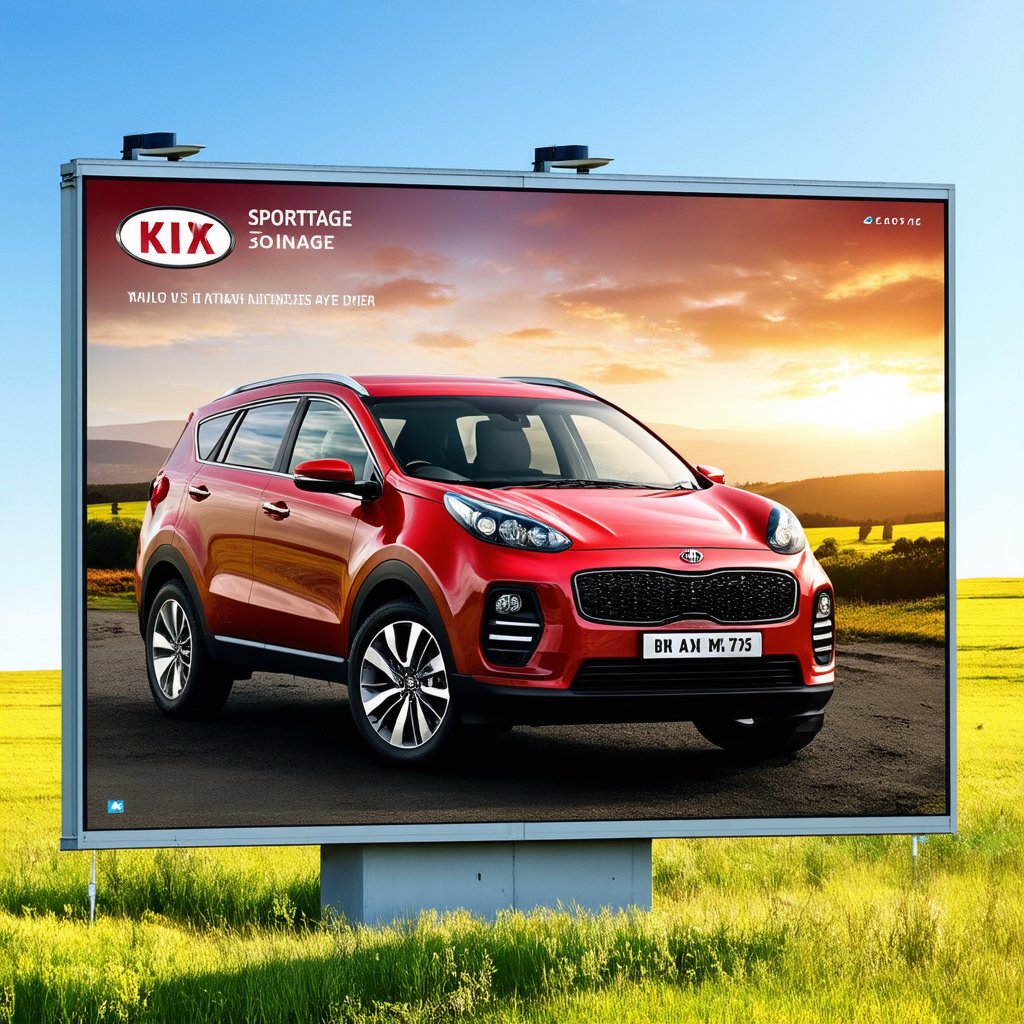} &
        \includegraphics[width=0.1\textwidth]{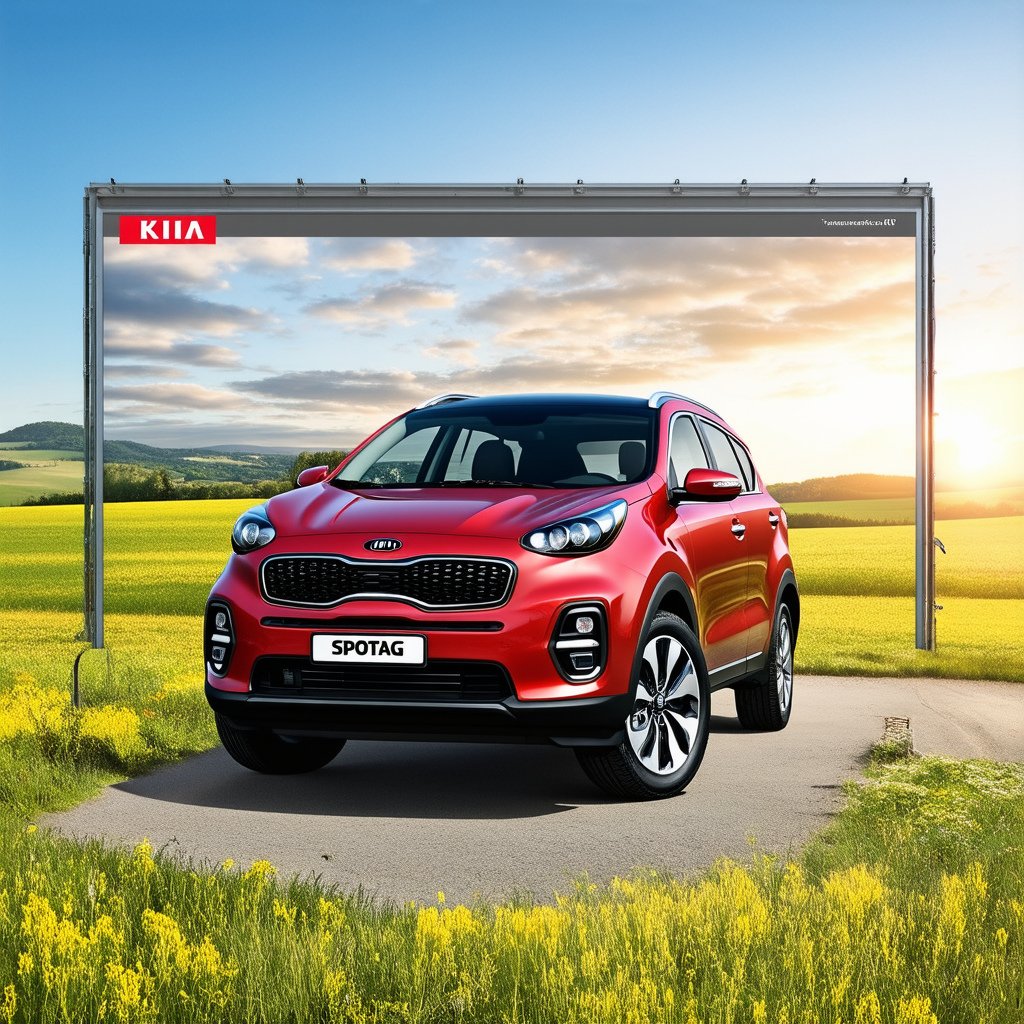} \\

    \end{tabular}

    \caption{More visual results over Boeing, DELL, EXXON MOBIL, FedEx, Goldman-Sachs, HP, Intel, Johnson, KIA.}

\label{tab:results_supp_1}
\end{figure*}

\begin{figure*}[ht]
    \centering
    \renewcommand{\arraystretch}{1.2} 
    \setlength{\tabcolsep}{0pt} 
    \begin{tabular}{ccccccccc}

    \multicolumn{1}{c}{\tiny \textbf{Before}} &
        \multicolumn{1}{c}{\tiny \textbf{NP}} &
        \multicolumn{1}{c}{\tiny \textbf{\sldzero}} &
        \multicolumn{1}{c}{\tiny \textbf{\sldhalf}} &
        \multicolumn{1}{c}{\tiny \textbf{\sldone}} &
        \multicolumn{1}{c}{\tiny \textbf{\segazero}} &
        \multicolumn{1}{c}{\tiny \textbf{\segahalf}} &
        \multicolumn{1}{c}{\tiny \textbf{\segaone}} &
        \multicolumn{1}{c}{\tiny \textbf{\ourBaseline}} \\

        \includegraphics[width=0.111\textwidth]{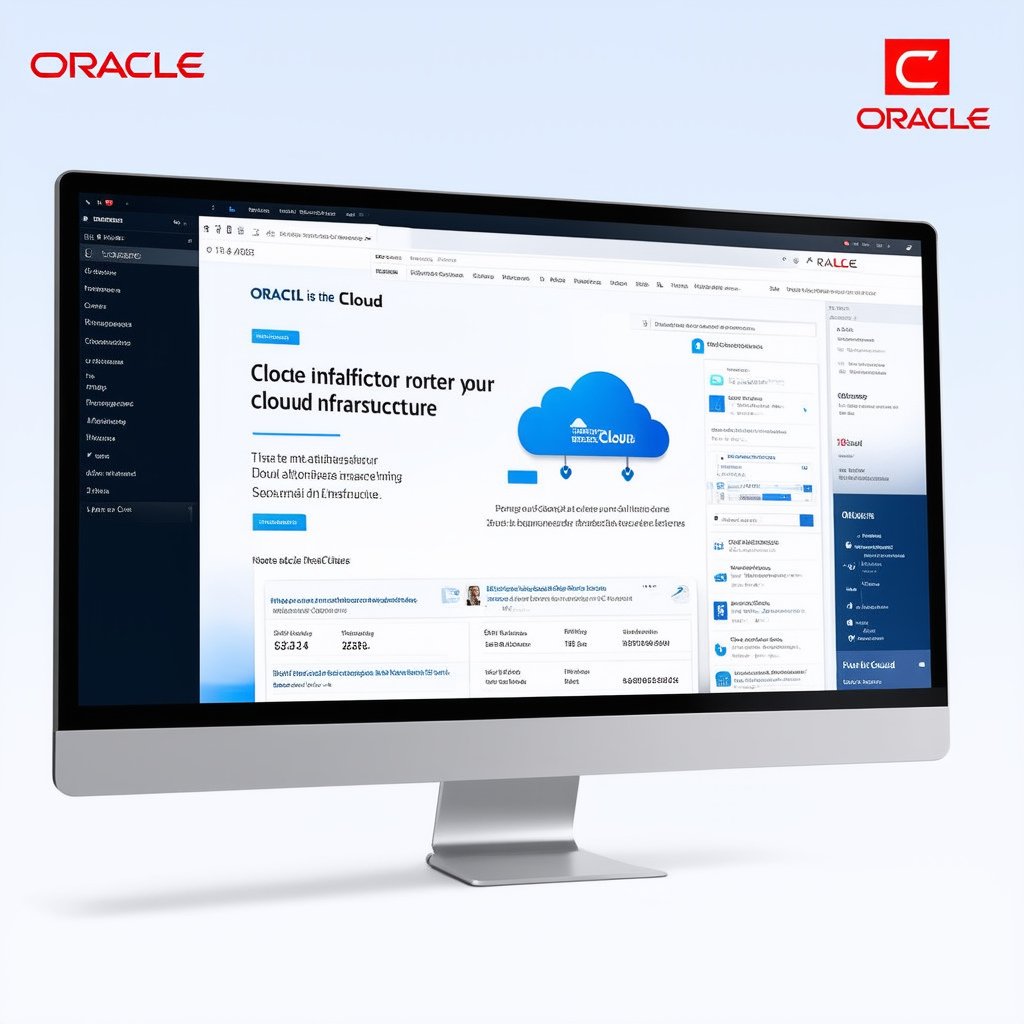} &
        \includegraphics[width=0.111\textwidth]{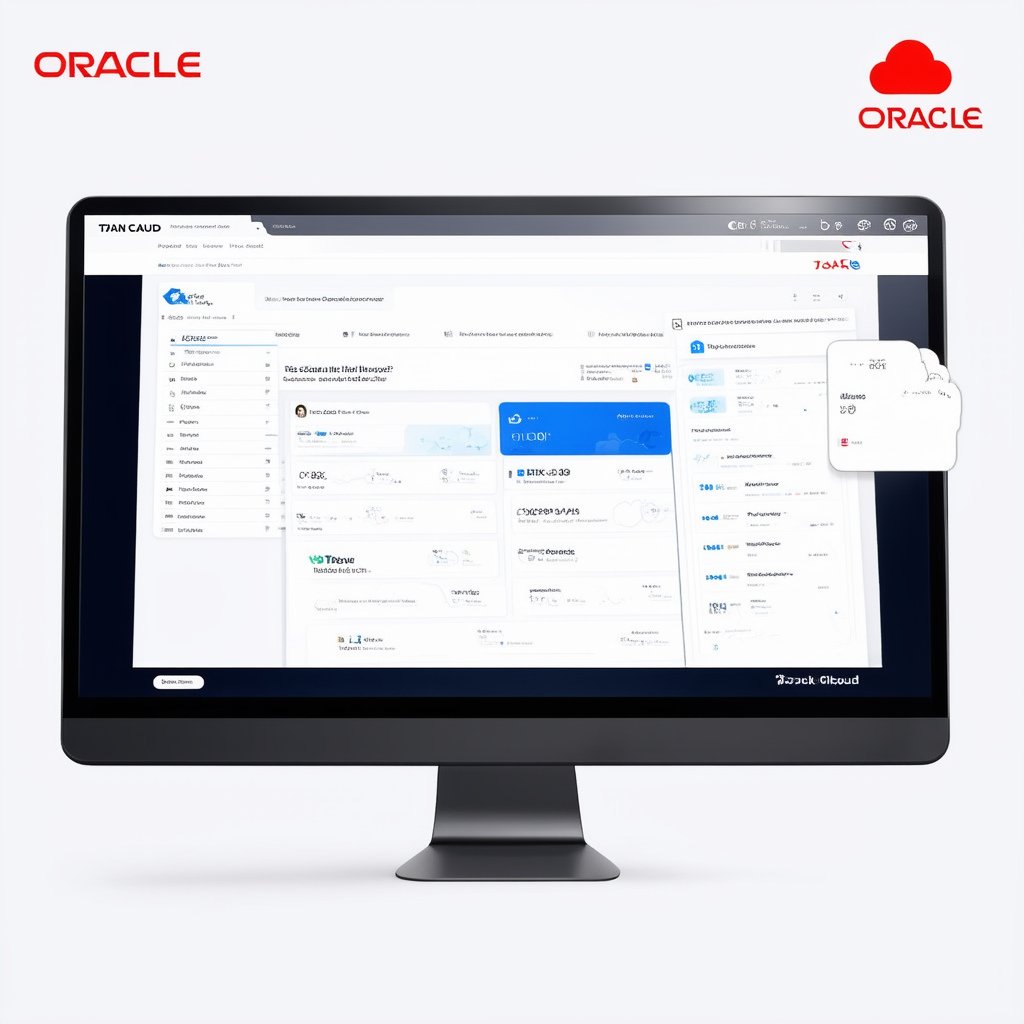} &
        \includegraphics[width=0.111\textwidth]{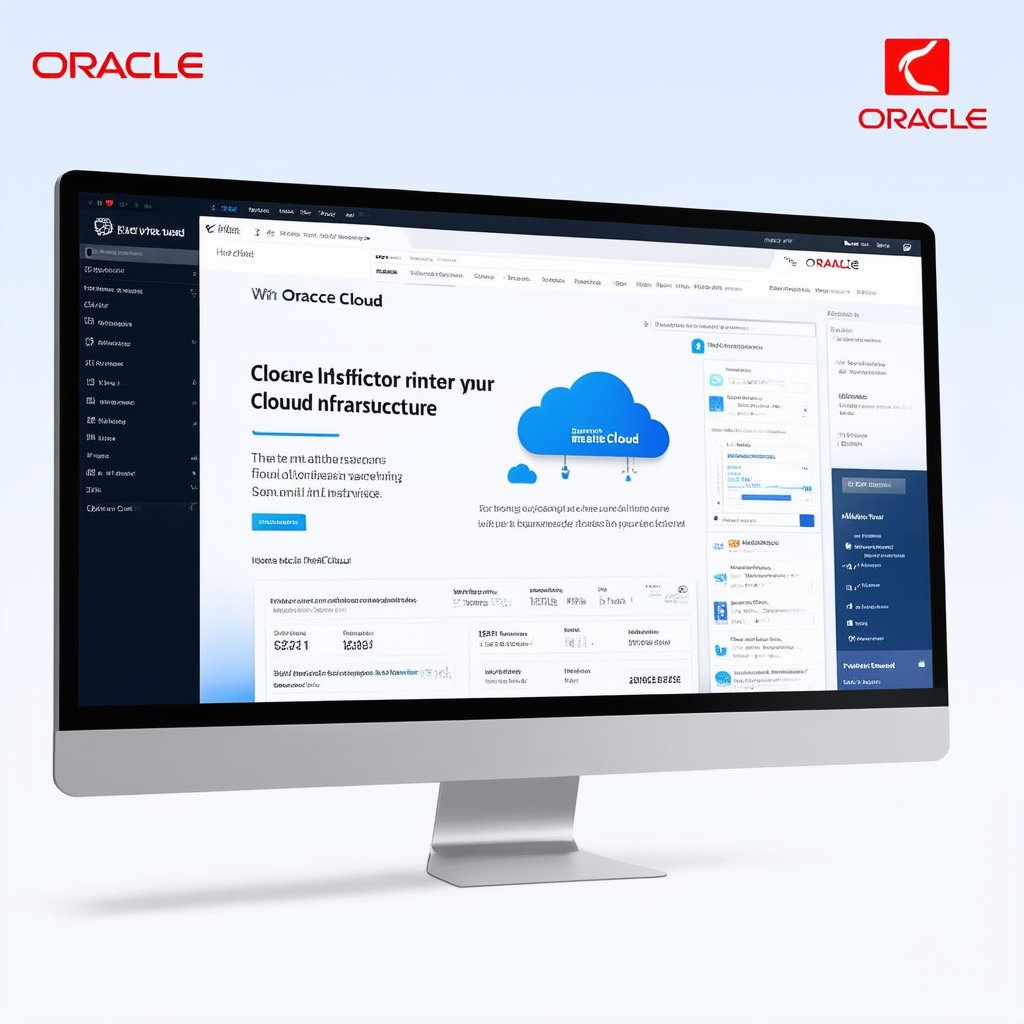} &
                \includegraphics[width=0.111\textwidth]{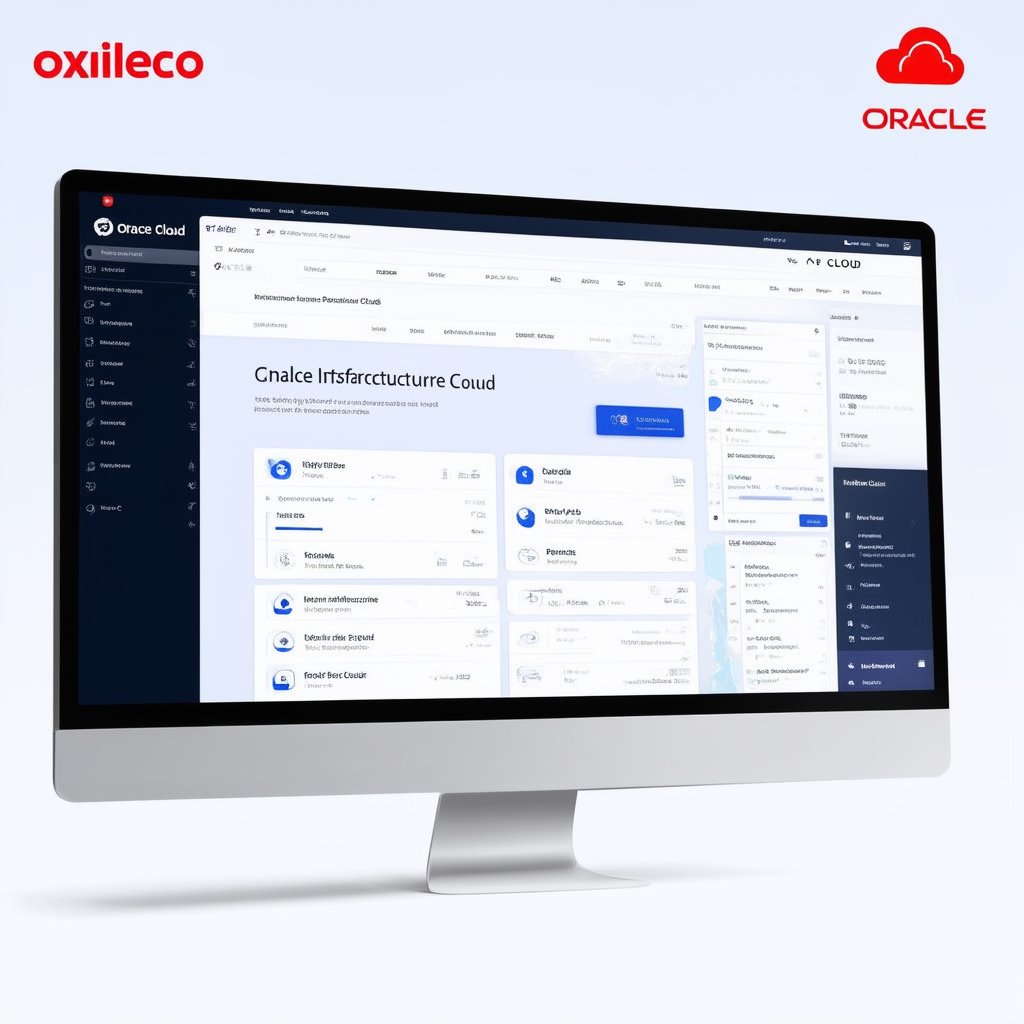} &
                        \includegraphics[width=0.111\textwidth]{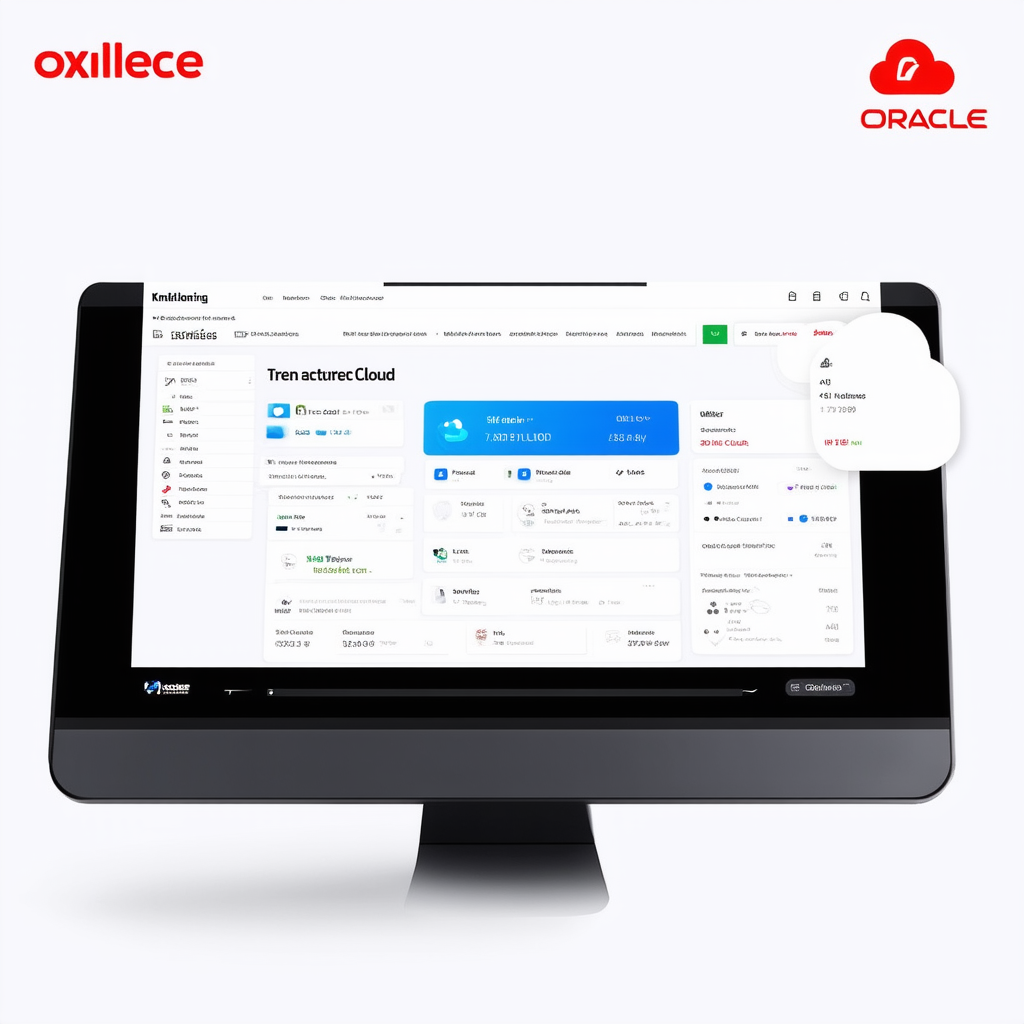} &
        \includegraphics[width=0.111\textwidth]{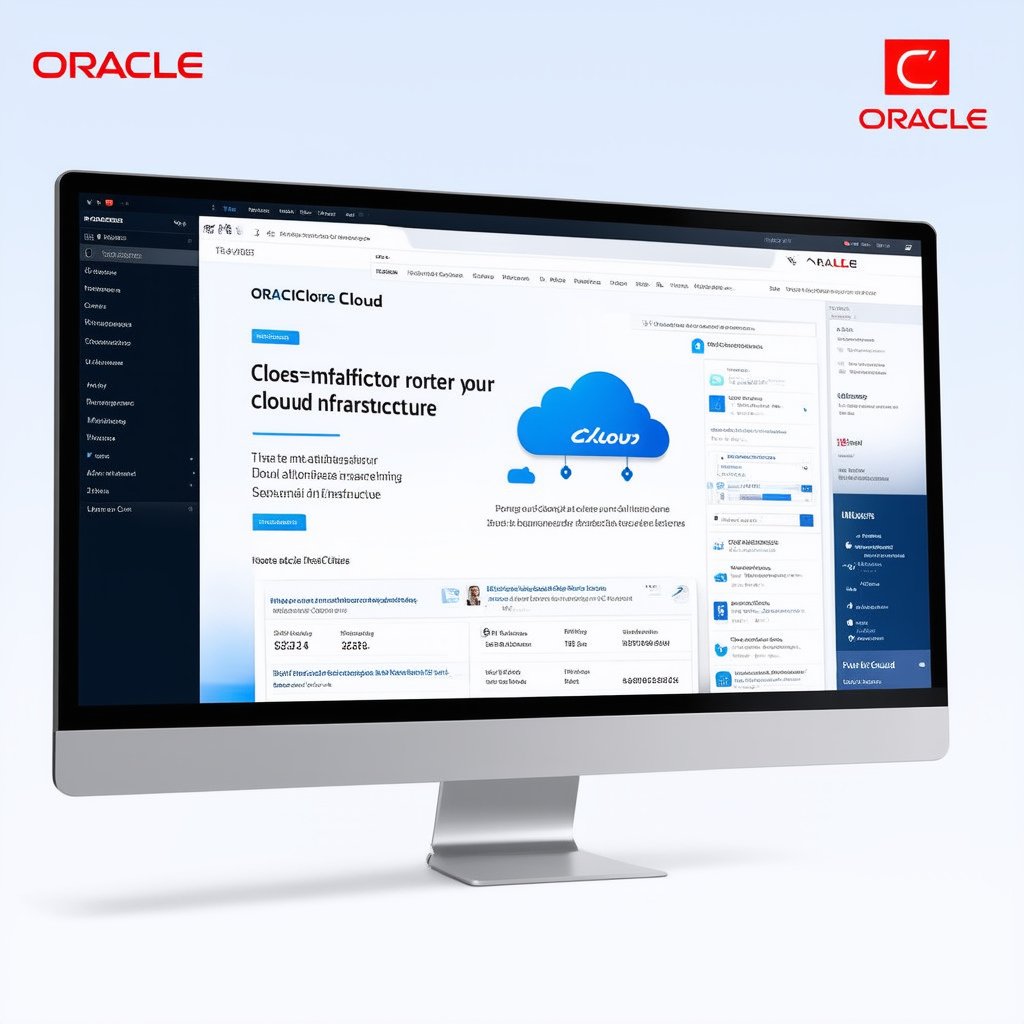} &
                \includegraphics[width=0.111\textwidth]{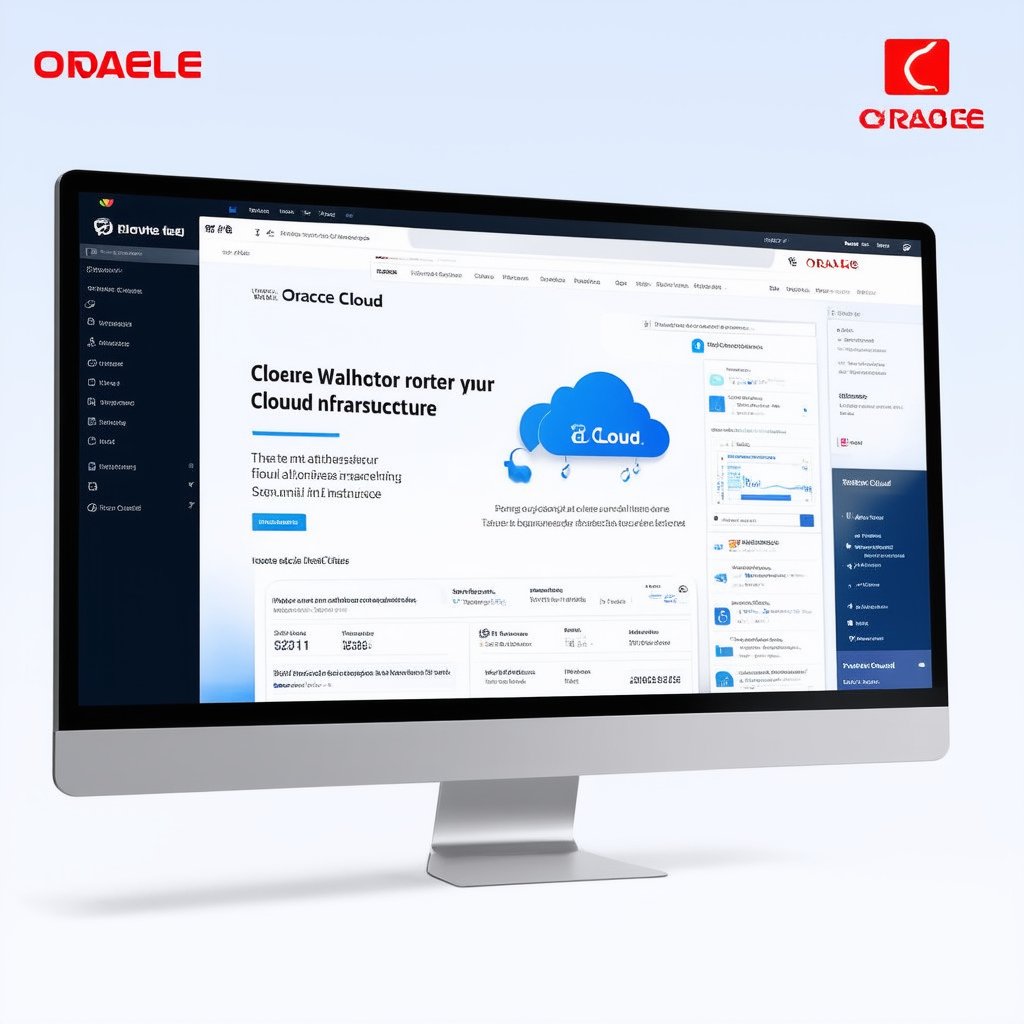} &
                        \includegraphics[width=0.111\textwidth]{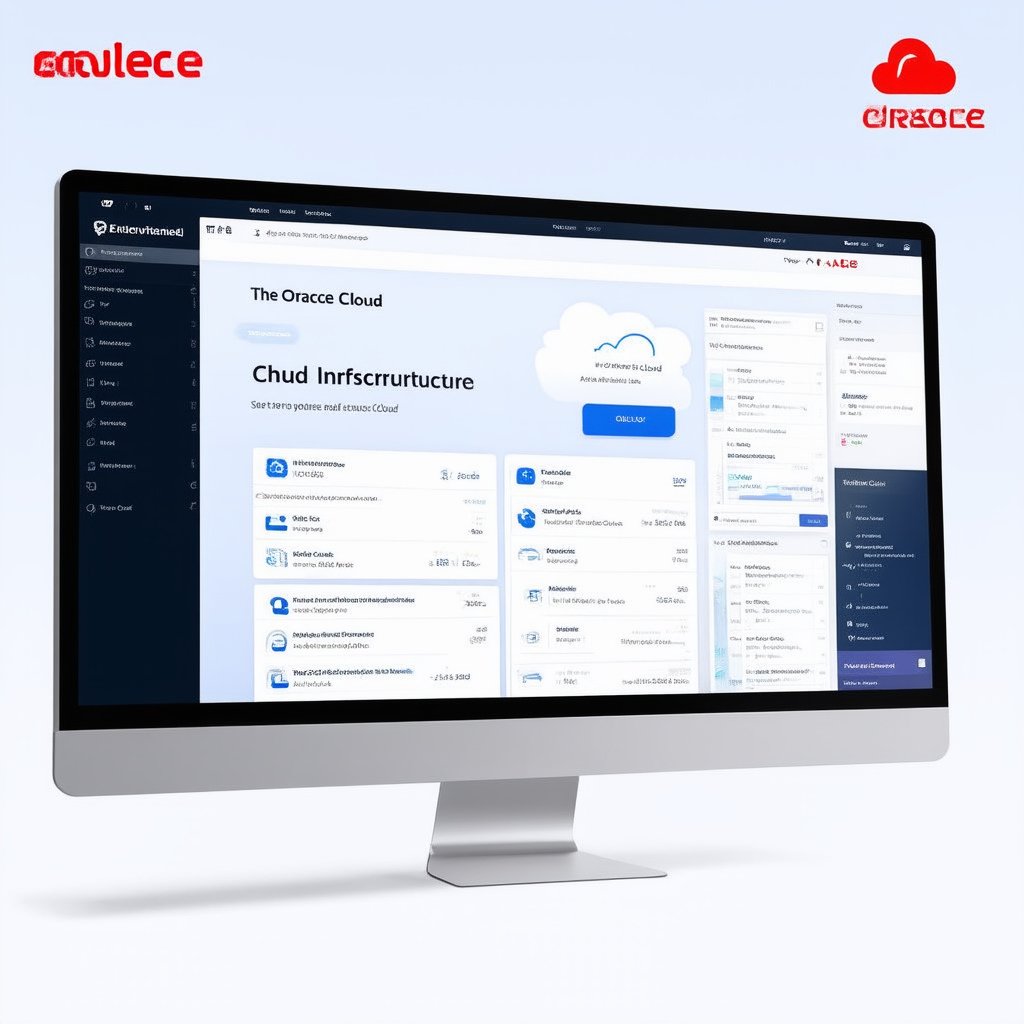} &
        \includegraphics[width=0.111\textwidth]{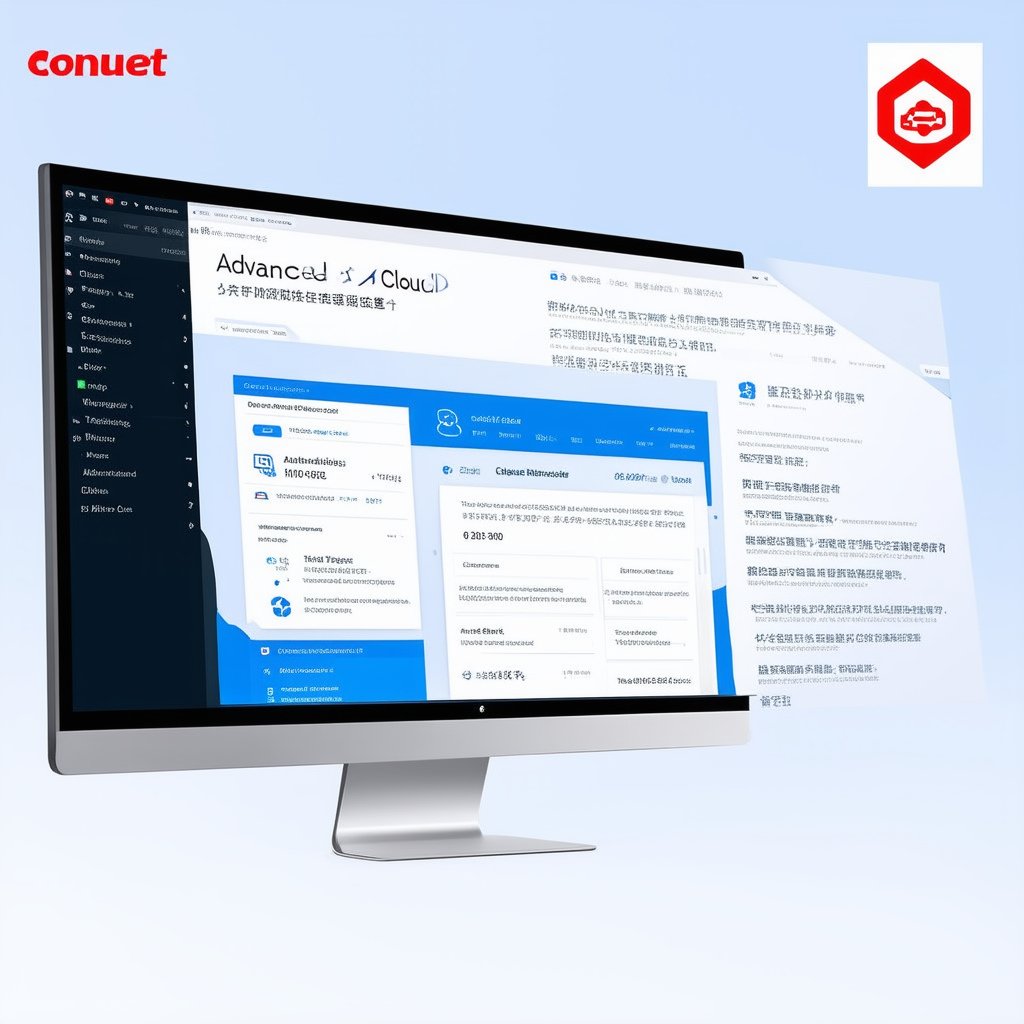} \\

        \includegraphics[width=0.111\textwidth]{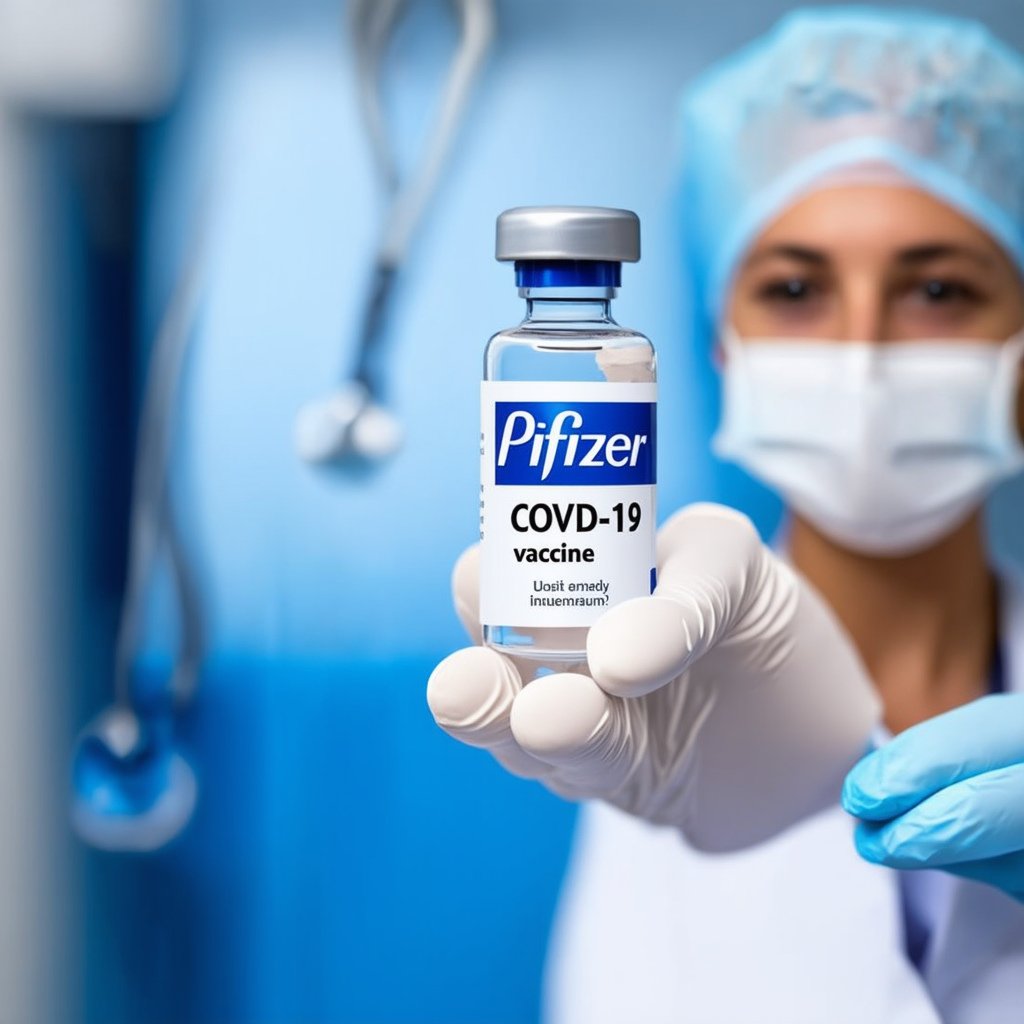} &
        \includegraphics[width=0.111\textwidth]{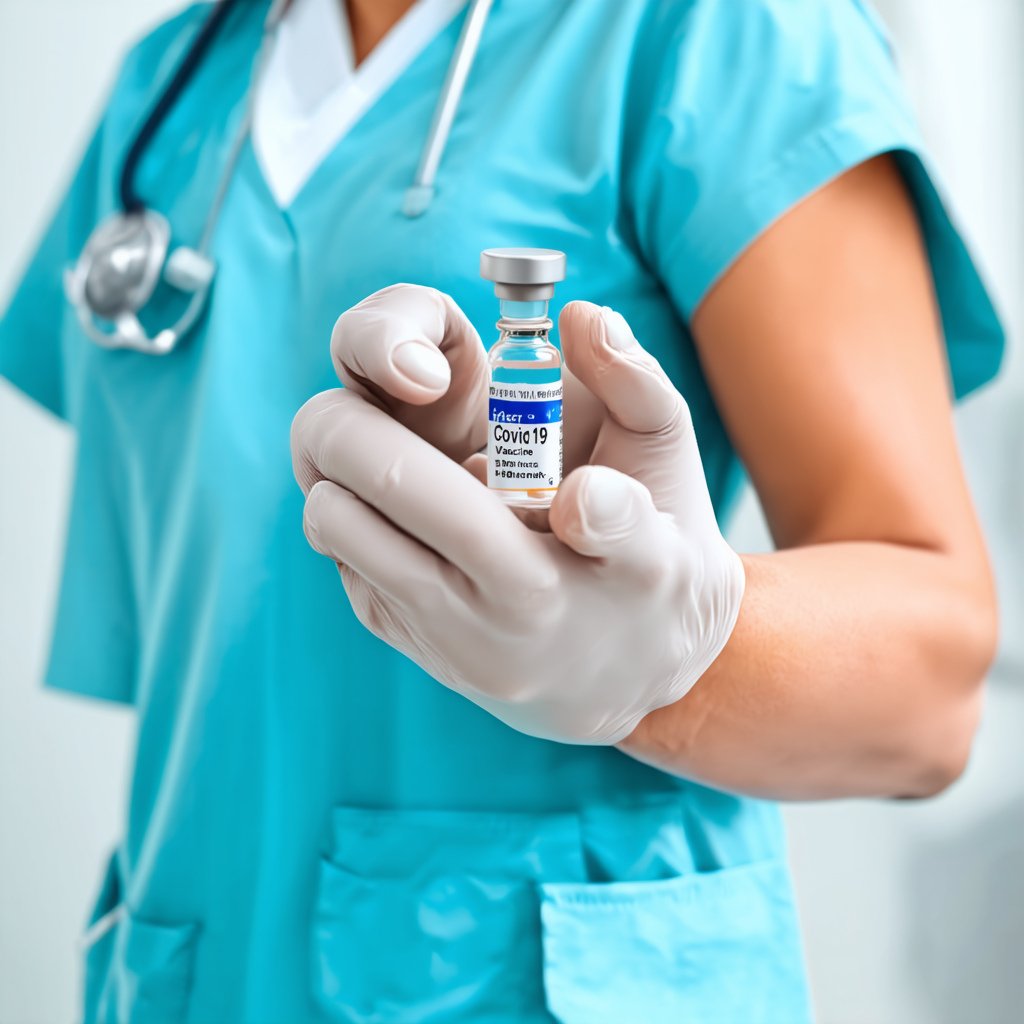} &
        \includegraphics[width=0.111\textwidth]{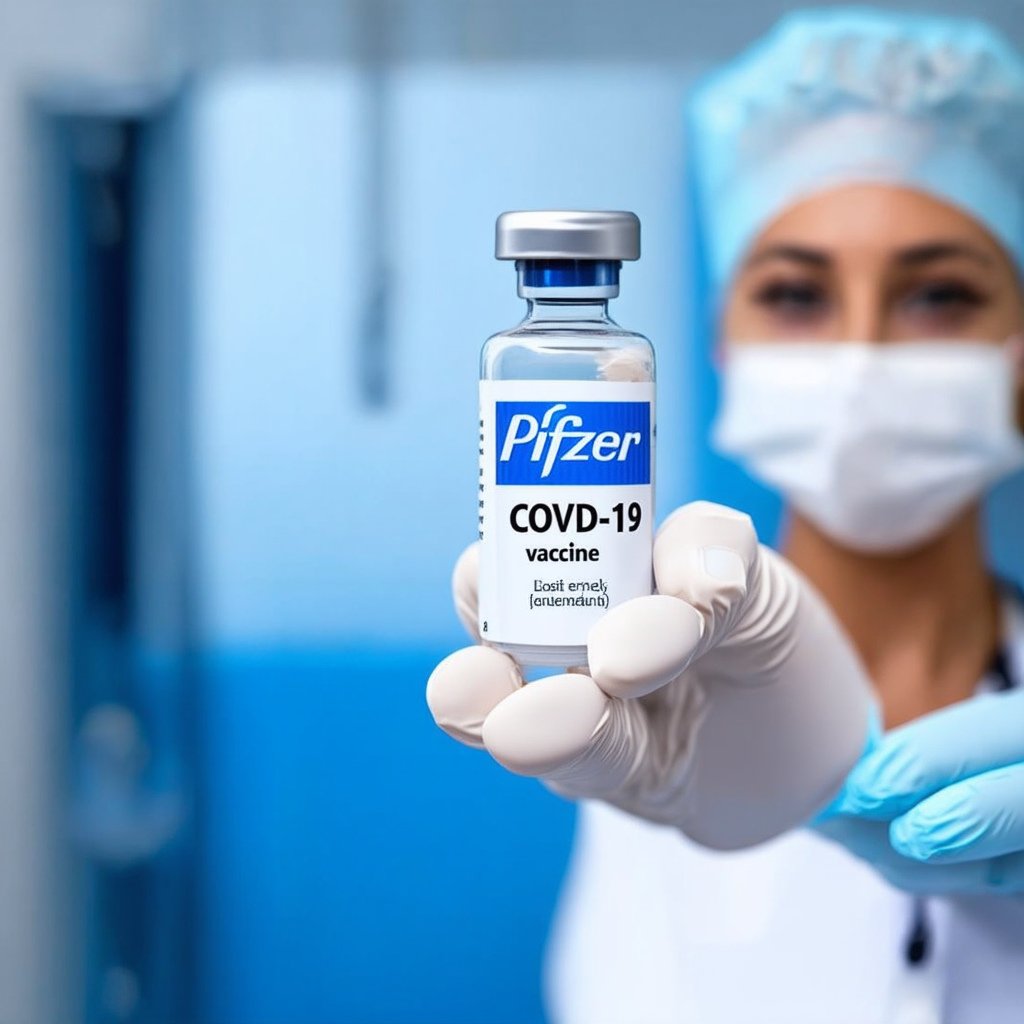} &
                \includegraphics[width=0.111\textwidth]{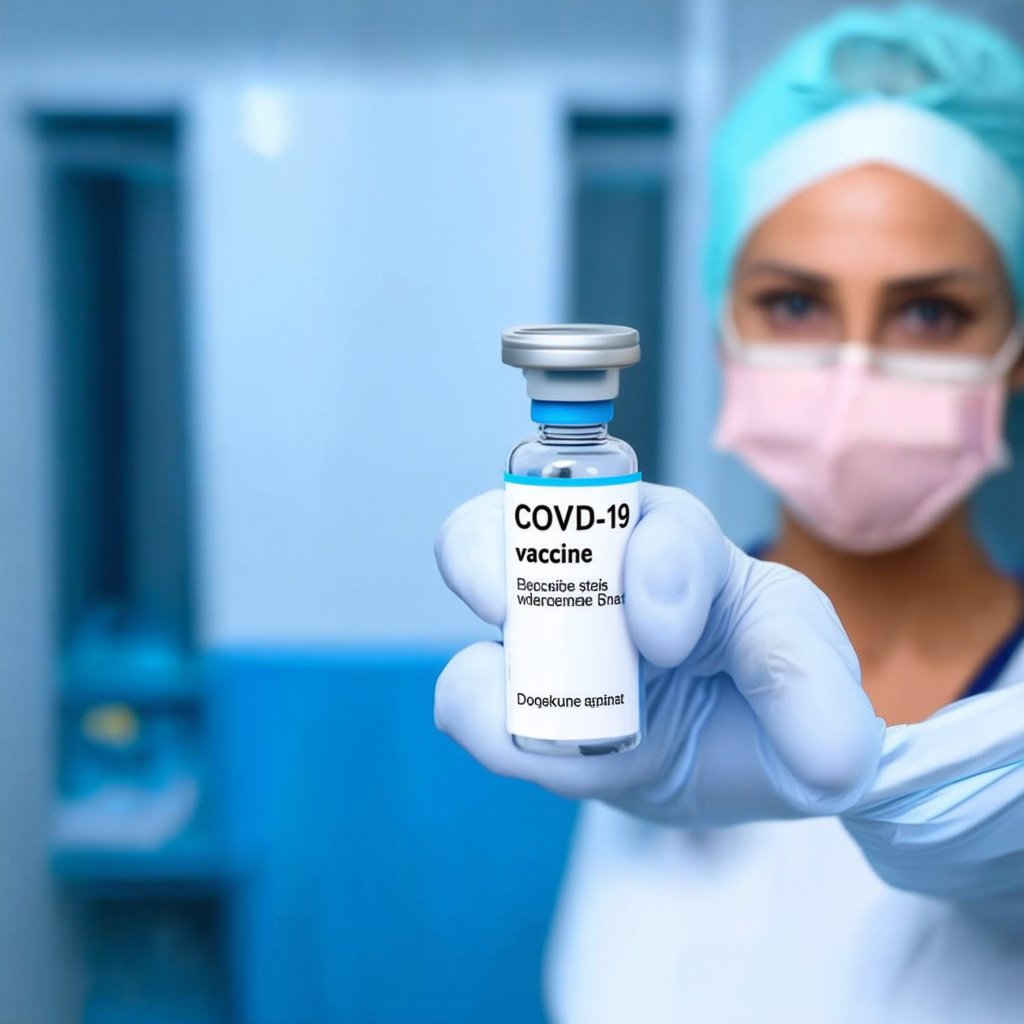} &
                        \includegraphics[width=0.111\textwidth]{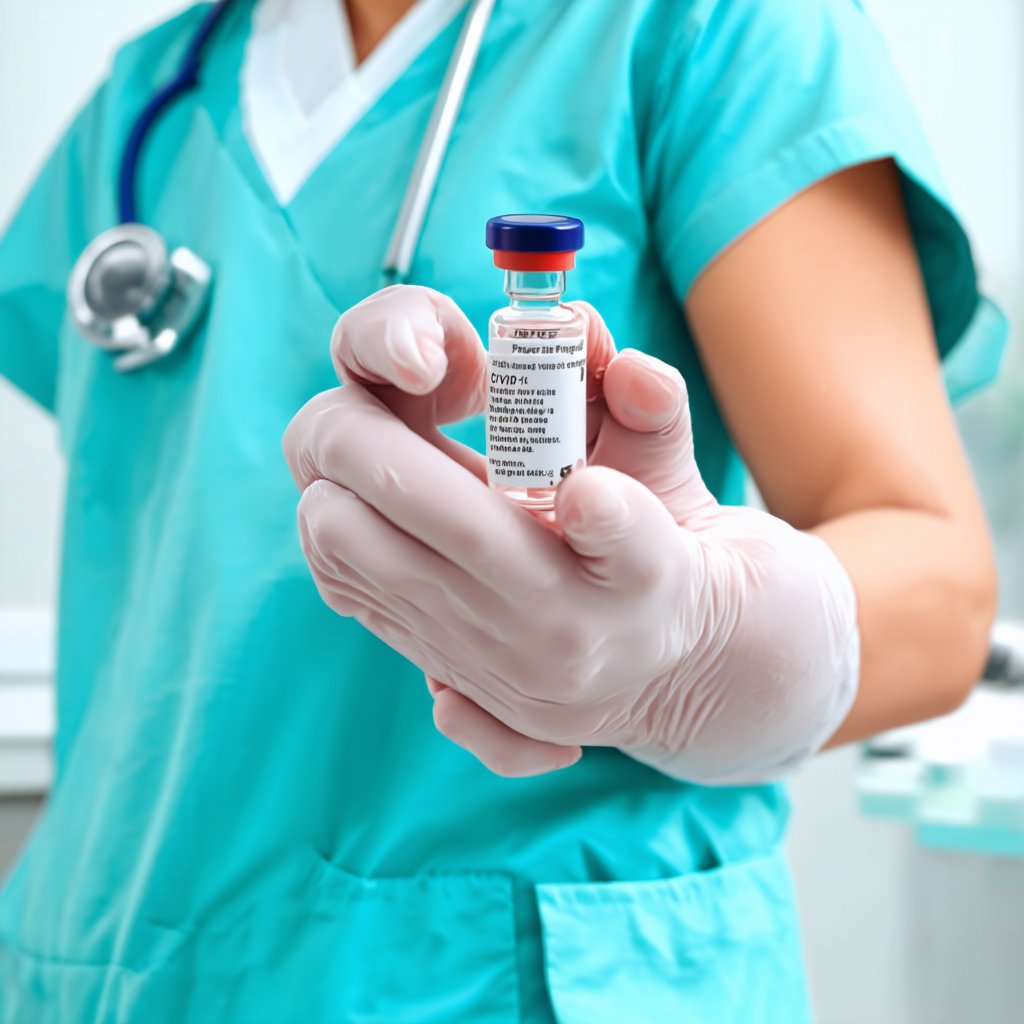} &
        \includegraphics[width=0.111\textwidth]{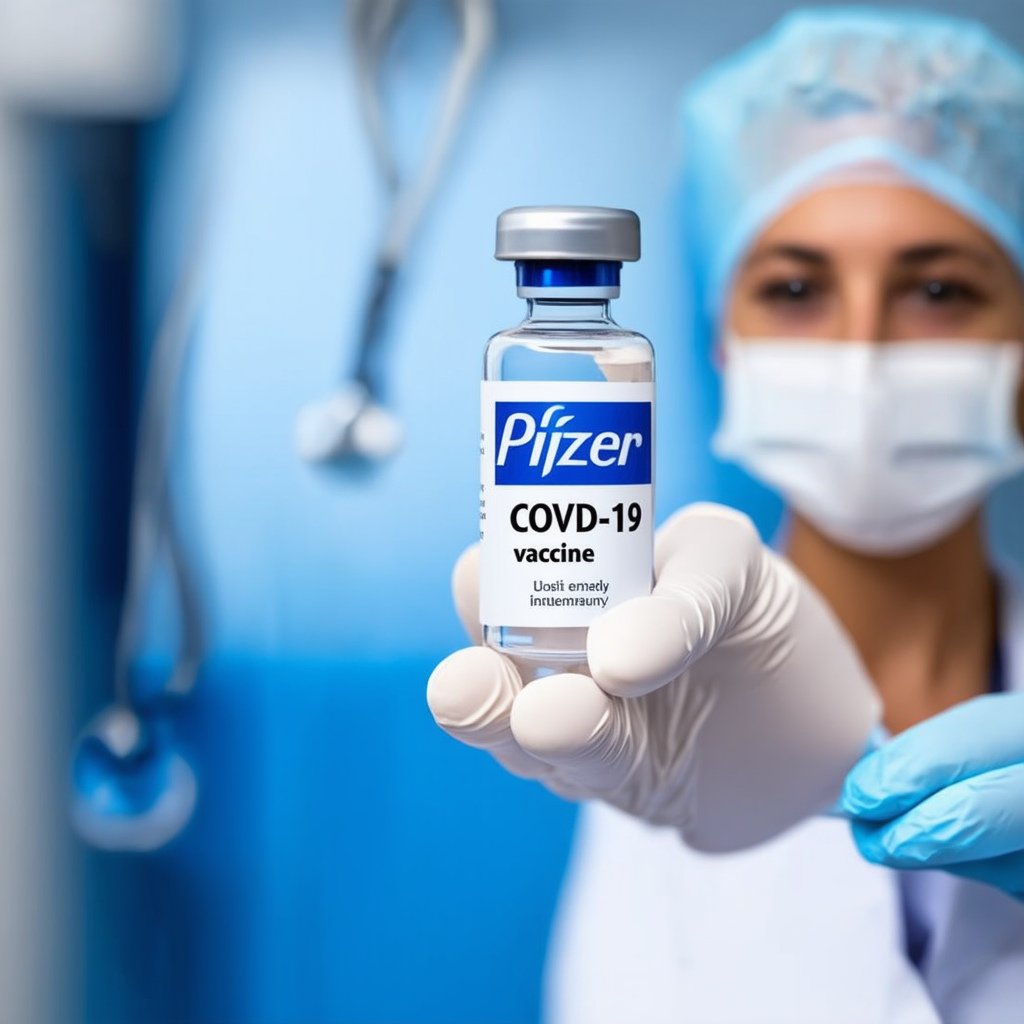} &
                \includegraphics[width=0.111\textwidth]{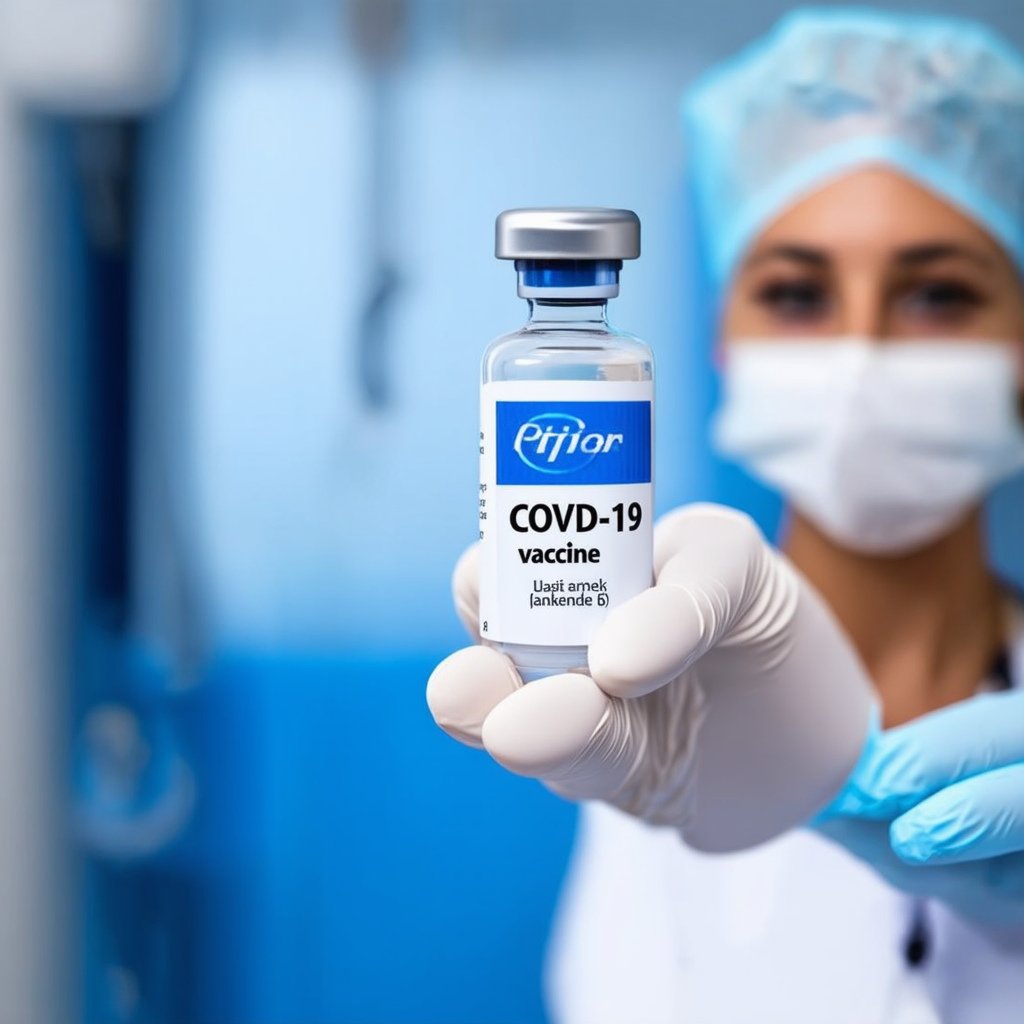} &
                        \includegraphics[width=0.111\textwidth]{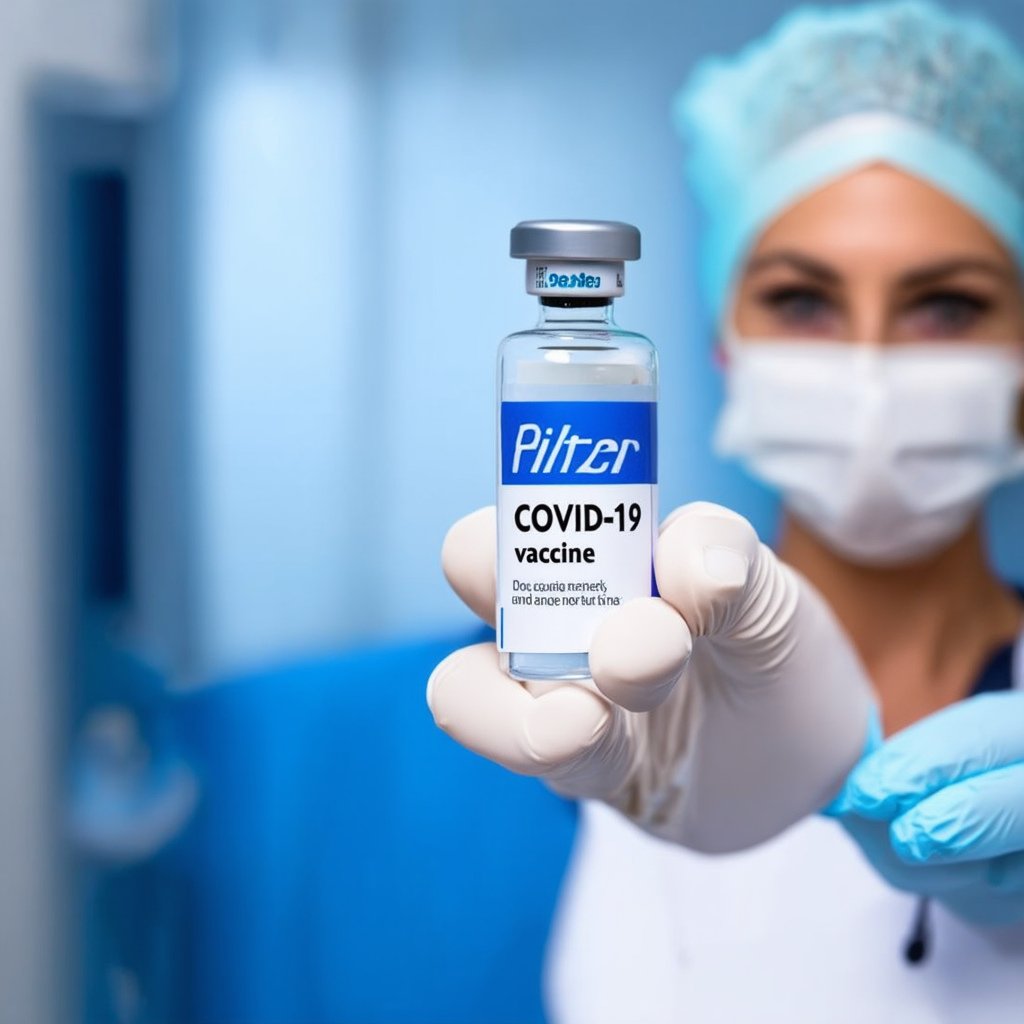} &
        \includegraphics[width=0.111\textwidth]{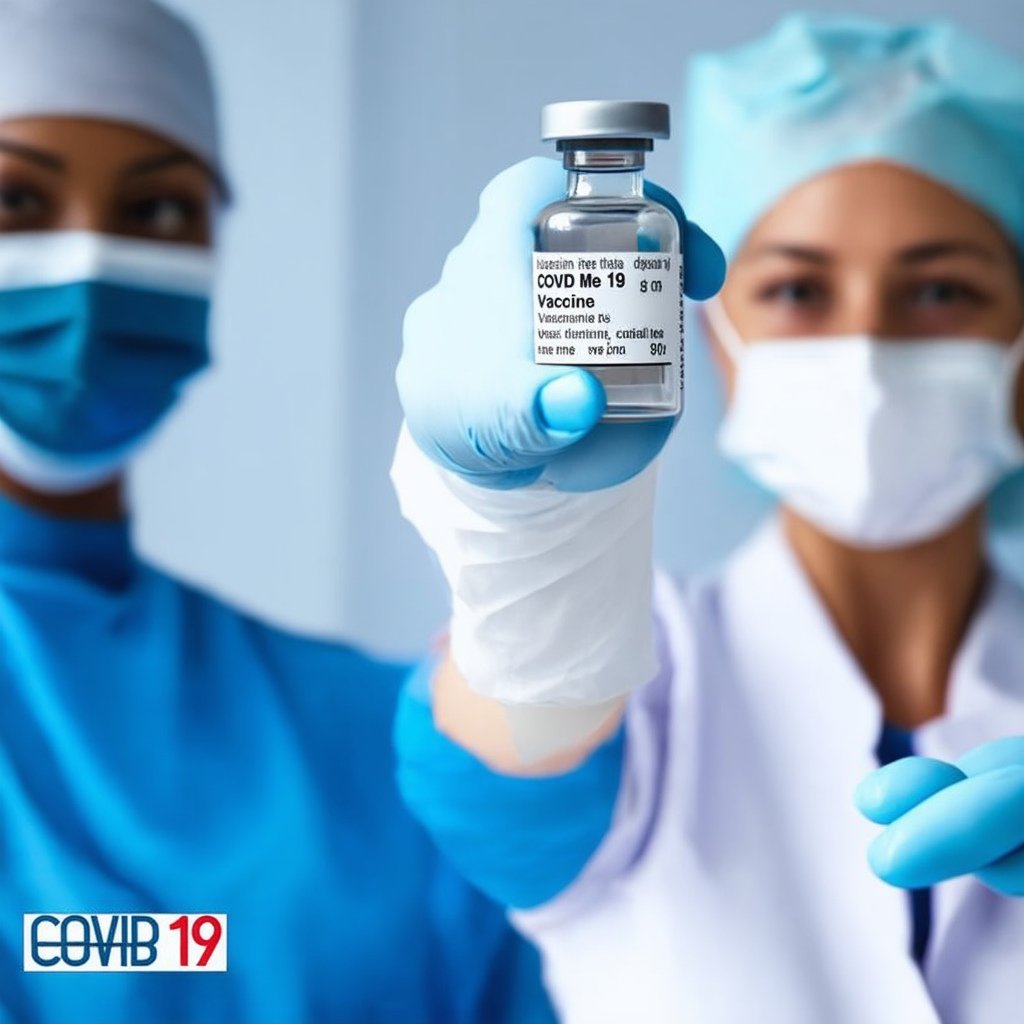} \\

        \includegraphics[width=0.111\textwidth]{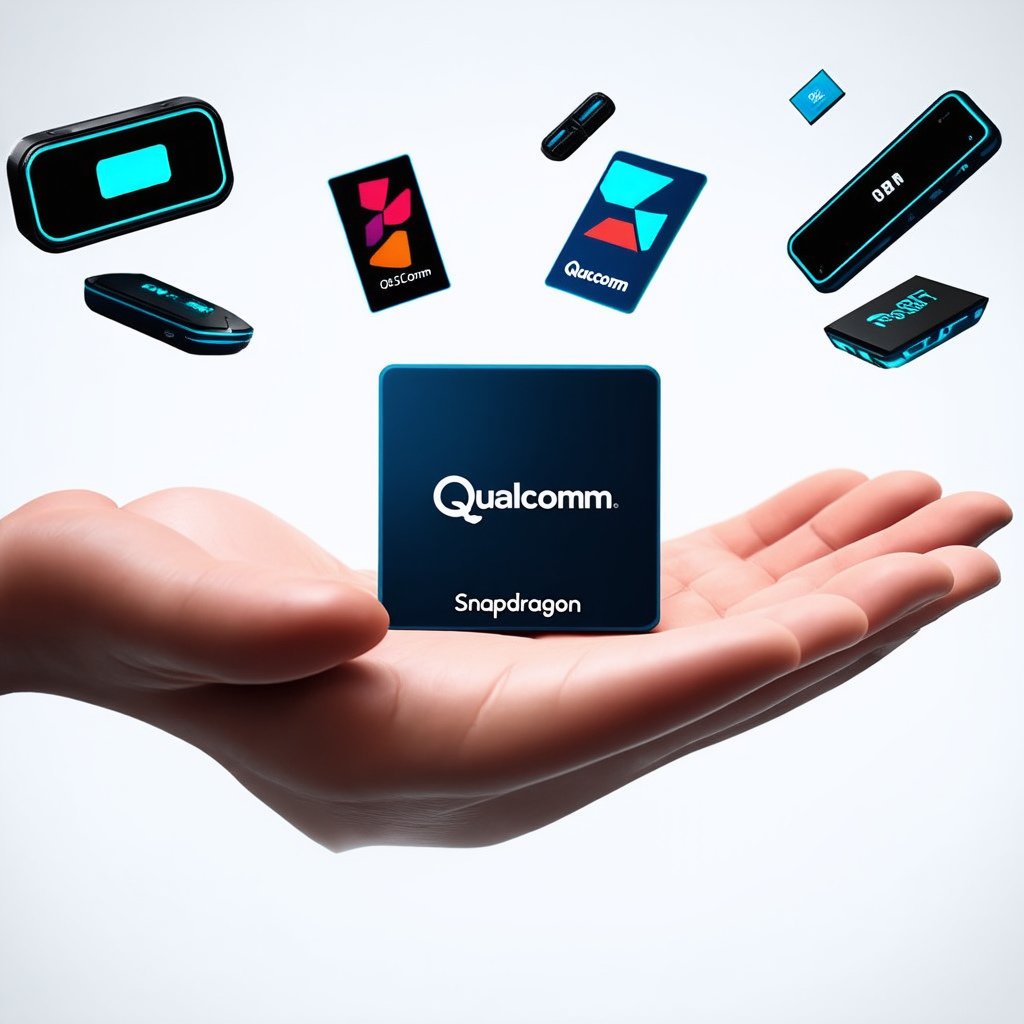} &
        \includegraphics[width=0.111\textwidth]{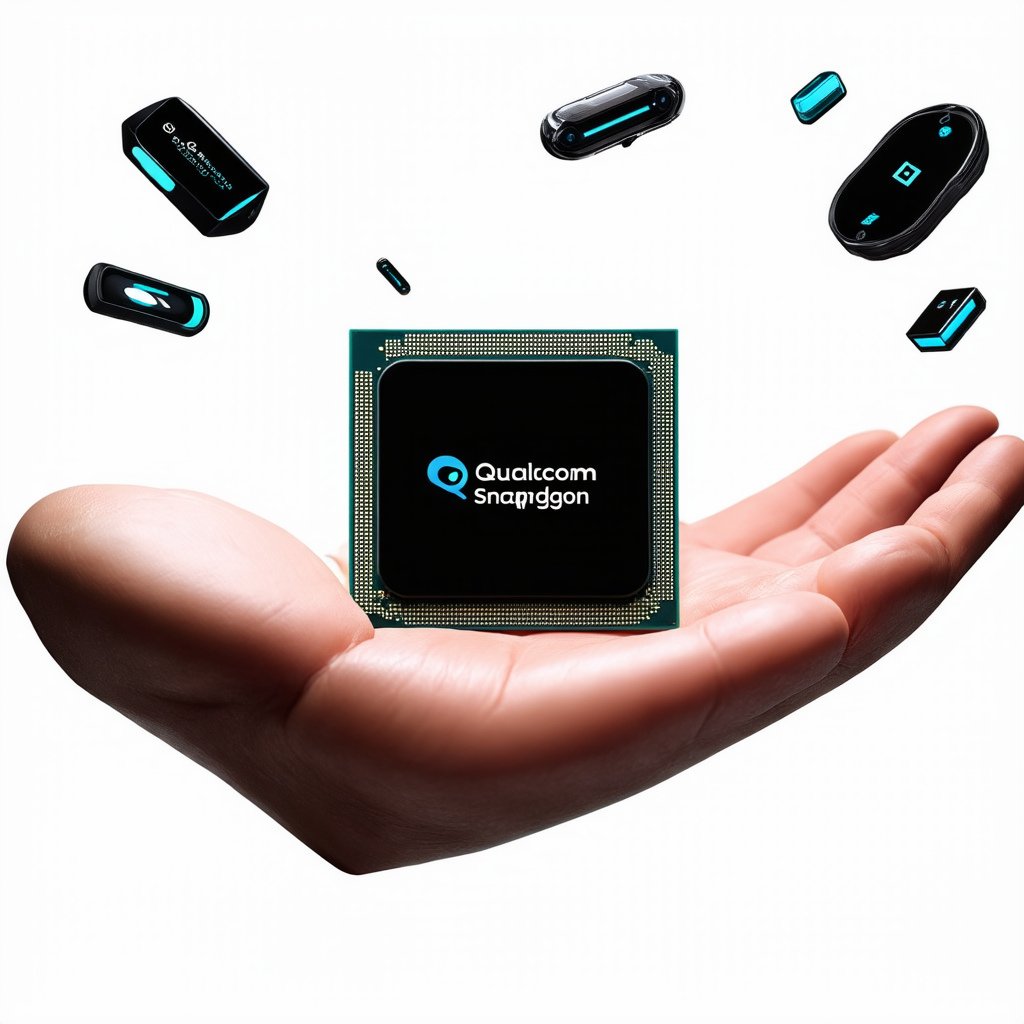} &
        \includegraphics[width=0.111\textwidth]{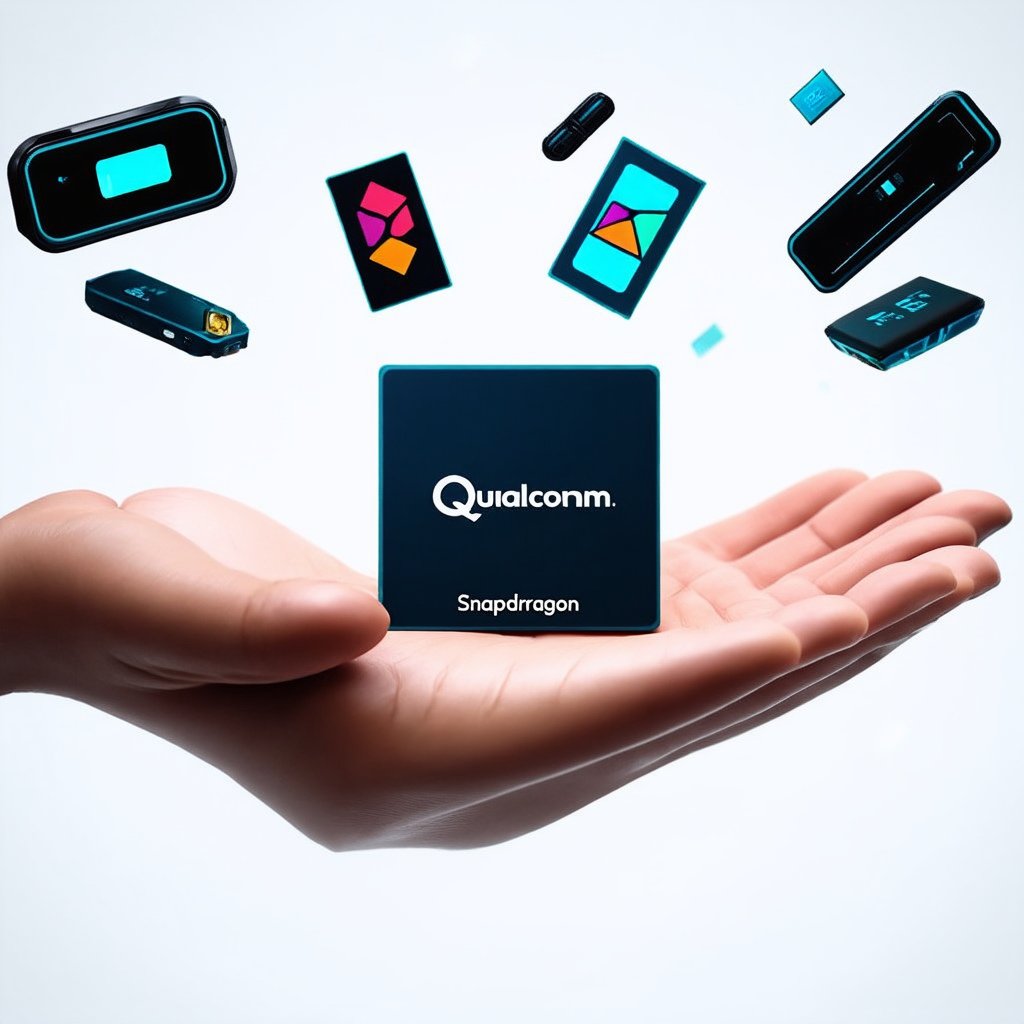} &
                \includegraphics[width=0.111\textwidth]{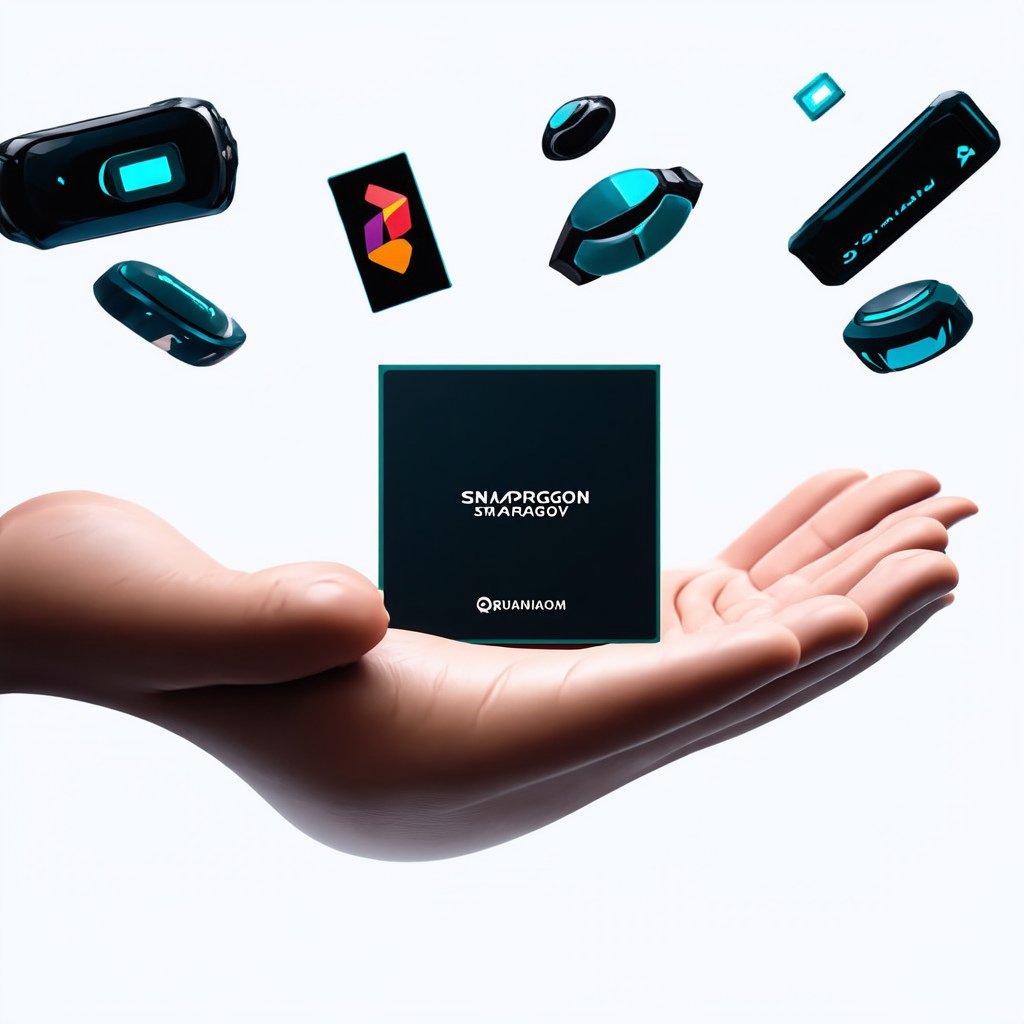} &
                        \includegraphics[width=0.111\textwidth]{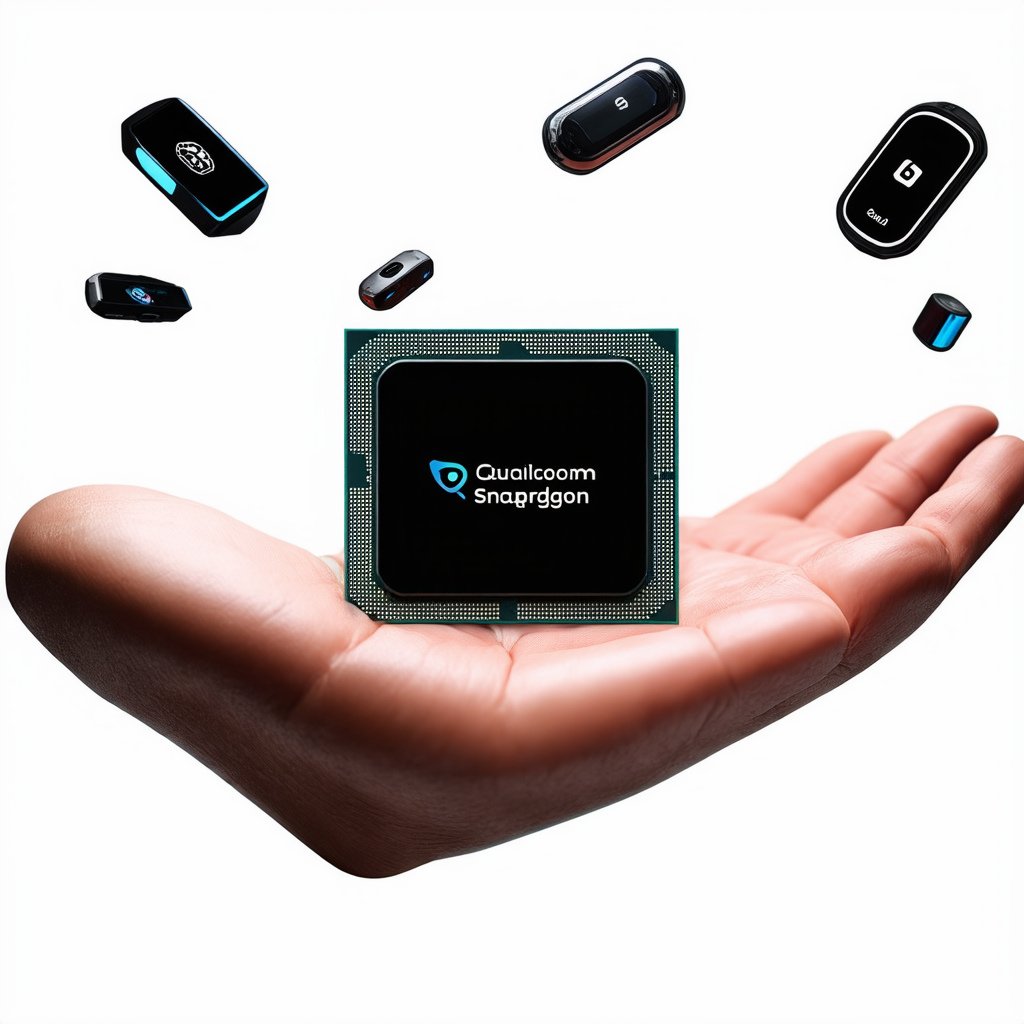} &
        \includegraphics[width=0.111\textwidth]{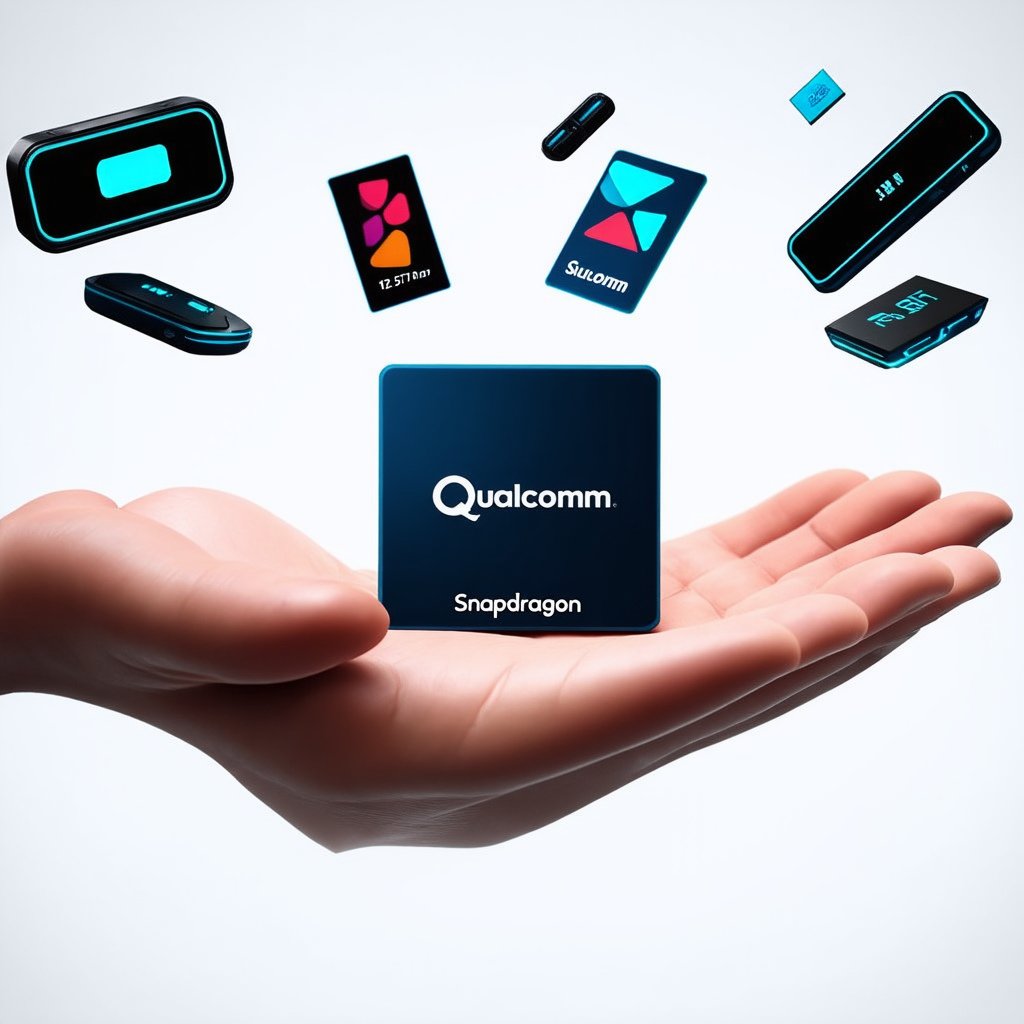} &
                \includegraphics[width=0.111\textwidth]{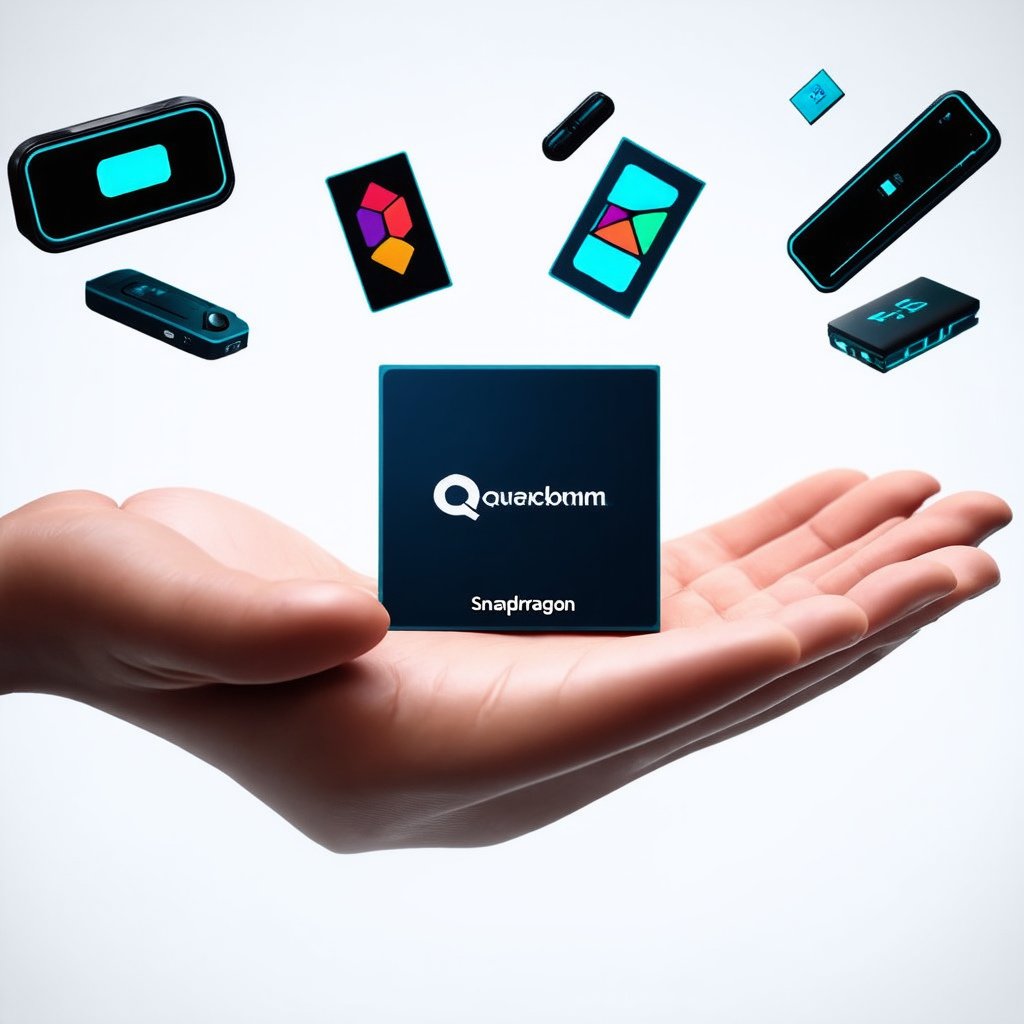} &
                        \includegraphics[width=0.111\textwidth]{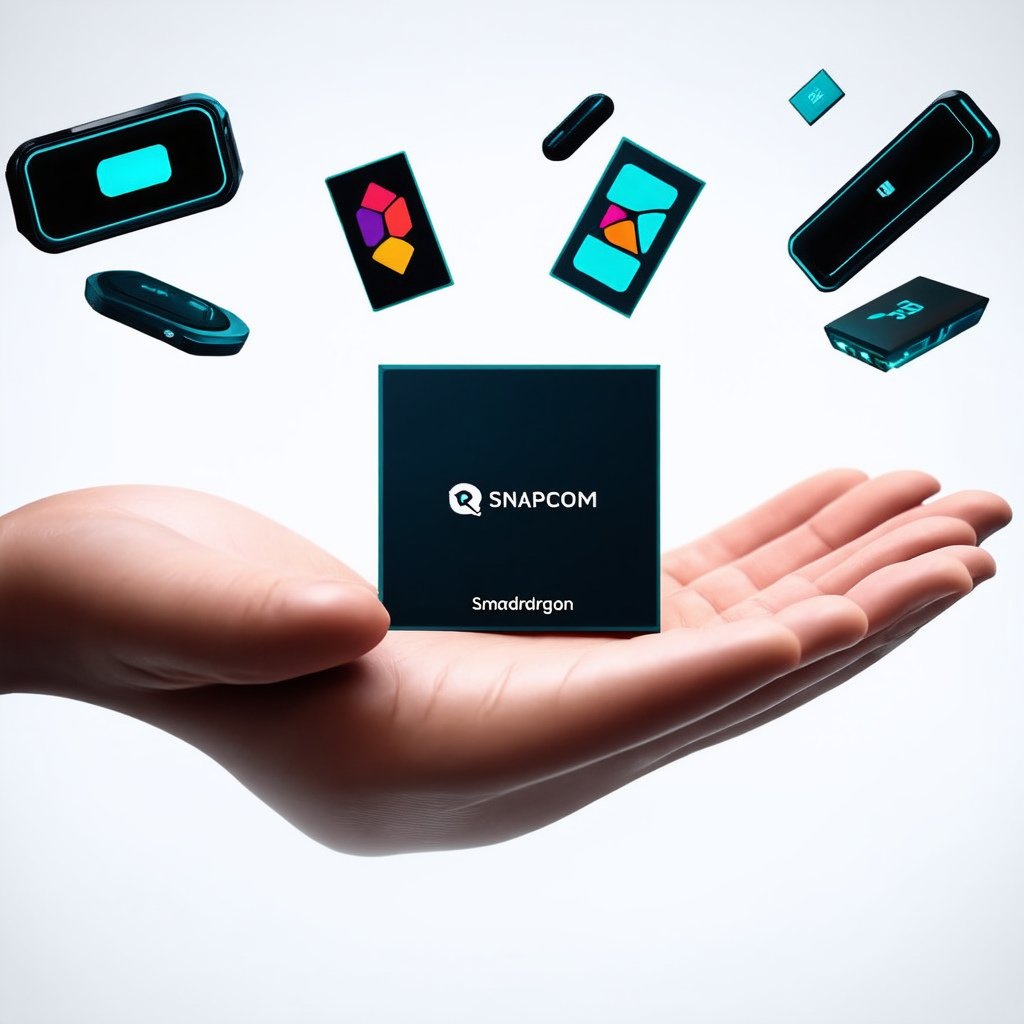} &
        \includegraphics[width=0.111\textwidth]{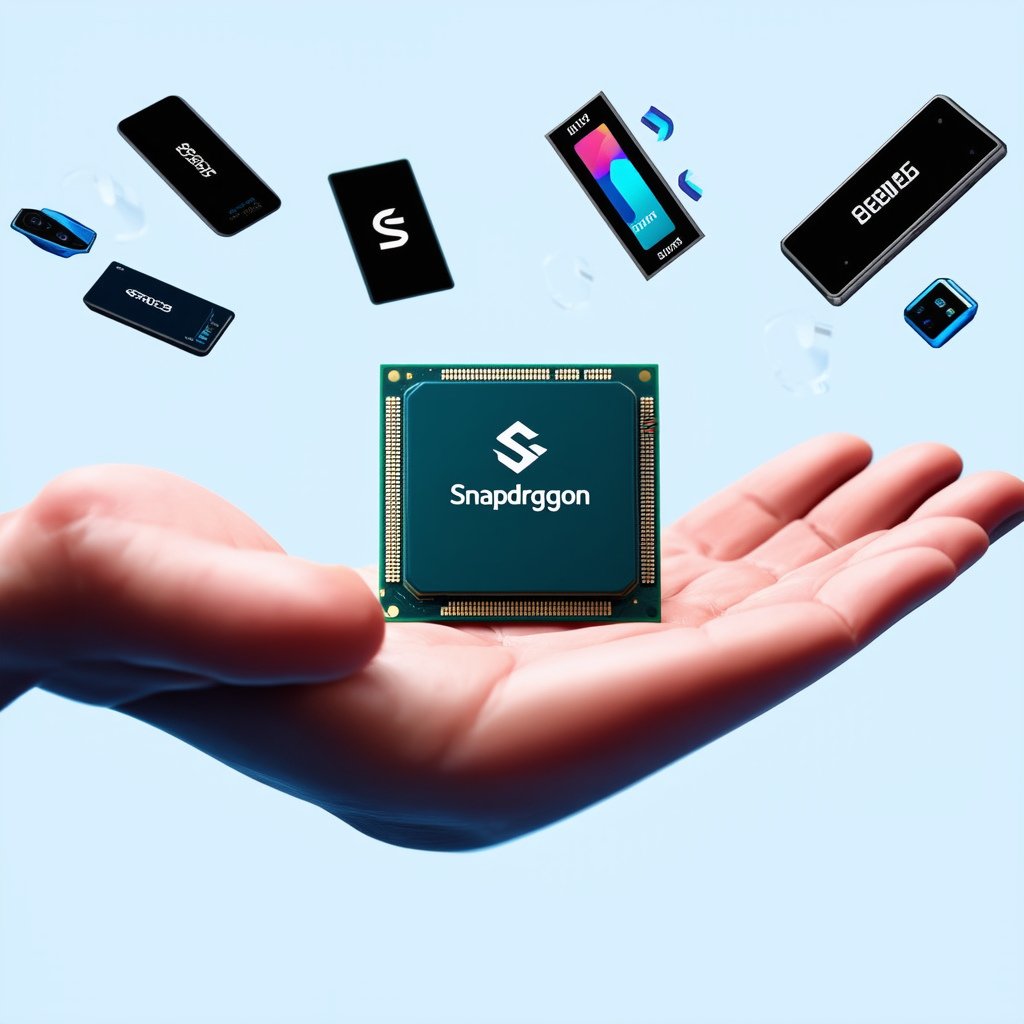} \\

        \includegraphics[width=0.111\textwidth]{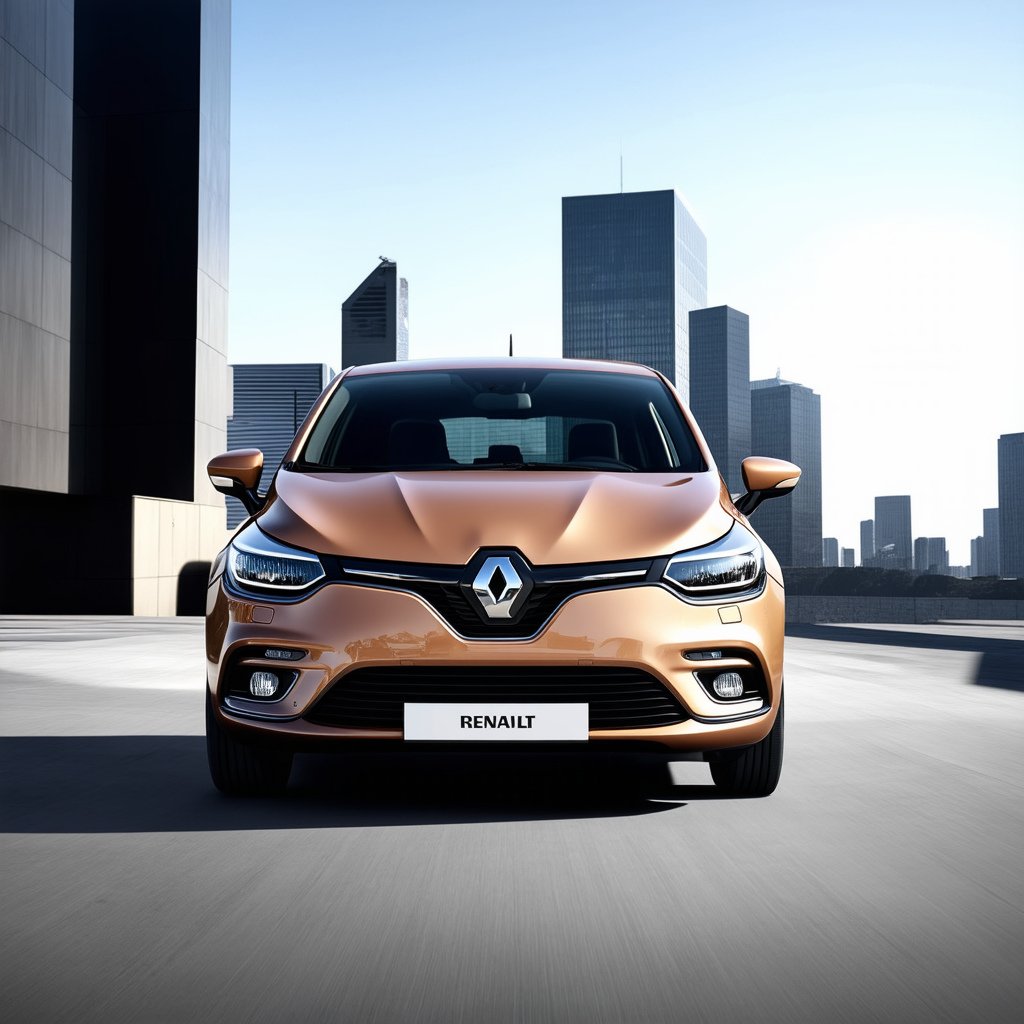} &
        \includegraphics[width=0.111\textwidth]{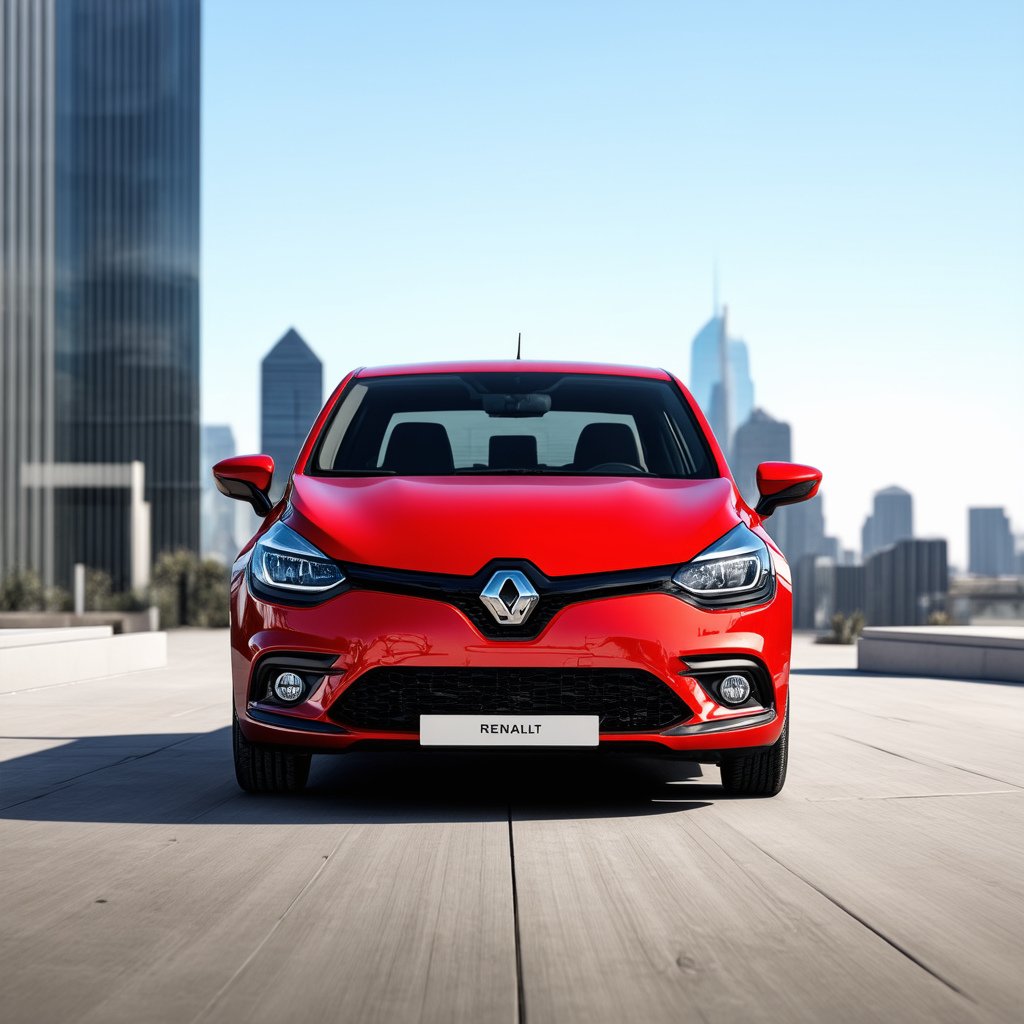} &
        \includegraphics[width=0.111\textwidth]{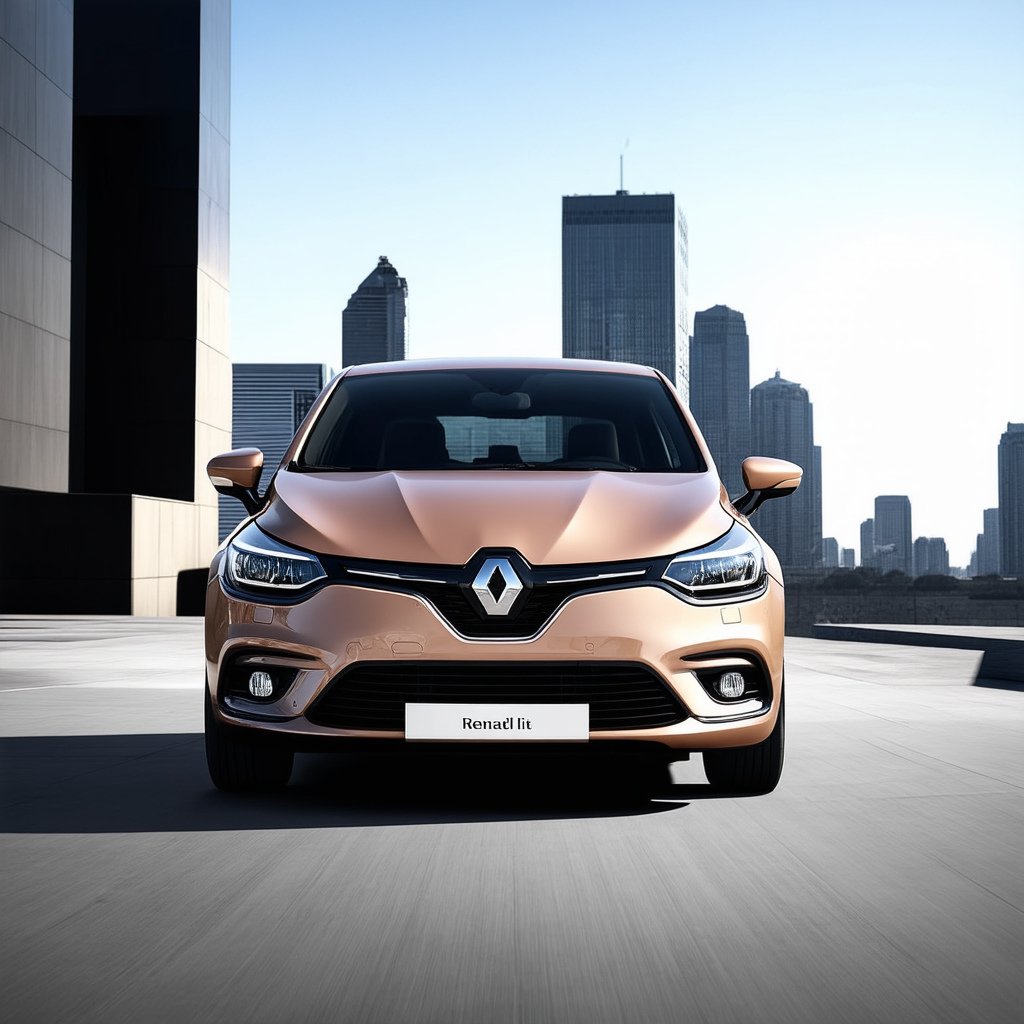} &
                \includegraphics[width=0.111\textwidth]{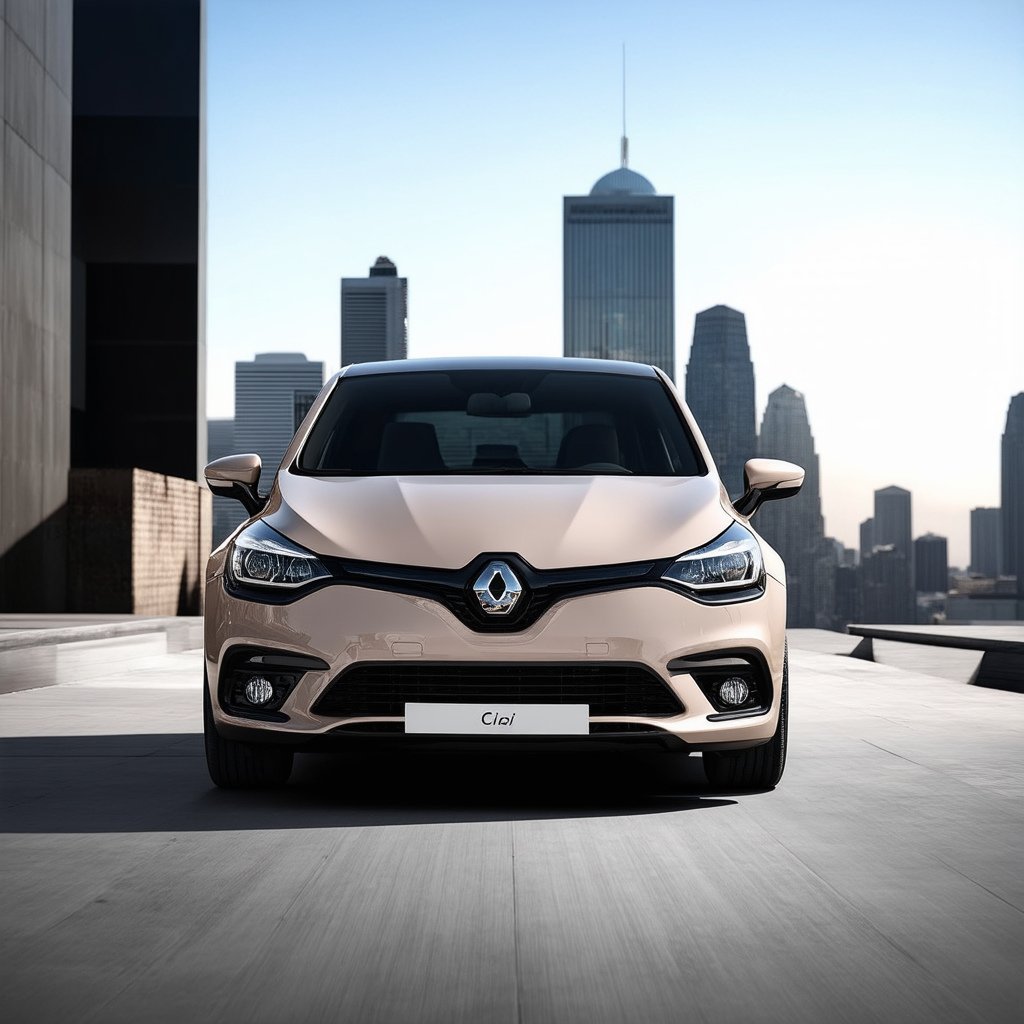} &
                        \includegraphics[width=0.111\textwidth]{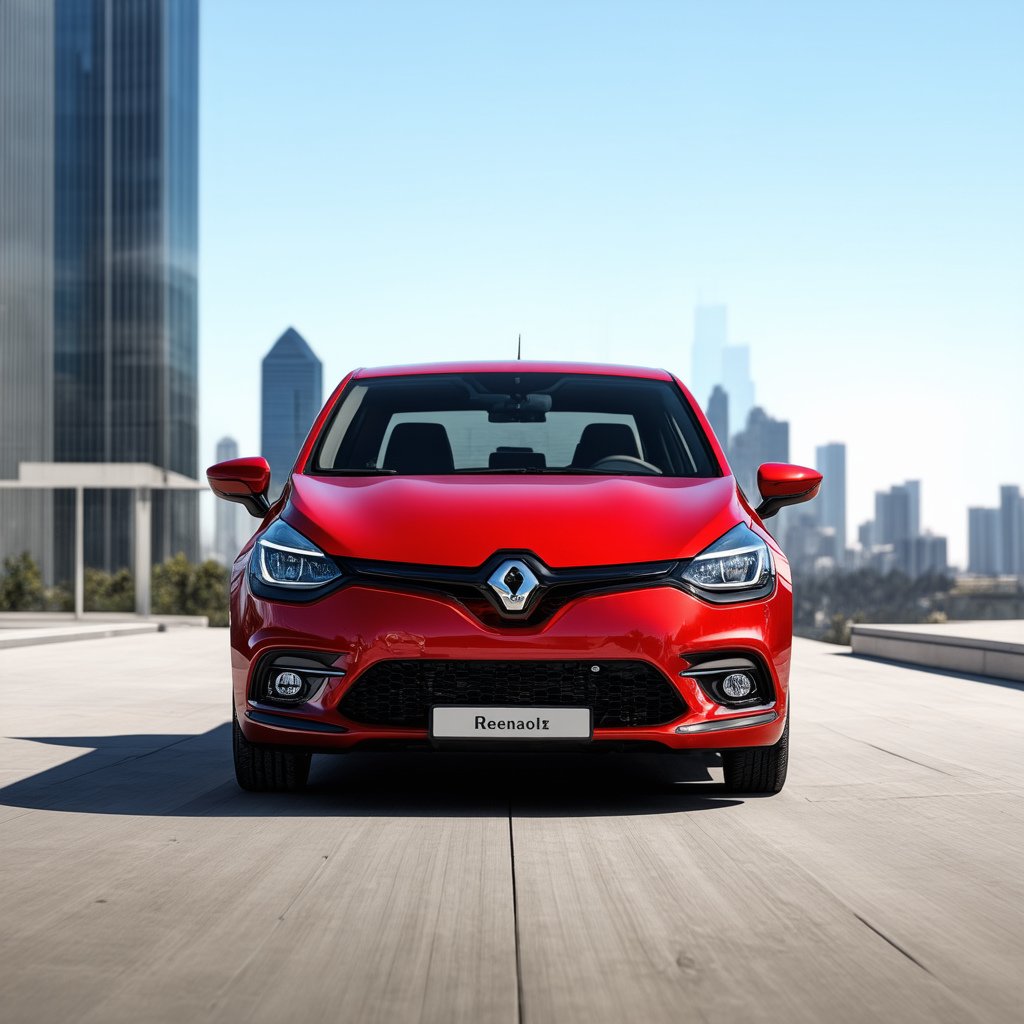} &
        \includegraphics[width=0.111\textwidth]{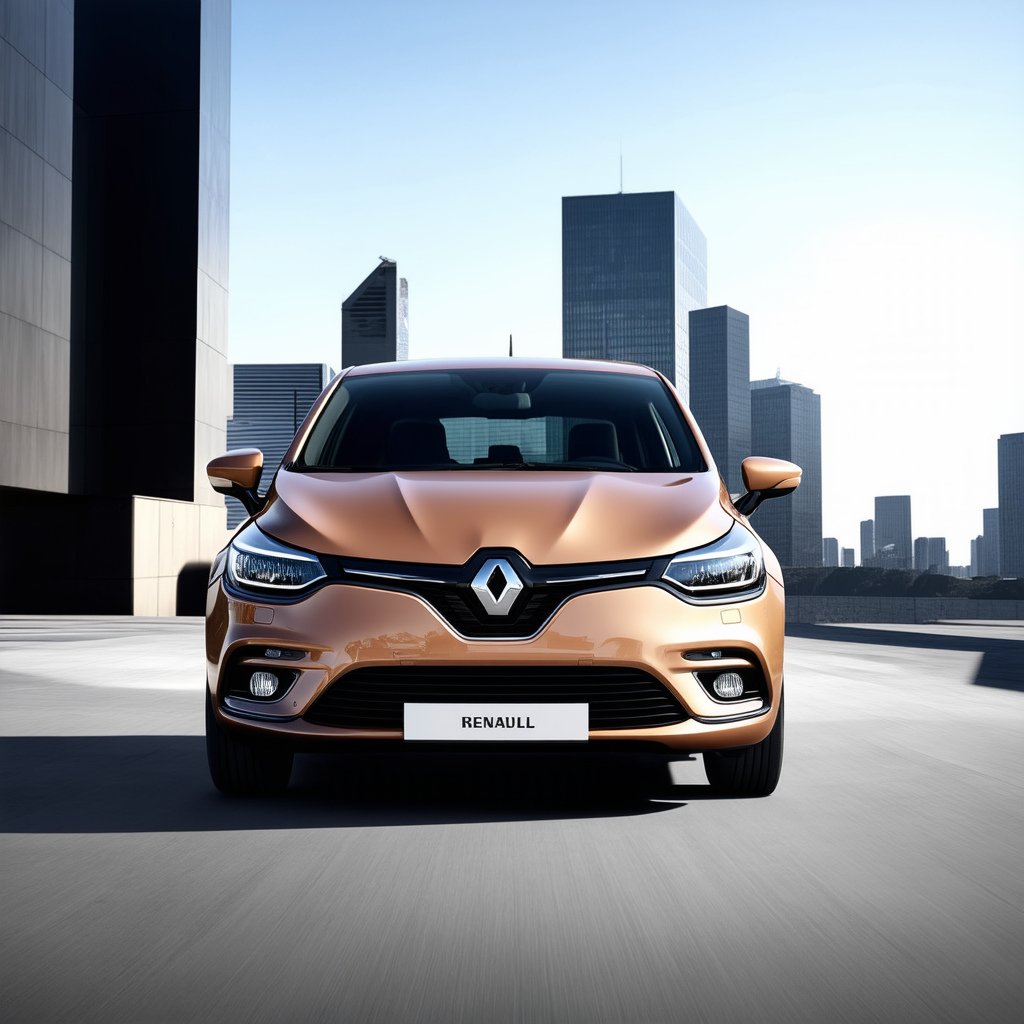} &
                \includegraphics[width=0.111\textwidth]{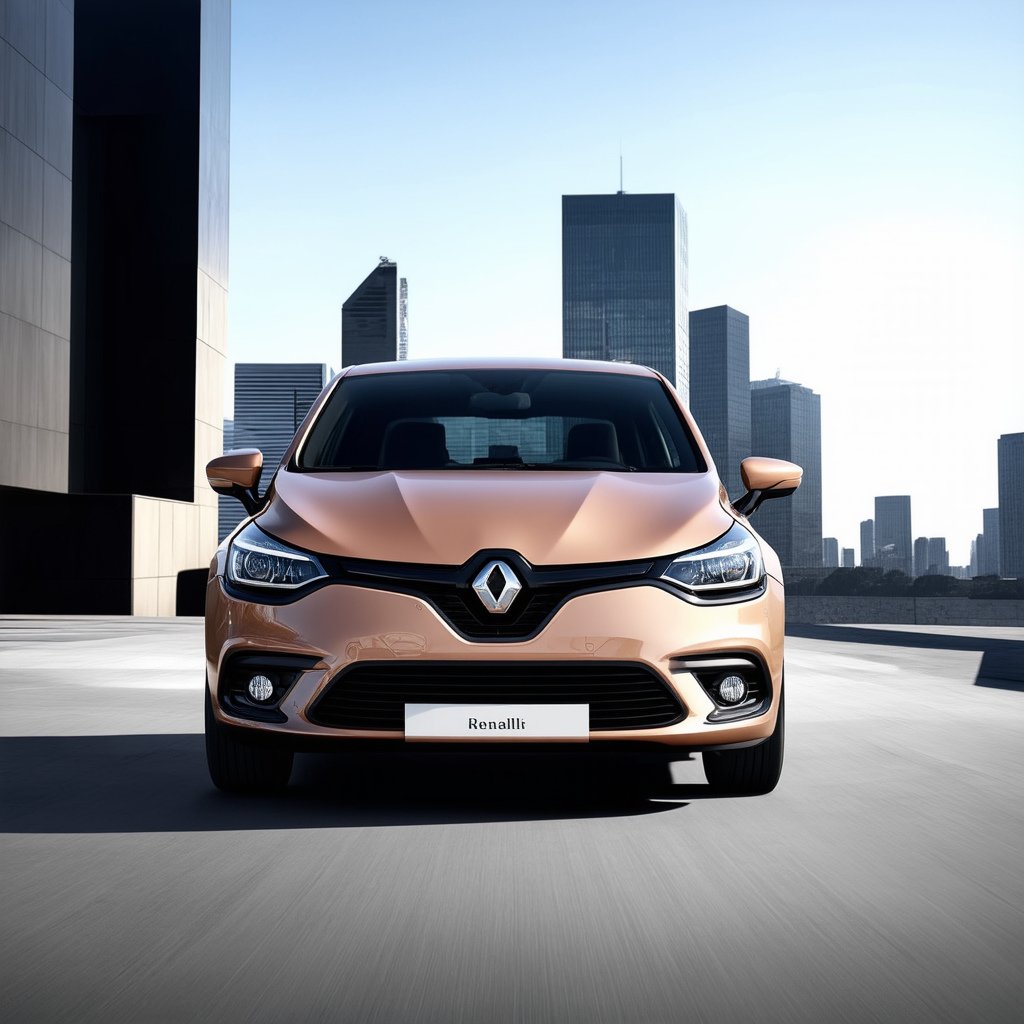} &
                        \includegraphics[width=0.111\textwidth]{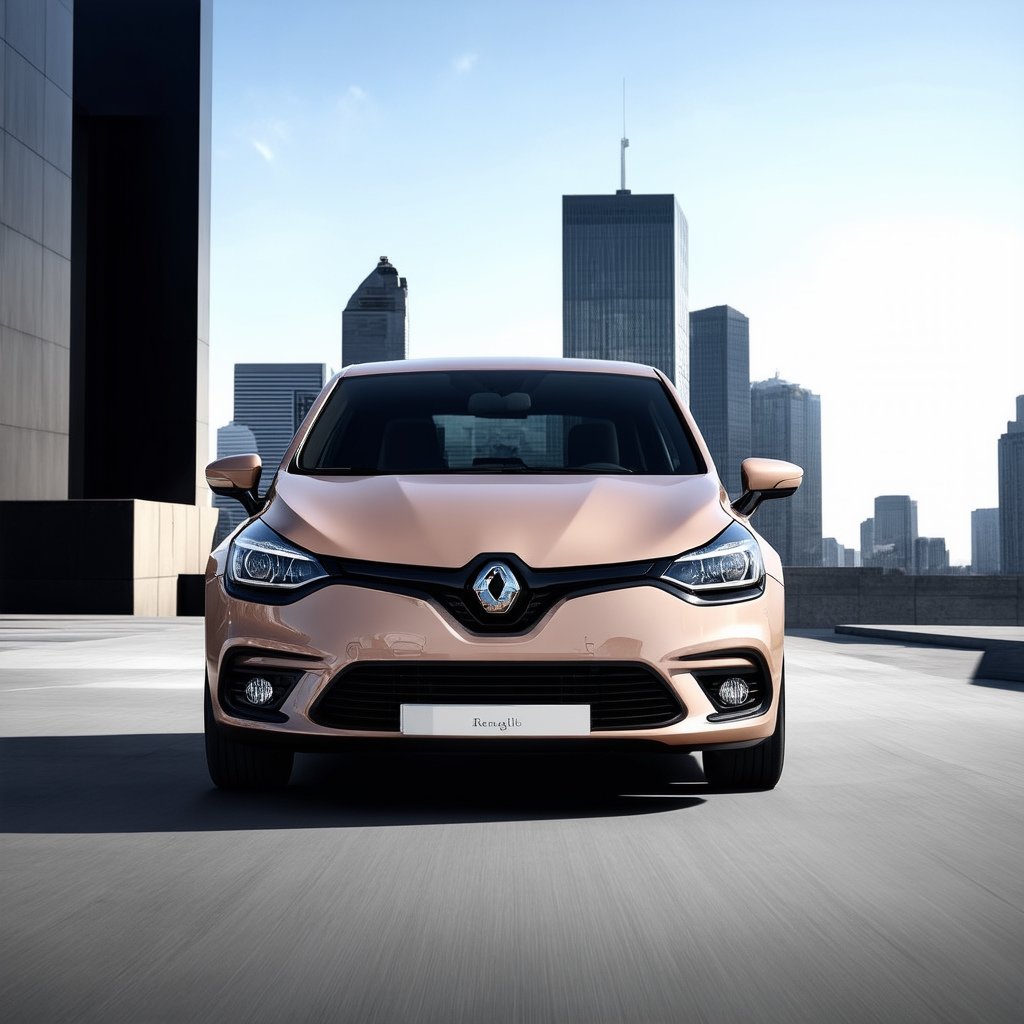} &
        \includegraphics[width=0.111\textwidth]{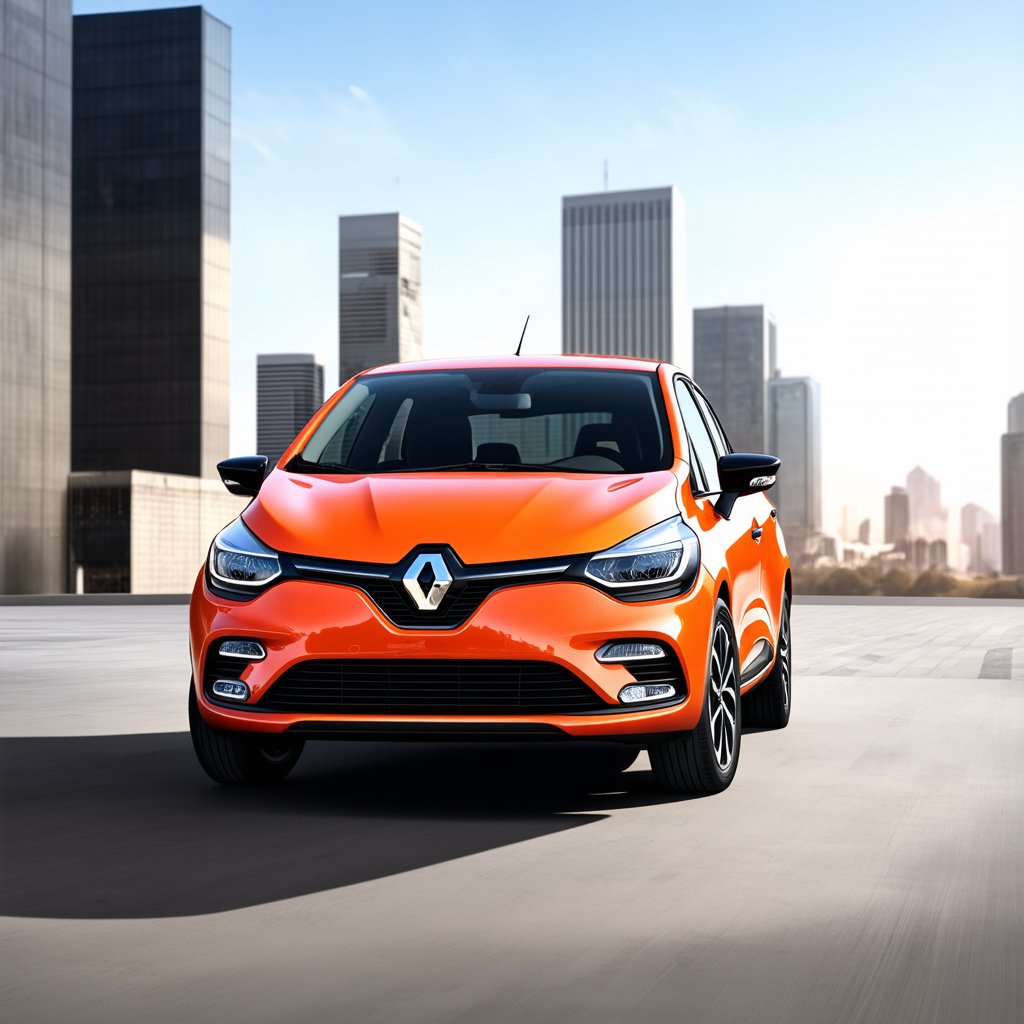} \\
        
        \includegraphics[width=0.111\textwidth]{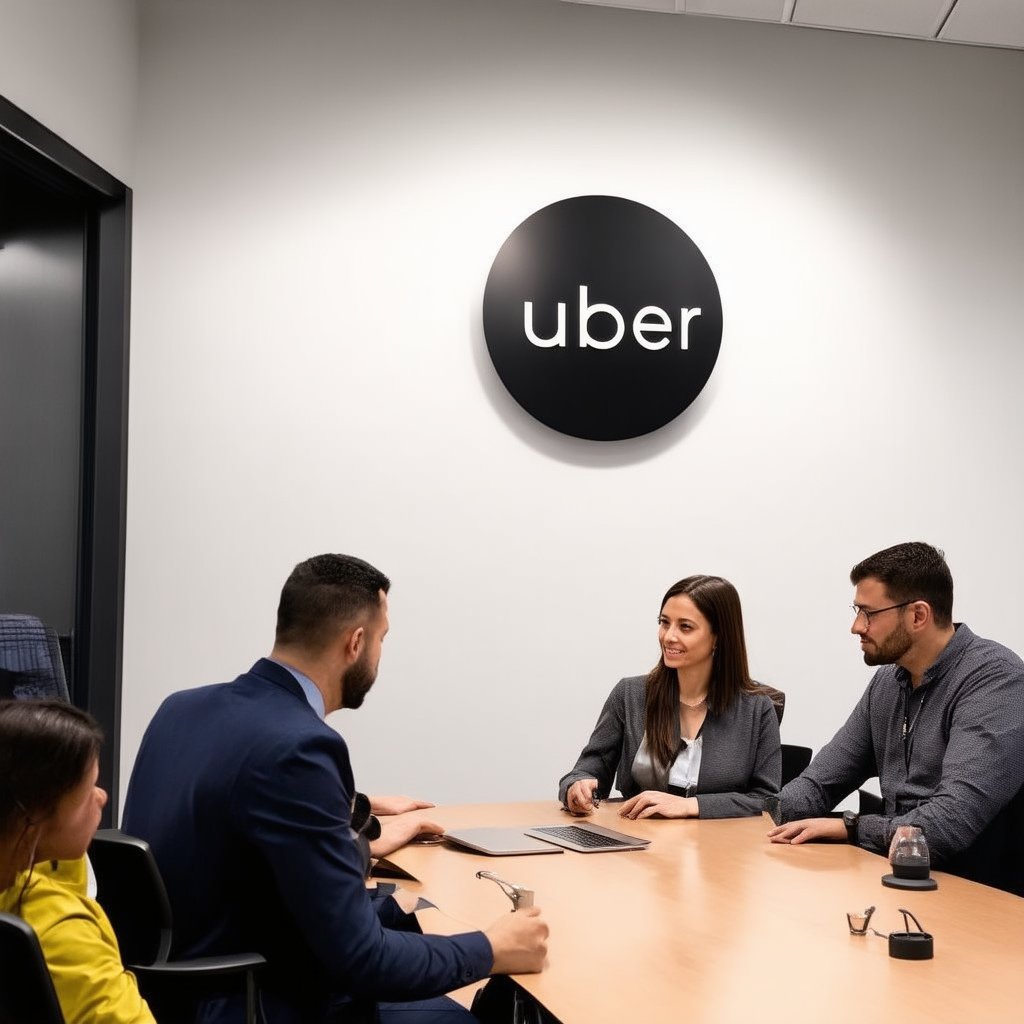} &
        \includegraphics[width=0.111\textwidth]{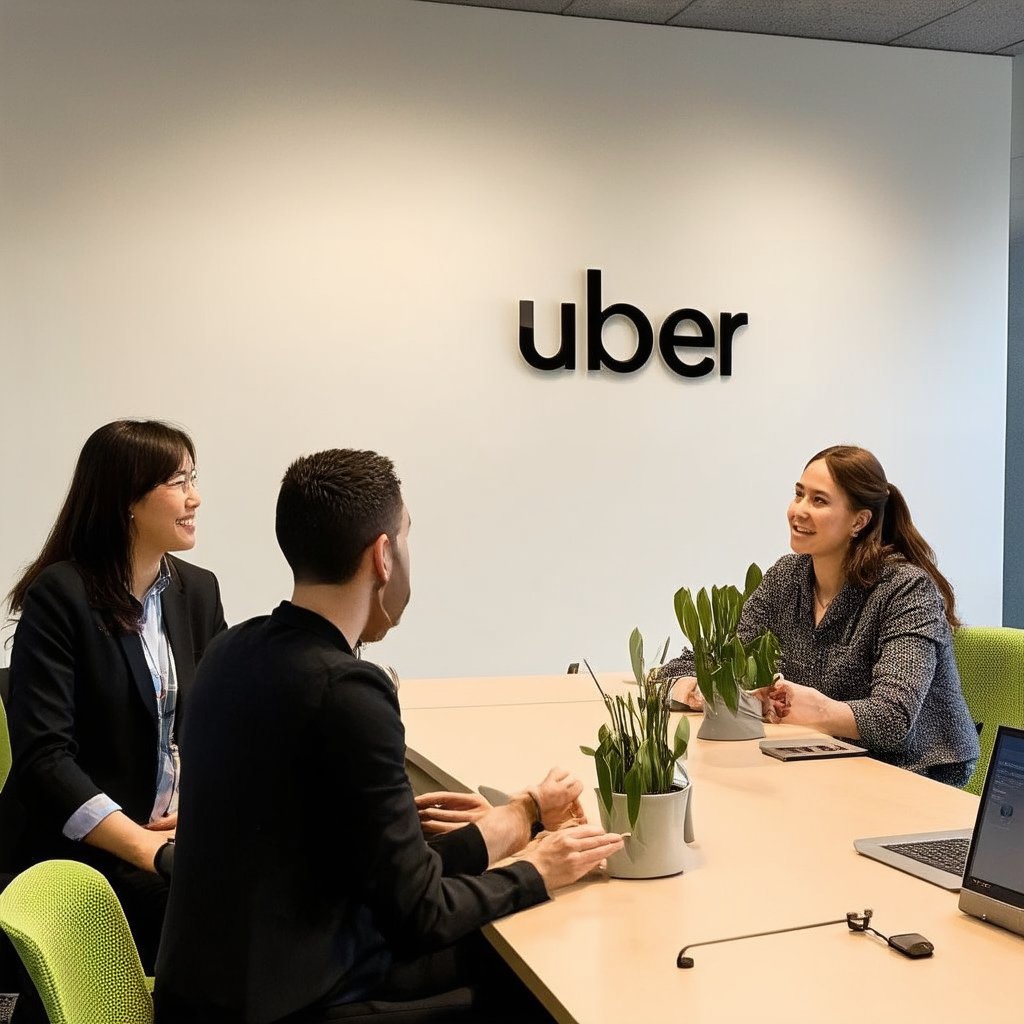} &
        \includegraphics[width=0.111\textwidth]{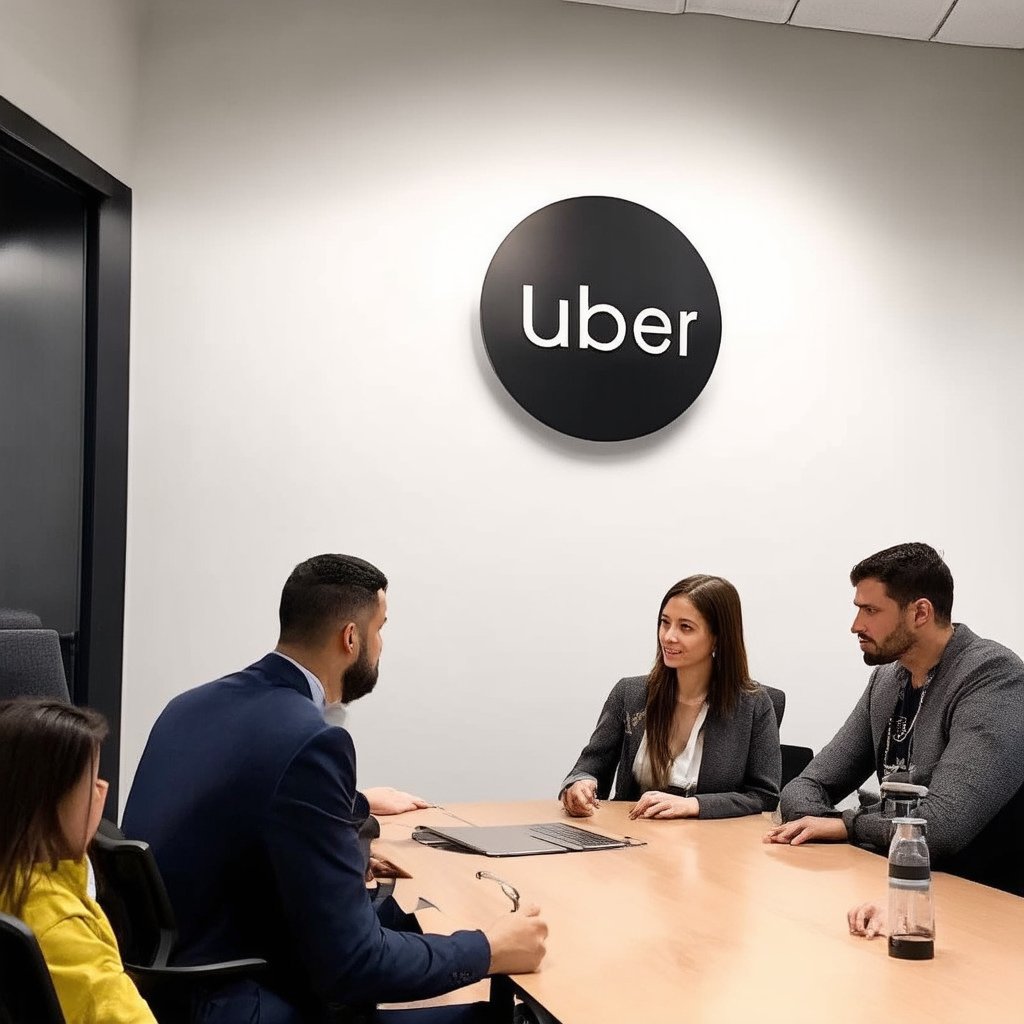} &
                \includegraphics[width=0.111\textwidth]{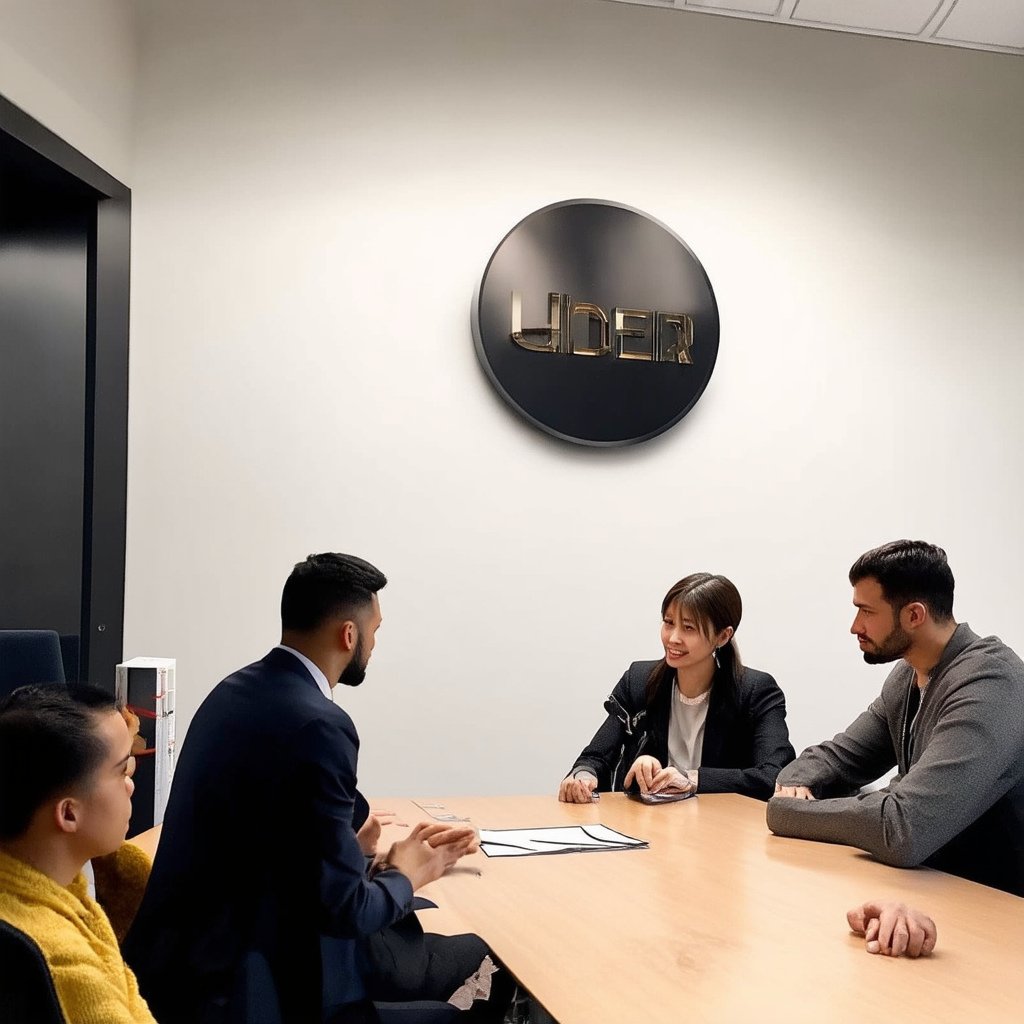} &
                        \includegraphics[width=0.111\textwidth]{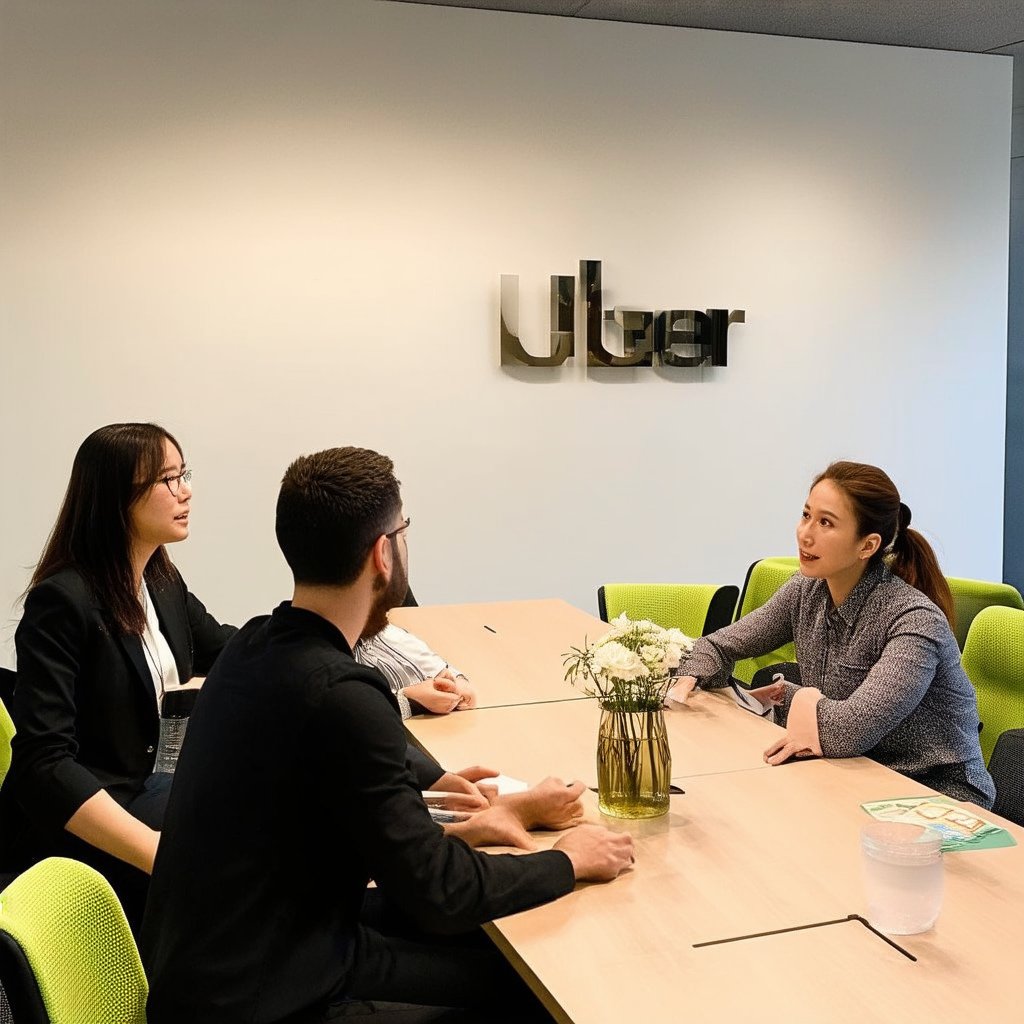} &
        \includegraphics[width=0.111\textwidth]{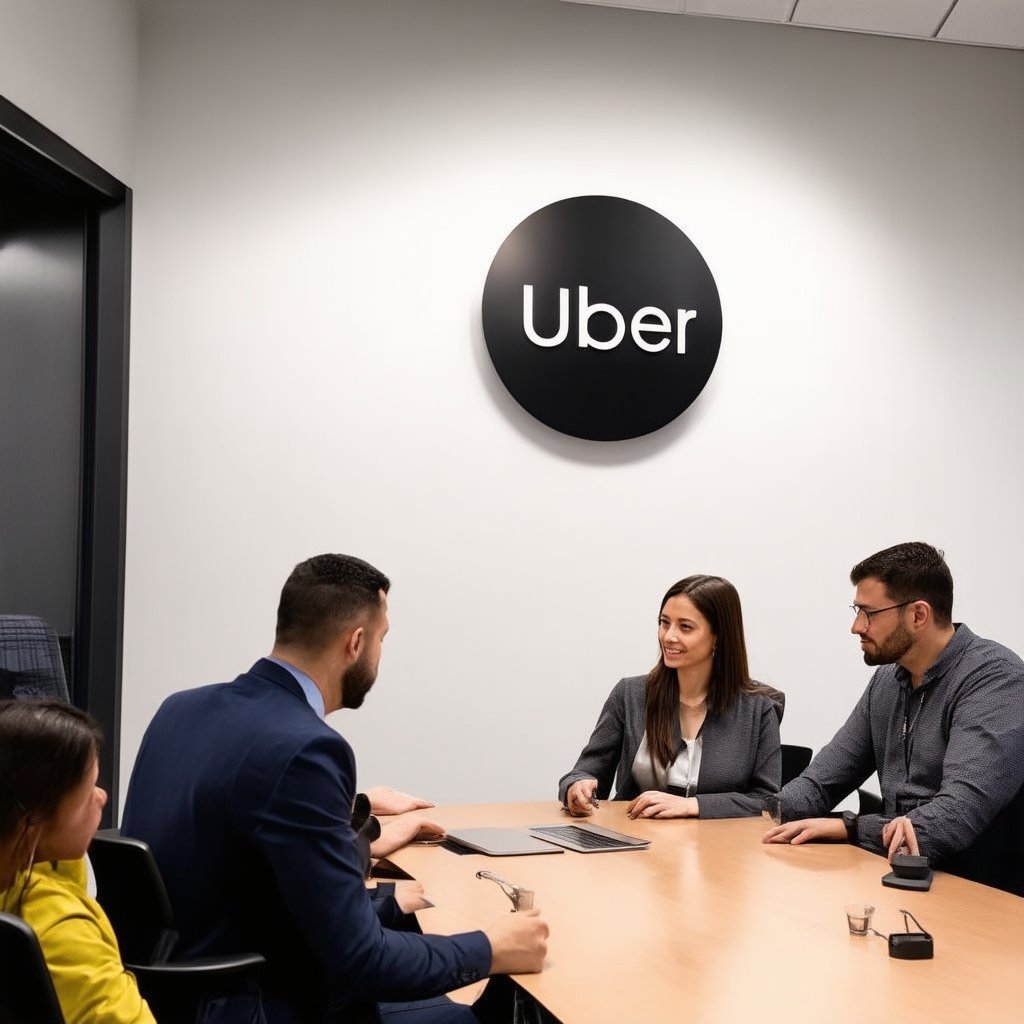} &
                \includegraphics[width=0.111\textwidth]{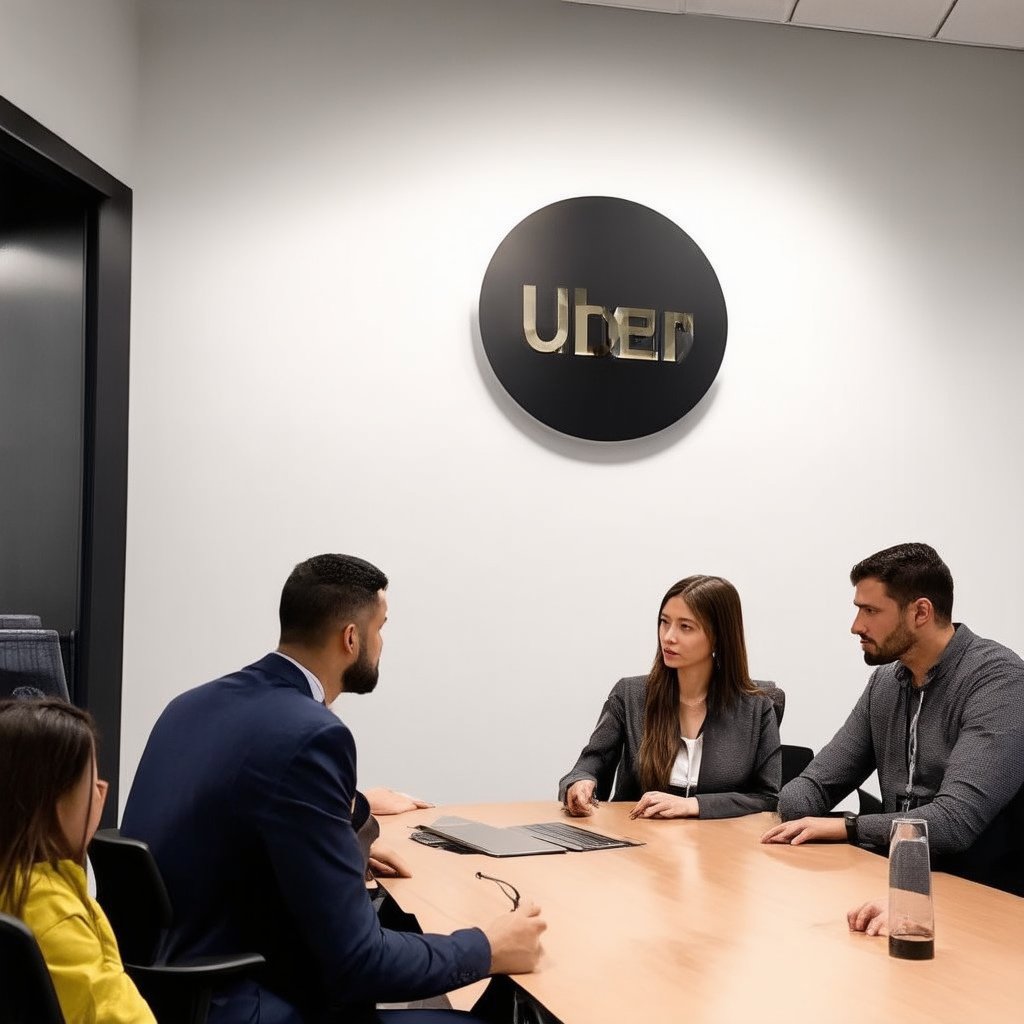} &
                        \includegraphics[width=0.111\textwidth]{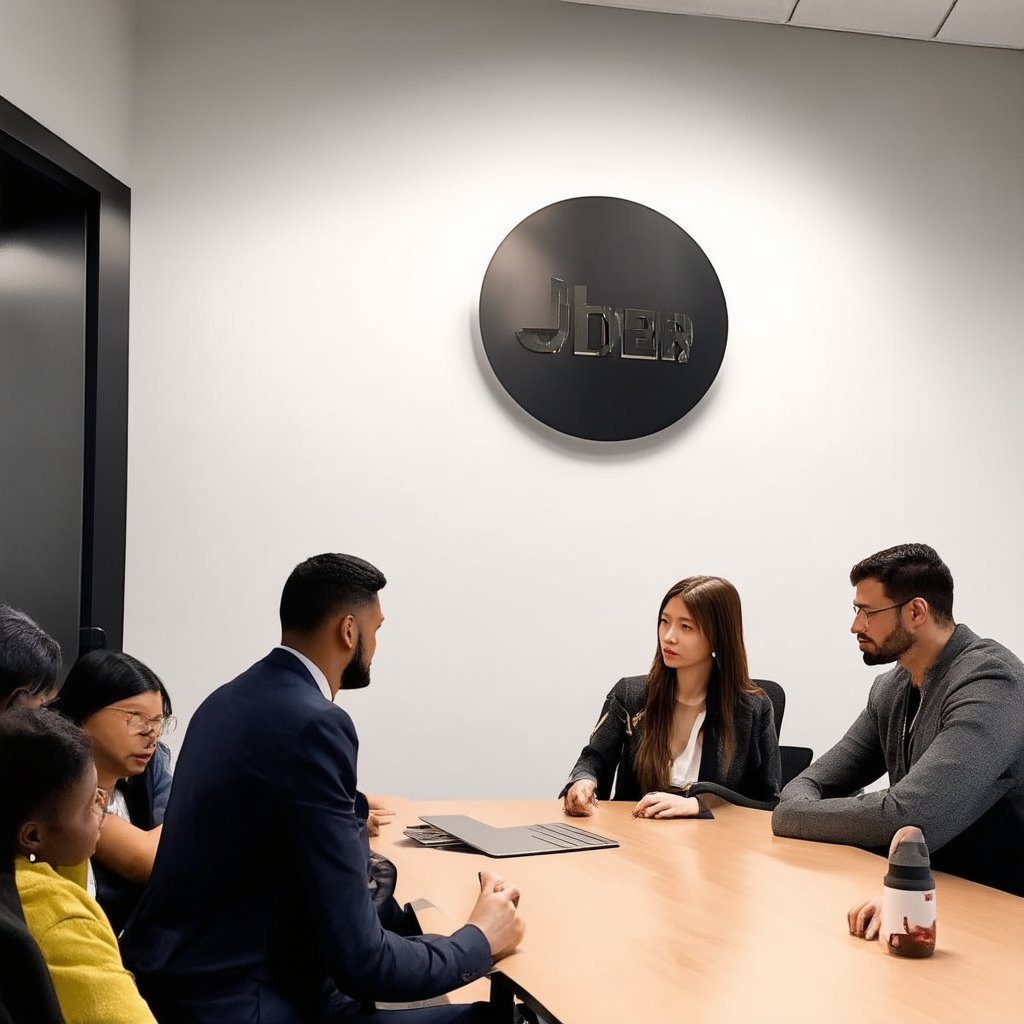} &
        \includegraphics[width=0.111\textwidth]{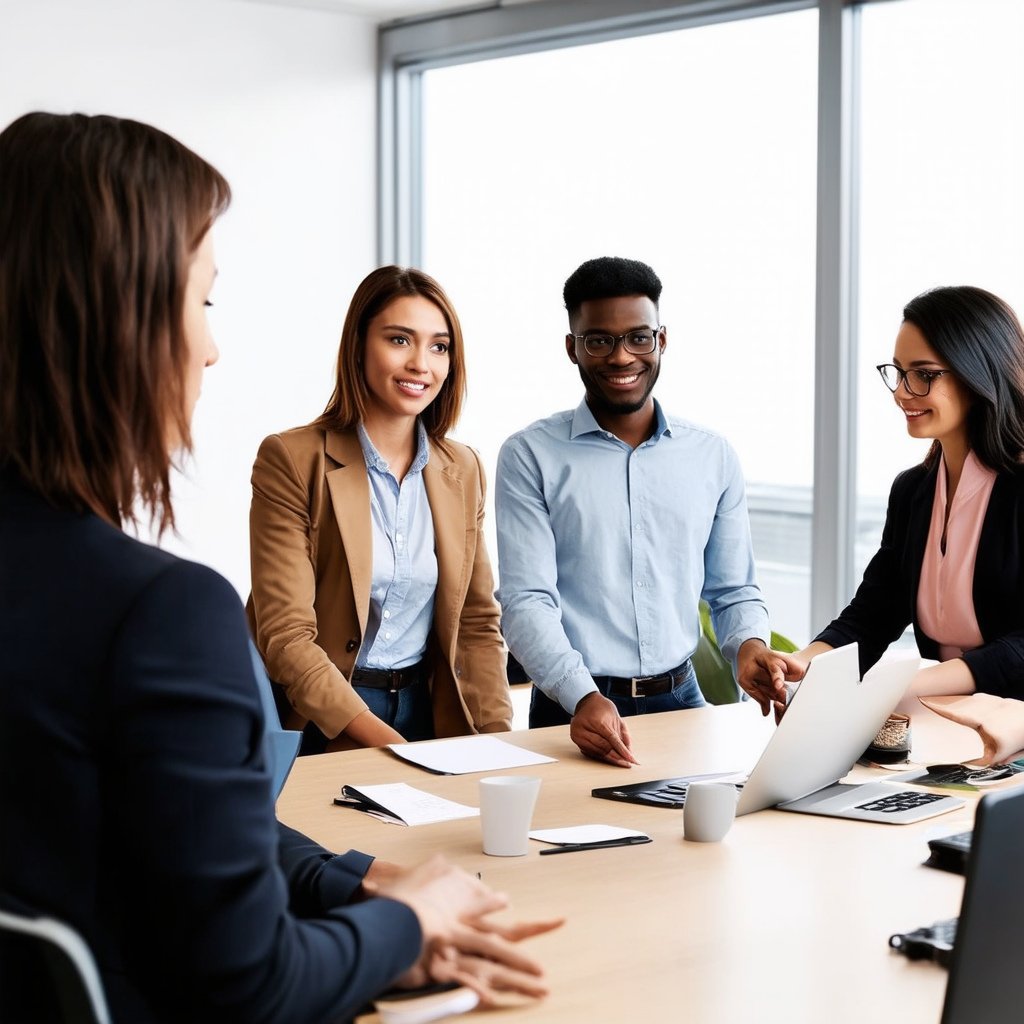} \\

        \includegraphics[width=0.111\textwidth]{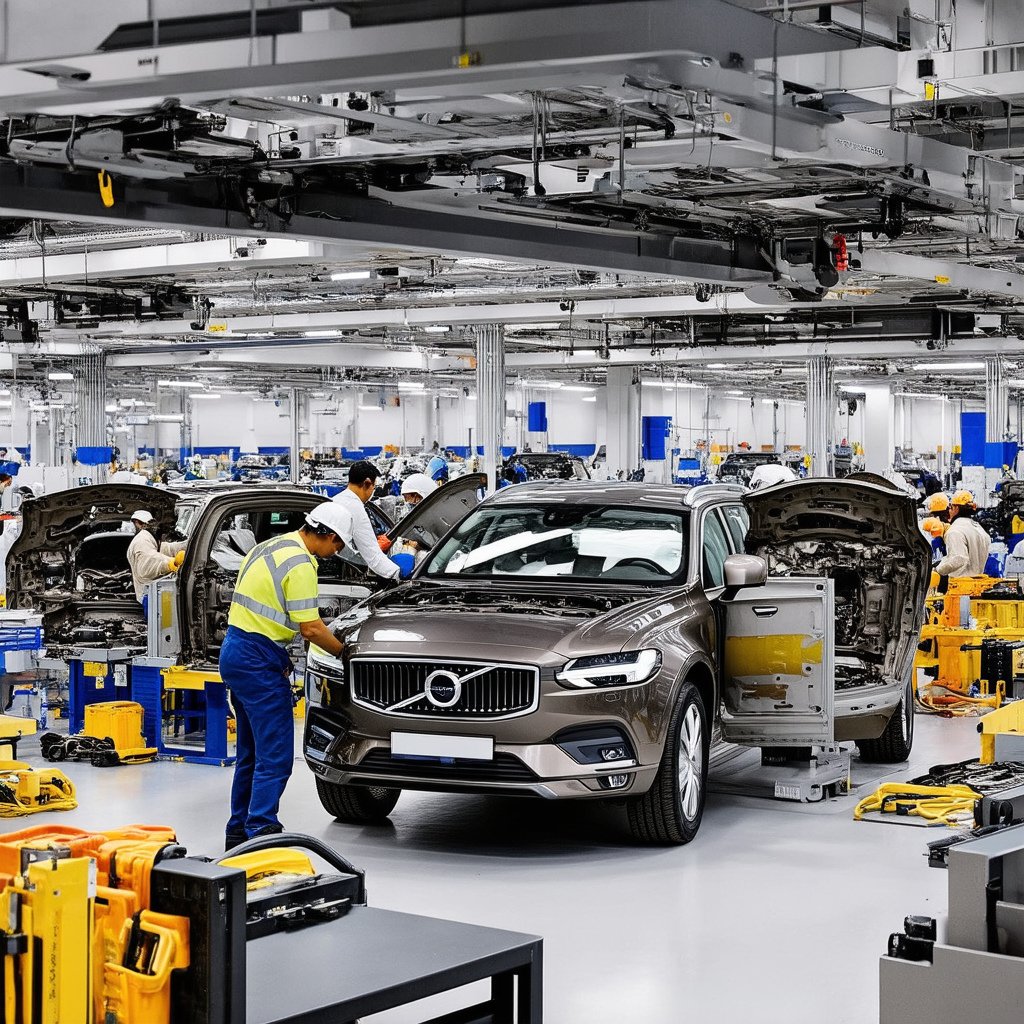} &
        \includegraphics[width=0.111\textwidth]{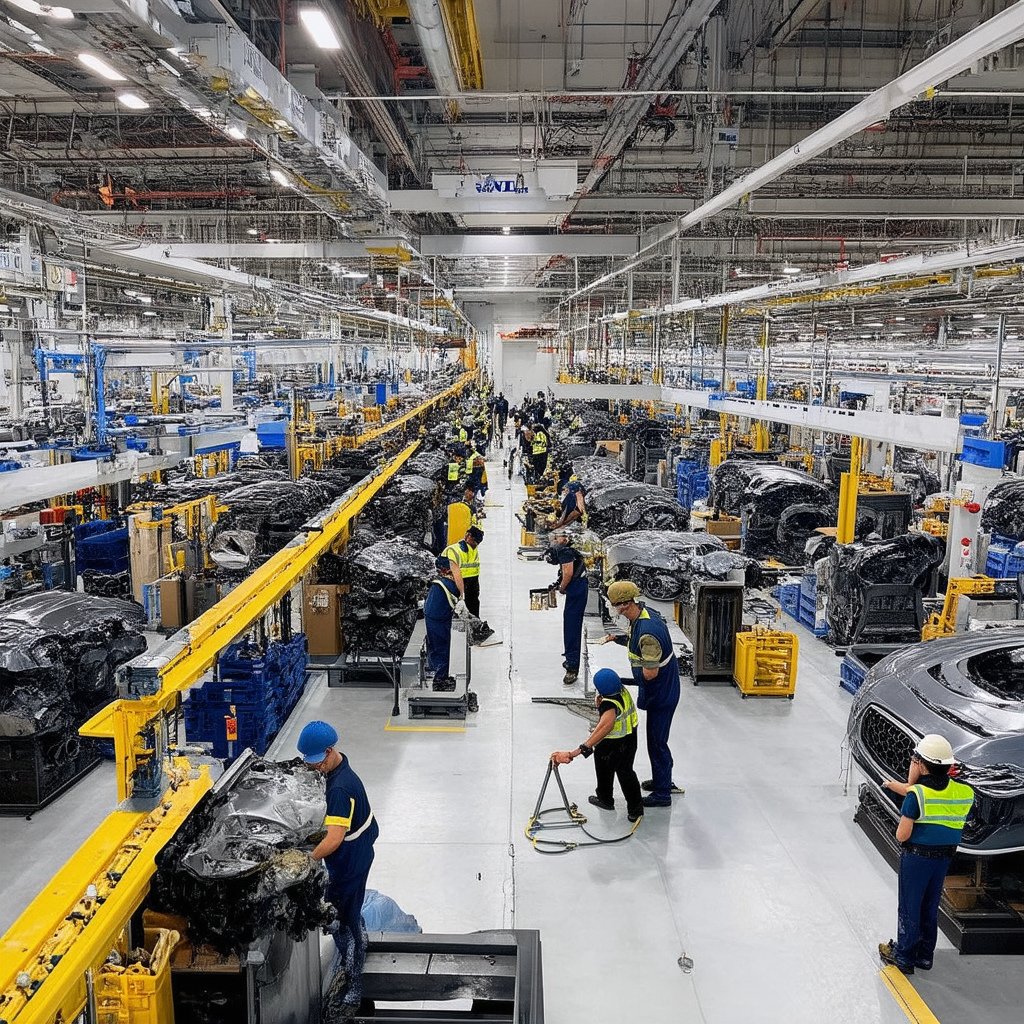} &
        \includegraphics[width=0.111\textwidth]{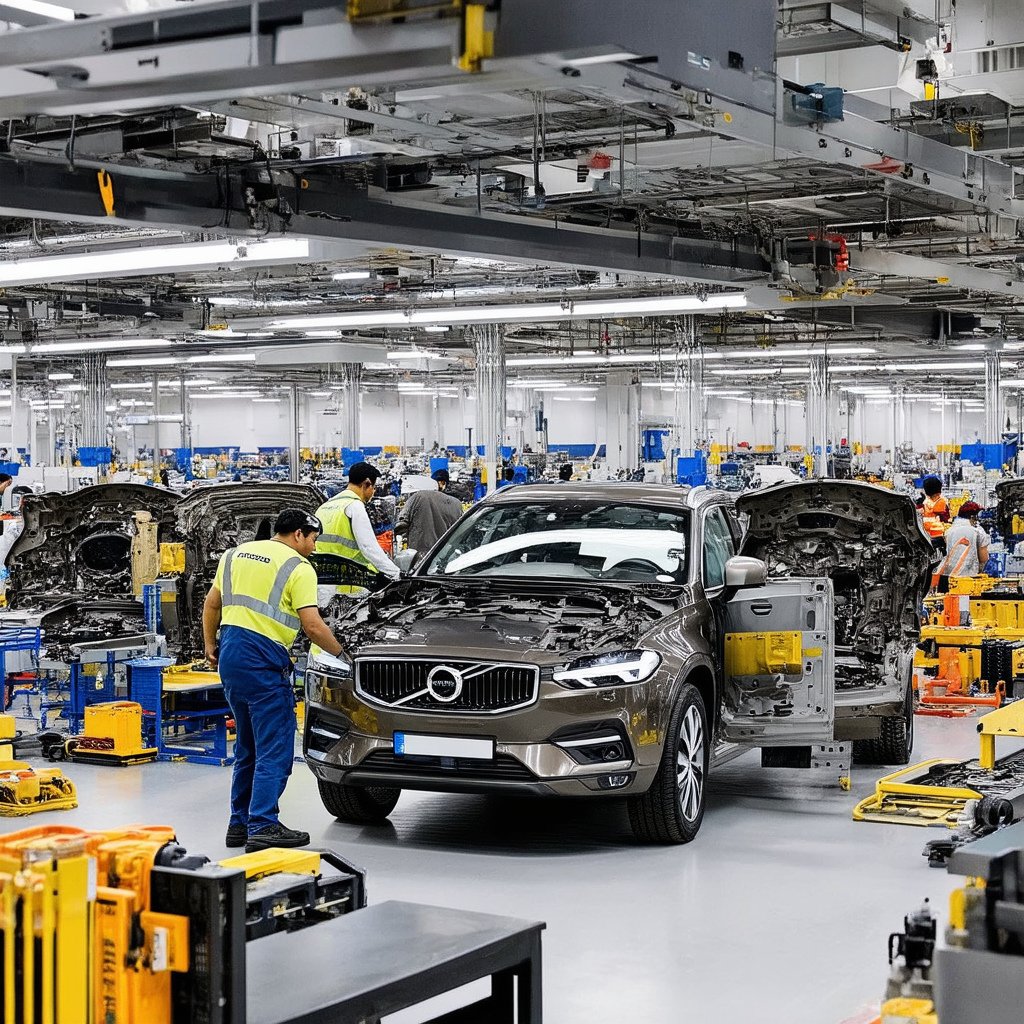} &
                \includegraphics[width=0.111\textwidth]{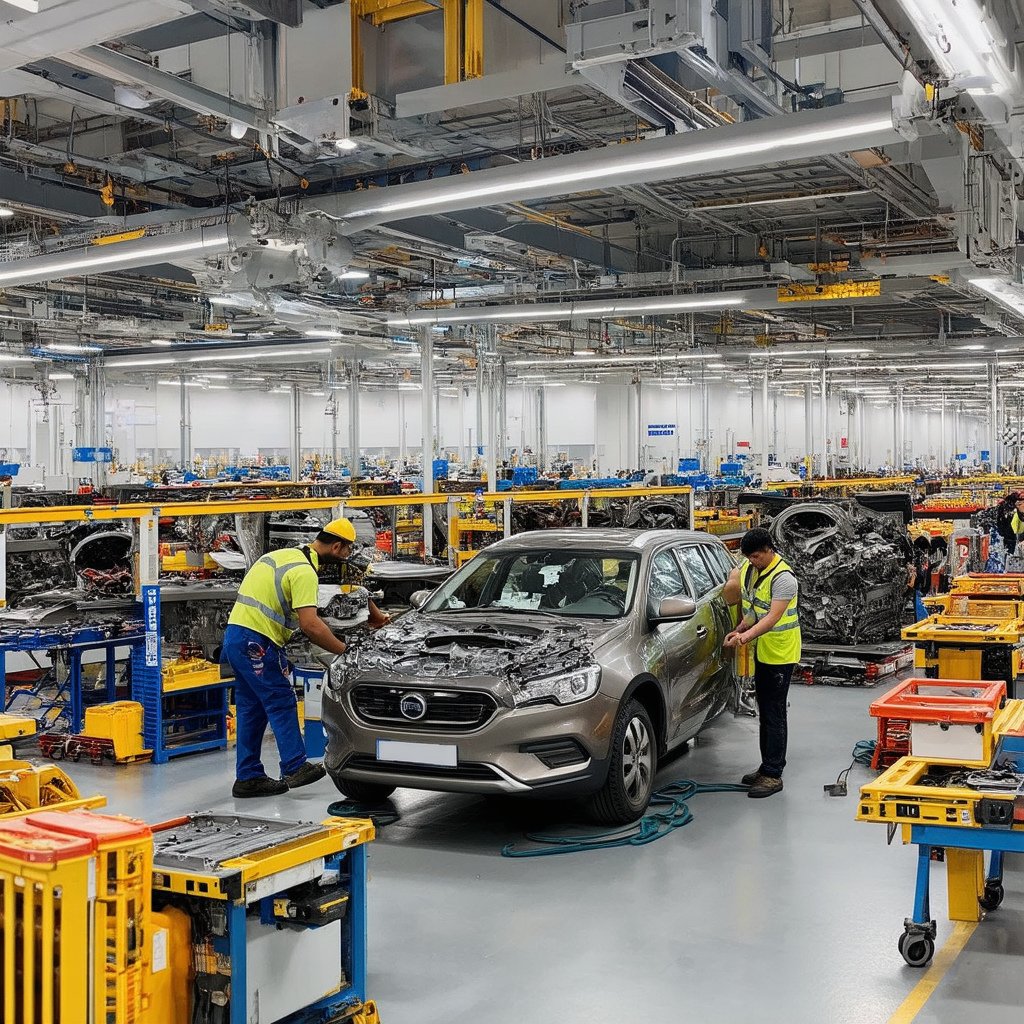} &
                        \includegraphics[width=0.111\textwidth]{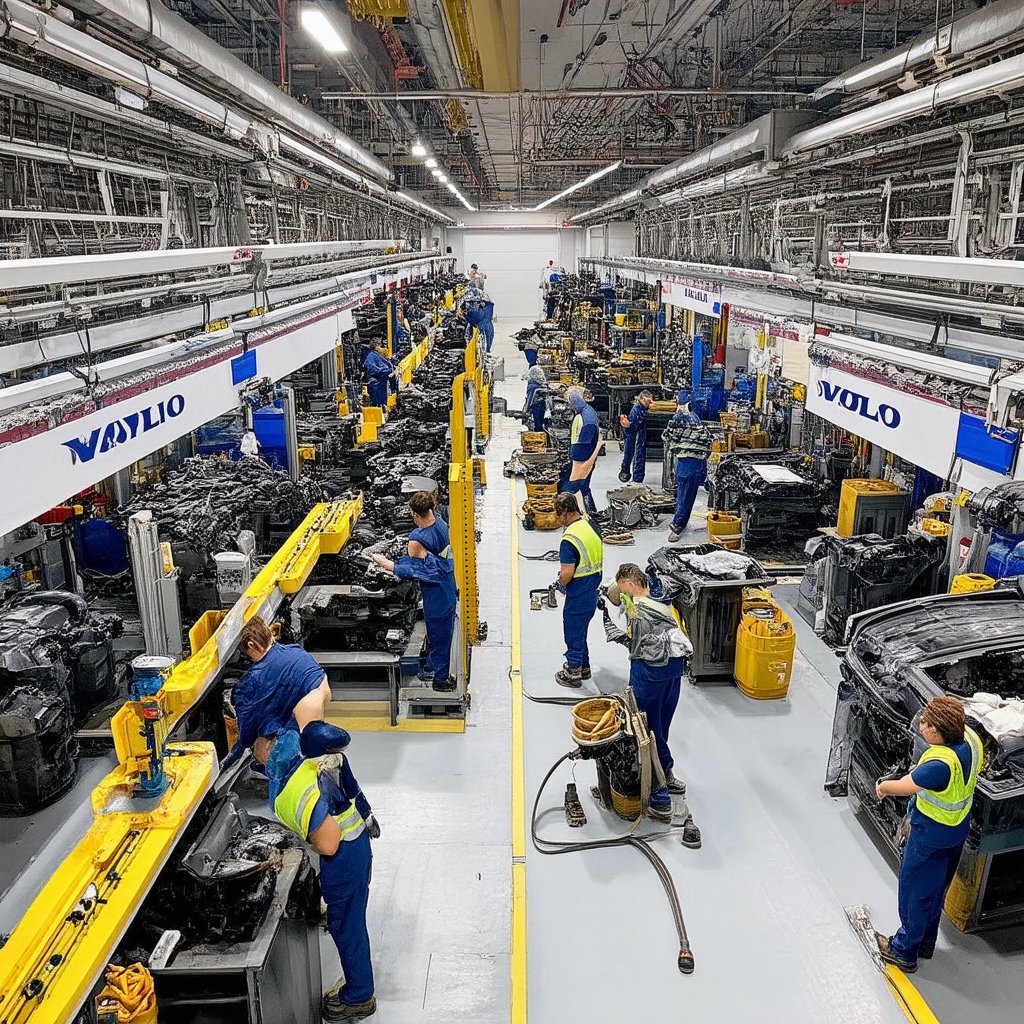} &
        \includegraphics[width=0.111\textwidth]{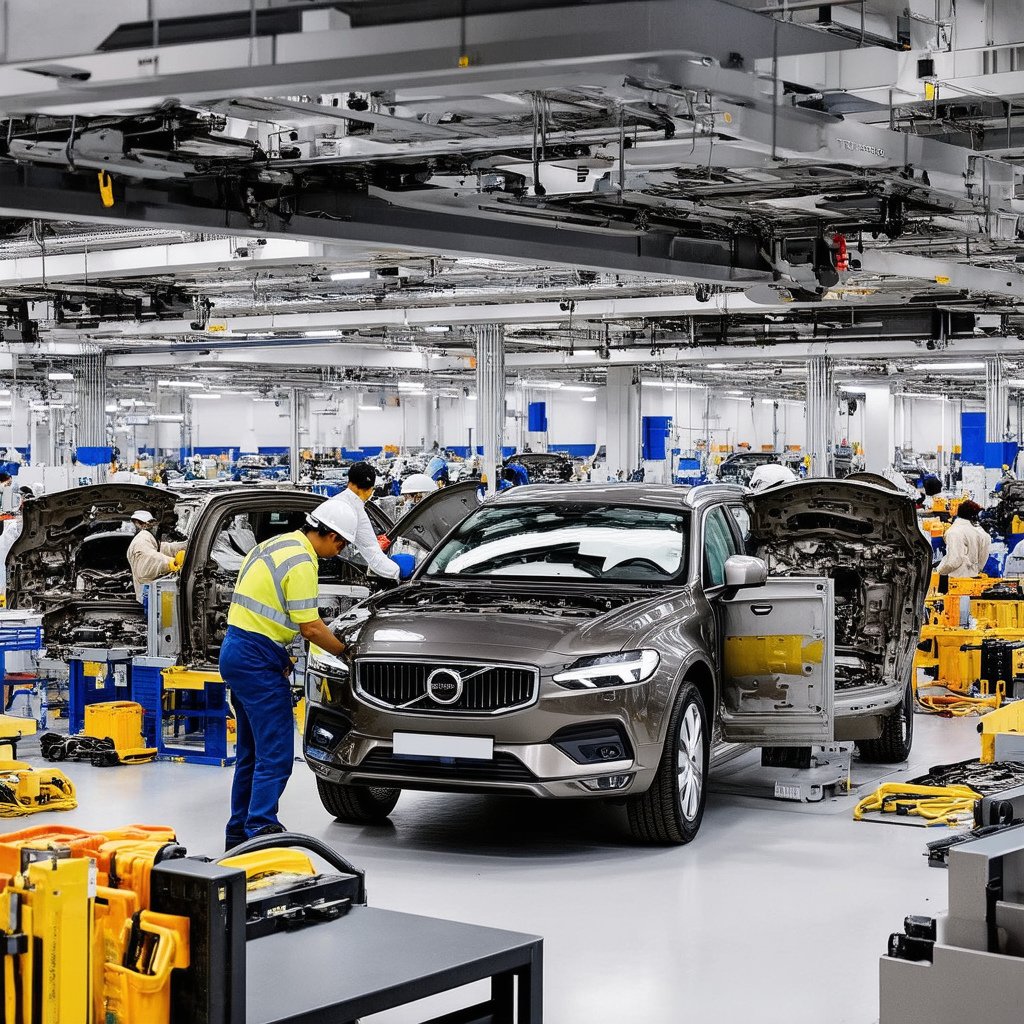} &
                \includegraphics[width=0.111\textwidth]{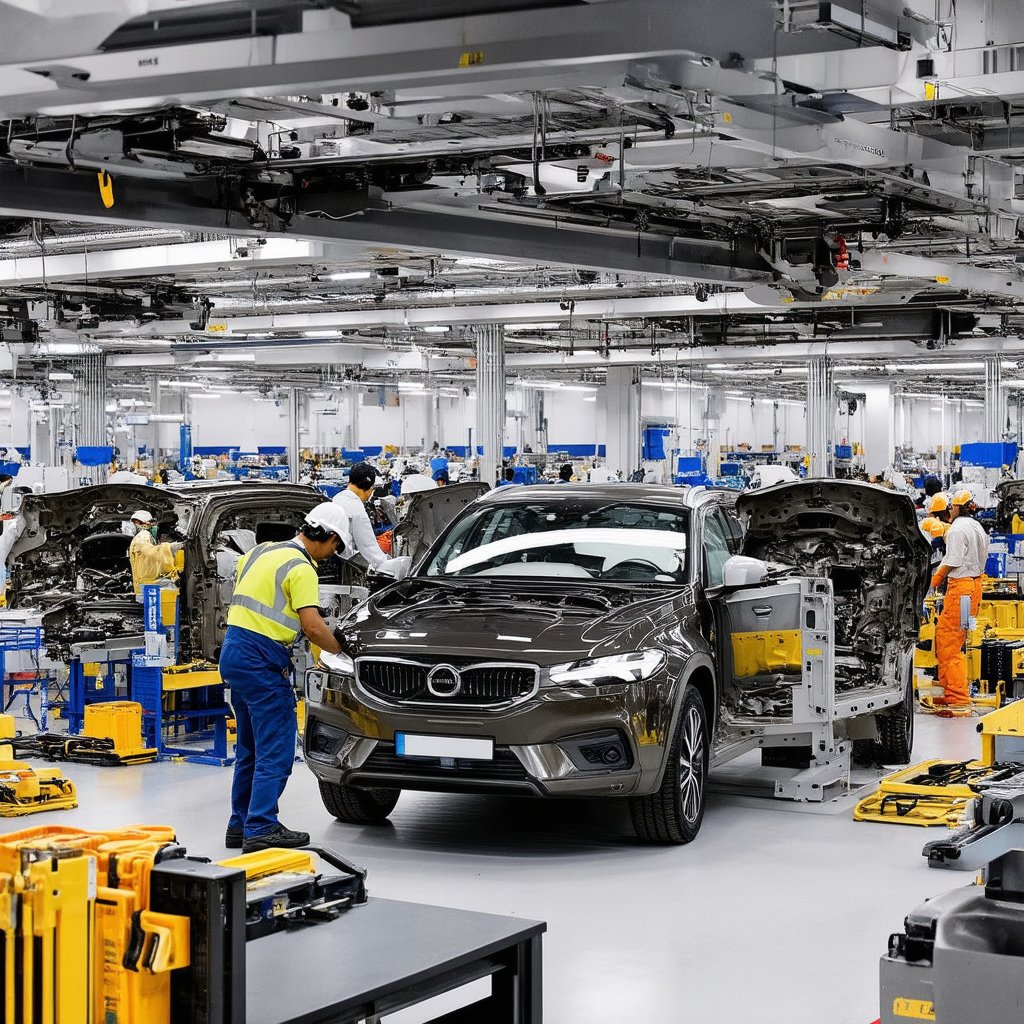} &
                        \includegraphics[width=0.111\textwidth]{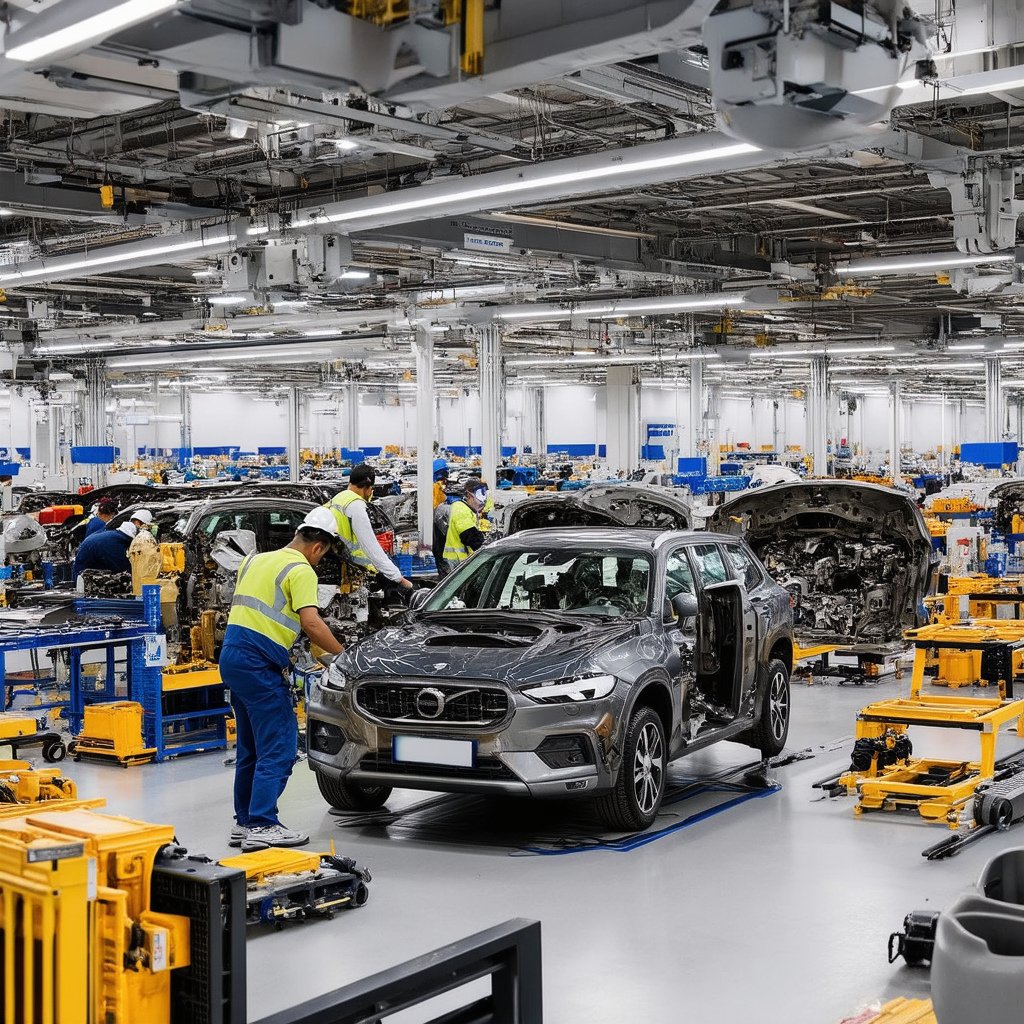} &
        \includegraphics[width=0.111\textwidth]{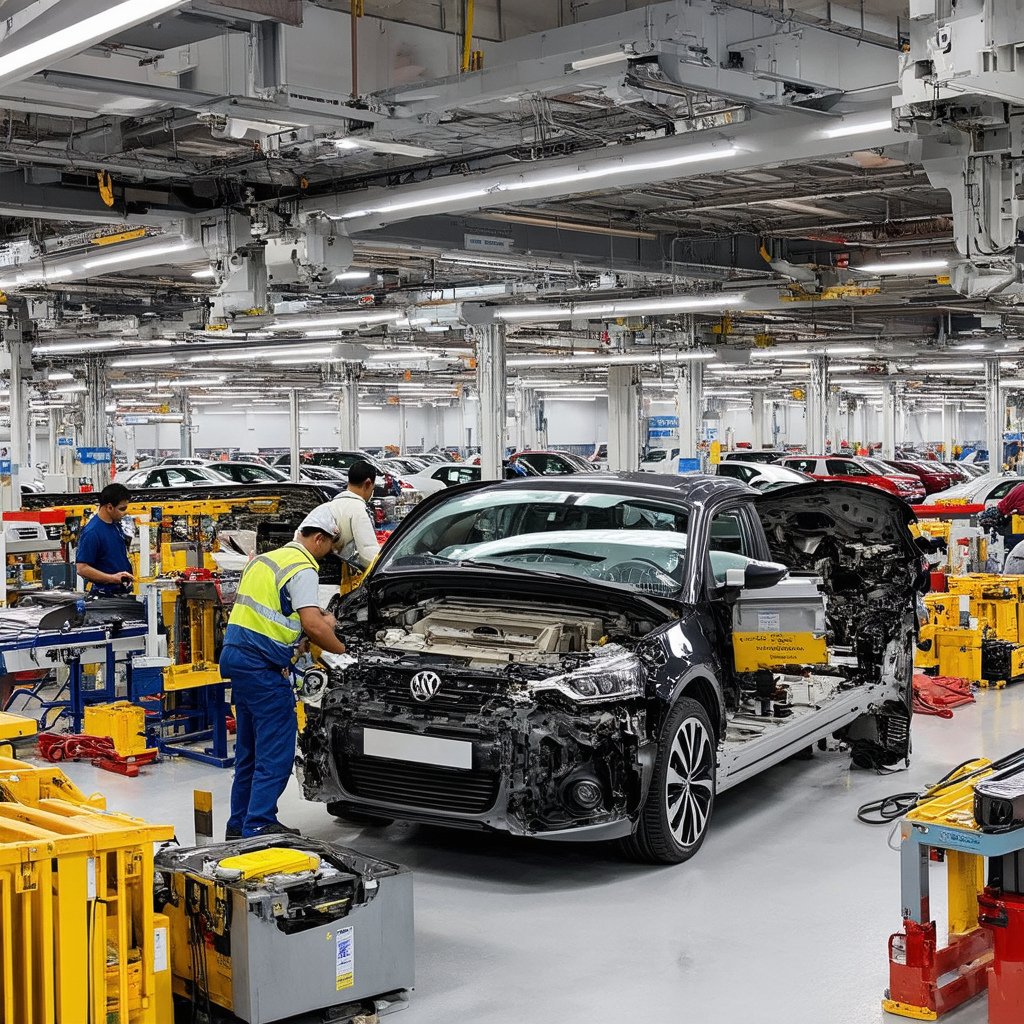} \\        

        \includegraphics[width=0.111\textwidth]{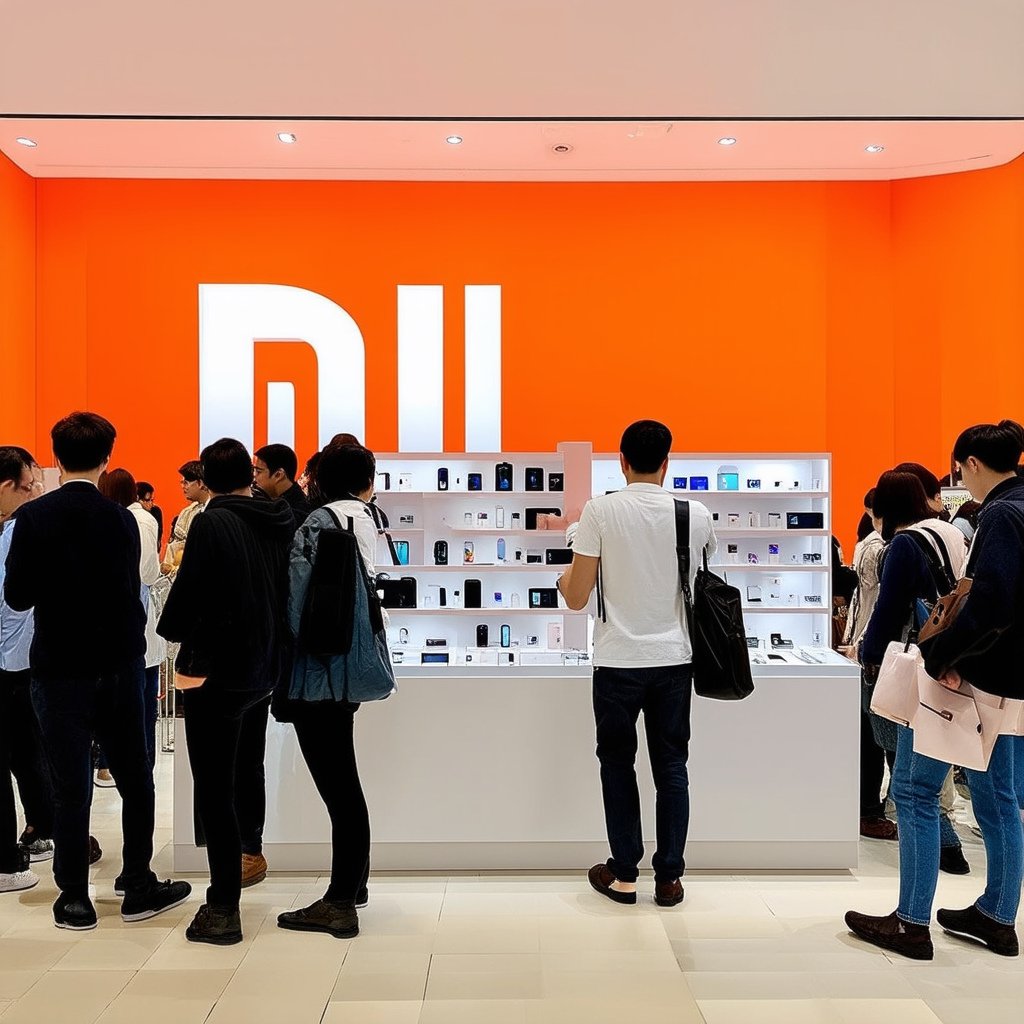} &
        \includegraphics[width=0.111\textwidth]{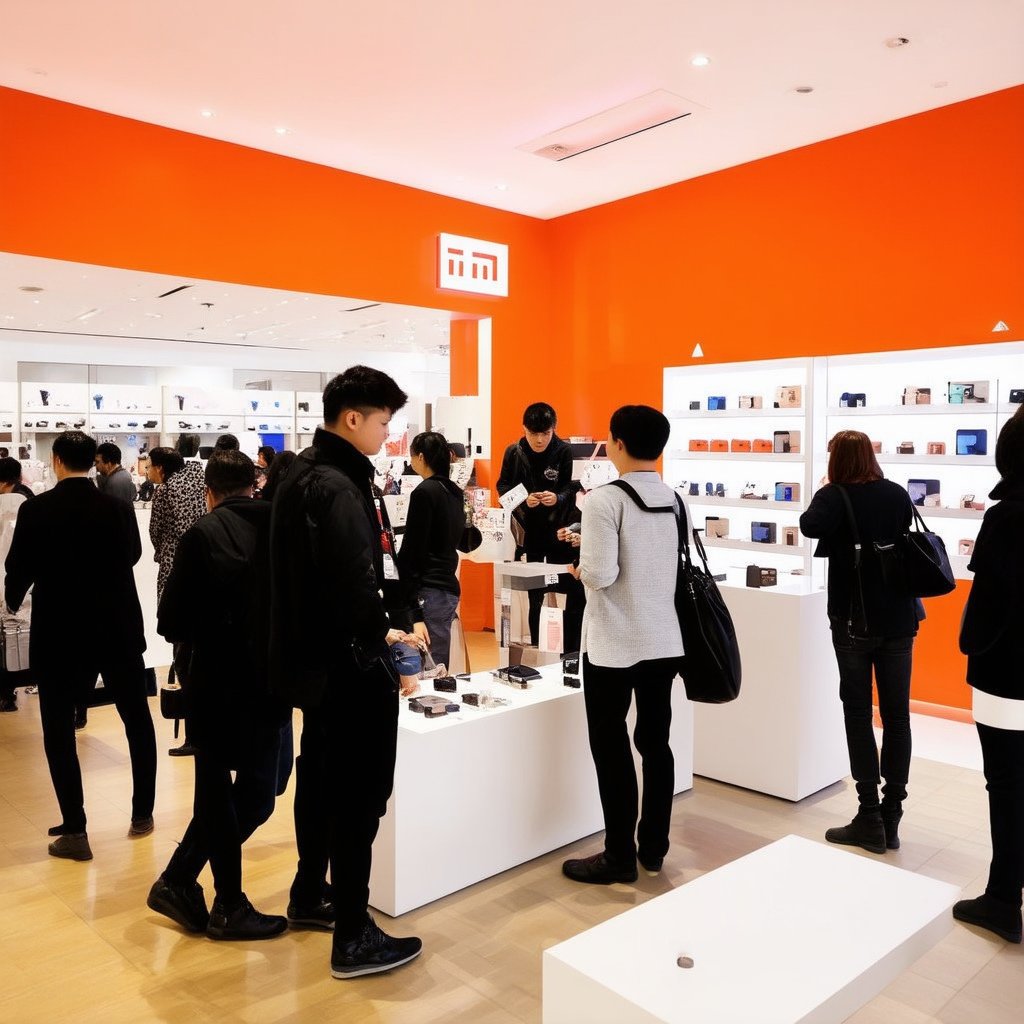} &
        \includegraphics[width=0.111\textwidth]{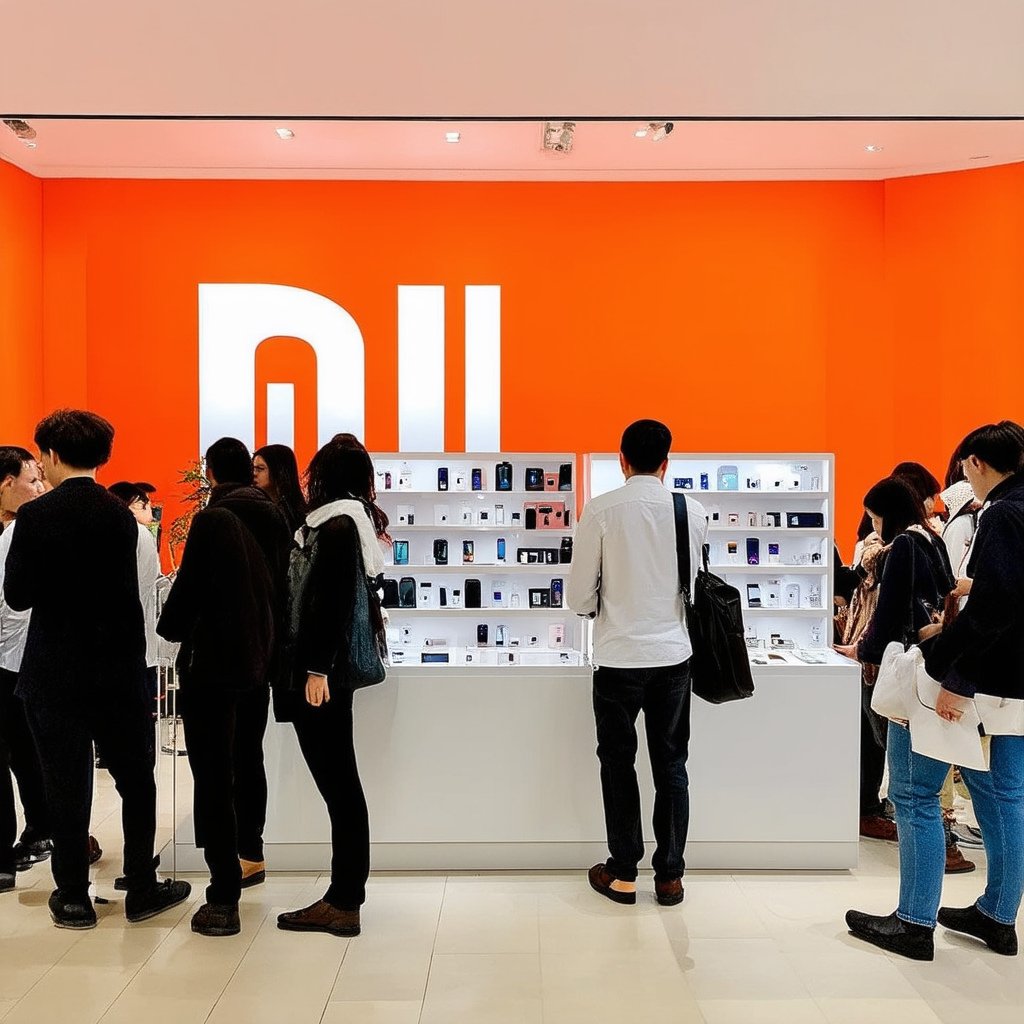} &
                \includegraphics[width=0.111\textwidth]{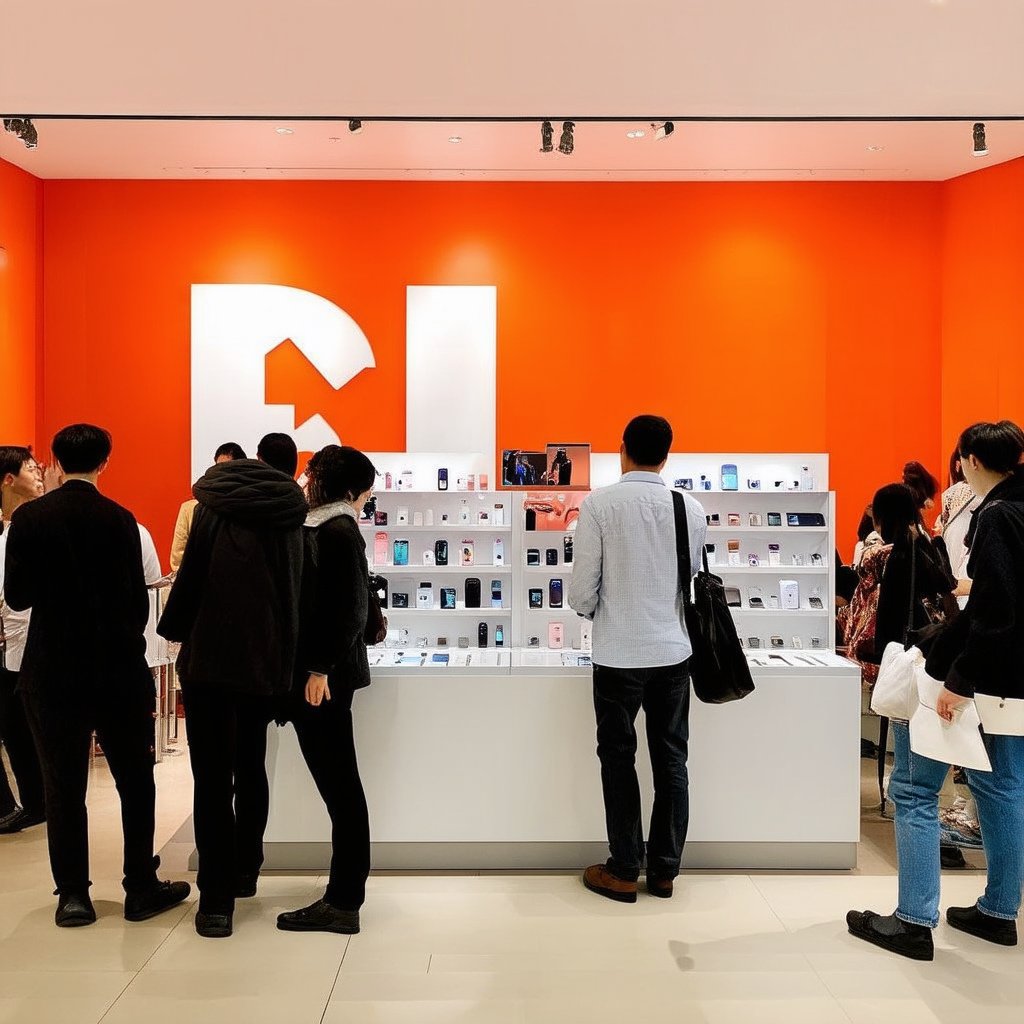} &
                        \includegraphics[width=0.111\textwidth]{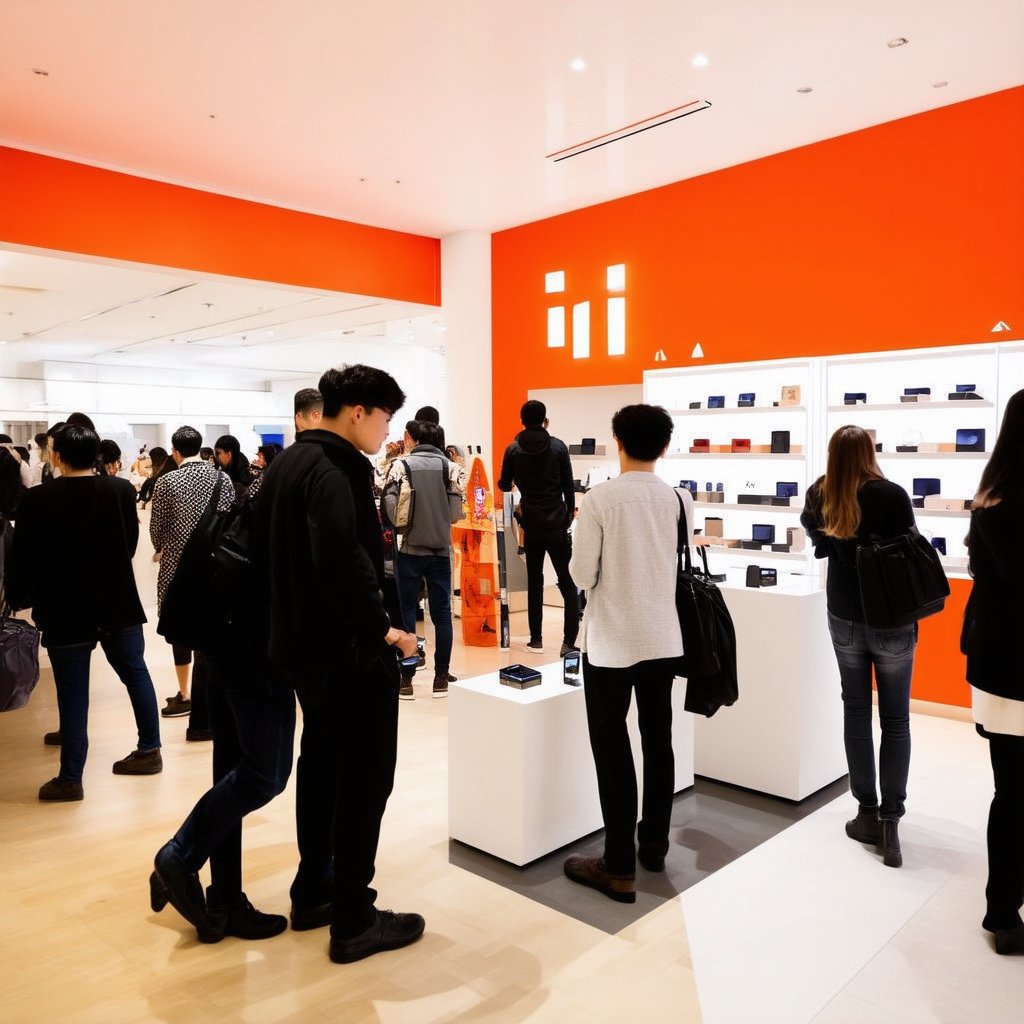} &
        \includegraphics[width=0.111\textwidth]{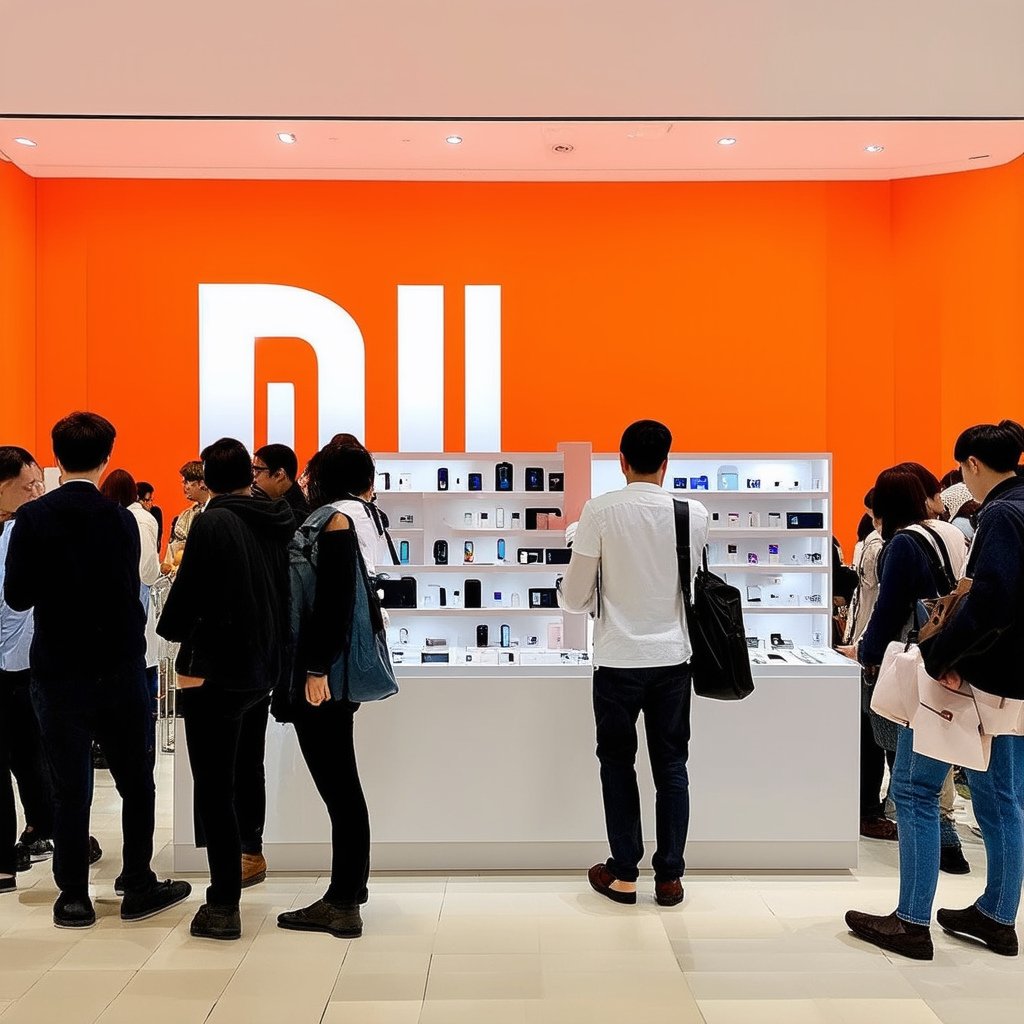} &
                \includegraphics[width=0.111\textwidth]{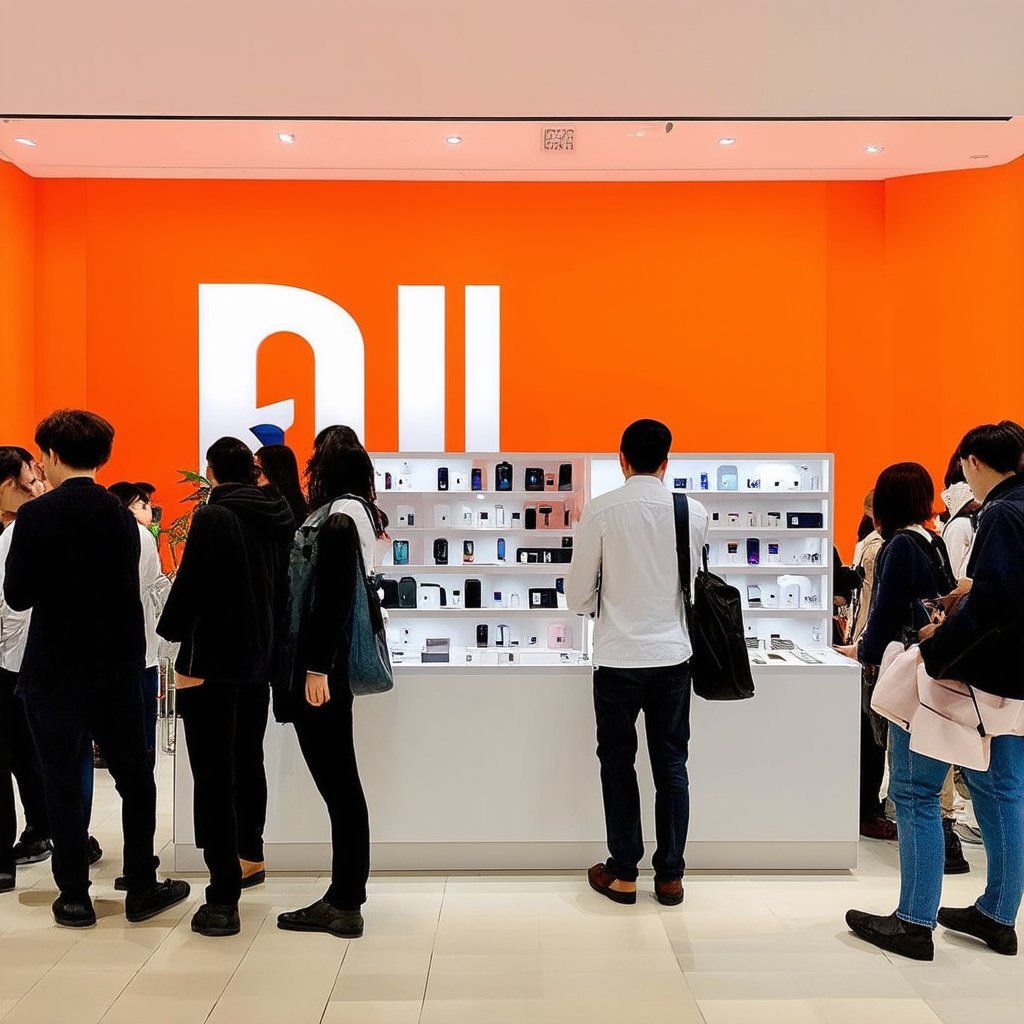} &
                        \includegraphics[width=0.111\textwidth]{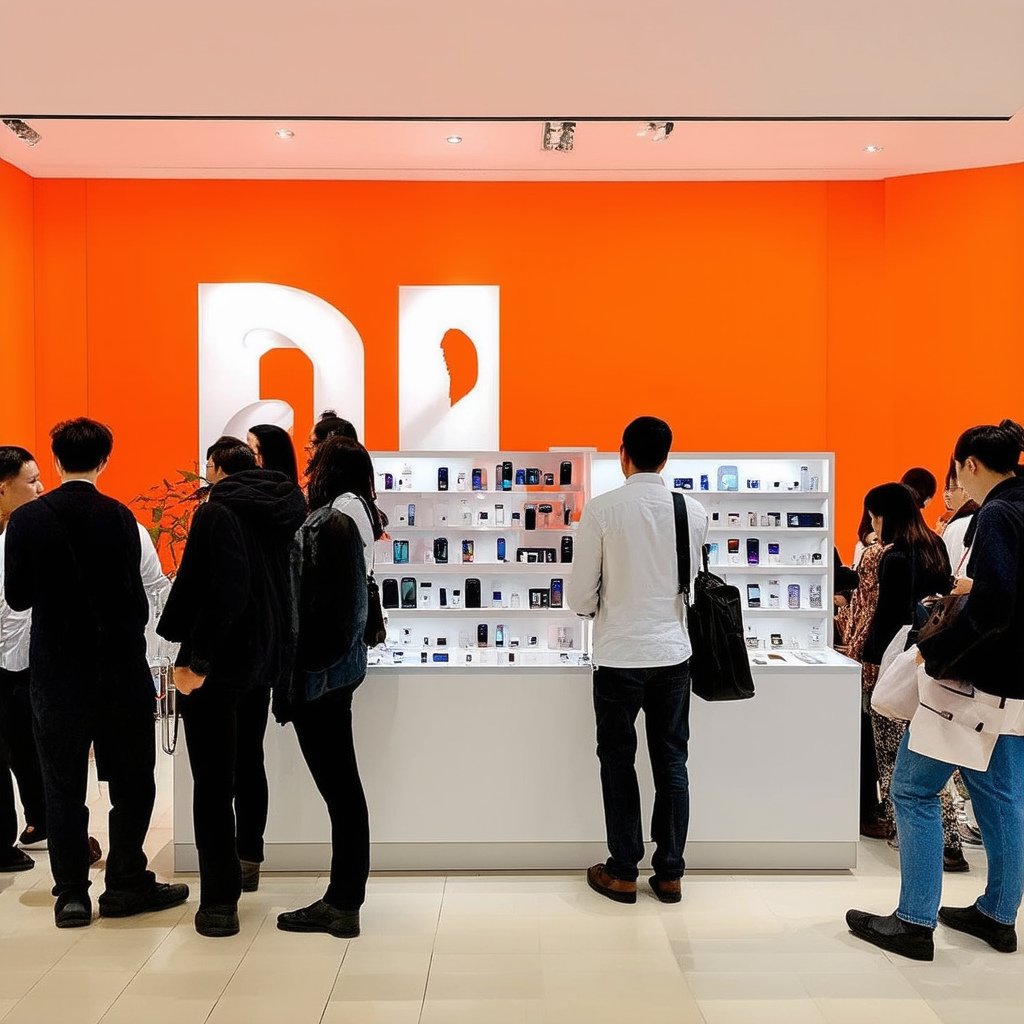} &
        \includegraphics[width=0.111\textwidth]{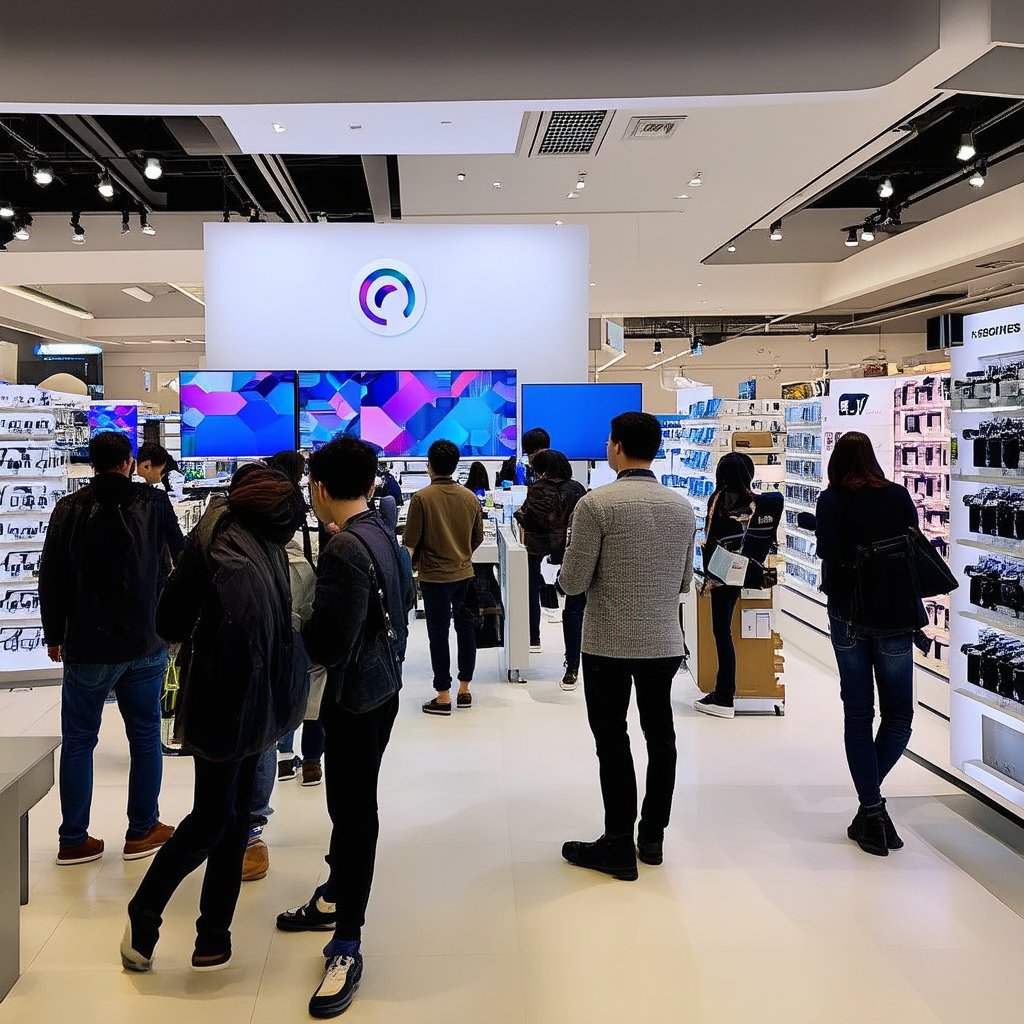} \\

        \includegraphics[width=0.111\textwidth]{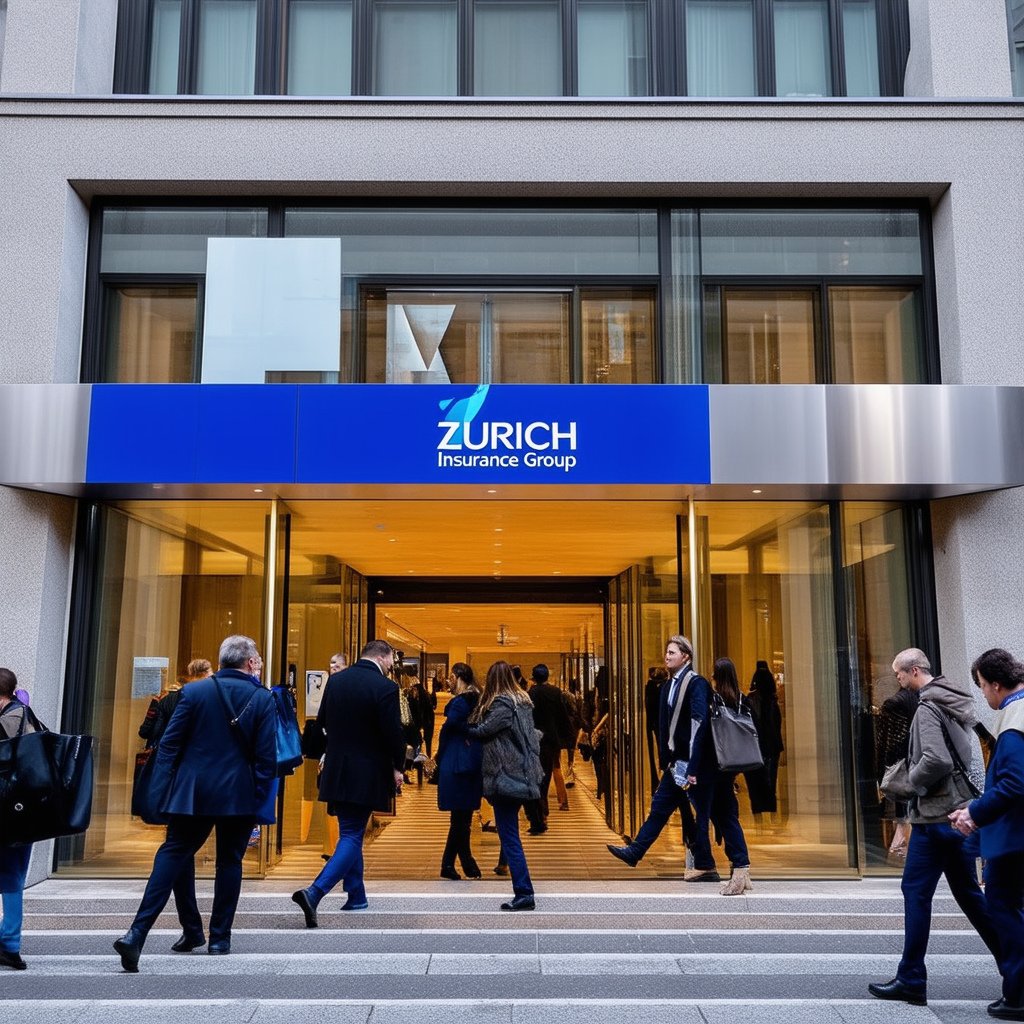} &
        \includegraphics[width=0.111\textwidth]{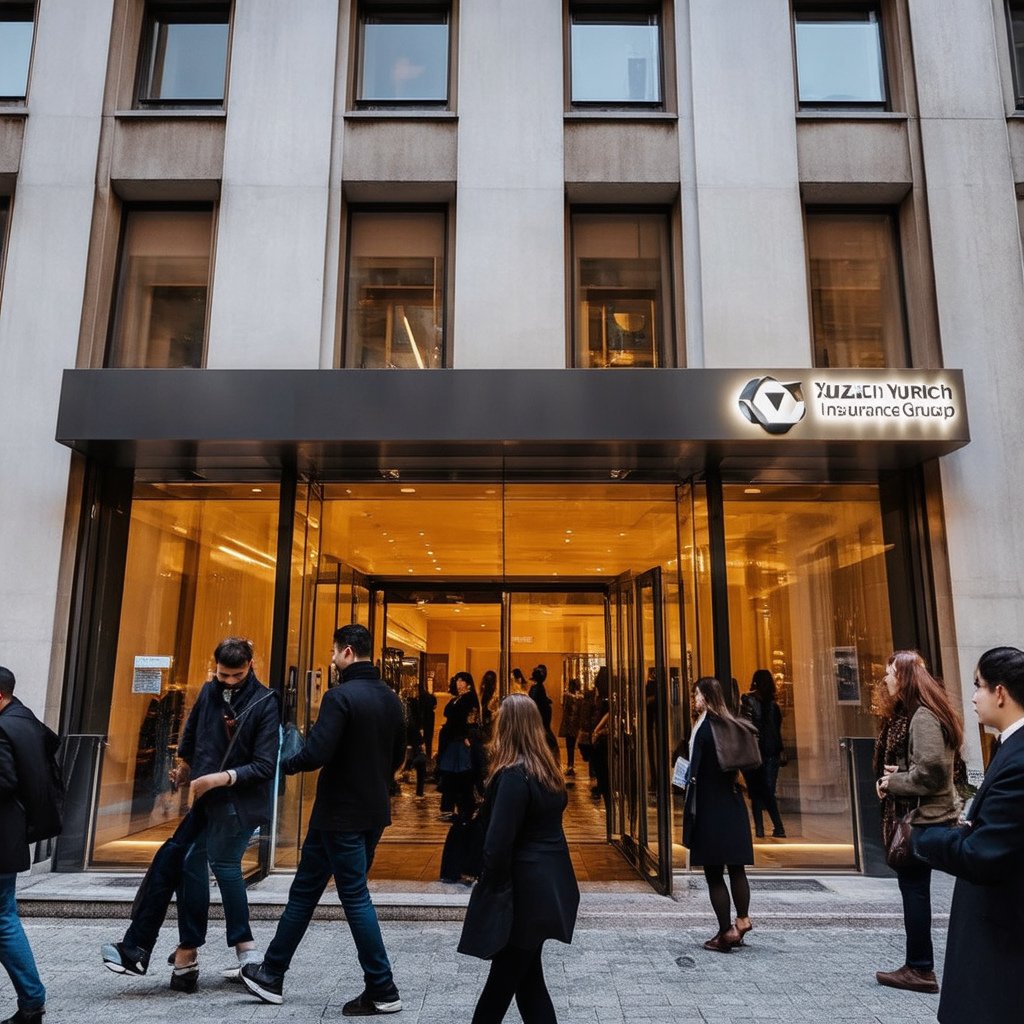} &
        \includegraphics[width=0.111\textwidth]{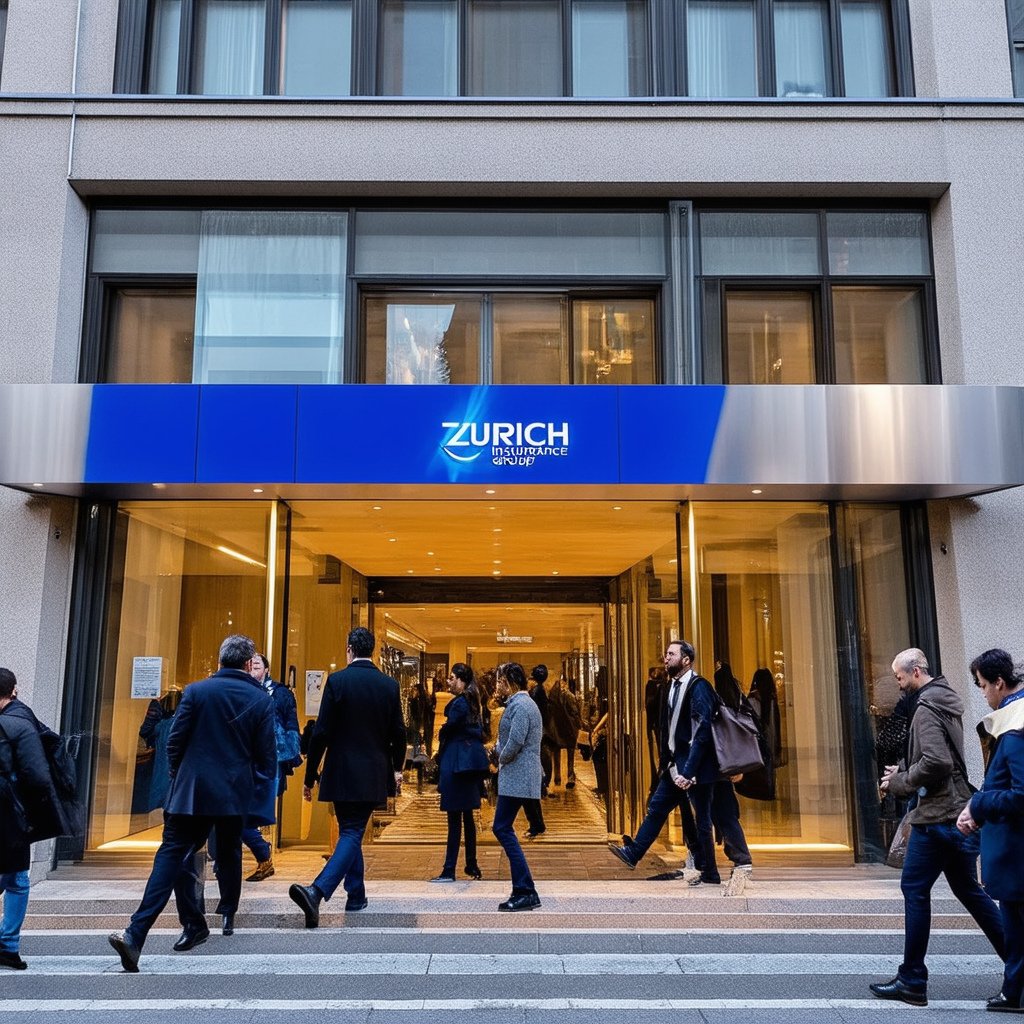} &
                \includegraphics[width=0.111\textwidth]{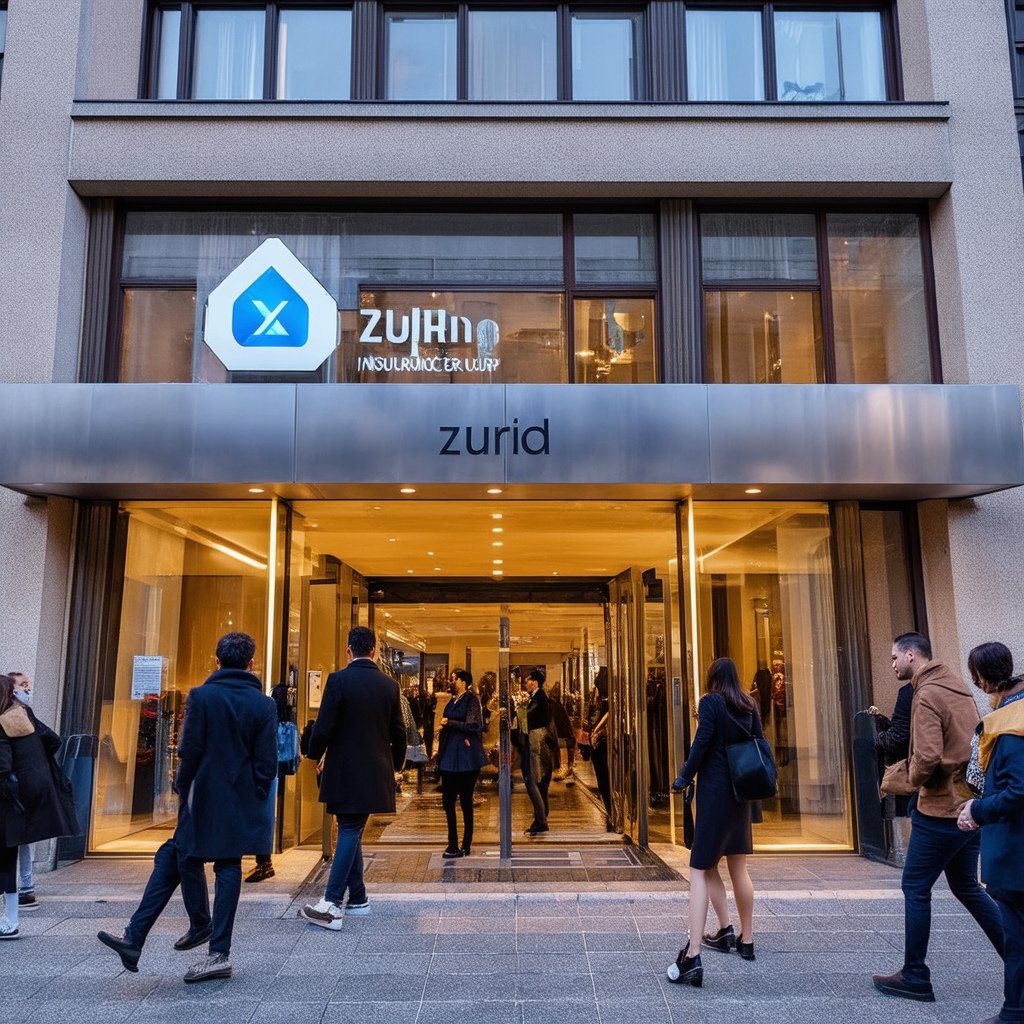} &
                        \includegraphics[width=0.111\textwidth]{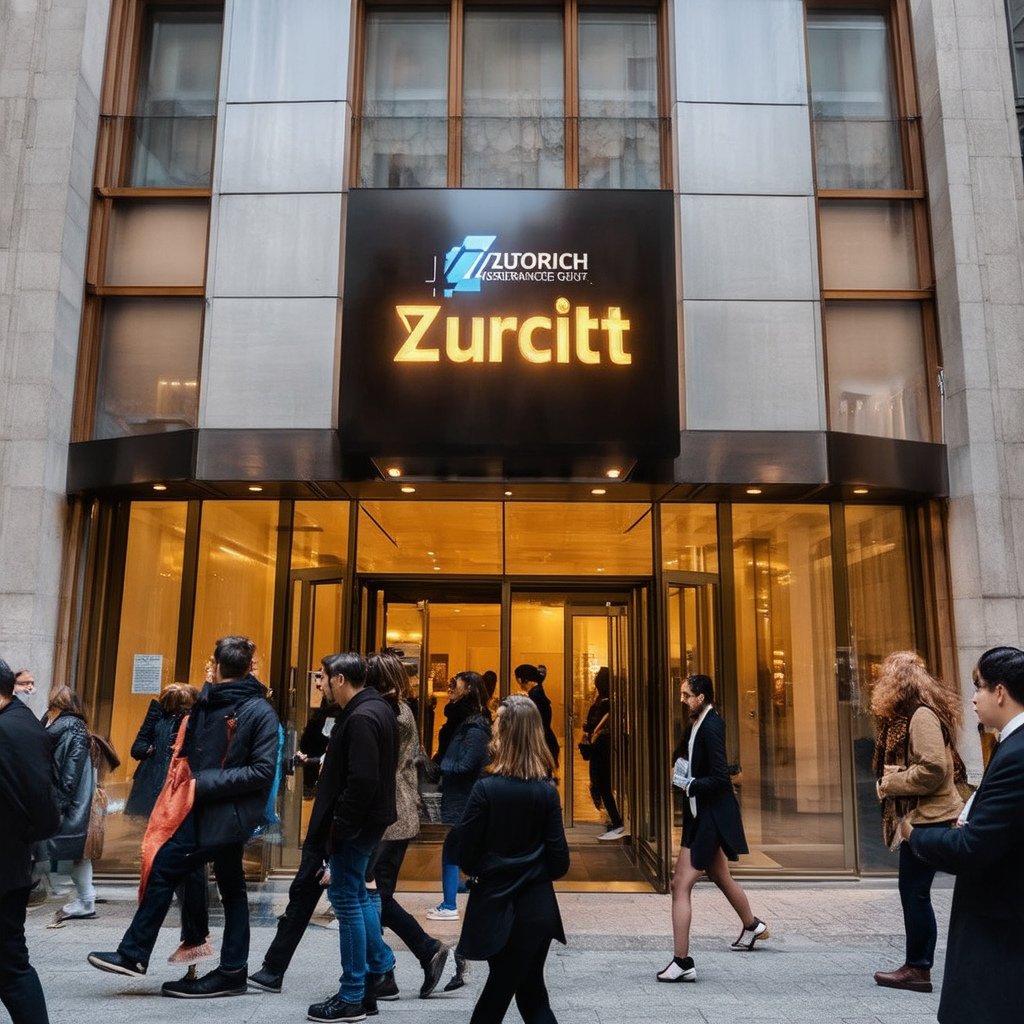} &
        \includegraphics[width=0.111\textwidth]{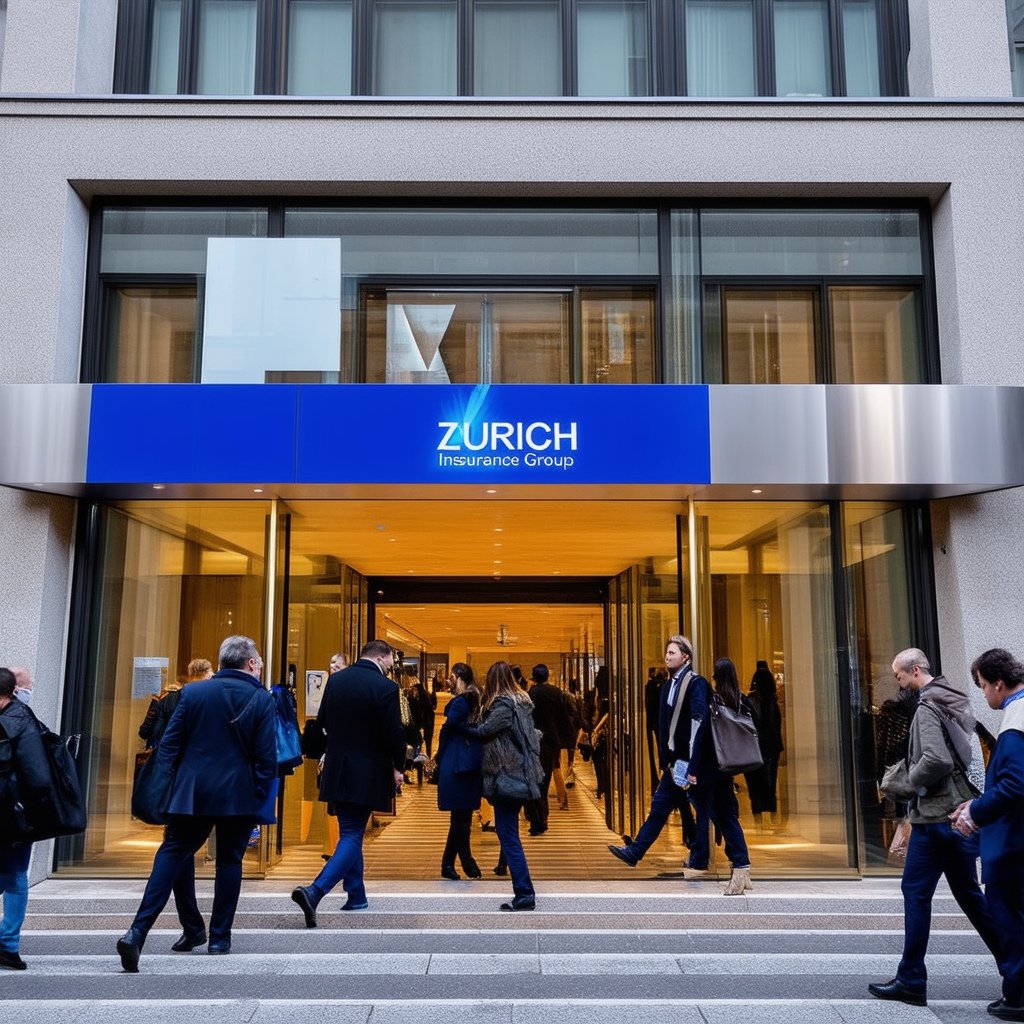} &
                \includegraphics[width=0.111\textwidth]{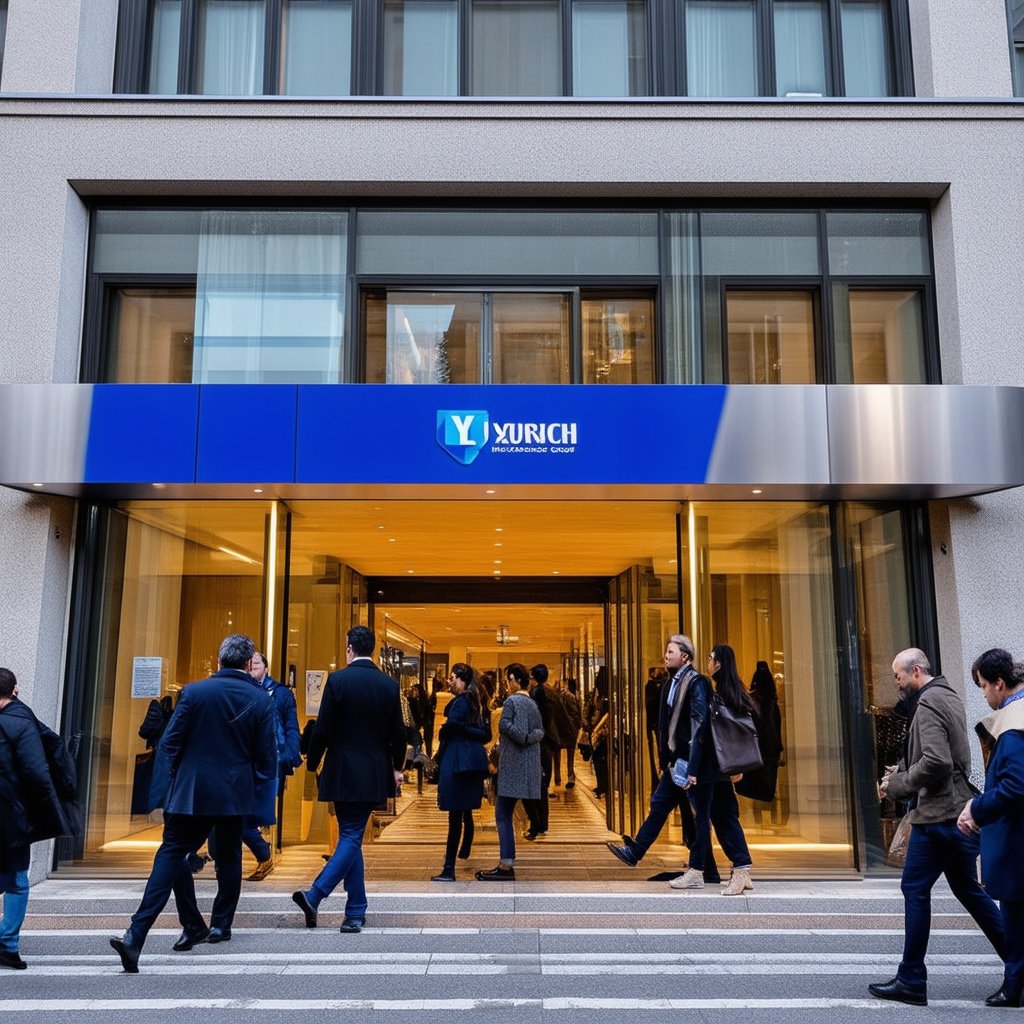} &
                        \includegraphics[width=0.111\textwidth]{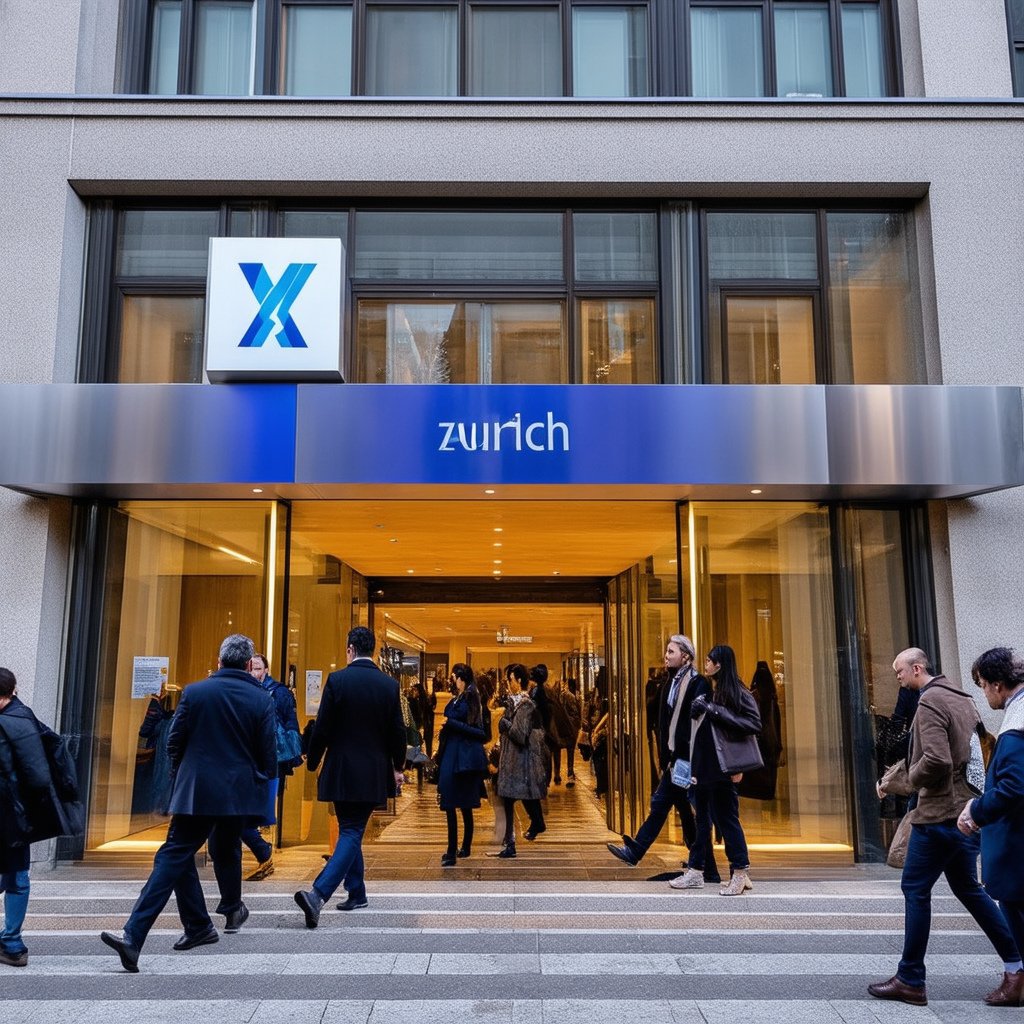} &
        \includegraphics[width=0.111\textwidth]{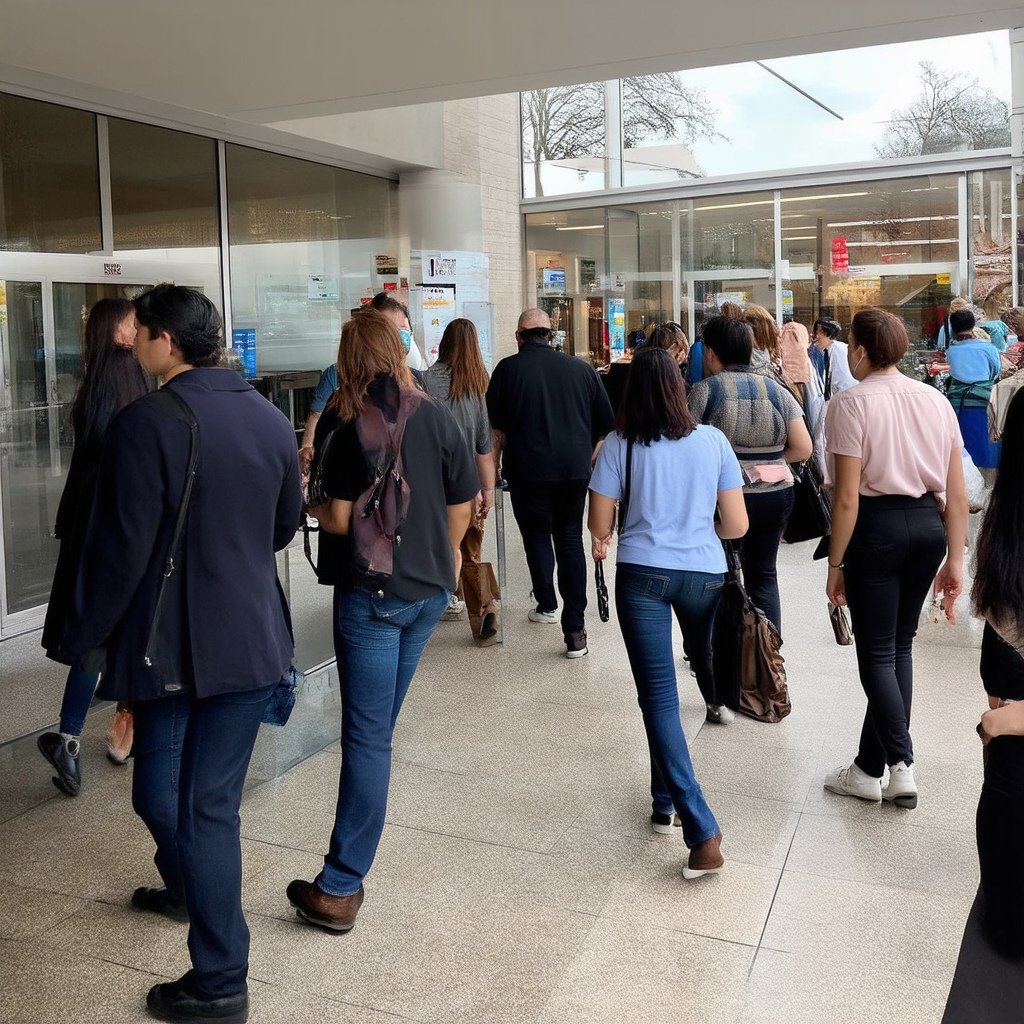} \\

    \end{tabular}

    \caption{More visual results on Oracle, Pfizer, Qualcomm, Renault, Uber, Volvo, Walt Disney, Xiaomi, and Zurich Insurance Group.}

\label{tab:results_supp_2}
\end{figure*}

Finally, to provide an exhaustive demonstration of unlearning dynamics across a diverse topological and semantic spectrum, we supply an extended gallery of supplementary visual results. These comprehensive qualitative evaluations are indexed across \Cref{tab:results_supp_1} and \Cref{tab:results_supp_2}, serving to further validate the behavioral boundaries of our approach.

To ensure an unbiased and comprehensive representation of the Fortune Global 500 distribution—encompassing both text-heavy and geometrically complex designs—we systematically sampled one representative company for nearly every letter of the alphabet from the \ourBench corpus (excluding 'Y', which contained no valid entries). The evaluated cohort explicitly includes: Apple, Boeing, Coca-Cola, DELL, EXXON MOBIL, FedEx, Goldman-Sachs, HP, Intel, Johnson \& Johnson, KIA, L'Oreal, Mercedes-Benz, Nike, Oracle, Pfizer, Qualcomm, Renault, Starbucks, Tesla, Uber, Volvo, Walt Disney, Xiaomi, and Zurich Insurance Group.

\end{document}